%% file: main.tex
\documentclass[11pt,letterpaper]{mystyle}
\usepackage{fancyhdr}

\usepackage[utf8]{inputenc} 
\usepackage[T1]{fontenc}
\usepackage[numbers]{natbib}
\usepackage{adjustbox}
\usepackage{breakurl}    
\usepackage{hyperref}
\usepackage[edges]{forest}
\usepackage{subcaption}
\usepackage{soul}
\usepackage{multirow}
\usepackage[utf8]{inputenc} 
\usepackage{booktabs}       
\usepackage{amsfonts}       
\usepackage{nicefrac}      
\usepackage{microtype}     
\usepackage{xcolor}        
\usepackage{colortbl}
\usepackage{enumitem}
\usepackage{amssymb}
\usepackage{wrapfig}
\usepackage{courier}
\usepackage{bxcoloremoji}
\usepackage{CJKutf8}

\newcommand{\github}{\raisebox{-1.5pt}{\includegraphics[height=1.05em]{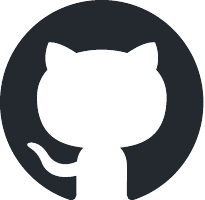}}}

\fancypagestyle{headstyle}{
    \fancyhead[L]{
        \includegraphics[width=80pt]{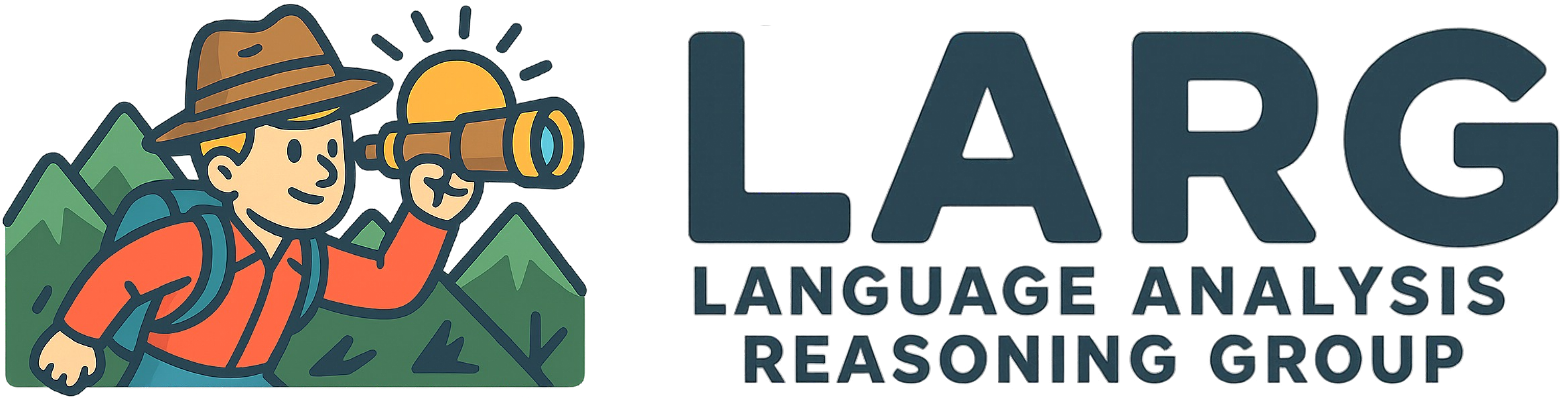}
    }
    \fancyhead[C]{}
    \fancyhead[R]{}

}

\definecolor{hidden-red}{RGB}{205, 44, 36}
\definecolor{hidden-blue}{RGB}{194,232,247}
\definecolor{hidden-orange}{RGB}{243,202,120}
\definecolor{hidden-green}{RGB}{34,139,34}
\definecolor{hidden-pink}{RGB}{255,245,247}
\definecolor{hidden-black}{RGB}{20,68,106}
\definecolor{purple}{RGB}{144,153,196}
\definecolor{yellow}{RGB}{255,228,123}
\definecolor{hidden-yellow}{RGB}{255,248,203}
\definecolor{tkcolor}{RGB}{224,223,255}
\definecolor{darkblue}{rgb}{0, 0.40, 0.75}
\newcommand{\eg}{\textit{e.g.,}}
\hypersetup{colorlinks=true, citecolor=darkblue, linkcolor=darkblue, urlcolor=darkblue}
\tcbset{
  aibox/.style={
    width=\linewidth,
    top=8pt,
    bottom=4pt,
    colback=blue!6!white,
    colframe=black,
    colbacktitle=black,
    enhanced,
    center,
    attach boxed title to top left={yshift=-0.1in,xshift=0.15in},
    boxed title style={boxrule=0pt,colframe=white,},
  }
}

\newtcolorbox{AIbox}[2][]{aibox,title=#2,#1}

\tcbset{
  takeawaybox/.style={
    width=\linewidth,
    top=8pt,
    bottom=4pt,
    colback=hidden-yellow,
    colframe=black,
    colbacktitle=black,
    enhanced,
    center,
    attach boxed title to top left={yshift=-0.1in,xshift=0.15in},
    boxed title style={boxrule=0pt,colframe=white,},
  }
}

\newtcolorbox{TakeawayBox}[2][]{takeawaybox,title=#2,#1}

\title{AI4Research: A Survey of Artificial Intelligence for Scientific Research}

\author{
  Qiguang Chen$^{1*}$ \quad Mingda Yang$^{1*}$ \quad Libo Qin$^{2, \coloremojicode{2709}}$ \quad Jinhao Liu$^1$ \quad Zheng Yan$^1$ \quad Jiannan Guan$^1$ \quad Dengyun Peng$^1$ \quad Yiyan Ji$^1$ \quad Hanjing Li$^1$ \quad Mengkang Hu$^3$ \quad Yimeng Zhang$^4$ \quad Yihao Liang$^5$ \quad Yuhang Zhou$^{6}$ \quad Jiaqi Wang$^{7}$ \quad Zhi Chen$^{8}$  \quad Wanxiang Che$^{1, \coloremojicode{2709}}$\\
\normalfont{$^1$ LARG, Research Center for Social Computing and Interactive Robotics, Harbin Institute of Technology,\vspace{-5pt}\\
$^2$ School of Computer Science and Engineering, Central South University,
$^3$ The University of Hong Kong,\vspace{-5pt}\\
$^4$ University of Illinois Urbana-Champaign,
$^5$ Princeton University,
$^6$ Fudan University, \vspace{-5pt}\\
$^7$ Chinese University of Hong Kong, 
$^8$ ByteDance Seed (China) \\
}}

\begin{document}

\begin{abstract}
  \vspace{5mm}
  \textbf{\large Abstract:}
  \vspace{2mm}

  Recent advancements in artificial intelligence (AI), particularly in large language models (LLMs) such as OpenAI-o1 and DeepSeek-R1, have demonstrated remarkable capabilities in complex domains such as logical reasoning and experimental coding. Motivated by these advancements, numerous studies have explored the application of AI in the innovation process, particularly in the context of scientific research. These AI technologies primarily aim to develop systems that can autonomously conduct research processes across a wide range of scientific disciplines. Despite these significant strides, a comprehensive survey on AI for Research (AI4Research) remains absent, which hampers our understanding and impedes further development in this field. To address this gap, we present a comprehensive survey and offer a unified perspective on AI4Research. Specifically, the main contributions of our work are as follows:  (1) \textbf{\textit{Systematic taxonomy:}} We first introduce a systematic taxonomy to classify five mainstream tasks in AI4Research. (2) \textbf{\textit{New frontiers:}} Then, we identify key research gaps and highlight promising future directions, focusing on the rigor and scalability of automated experiments, as well as the societal impact. (3) \textbf{\textit{Abundant applications and resources:}} Finally, we compile a wealth of resources, including relevant multidisciplinary applications, data corpora, and tools. We hope our work will provide the research community with quick access to these resources and stimulate innovative breakthroughs in AI4Research.
  \vspace{5mm}

  $^{*}$ \textit{Equal Contribution}

  $^{\coloremojicode{2709}}$ \textit{Corresponding Author}

  \vspace{5mm}
  \vspace{5mm}

  \coloremojicode{1F4C5} \textbf{Date}: July 12, 2025

  \coloremojicode{1F3E0} \textbf{Projects}: \href{https://ai-4-research.github.io}{https://ai-4-research.github.io}

  \github{} \textbf{Code Repository}: \href{https://github.com/LightChen233/Awesome-AI4Research}{https://github.com/LightChen233/Awesome-AI4Research}

  \coloremojicode{1F4E7} \textbf{Contact}: \href{mailto:qgchen@ir.hit.edu.cn}{qgchen@ir.hit.edu.cn}, \href{mailto:car@ir.hit.edu.cn}{car@ir.hit.edu.cn}, \href{mailto:lbqin@csu.edu.cn}{lbqin@csu.edu.cn}

\end{abstract}
\maketitle

\vspace{3mm}
\pagestyle{headstyle}
\thispagestyle{empty}
\newpage
\tableofcontents

\input{sections/introduction}

\input{sections/definition}

\input{sections/comprehension}

\input{sections/survey}

\input{sections/discovery}

\input{sections/writing}

\input{sections/review}

\input{sections/application}

\input{sections/resources}

\input{sections/future}

\input{sections/related_work}

\input{sections/conclusion}

\bibliographystyle{refstyle}
\bibliography{ref}

\end{document}

%% file: sections/introduction.tex
\vspace{-2mm}\section{Introduction}\vspace{-1mm}
\begin{figure*}[!b]
	\centering
	\includegraphics[width=0.99\textwidth]{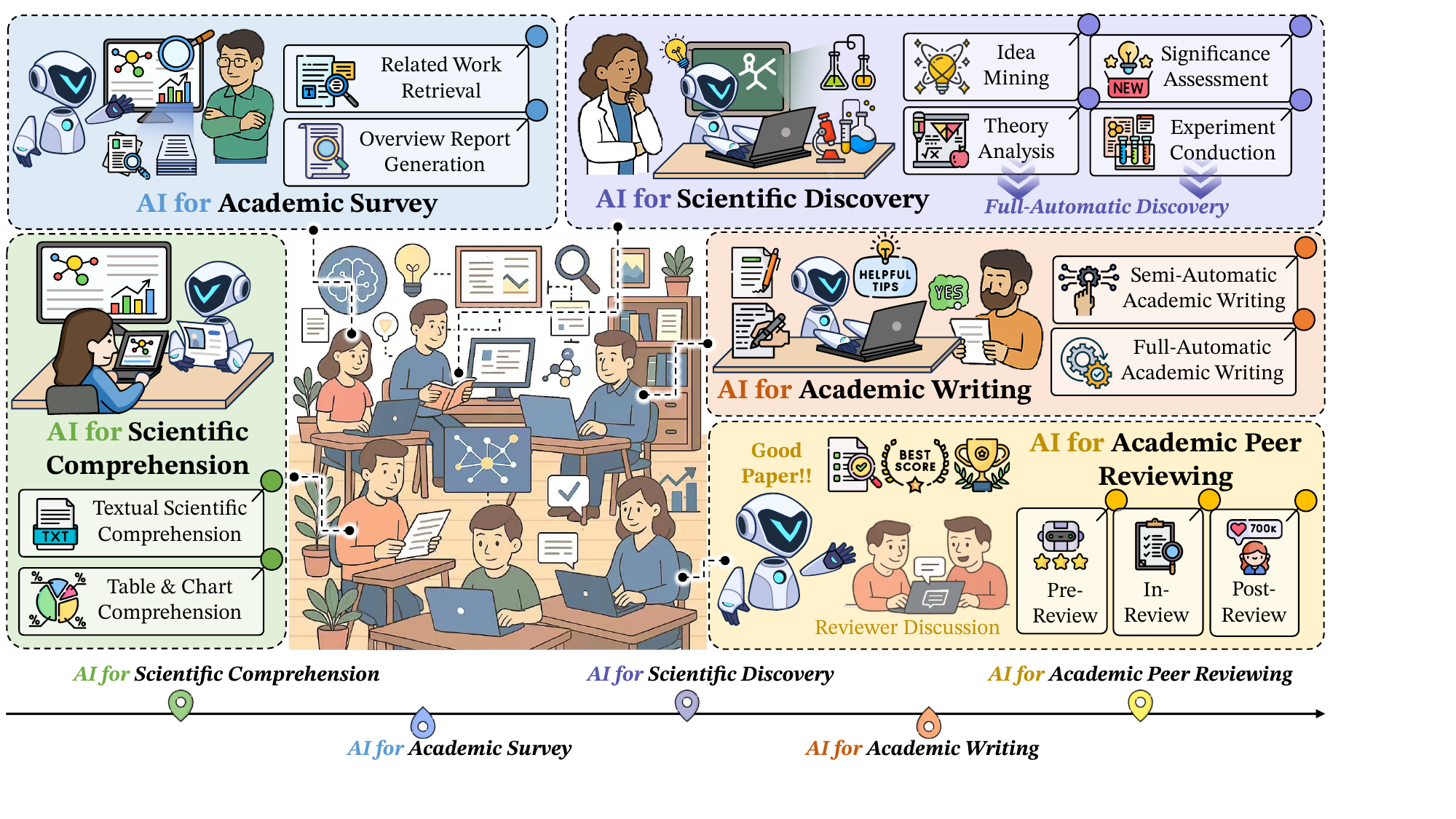}
	\caption{The mainstream processes and categories of AI4Research, which can be divided into five key areas: (1) AI for Scientific Comprehension, (2) AI for Academic Survey, (3) AI for Scientific Discovery, (4) AI for Academic Writing, and (5) AI for Academic Peer Review. Each of these areas contributes to improving the effectiveness and efficiency of AI-integrated research and publication.}
	\label{fig:main}
\end{figure*}
In recent years, the rise of artificial intelligence (AI), particularly large language models (LLMs) like DeepSeek-R1~\citep{guo2025deepseek}, has stimulated significant research in the field of reasoning. These breakthroughs have notably improved the models' performance across diverse areas, including mathematical reasoning, programming, and interdisciplinary knowledge~\citep{sun2023survey, team2025kimi,qin2024large,zhong2024evaluation, zhuang2023through,chen2025towards}. Some of these models have even surpassed the Turing Test~\citep{jones2025large}, marking a pivotal achievement in AI development.
Inspired by these, a series of works attempts to explore advanced AI systems for innovative tasks, especially in the scientific discovery of new research~\citep{yax2024studying,zeng2024exploring,yager2024towards,zimmermann202534}.
Earlier, the AI Scientist~\citep{lu2024ai} introduces the concept of a fully automatic AI for research system, which divides the research process into three key stages: idea mining, experiment conduction, and academic writing. Initially, the system generates and evaluates novel ideas and hypotheses. Once a hypothesis is formulated, experiments are conducted automatically, producing results that include numerical data and visual summaries.
These results are presented in tables and images, followed by an interpretation with a convincing description, culminating in a LaTeX report. In the final stage, the AI Scientist generates an automated review that refines the project and provides feedback for future open scientific discoveries.
Similarly, other classic models, such as Carl~\citep{carl2025} and Zochi~\citep{zochi2025}, follow broadly analogous workflows.
Notably, AgentArxiv~\citep{schmidgall2025agentrxiv} and AgentLab~\citep{schmidgall2025agent} assign distinct roles to multiple agents to simulate the collaborative nature of scientific research teams, incorporating additional peer review, academic survey, enabling semi-automatic and even full-automatic collaboration rather than relying on a single agent~\citep{liu2025advances,yu2024natural,chen2025ai,saetra2025rise,beel2025evaluating}.
More recently, \citet{swanson2025virtual} extend LLMs’ scientific reach by presenting the Virtual Lab, an AI–human platform that designs nanobody binders for emerging SARS-CoV-2 variants, underscoring the promise of such collaboration for research.
Despite these advancements, there remains a lack of comprehensive surveys to systematically analyze the key factors and recent developments in AI-driven research, which significantly impedes the continued progress of this field.

To address this gap, we first define and present a comprehensive survey of AI for research, termed AI4Research. As shown in Figure~\ref{fig:main}, we \textbf{introduce a systematic taxonomy of AI4Research}, focusing on the following areas: (1) \textit{AI for Scientific Comprehension:} AI systems' ability to extract relevant information from scientific literature is crucial; (2) \textit{AI for Academic Surveys:} This involves AI techniques for systematically reviewing and summarizing scientific literature; (3) \textit{AI for Scientific Discovery:} AI is used to generate hypotheses, theories, or models based on existing scientific knowledge; (4) \textit{AI for Academic Writing:} AI tools support researchers in drafting, editing, and formatting manuscripts; (5) \textit{AI for Academic Reviewing:} AI assists in evaluating and providing feedback on scientific manuscripts.
Given the vast literature, we \textbf{highlight promising future research in AI4Research}. Future work should prioritize interdisciplinary AI models that integrate knowledge across scientific domains to encourage cross-disciplinary collaboration. Addressing ethical concerns and biases within AI systems is crucial for ensuring fairness and transparency in research. Improving the explainability of AI models and exploring adaptive, real-time systems for dynamic scientific experiments will be vital for advancing AI’s role in research.
Additionally, we \textbf{suggest key applications and valuable resources in AI4Research}, such as representative multidisciplinary applications, open-source frameworks, and datasets repositories to support further studies. We introduce AI for Natural Science research, AI for Applied Science and Engineering research, and AI for Social Science research. Finally, we review tools essential for model development and public benchmarks that provide rich data for training and experimentation.

The main contributions of this work are as follows:
\begin{itemize}[left=2pt,topsep=1pt,itemsep=2pt, parsep=1pt]
	\item \textbf{Systematic Taxonomy for AI in Research:} This paper introduces a comprehensive taxonomy of AI applications in research, spanning five areas: scientific comprehension, academic surveys, scientific discovery, academic writing, and academic reviewing. It categorizes AI tools that enhance and even automatically execute various stages of the research process.
	\item \textbf{Emerging Future Research Areas:} The paper identifies key future research avenues for AI in academia, including the development of interdisciplinary AI models, addressing ethical concerns and biases, improving model explainability, and exploring adaptive AI systems for dynamic scientific experiments.
	\item \textbf{Key Applications and Abundant Trending Resources:} We present multidisciplinary applications of AI4Research across natural sciences, applied science, and social sciences. It also identifies essential resources, open-source frameworks, public datasets, collaborative platforms, and academic tools, that facilitate discovery management and AI-driven research.
\end{itemize}

%% file: sections/definition.tex
\vspace{-2mm}\section{The Definition of AI4Research}\vspace{-1mm}
\input{figures/tree_taxonomy.tex}
AI4Research denotes the application of artificial intelligence methods to improve, accelerate, and partially automate research across disciplines. To clarify this paradigm, as shown in Figure~\ref{fig:ai4research-taxonomy}, we identify six core capabilities: AI for Scientific Comprehension, AI for Academic Survey, AI for Scientific Discovery, AI for Academic Writing, AI for Academic Peer Reviewing. Each of them illustrates a distinct way that AI advances the research process.
Formally, let $
    \mathcal{T} = \bigl\{\,T_{\mathrm{SC}},\,T_{\mathrm{AS}},\,T_{\mathrm{SD}},\,T_{\mathrm{AW}},\,T_{\mathrm{PR}}\,\bigr\}
$ be the set of research tasks, Scientific Comprehension, Academic Survey, Scientific Discovery, Academic Writing, and Peer Reviewing. For each task $T_i\in\mathcal{T}$, there exists a corresponding AI model $A_i$ that is specifically tailored to address the requirements of that task. Then the overall AI4Research system can be expressed as the functional composition:
\begin{equation}
    \mathcal{A} = A_{\mathrm{PR}} \circ A_{\mathrm{AW}} \circ A_{\mathrm{SD}} \circ A_{\mathrm{AS}} \circ A_{\mathrm{SC}}\text{,}
\end{equation}
where $\circ$ denotes the function composition operator, meaning that the output of one function becomes the input of the next.
Furhter, applied to a research query $q$ (or more generally to an interactive research‐query lifecycle $\mathcal{Q}$), we can obtain:
\begin{equation}
    \mathcal{A}(q) = \bigl(A_{\mathrm{PR}}\circ A_{\mathrm{AW}}\circ A_{\mathrm{SD}}\circ A_{\mathrm{AS}}\circ A_{\mathrm{SC}}\bigr)(q).
\end{equation}

The objective of an AI4Research system is to maximize research lifecycle efficiency, application performance, and innovation capacity, namely:
\begin{equation}
    \max\;\bigl\{\,
    \eta(\mathcal{A}(\mathcal{Q})),\;\alpha(\mathcal{A}(\mathcal{Q})),\;\tau(\mathcal{A}(\mathcal{Q}))
    \bigr\}\text{,}
\end{equation}
where $\eta(\cdot)$, $\tau(\cdot)$, and $\alpha(\cdot)$ evaluates the  efficiency, performance, and innovation of the generated research publications $\mathcal{A}(\mathcal{Q})$, respectively.

\vspace{-2mm}\subsection{Component-wise Definition of AI4Research}\vspace{-1mm}

We now define and formalize each core module in the AI4Research framework.

\vspace{-2mm}\subsubsection{AI for Scientific Comprehension}\vspace{-1mm}
AI for Scientific Comprehension (AI4SC) is central to AI4Research, enabling extraction, interpretation, and synthesis of information from a single scientific literature. This accelerates human knowledge acquisition and improves the efficiency of automated analysis.
Formally, we define this module to gain knowledge $K$ after comprehension as a composite reasoning function:
\begin{equation}
    \hat{\mathcal{K}} = A_{SC}(\mathcal{K}) = f_{SC}(\mathcal{K}| D_{SC}, \Phi_{SC})  =  f_{\text{TCSC}} \circ f_{\text{TSC}}(\mathcal{K}| D_{SC}, \Phi_{SC})\text{,}
\end{equation}
where $A_{SC}$ is the comprehension AI model to extract the possible knowledge $\mathcal{K}$; the documents
$D_{SC} = \{D_T, D_F, D_M\}$ comprises texts ($D_T$), figures ($D_F$), and other metadata ($D_M$);
$\Phi_{SC}$ includes model parameters and domain priors;
and $f_{SC}$ means the specific comprehension algorithms, including a textual comprehension function $f_{\text{TSC}}$ that extracts and interprets textual content, and a table \& chart comprehension function $f_{\text{TCSC}}$ that analyzes tables and charts.

The goal of AI4SC is to maximize scientific understanding $\sigma$ with extracted knowledge $\hat{\mathcal{K}}$ from the original documents $D_{SC}$:
\begin{equation}
    \max\{\sigma\} = \max \{\mathbb{E}_{\hat{\mathcal{K}} \sim A_{SC}}[\text{Coherence}(\hat{\mathcal{K}},D_{SC}) + \text{Coverage}(\hat{\mathcal{K}},D_{SC})]\}\text{,}
\end{equation}
where Coherence measures logical consistency; Coverage quantifies concept completeness between them.

\vspace{-2mm}\subsubsection{AI for Academic Survey}\vspace{-1mm}

AI for Academic Survey (AI4AS) is designed to synthesize and structure multiple existing literature, providing a comprehensive overview of a research domain. This enhances the ability to identify trends, gaps, and key contributions in scientific fields. Formally, we define this module to generate a structured literature survey $S$ as a functional synthesis function:
\begin{equation}
    \hat{\mathcal{S}} = A_{AS}(\mathcal{S}) = f_{AS}(\mathcal{S}| R_{AS}, \Phi_{AS}) = f_{\text{Gen}} \circ f_{\text{Retrieval}}(\mathcal{S}| R_{AS}, \Phi_{AS})\text{,}
\end{equation}
where $A_{AS}$ is the survey AI model to generate the possible survey $\mathcal{S}$; $R_{AS}$ comprising survey domain requirements; $\Phi_{AS}$ includes model parameters and domain priors; $f_{AS}$ means the specific survey algorithms, which include a retrieval function $f_{\text{Retrieval}}$ that retrieves relevant literature based on the query, and a generative function $f_{\text{Gen}}$ that produces thematic clusters and summaries.

The objective of AI4AS is to maximize survey quality $\rho$ of the generated survey $\hat{\mathcal{S}}$ with respect to the requirement $R_{AS}$:
\begin{equation}
    \max\{\rho\} = \max \{\mathbb{E}_{\hat{\mathcal{S}} \sim A_{AS}}[\text{Relevance}(\hat{\mathcal{S}},R_{AS}) + \text{Coverage}(\hat{\mathcal{S}},R_{AS}) + \text{Clarity}(\hat{\mathcal{S}},R_{AS})]\},
\end{equation}
where Relevance measures the match between documents and the target topic; Coverage assesses the breadth and depth of the domain; Clarity reflects the coherence, abstraction quality, and utility of the synthesized representation based on the generated survey and requirements.

\vspace{-2mm}\subsubsection{AI for Scientific Discovery}\vspace{-1mm}

AI for Scientific Discovery (AI4SD) is focused on generating, and validating novel scientific hypotheses or ideas and conducting experiments or simulations. This module enhances the ability to explore uncharted scientific territories and accelerate innovation. Formally, we define this module to generate, validate, and implement scientific innovations $\hat{\mathcal{I}}$ as a discovery-oriented function:
\begin{equation}
    \hat{\mathcal{I}} = A_{SD}(\mathcal{I}) = f_{SD}(\mathcal{I}| K_{SD}, R_{SD}, \Phi_{SD}) = f_{\text{ED}} \circ f_{\text{TA}} \circ f_{\text{NSA}} \circ f_{\text{IM}}(\mathcal{I}| K_{SD}, R_{SD}, \Phi_{SD})\text{,}
\end{equation}
where $A_{SD}$ is a discovery-oriented AI to explore possible innovation $\mathcal{I}$; scientific knowledge $K_{SD} = \{ K_D, K_{AS} \}$ is the given domain knowledge $(K_D)$ and recent related-work summarized knowledge $(K_{AS})$ derived from upstream comprehension and survey stages; $R_{SD}$ means the research requirement; $\Phi_{SD}$ includes model parameters and domain priors; $f_{SD}$ means the specific discovery algorithms, which include a generative function $f_{\text{IM}}$ that mines candidate ideas, a novelty and significance assessment function $f_{\text{NSA}}$ that evaluates the quality and importance of each idea candidates, a theory analysis function $f_{\text{TA}}$ that checks theoretical soundness, and an experiment conduction function $f_{\text{ED}}$ that makes plans and executes experiments then finally complete the scientific discovery.

The goal of AI4SD is to maximize the total discovery quality $\delta$ of the generated innovations $\hat{\mathcal{I}}$:
\begin{equation}
    \max\{\delta\} = \max \{\mathbb{E}_{\hat{\mathcal{I}} \sim A_{SD}}[\text{Novelty}(\hat{\mathcal{I}}) + \text{Validity}(\hat{\mathcal{I}}) + \text{Significance}(\hat{\mathcal{I}})]\},
\end{equation}
where Novelty evaluates innovativeness; Validity assesses experimental and theoretical soundness; Significance reflects the follow-up impact of the study.

\vspace{-2mm}\subsubsection{AI for Academic Writing}\vspace{-1mm}

AI for Academic Writing (AI4AW) is a highlight section of AI4Research, assisting researchers in generating, revising, and formatting scientific manuscripts. This module enhances the quality and efficiency of academic writing, ensuring that manuscripts are well-structured and compliant with publication standards. Formally, we define this module to generate a publication-ready manuscript $\mathcal{M}$ as a collaborative writing function:
\begin{equation}
    \hat{\mathcal{M}} = A_{AW}(\mathcal{M}) = f_{AW}(\mathcal{M}| K_{AS}, \text{Info}_I, \Phi_{AW}) = f_{\text{DWP}} \circ f_{\text{DMW}} \circ f_{\text{AWC}}(\mathcal{M}| K_{AS}, \text{Info}_I, \Phi_{AW})\text{,}
\end{equation}
where $A_{AW}$ denotes a writing-oriented AI to generate the possible manuscript $\mathcal{M}$; $\text{Info}_I$ is all information in the scientific discovery stage, including ideas, experimental designs, and attachments such as codes and data; $\Phi_{AW}$ includes model parameters and domain priors; $f_{AW}$ means the specific writing algorithms, which include a during-manuscript-preparation function $f_{\text{DWP}}$ that prepares the manuscript structure, a during-manuscript-writing function $f_{\text{DMW}}$ that generates the manuscript content, and a after-manuscript-completion function $f_{\text{AWC}}$ that completes grammatical corrections, expressions and logical modifications.

The objective of AI4AW is to maximize writing quality and effectiveness $\omega$ of the manuscript $\hat{\mathcal{M}}$:
\begin{equation}
    \max\{\omega\} = \max \{\mathbb{E}_{\hat{\mathcal{M}} \sim A_{AW}}[\text{Consistency}(\hat{\mathcal{M}}) + \text{Readability}(\hat{\mathcal{M}}) + \text{Compliance}(\hat{\mathcal{M}})]\}\text{,}
\end{equation}
where Consistency reflects logical flow and internal coherence; Readability measures linguistic clarity and ease of understanding; Compliance assesses adherence to formatting and stylistic requirements.

\vspace{-2mm}\subsubsection{AI for Academic Peer Reviewing}\vspace{-1mm}

AI for Academic Peer Reviewing (AI4PR) is a critical component of AI4Research, automating and enhancing the peer review process. This module aims to provide structured, objective, and constructive reviews of scientific manuscripts, improving the quality and efficiency of the review cycle. Formally, we define this module to generate a structured review result $R$ as an evaluative reasoning function:
\begin{equation}
    \hat{\mathcal{R}} = A_{PR}(\mathcal{R}) = f_{PR}(\mathcal{R}| P, \Phi_{PR}) = f_{\text{PostP}} \circ f_{\text{InP}} \circ f_{\text{PreP}}(\mathcal{R}| \hat{\mathcal{M}}, \Phi_{PR}),
\end{equation}
where $A_{PR}$ denotes a review-oriented AI to generate the possible review $\mathcal{R}$; $\Phi_{PR}$ includes model parameters and domain priors; $f_{PR}$ means the specific review algorithms, which include a pre-review function $f_{\text{PreP}}$ that completes pre-review preparations, an in-review function $f_{\text{InP}}$ that generates or augments review reports, and a post-review function $f_{\text{PostP}}$ that completes post-review analysis of papers.

The goal of AI4PR is to maximize review quality $\theta$ of the review result $\hat{\mathcal{R}}$ based on the manuscript $\hat{\mathcal{M}}$:
\begin{equation}
    \max\{\theta\} = \max \{\mathbb{E}_{\hat{\mathcal{R}} \sim A_{PR}}[\text{Correctness}(\hat{\mathcal{R}},\hat{\mathcal{M}}) + \text{Helpfulness}(\hat{\mathcal{R}},\hat{\mathcal{M}}) + \text{Consistency}(\hat{\mathcal{R}},\hat{\mathcal{M}})]\}\text{,}
\end{equation}
where Correctness means the review can correctly reflect the pros and cons of research; Helpfulness measures the depth, constructiveness, and usefulness of feedback; Consistency quantifies the alignment of the review with established evaluation criteria and domain standards.

\begin{table*}[t]
    \centering
    \renewcommand{\arraystretch}{1.25}
    \resizebox{\textwidth}{!}{
        \begin{tabular}{l|l|l}
            \toprule
                         & \textbf{AI4Science}                                                                                                           & \textbf{AI4Research}                                                                                                   \\
            \midrule
            Scope        & \coloremojicode{1F52C} Scientific Discovery, \coloremojicode{1F4CA} Data Analysis.                                            & \coloremojicode{1F4DA} Broader Research workflows.                                                                     \\
            Goal         & \coloremojicode{1F9EC} Scientific Breakthroughs.                                                                              & \coloremojicode{1F4DD} Publications, \coloremojicode{1F527} Methods, \coloremojicode{1F4C8} Overall Productivity.      \\
            Applications & \coloremojicode{1F9EA} Material Discovery, \coloremojicode{1F48A} Drug Design, \coloremojicode{1F9EC} Genomics, \textit{etc.} & \coloremojicode{1F4D6} Comprehension, \coloremojicode{FE0F} Writing, \coloremojicode{1F9D0} Peer Review, \textit{etc.} \\
            Target Users & \coloremojicode{1F947} Research Experts.                                                                                      & Both \coloremojicode{1F947} Research Experts and \coloremojicode{1F948} New Scientists.                                \\
            \bottomrule
        \end{tabular}
    }
    \caption{Comparison and discussion between AI4Science and AI4Research, especially in terms of scope, goal, applications, and target users.}
    \label{tab:comparison}
\end{table*}

\vspace{-2mm}\subsection{Discussion About AI4Science and AI4Research}\vspace{-1mm}
Since the concepts of AI4Science and AI4Research share many similarities, as outlined in Table~\ref{tab:comparison}, it is important to distinguish the key differences between the two. \textbf{AI4Science (Artificial Intelligence for Science)} focuses on applying AI technologies to accelerate scientific discovery and data analysis across various fields, including material discovery, drug design, and genomic analysis. Its primary objective is to integrate AI into research workflows to support experts in achieving significant scientific advancements.
In contrast, \textbf{AI4Research (Artificial Intelligence for Research)} adopts a broader perspective, addressing publications, methodologies, and overall research productivity. It emphasizes AI’s role in enhancing research methods and supporting the academic environment for both established researchers and emerging scientists. Key applications in this domain include AI-driven tools for literature comprehension, academic writing assistance, and peer review processes.

The core distinction between these frameworks lies in their focus: AI4Science targets specific scientific problems and experimental protocols, while AI4Research addresses broader research methodologies and academic infrastructure. However, as LLMs develop more advanced reasoning and generative capabilities, a unified workflow is emerging that can address both specialized scientific challenges and general research processes. Consequently, AI4Science tools are increasingly integrated into AI4Research environments, often serving as callable components in LLM-based systems for scientific exploration. Subsequently, we will provide a detailed analysis of our taxonomy and the relevant literature.

%% file: figures/tree_taxonomy.tex
\tikzstyle{my-box}=[
rectangle,
draw=hidden-black,
rounded corners,
text opacity=1,
minimum height=1.5em,
minimum width=5em,
inner sep=2pt,
align=center,
fill opacity=.5,
]
\tikzstyle{leaf3}=[
my-box,
minimum height=1.5em,
fill=yellow!32,
text=black,
align=left,
font=\normalsize,
inner xsep=5pt,
inner ysep=4pt,
align=left,
text width=45em,
]
\tikzstyle{leaf6}=[
my-box,
minimum height=1.5em,
fill=purple!30,
text=black,
align=left,
font=\normalsize,
inner xsep=5pt,
inner ysep=4pt,
]
\tikzstyle{leaf4}=[
my-box,
minimum height=1.5em,
fill=hidden-blue!57,
text=black,
align=left,
font=\normalsize,
inner xsep=5pt,
inner ysep=4pt,
]
\tikzstyle{leaf2}=[
my-box,
minimum height=1.5em,
fill=hidden-green!20,
text=black,
align=left,
font=\normalsize,
inner xsep=5pt,
inner ysep=4pt,
]
\tikzstyle{leaf}=[
my-box,
minimum height=1.5em,
fill=hidden-red!20,
text=black,
align=left,
font=\normalsize,
inner xsep=5pt,
inner ysep=4pt,
]
\tikzstyle{leaf5}=[
my-box,
minimum height=1.5em,
fill=darkblue!15,
text=black,
align=left,
font=\normalsize,
inner xsep=5pt,
inner ysep=4pt,
]
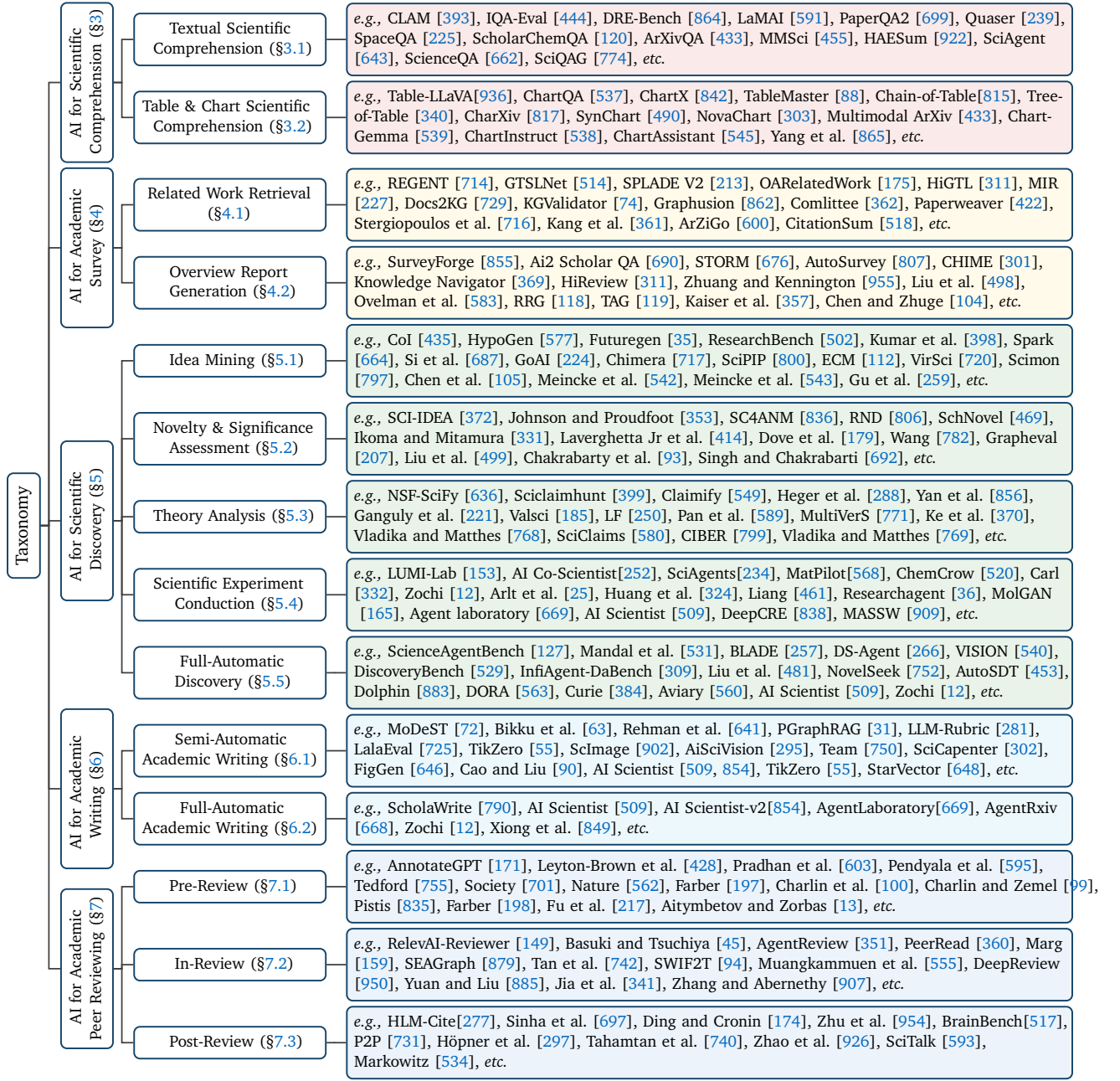
\begin{figure*}[!t]
        \vspace{-2mm}
        \centering
        \resizebox{0.96\textwidth}{!}{
                \begin{forest}
                        forked edges,
                        for tree={
                        grow=east,
                        reversed=true,
                        anchor=base west,
                        parent anchor=east,
                        child anchor=west,
                        base=left,
                        font=\large,
                        rectangle,
                        draw=hidden-black,
                        rounded corners,
                        align=left,
                        minimum width=4em,
                        edge+={darkgray, line width=1pt},
                        s sep=3pt,
                        inner xsep=2pt,
                        inner ysep=4pt,
                        line width=1.1pt,
                        ver/.style={rotate=90, child anchor=north, parent anchor=south, anchor=center},
                        },
                        where level=1{text width=9.5em,font=\normalsize,}{},
                        where level=2{text width=11.5em,font=\normalsize,}{},
                        where level=3{text width=12em,font=\normalsize,}{},
                        where level=4{text width=50em,font=\normalsize,}{},
                        [\ \ Taxonomy\ \ \ , ver
                        [\ \ \ \ AI for Scientific \\  \ Comprehension~(\S\ref{sec:ai4scientific-comprehension}),ver
                        [\ \ \ \ \ \ Textual Scientific \\ \ \ \ Comprehension~(\S\ref{sec:textual-scientific-comprehension})
                        [\eg ~CLAM~\citep{kuhn2022clam}{,} IQA-Eval~\citep{li2024iqa}{,} DRE-Bench~\citep{yang2025truly}{,} LaMAI~\citep{pang2024empowering}{,} PaperQA2~\citep{skarlinski2024language}{,} Quaser~\citep{ghoshal2022quaser}{,} \\SpaceQA~\citep{garcia2022spaceqa}{,} ScholarChemQA~\citep{chen2024scholarchemqa}{,} ArXivQA~\citep{li-etal-2024-multimodal-arxiv}{,} MMSci~\citep{li2024mmsci}{,}  HAESum~\citep{zhao2024hierarchical}{,} SciAgent\\
                        \citep{roberts2024scifibench}{,} ScienceQA~\citep{saikh2022scienceqa}{,}   SciQAG~\citep{wan2024sciqag}{,}  \textit{etc.}, leaf, text width=44em]
                        ]
                        [\ Table \& Chart Scientific \\ \ \ \ Comprehension~(\S\ref{sec:table-chart-scientific-comprehension})
                        [\eg ~Table-LLaVA\citep{zheng2024multimodal}{,} ChartQA~\citep{masry-etal-2022-chartqa}{,} ChartX~\citep{xia2024chartx}{,} TableMaster~\citep{cao2025tablemaster}{,} Chain-of-Table\citep{wang2024chain}{,} Tree-\\ of-Table~\citep{ji2024tree}{,} CharXiv~\citep{wang2024charxiv}{,}  SynChart~\citep{liu2024synchart}{,} NovaChart~\citep{hu2024novachart}{,} Multimodal ArXiv~\citep{li-etal-2024-multimodal-arxiv}{,} Chart-\\ Gemma~\citep{masry-etal-2025-chartgemma}{,}  ChartInstruct~\citep{masry-etal-2024-chartinstruct}{,} ChartAssistant~\citep{meng-etal-2024-chartassistant}{,} \citet{yang2025multimodal}{,} \textit{etc.}, leaf, text width=44em]
                        ]
                        ]
                        [\ \ \ \  AI for Academic \\ \ \ \  \ \ \ \ Survey~(\S\ref{sec:ai4academic-survey}), ver
                                [\ \ Related Work Retrieval\\ \ \ \ \ \ \ \ \ \ \ \ \ \ (\S\ref{sec:related-work-retrieval})
                                        [\eg ~REGENT~\citep{sridhar2024regent}{,} GTSLNet~\citep{luo2024clinical}{,} SPLADE V2~\citep{formal2021splade}{,} OARelatedWork~\citep{docekal2024oarelatedwork}{,}  HiGTL~\citep{hu2024taxonomy}{,} MIR\\\citep{garikaparthi2025mir}{,} Docs2KG~\citep{sun2025docs2kg}{,} KGValidator~\citep{boylan2024kgvalidator}{,} Graphusion~\citep{yang2024graphusion}{,} Comlittee~\citep{kang2023comlittee}{,} Paperweaver~\citep{lee2024paperweaver}{,}\\ \citet{stergiopoulos2024academic}{,}  \citet{kang2022you}{,} ArZiGo~\citep{pinedo2024arzigo}{,}    CitationSum~\citep{luo2023citationsum}{,}
                                                        \textit{etc.}, leaf3, text width=44em]SurveyForge
                                ]
                                [\ \ \ \ \ \ Overview Report \\ \ \ \ \ \ \ Generation~(\S\ref{sec:overview-report-generation})
                                        [\eg ~SurveyForge~\citep{yan2025surveyforge}{,} Ai2 Scholar QA~\citep{singh2025ai2}{,} STORM~\citep{shao2024assisting}{,} AutoSurvey~\citep{wang2024autosurvey}{,} CHIME~\citep{hsu2024chime}{,} \\ Knowledge Navigator~\citep{katz2024knowledge}{,} HiReview~\citep{hu2024taxonomy}{,} \citet{zhuang-kennington-2024-understanding}{,} \citet{liu2025select}{,} \\ \citet{ovelman2024use}{,}  RRG~\citep{chen2021capturing}{,} TAG~\citep{chen2022target}{,} \citet{kaiser2025376}{,} \citet{chen2019automatic}{,}
                                                        \textit{etc.}, leaf3, text width=44em]
                                ]
                        ]
                        [\ \ \ AI for Scientific \\ \ \ \ \ \ Discovery~(\S\ref{sec:ai4scientific-discovery}), ver
                                [\ \ \ \ \ \ Idea Mining (\S\ref{sec:idea-mining})
                                        [\eg ~CoI~\citep{li2024chain}{,}  HypoGen~\citep{o2025sparks}{,} Futuregen~\citep{azher2025futuregen}{,} ResearchBench~\citep{liu2025researchbench}{,} \citet{kumar2024can}{,} Spark\\ \citep{sanyal2025spark}{,} \citet{si2024can}{,} GoAI~\citep{gao2025graph}{,} Chimera~\citep{sternlicht2025chimera}{,} SciPIP~\citep{wang2024scipip}{,} ECM~\citep{chen2025ecm}{,}  VirSci~\citep{su2024two}{,} Scimon \\ \citep{wang2024scimon}{,} \citet{chen2025structuring}{,} \citet{meincke2024innovation}{,} 
                                        \citet{meincke2024prompting}{,}
                                        \citet{gu2024llms}{,}
                                                        \textit{etc.}, leaf2, text width=44em]
                                ]
                                [\ \ \ Novelty \& Significance \\ \ \ \ \ \ \ Assessment (\S\ref{sec:novelty-judgment})
                                        [\eg  ~SCI-IDEA~\citep{keya2025sci}{,} \citet{johnson2024greater}{,} SC4ANM~\citep{wu2025sc4anm}{,} RND~\citep{wang2025enabling}{,} SchNovel~\citep{lin-etal-2025-evaluating}{,} \\ \citet{ikoma2025can}{,}  \citet{laverghetta2025humans}{,} \citet{dove2025semi}{,} \citet{wang2024content}{,} Grapheval\\ \citep{feng2025grapheval}{,} \citet{liu2025harnessing}{,}  \citet{chakrabarty2024art}{,} \citet{singh2024supporting}{,} \textit{etc.}, leaf2, text width=44em]
                                ]
                                [\ \ \ Theory Analysis (\S\ref{sec:theory-analysis})
                                        [\eg ~NSF-SciFy~\citep{rao2025nsf}{,} Sciclaimhunt~\citep{kumar2025sciclaimhunt}{,} Claimify~\citep{metropolitansky2025towards}{,} \citet{heger2024natural}{,} \citet{yan2025position}{,} \\ \citet{ganguly2025grammars}{,} Valsci~\citep{edelman2025valsci}{,} LF~\citep{goodsell2024lf}{,}  \citet{pan2023investigating}{,} MultiVerS~\citep{wadden2021multivers}{,} \citet{ke2024can}{,} \\ \citet{vladika2024comparing}{,} SciClaims~\citep{ortega2025sciclaims}{,} CIBER~\citep{wang2025llmbased}{,} \citet{vladika2024improving}{,}
                                                        \textit{etc.}, leaf2, text width=44em]
                                ]
                                [\ \ \ Scientific Experiment \\ \ \ \ \ \ \ \ Conduction (\S\ref{sec:scientific-experiment-conduction})
                                        [\eg ~LUMI-Lab~\citep{cui2025lumi}{,} AI Co-Scientist\citep{gottweis2025towards}{,} SciAgents\citep{ghafarollahi2024sciagents}{,} MatPilot\citep{ni2024matpilot}{,} ChemCrow~\citep{m2024augmenting}{,} Carl\\ \citep{carl2025}{,} Zochi~\citep{zochi2025}{,} \citet{arlt2024meta}{,} \citet{huang2024ai}{,} \citet{liang2024application}{,} Researchagent~\citep{baek2024researchagent}{,} MolGAN\\~\citep{de2018molgan}{,} Agent  laboratory~\citep{schmidgall2025agent}{,} AI Scientist~\citep{lu2024ai}{,}  DeepCRE~\citep{wu2024deepcre}{,} MASSW~\citep{zhang2024massw}{,}
                                                        \textit{etc.}, leaf2, text width=44em]
                                ]
                                [ \ \ \ \ \ \ \ \ Full-Automatic \\ \ \ \ \ \ \ \ Discovery (\S\ref{sec:full-automatic-discovery})
                                        [\eg ~ScienceAgentBench~\citep{chen2024scienceagentbench}{,} \citet{mandal2024autonomous}{,} BLADE~\citep{gu2024blade}{,} DS-Agent~\citep{guo2024ds}{,} VISION~\citep{mathur2025vision}{,}  \\ DiscoveryBench~\citep{majumder2024discoverybench}{,} InfiAgent-DaBench~\citep{hu2024infiagent}{,} \citet{liu2025vision}{,}  NovelSeek~\citep{team2025novelseek}{,} AutoSDT~\citep{li2025autosdt}{,} \\ Dolphin~\citep{yuan2025dolphin}{,} DORA~\citep{naumov2025dora}{,} Curie~\citep{kon2025curie}{,} Aviary~\citep{narayanan2024aviary}{,} AI Scientist~\citep{lu2024ai}{,} Zochi~\citep{zochi2025}{,}
                                                        \textit{etc.}, leaf2, text width=44em]
                                ]
                        ]
                        [\ \ \  AI for Academic  \\ \ \ \ \ \ Writing~(\S\ref{sec:ai4writing}), ver
                                [\ \ \ \ \ \ \ Semi-Automatic \\ \ Academic Writing (\S\ref{sec:semi-automatic-academic-writing})
                                        [\eg ~MoDeST~\citep{bolucu2025modest}{,} \citet{bikku2025generating}{,} \citet{rehman2025can}{,} PGraphRAG~\citep{au2025personalized}{,} LLM-Rubric~\citep{hashemi-etal-2024-llm}{,} \\ LalaEval~\citep{sun2024lalaeval}{,} TikZero~\citep{belouadi2025tikzero}{,} ScImage~\citep{zhang2024scimage}{,} AiSciVision~\citep{hogan2024aiscivision}{,} \citet{illustrae2025how}{,} SciCapenter~\citep{hsu2024scicapenter}{,} \\ FigGen~\citep{rodriguez2023figgen}{,} \citet{cao2024figuring}{,}  AI Scientist~\citep{lu2024ai,yamada2025ai}{,} TikZero~\citep{belouadi2025tikzero}{,} StarVector~\citep{rodriguez2025starvector}{,} \textit{etc.}, leaf4, text width=44em]
                                ]
                                [\ \ \ \ \ \ \ \ Full-Automatic \\ \ Academic Writing (\S\ref{sec:full-automatic-academic-writing})
                                        [\eg ~ScholaWrite~\citep{wang2025scholawrite}{,} AI Scientist~\citep{lu2024ai}{,}  AI Scientist-v2\citep{yamada2025ai}{,}  AgentLaboratory\citep{schmidgall2025agent}{,} AgentRxiv\\\citep{schmidgall2025agentrxiv}{,} Zochi~\citep{zochi2025}{,} \citet{xiong2025beyond}{,}
                                                        \textit{etc.}, leaf4, text width=44em]
                                ]
                        ]
                        [\ \ \ \  AI for Academic   \\ \  Peer Reviewing~(\S\ref{sec:ai4reviewing}), ver
                                [\ \ \ \ \ \ Pre-Review (\S\ref{sec:pre-review})
                                        [\eg ~AnnotateGPT~\citep{diaz2024streamlining}{,}  \citet{leyton2024matching}{,} \citet{pradhan2020automated}{,} \citet{pendyala2025automated}{,} \\ \citet{tedford2015helping}{,} \citet{ieeecomputersociety}{,} \citet{nature}{,} \citet{farber2024enhancing}{,} \citet{charlin2011framework}{,} \citet{charlin2013toronto}{,} \\ Pistis~\citep{wu2018pistis}{,} \citet{farber2025enhancing}{,} \citet{fu2025peer}{,}  \citet{aitymbetov2025autonomous}{,} \textit{etc.}, leaf5, text width=44em]
                                ]
                                [\ \ \ \ \ \ \ In-Review (\S\ref{sec:in-review})
                                        [\eg ~RelevAI-Reviewer~\citep{couto2024relevai}{,}  \citet{basuki2022quality}{,} AgentReview~\citep{jin2024agentreview}{,} PeerRead~\citep{kang2018dataset}{,} Marg\\ \citep{d2024marg}{,} SEAGraph~\citep{yu2024seagraph}{,}  \citet{tan2024peer}{,} SWIF2T~\citep{chamoun-etal-2024-automated}{,} \citet{muangkammuen2022exploiting}{,}  DeepReview\\\citep{zhu2025deepreview}{,} \citet{yuan2022kid}{,}  \citet{jia2021all}{,}   \citet{zhang2025reviewing}{,} \textit{etc.}, leaf5, text width=44em]
                                ]
                                [\ \ \ \ \ \ Post-Review (\S\ref{sec:post-review})
                                        [\eg ~HLM-Cite\citep{hao2024hlm}{,} \citet{sinha2015overview}{,} \citet{ding2011popular}{,} \citet{zhu2015measuring}{,} BrainBench\citep{luo2025large}{,} \\P2P~\citep{sun2025p2p}{,}  \citet{hopner2025automatic}{,} \citet{tahamtan2016factors}{,}  \citet{zhao2025words}{,} SciTalk~\citep{park2025stealing}{,} \\ \citet{markowitz2024complexity}{,}
                                                        \textit{etc.}, leaf5, text width=44em]
                                ]
                        ]
                        ]
                \end{forest}
        }
        \caption{The taxonomy of AI for research (AI4Research) is categorized into five key areas. Each area is subdivided into specific tasks, underscoring the varied roles of AI in the entire research process.}
        \label{fig:ai4research-taxonomy}

\end{figure*}

%% file: sections/comprehension.tex
\vspace{-2mm}\section{AI for Scientific Comprehension}
\begin{figure*}[t]
	\centering
	\includegraphics[width=0.99\textwidth]{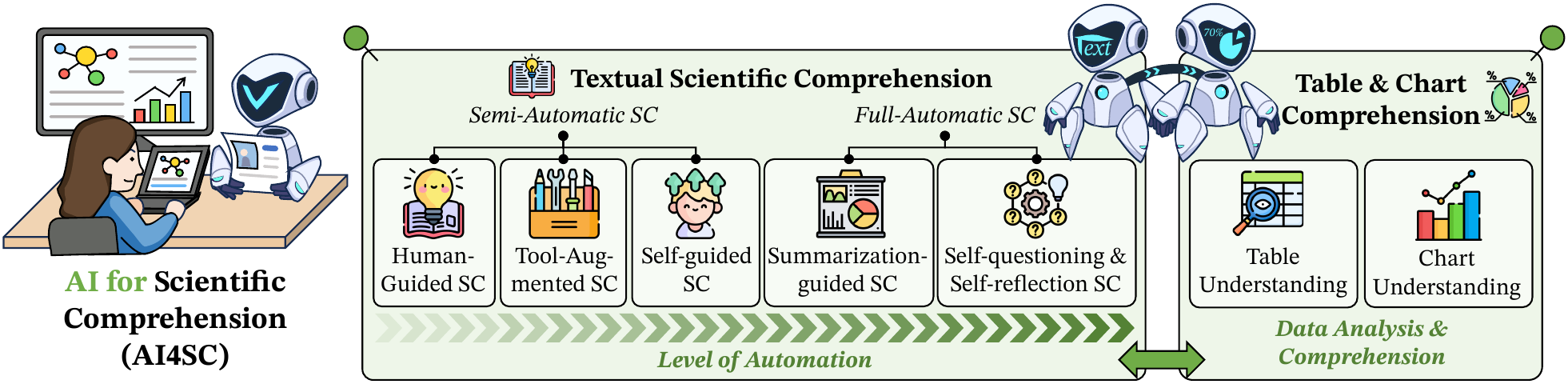}
	\caption{The primary paradigms of AI for Scientific Comprehension. These include: (1) Textual Scientific Comprehension, which is further categorized into Semi-Automatic and Fully-Automatic Scientific Comprehension; and (2) Table \& Chart Scientific Comprehension, encompassing Table and Chart Understanding.}
	\label{fig:scientific-comprehension}
\end{figure*}

\vspace{-1mm}
\label{sec:ai4scientific-comprehension}
Scientific comprehension plays a pivotal role in advancing AI4Research, encompassing the ability to extract, understand, and synthesize information from the scientific literature. This capability not only accelerates human understanding and knowledge acquisition but also enhances the efficiency of automatic analysis, enabling more effective research processing. As shown in Figure~\ref{fig:scientific-comprehension}, it contains two main categories: Textual Scientific Comprehension ($\S$~\ref{sec:textual-scientific-comprehension}) and Table \& Chart Scientific Comprehension ($\S$~\ref{sec:table-chart-scientific-comprehension}).

\vspace{-2mm}\subsection{Textual Scientific Comprehension}\vspace{-1mm}
\label{sec:textual-scientific-comprehension}
Textual Scientific Comprehension refers to the ability to understand, interpret, and critically evaluate scientific texts. It involves identifying key concepts, grasping complex terminology, and synthesizing information to form a cohesive understanding of scientific principles and findings~\citep{qu2020open, bolotova2022non,mansour-etal-2025-well}. As depicted in Figure~\ref{fig:scientific-comprehension} (middle), we categorize the corresponding comprehension technologies into two types based on automation level: Semi-Automatic and Fully-Automatic Scientific Comprehension.

\vspace{-2mm}\subsubsection{Semi-Automatic Scientific Comprehension}\vspace{-1mm}
Semi-automatic scientific comprehension denotes systems in which, give a manually created question, the AI produces comprehensive question-related comprehension of long-context scientific content. Such systems support both researchers and AI models in deepening their grasp of complex scientific concepts \citep{chen2024scholarchemqa, hilgert-etal-2024-evaluating,wen2025plain}. Specifically, these systems comprise three main categories:\vspace{-15pt}

\paragraph{Human-Guided Scientific Comprehension} is an interactive approach where researchers and language models engage in iterative dialogues to produce a deepened understanding of questions on complex scientific literature step-by-step~\citep{kuhn2022clam,zhang2023clarify,li2024iqa,yang2025truly,qin2020dynamic}. LaMAI~\citep{pang2024empowering} equips language models with ``active inquiry'' capabilities: before providing a definitive answer, the model asks clarifying questions to resolve ambiguities in user queries, reducing misinterpretations and enhancing relevance~\citep{yamada2025ai}. These platforms illustrate that embedding structured human feedback loops within LLM-based tools improves output reliability and enriches the scientific discovery process by uncovering latent questions and assumptions. However, the approach requires significant human-AI interaction, which can increase costs.\vspace{-15pt}

\paragraph{Tool-Augmented Scientific Comprehension} refers to cases where a researcher’s query surpasses a language model’s knowledge base or its context-window limit \citep{wright2021citeworth}. The model then invokes several external tools to ensure accurate output:
\textit{\textbf{(1) Knowledge Retrieval Tool}} uses retrieval-augmented generation to inject knowledge beyond the model’s training \citep{kim2025medbiolm,saikh2022scienceqa}. Early systems like document-centric agents \citep{lala2023paperqa} extract key findings, note limitations, and propose future directions. Graphusion \citep{yang2024graphusion} advances this by building scientific knowledge graphs and extracting entity triples, resolving conflicts across disciplines without manual effort. SiGIR \citep{chu2025self} uses self-critique feedback to guide the iterative reasoning process during knowledge-intensive multi-hop reasoning tasks.
\textit{\textbf{(2) Fact Checking Tool}} mitigates hallucinations and factual errors by applying verification modules to reduce the AI's hallucinations~\citep{juneja2022human,hartley2024efficacy,gosmar2025hallucination,zhang2025cchall}. PaperQA2 \citep{skarlinski2024language} integrates rigorous factuality checks and matches or exceeds expert accuracy on literature-review tasks, all without unrestricted Internet access or human oversight.
\textit{\textbf{(3) Reasoning-Augmentation Tool}} addresses limited logical reasoning and computation in standalone models to deepen the AI's theory-level comprehension~\citep{chen2025towards}. For example, SciAgent \citep{ma-etal-2024-sciagent} dynamically selects calculators and formula evaluators to deliver precise, domain-specific reasoning.
Collectively, these advances show how coupling AI with specialized tools transforms scientific workflows from passive consumption into an interactive, tool-powered process that accelerates discovery while preserving rigor.\vspace{-15pt}

\paragraph{Self-guided Scientific Comprehension} refers to a model's capacity to respond to a single-turn query regarding a scientific publication with a comprehensive, context-sensitive answer~\citep{beltagy-etal-2019-scibert,rostam2024fine,bolton2024biomedlm}. Earlier, \citet{clark2019boolq} demonstrate that even seemingly factual questions about academic papers require deep contextual understanding and meticulous attention to document-specific details~\citep{ghoshal2022quaser}. To address these challenges, subsequent studies focus more on enhanced semi-automatic scientific comprehension in long-context papers~\citep{liu2025comprehensive,tang2024citeeval}, particularly in specialized fields such as aerospace science~\citep{garcia2022spaceqa}, chemistry~\citep{peretz2023if,chen2024scholarchemqa}, and clinical medicine~\citep{raza2022coquad,singhal2025toward}. It illustrates that enhancing models to align with the linguistic and conceptual conventions of each discipline, particularly those with improved long-context capabilities, leads to significant advancements~\citep{chen2024essential,liu2025survey}. Furthermore, recognizing the inherently multimodal nature of scientific papers, several studies have begun to integrate textual analysis with figures and charts~\citep{li-etal-2024-multimodal-arxiv,li2024mmsci,roberts2024scifibench}, thereby advancing towards a more holistic, paper-wide comprehension of scientific content.

\vspace{-2mm}\subsubsection{Full-Automatic Scientific Comprehension}\vspace{-1mm}
AI for full-automatic scientific comprehension refers to the ability of an AI system to read and understand scientific knowledge independently without human questions or other intervention. The goal of such systems is to fully automate the processing of scientific literature, the formulation and answering of complex questions, and even, to some extent, scientific discovery or idea mining.\vspace{-15pt}

\paragraph{Summarization-guided Automatic Scientific Comprehension} refers to the capability of LLMs to autonomously generate summaries of scientific articles and, based on these summaries, construct a comprehensive narrative of the research~\citep{kerwer2021straight,fonseca-cohen-2024-large-language}. This process enhances the model's holistic understanding of lengthy scientific texts and mitigates comprehension biases that arise from processing extensive documents in a purely token-by-token manner \citep{zhao2024hierarchical}.
Furthermore, \citet{ifargan2025autonomous} suggest that LLMs can further enhance their overall comprehension of lengthy scientific documents through the generation of autonomous summaries. Their approach utilizes a system of multiple agents, such as a Summary Agent and a Proofreading Agent, working collaboratively to extract and refine experimental results and research methodologies without human intervention. This ultimately produces a refined abstract suitable for peer review.\vspace{-15pt}

\paragraph{Self-Questioning \& Self-Reflection Automatic Scientific Comprehension}
involve an AI generating and answering its own questions or reflection to deepen its understanding of scientific content~\citep{huang2022large,miao2023selfcheck,yu2025frame}. Earlier, SciInstruct \citep{zhang2024sciglm} proposes a self-reflective annotation framework, where a model generates step-by-step reasoning for unlabeled scientific questions and then refines its output through self-critique, producing high-quality annotations. Building on this, several studies~\citep{luo2024generating} have focused on prompting models to autonomously create question sequences that enhance their comprehension of scientific texts.
One notable example is SciQAG~\citep{wan2024sciqag}, which proposes a pipeline where a ``question generator'' and an ``answer evaluator'' collaborate to extract diverse, research-level comprehension from scientific papers.

More recently, LLMs have been directed to self-improve by posing clarifying questions and decomposing complex problems in a Socratic style, strengthening reasoning and conceptual understanding~\citep{qu2024recursive,song2024mind,wang2023enabling}. The Introspective Growth \citep{wu2025introspective} further refines this approach, prompting smaller models to generate fundamental, open-ended questions that guide larger models toward better task comprehension. This process integrates external text retrieval to refine the understanding of technical semantics.

\vspace{-2mm}\subsection{Table \& Chart Scientific Comprehension}\vspace{-1mm}
\label{sec:table-chart-scientific-comprehension}
Beyond pure textual content, LLMs are employing various techniques to more efficiently interpret and leverage information from tables and figures, thereby achieving a deeper and more comprehensive understanding of scientific literature~\citep{circi2024well,newman-etal-2024-arxivdigestables,guo2025sciverse}.

\vspace{-2mm}\subsubsection{Table Understanding}\vspace{-1mm}
\label{sec:table-understanding}
Table understanding involves methods that enable LLMs to extract, interpret, and infer data from tables in scientific literature~\citep{si2024table,wu2025tablebench,ashury2025mighty}.
\textit{\textbf{(1) Data Augmentation:}} The most direct way is to add higher quality table understanding data. For instance, \citet{zheng2024multimodal} introduce the MMTab dataset for large-scale multimodal table understanding in a generative format and propose Table-LLaVA, which reasons directly on table images through instruction tuning, demonstrating the great advantage of visually grounded table representations.
\textit{\textbf{(2) Reasoning Paradigm Augmentation:}} Subsequent work explores suitable reasoning paradigms~\citep{cao2025tablemaster,zhang2025survey,wang2022survey,wang2024improving_a}. \citet{wang2024chain} propose Chain-of-Table, which incrementally constructs and updates tables within an LLM's reasoning chain to improve comprehension of complex tables. \citet{ji2024tree} introduce Tree-of-Table, hierarchically condensing and decomposing large tables into a tree structure to facilitate LLM reasoning.  \citet{cao2025tablemaster} present TableMaster, a framework that enhances LLM table understanding by extracting and verbalizing relevant table content with enriched semantic context and adaptively switching between textual and symbolic reasoning.

\vspace{-2mm}\subsubsection{Chart Understanding}\vspace{-1mm}
Chart Understanding involves techniques enabling multimodal large language models to directly process and interpret chart images in scientific papers, supporting tasks such as question answering and summarization based on chart content~\citep{pramanickspiqa,liang-etal-2024-scemqa,meng2024chartassisstant,huang2025chartsketcher}.
Furthermore, several studies focus on assembling and synthesizing large, diverse chart datasets to improve chart understanding~\citep{liu2024synchart,hu2024novachart,li-etal-2024-multimodal-arxiv}. \citet{masry-etal-2024-chartinstruct} and \citet{meng-etal-2024-chartassistant} present vision-language instruction datasets for charts, and train both end-to-end and pipeline models that achieve state-of-the-art results on scientific chart understanding tasks~\citep{masry-etal-2025-chartgemma}.
Further, \citet{yang2025multimodal} propose the Formalized Description for Visualization (FDV), a structured textual representation of charts that enables large language models to learn for diverse and deeper comprehension.

%% file: sections/survey.tex
\vspace{-2mm}\section{AI for Academic Survey}
\begin{figure*}[t]
	\centering
	\includegraphics[width=0.99\textwidth]{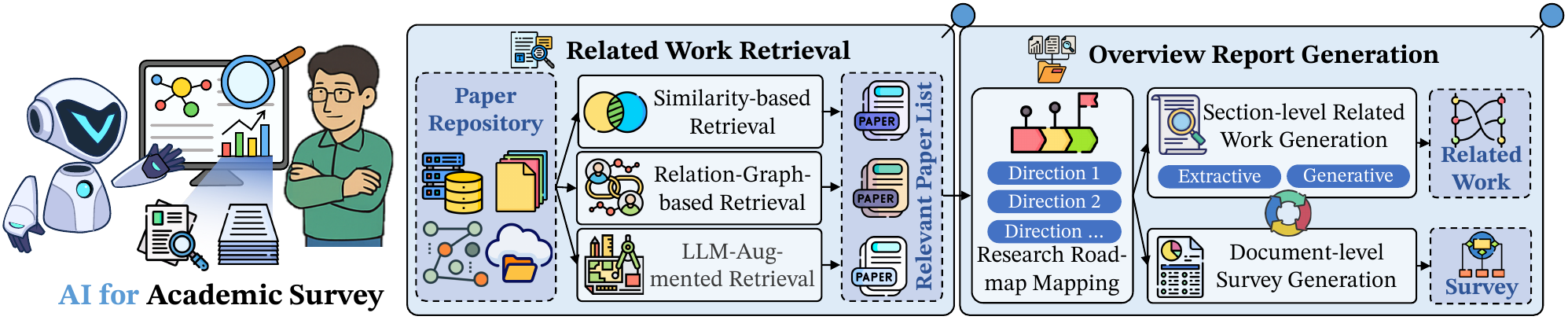}
	\caption{The two primary stages in AI-driven academic surveys: Related Work Retrieval and Overview Report Generation. Related Work Retrieval is further subdivided into Semantic-Guided Retrieval, Graph-Guided Retrieval, and LLM-Augmented Retrieval. Overview Report Generation encompasses Research Roadmap Mapping, Section-level Related Work Generation, and Document-level Survey Generation.
	}
	\label{fig:academic-survey}
\end{figure*}

\vspace{-1mm}
\label{sec:ai4academic-survey}
It is widely acknowledged that a thorough and well-conducted pre-writing survey and research phase forms the cornerstone of a successful academic article \citep{rohman1965pre}. Inspired by this, AI for Academic Survey is proposed to systematically review and summarize scientific literature through the application of artificial intelligence techniques. This process plays a crucial role in ensuring that researchers and automated systems remain current with the latest advancements in their field and can efficiently identify relevant studies to inform their own work. As shown in Figure~\ref{fig:academic-survey}, it contains two main stages: Related Work Retrieval ($\S$~\ref{sec:related-work-retrieval}) and Overview Report Generation ($\S$~\ref{sec:overview-report-generation}).

\vspace{-2mm}\subsection{Related Work Retrieval}\vspace{-1mm}
\label{sec:related-work-retrieval}
Related Work Retrieval entails AI proactively identifying foundational and novel research papers aligned with their evolving scientific objectives~\citep{beel2016paper,li2019review,shahid2020insights,fan2024survey}. Existing research divides into three paradigms:\vspace{-15pt}

\paragraph{Semantic-Guided Retrieval} involves identifying relevant literature by matching the semantic representation extracted from a user query to the terms present in documents based on similarity~\citep{bai2019scientific, kreutz2022scientific, li-ouyang-2024-related}. In the biomedical domain, GTSLNet \citep{luo2024clinical} enhances semantic-guided retrieval by utilizing a group-based keyword similarity learning network, which automatically selects clinically analogous studies. Similarly, SPLADE V2 \citep{formal2021splade} advances neural retrieval by integrating sparse lexical signals with dense expansion models, achieving significant results. Moreover, \citet{garikaparthi2025mir} concentrate on inspiration retrieval to provide enhanced support for idea mining.\vspace{-15pt}

\paragraph{Graph-Guided Retrieval} models scholarly entities (e.g., papers, authors, citations) as a graph~\citep{boylan2024kgvalidator,sun2025docs2kg,yang2024graphusion,stergiopoulos2024academic}. Based on the types and granularity of nodes, this search method can be categorized into three types:
\textit{\textbf{(1) Author Relationship Graph}} captures the connections between researchers, enabling searches based on author relationships. For example, author-relationship graphs can effectively model collaboration networks and researcher influence~\citep{kang2023comlittee,pinedo2024arzigo}. Building on this concept, \citet{kang2022you} developed a ``user-recommended paper'' knowledge graph that traces users' interactions with literature, enhancing recommendation transparency and trust.
\textit{\textbf{(2) Paper Relationship Graph}} is always constructed using citation relationships between papers to construct broader paper relationships~\citep{hu2024taxonomy}. CitationSum~\citep{luo2023citationsum} creates a citation graph linking target papers to their references with weighted relevance scores, then uses graph contrastive learning to produce abstractive summaries.
\textit{\textbf{(3) Entity Relationship Graph:}} can be constructed by modeling relationships between logical entities within papers, enabling more precise retrieval. For example, \citet{li2024explaining} propose a graph-based model that automatically identifies inter-paper entity relationships, such as contrast and support, guiding the construction of a structured Related Work section. Cross-domain graph methods are increasingly used in interdisciplinary research. Further, to address complex questions such as ``Which synthesis pathways enable material X to achieve optimal conductivity?'', \citet{ye2024construction} systematically extract entities and their relationships from the materials science literature, thus enhancing the depth of exploration.\vspace{-15pt}

\paragraph{LLM-Augmented Retrieval} involves leveraging the capabilities of LLMs to improve search effectiveness and result quality by integrating them with academic retrieval systems. 
\textit{\textbf{(1) Single-Agent Retrieval:}} The most straightforward approach is to employ a single AI model as a standalone agent to accomplish the retrieval task. For instance, \citet{lee2024paperweaver} introduce the PaperWeaver framework, which places an LLM-based agent atop a graph to enable deeper reasoning, thereby enhancing interpretability in classification and recommendation tasks.
\textit{\textbf{(2) Multi-Agent Retrieval:}} Beyond single-agent systems, several studies employ multiple specialized agents to simplify retrieval and increase accuracy~\citep{seabra2024dynamic,liu2025select,singh2025agentic}. LitLLMs~\citep{agarwal2025litllms} splits the automatic literature review into two subtasks: retrieval and generation. It proposes a two-phase LLM pipeline that extracts keywords from abstracts and reranks results to improve recall. \citet{liu2025select} propose a multi-agent framework for full-text related-work generation, which includes a selector to choose sections to read, a reader to update shared memory, and a writer to generate the related work section. The framework uses graph-aware strategies to optimize the reading order of references.
\textit{\textbf{(3) Deep Research:}} Recent research has advanced this paradigm towards more autonomous ``Deep Research'' \citep{openai2025deepresearch,openai2025gpt4osearch}, where AI agents perform the end-to-end research process, from exploration and synthesis to generating citation-rich reports~\citep{yang2025multimodal,du2025deepresearch,zhou2025academicbrowse}. This progress is enabled by novel agent architectures that emulate human research heuristics~\citep{wu2025webdancer}. For instance, the PaSa agent \cite{he2025pasa} discovers literature by actively traversing citation networks. Concurrently, the retrieval strategies themselves have become more intelligent; the ExSearch framework \cite{shi2025iterative} allows an agent to continuously optimize its search strategies through a self-incentivization loop, while CuriousLLM \cite{yang2025curiousllm} employs a "curiosity-driven" mechanism where the agent actively generates questions to guide its retrieval process of knowledge graphs.

\vspace{-2mm}\subsection{Overview Report Generation}\vspace{-1mm}
\label{sec:overview-report-generation}
Based on retrieved data, automated generation of structured, coherent overview reports has become essential in academic writing and AI4Research process~\citep{hoang2010towards}. According to the writing sequence, we need to first complete the research roadmap mapping, followed by the generation of section-level related work, and finally produce the complete document-level survey.

\vspace{-2mm}\subsubsection{Research Roadmap Mapping}\vspace{-1mm}
Research Roadmap Mapping refers to the process of cleaning, integrating, and depicting the developmental trajectories of a research topic by synthesizing insights from a broad corpus of literature~\citep{agarwal2024llms,yan2025surveyforge,singh2025ai2,wang2024autosurvey}.  This methodology is crucial for enhancing the rigor and completeness of literature surveys and meta-analyses, as it enables researchers to discern emerging trends, unresolved gaps, and potential future directions more systematically~\citep{chen2025towards,bolanos2024artificial,yan2025surveyforge}.
Specifically, \citet{zhu2023hierarchical} demonstrate that organizing a survey into a hierarchical structure significantly improves coherence~\citep{shao2024assisting}.

Recently, more interactive hierarchical frameworks have also emerged. CHIME~\citep{hsu2024chime}, for instance, refines LLM-generated structures through iterative human-AI collaboration. Similarly, \citet{katz2024knowledge} expand this to a two-tiered hierarchy, effectively organizing extensive surveys. Further, HiReview~\citep{hu2024taxonomy} illustrates the benefits of multilayered tree structures for systematic knowledge organization. Moreover, \citet{zhuang-kennington-2024-understanding} propose a graph-based taxonomy that categorizes LLM survey papers into defined classes, outperforming fine-tuned LLMs and providing a scalable framework for organizing survey literature.

\vspace{-2mm}\subsubsection{Section-level Related Work Generation}\vspace{-1mm}
Section-level Related Work Generation has been regarded as a prominent research~\citep{chen2021capturing,chen2022target,ovelman2024use,liu2025select,kaiser2025376,li-ouyang-2024-related}. Such section-level approaches are well-aligned with the actual structure of  scientific papers and can effectively fulfill the requirements of the related-work-section~\citep{hoang2010towards}.\vspace{-15pt}

\paragraph{Extractive Related Work.} Early automated methods for generating the ``Related Work'' section involve extracting key sentences from multiple papers, which were then rewritten and combined into a coherent narrative~\citep{hoang2010towards}. A subsequent approach refine this by selecting papers that cited similar references to the target work and extracting relevant sentences from them~\citep{chen2019automatic, wang2019toc}. Further research has focused on improving the organization and integration of these extracted sentences. Some methods explore optimal reference structures and sentence orderings~\citep{hu2014automatic, deng2021automatic}. For instance, ReWoS~\citep{hoang2010towards} and RWS-Cit~\citep{chen2019automatic} build topic trees to sequence sentences, while \citet{wang2019neural} employ a ranking mechanism based on predicted salience probabilities to enhance the quality of the extractive related work.\vspace{-15pt}

\begin{table*}
    \centering
    \resizebox{\textwidth}{!}{
        \begin{tabular}{l|l|cc|c|cccc}
            \toprule
            \multirow{2}{*}{\textbf{Methods}} & \multirow{2}{*}{\textbf{Model}} & \multicolumn{2}{c|}{\textbf{Reference Quality}} & \multirow{2}{*}{\textbf{Outline Quality}} & \multicolumn{4}{c}{\textbf{Content Quality}}           \\
 &    & \textbf{Input Cov.} & \textbf{Reference Cov.} & & \textbf{Structure} & \textbf{Relevance} & \textbf{Coverage} & \textbf{Avg} \\
            \midrule
            Human-Written  & -  & -  & 0.6294 & 87.62   & - & - & -  & -            \\
            \midrule
            AutoSurvey~\citep{wang2024autosurvey}  & Claude-3-Haiku~\citep{anthropic2024claude3} & 0.1153  & 0.2341 & 82.18   & 72.83 & 76.44 & 72.35   & 73.87        \\

            SurveyForge~\citep{yan2025surveyforge} & Claude-3-Haiku~\citep{anthropic2024claude3} & 0.2231  & 0.3960 & 86.85   & 73.82 & 79.62 & 75.59   & 76.34        \\
            \midrule
            AutoSurvey~\citep{wang2024autosurvey}  & GPT-4o-mini~\citep{robertson2023gpt4}  & 0.0665  & 0.2035 & 83.10   & 74.66 & 74.16 & 76.33   & 75.05        \\
            SurveyForge~\citep{yan2025surveyforge} & GPT-4o-mini~\citep{robertson2023gpt4}  & 0.2018  & 0.4236 & 86.62   & 77.10 & 76.94 & 77.15   & 77.06        \\
            \midrule
            SurveyForge~\citep{yan2025surveyforge} & DeepSeek-v3~\citep{liu2024deepseek}  & 0.2554  & 0.4553 & 87.42   & 79.20 & 80.17 & 81.07   & 80.15        \\
            \bottomrule
        \end{tabular}
    }
    \caption{A comparison of document-level survey generation capabilities on SurveyBench~\citep{yan2025surveyforge} using three key Survey Assessment Metrics: Reference quality, Outline quality, and Content quality. ``Input Cov.'' indicates the overlap between retrieved papers and benchmark references, while ``Reference Cov.'' evaluates the alignment of cited references with the benchmark. Data are sourced from~\citet{yan2025surveyforge}.}
    \label{tab:ai4survey-results}
\end{table*}

\paragraph{Generative Related Work.}
Recent studies have focused on methods to structure citations and generate cohesive connecting text for entire related work sections \citep{wang2022multi,liu2023causal,li2024explaining}. These approaches generally fall into three categories:
\textit{\textbf{(1) Human-Guided Generation:}} This approach incorporates human input, such as keywords, short abstracts, or paper groupings, to guide the generation process and maintain focus~\citep{li2024related,martin2024shallow}. For instance, \citet{gu2022controllable} and \citet{li2023cited} integrate user-provided or self-extracted keywords for better related-work generation.
\textit{\textbf{(2) Graph-Guided Generation:}} These methods utilize citation relationships through bibliographic graphs~\citep{ge2021baco,chen2021capturing,yu2024reinforced}. Specifically, \citet{wang2019neural} enhance related work generation by performing random walks on heterogeneous citation graphs. Similarly, \citet{chen2022target} use a graph to link references to the paper.
\textit{\textbf{(3) Model-Guided Generation:}} In this approach, models complete the task autonomously, without additional human input~\citep{shi2023towards,wang2024disentangling,pratapa2025estimating}. \citet{guo2021automated} and \citet{nishimura2024toward} treat related work generation as a summarization task with structured paragraphs and novelty statements. Additionally, \citet{pu2024rst} integrate Rhetorical Structure Theory into LoRA-based fine-tuning to identify discourse relations, and \citet{achkar2025ask} propose a customizable multi-stage pipeline (retrieval, citation extraction, context aggregation, polishing), further enhancing the related work generation processes.

\vspace{-2mm}\subsubsection{Document-level Survey Generation}\vspace{-1mm}
Document-level survey generation seeks to automate the creation of systematic literature reviews by leveraging existing research and established frameworks~\citep{webster2002analyzing,zhu2023hierarchical,bolanos2024artificial,galli2024intelligent,wu2025lag}. The detailed comparison results can be found in Table~\ref{tab:ai4survey-results}. For example, AutoSurvey \citep{wang2024autosurvey} employs cue-word guidance to direct LLMs through a staged generation process. Similarly, LitLLM \citep{agarwal2024litllm} enhances content structuring by implementing a plan-based search mechanism. SurveyX \citep{liang2025surveyx} strengthens logical coherence through the combination of online reference retrieval and AttributeTree preprocessing. Building on these approaches, SurveyForge \citep{yan2025surveyforge} retrieves high-quality papers via scholar-navigating intelligences and generates survey chapters from a predefined outline, followed by iterative refinement to maintain document-level quality~\citep{skarlinski2024language}. In contrast, STORM \citep{shao2024assisting} uses multi-agent dialogue to further enhance generation performance. Beyond training-free agent management, Bio-SIEVE \citep{robinson2023bio} and \citet{susnjak2025automating} fine-tune LLMs specifically for survey generation~\citep{lai2024instruct}, while OpenScholar \citep{asai2024openscholar} offers a pipeline for training models for survey writing without relying on specialized generation architectures.

%% file: sections/discovery.tex
\vspace{-2mm}\section{AI for Scientific Discovery}
\begin{figure*}[t]
	\centering
	\includegraphics[width=0.99\textwidth]{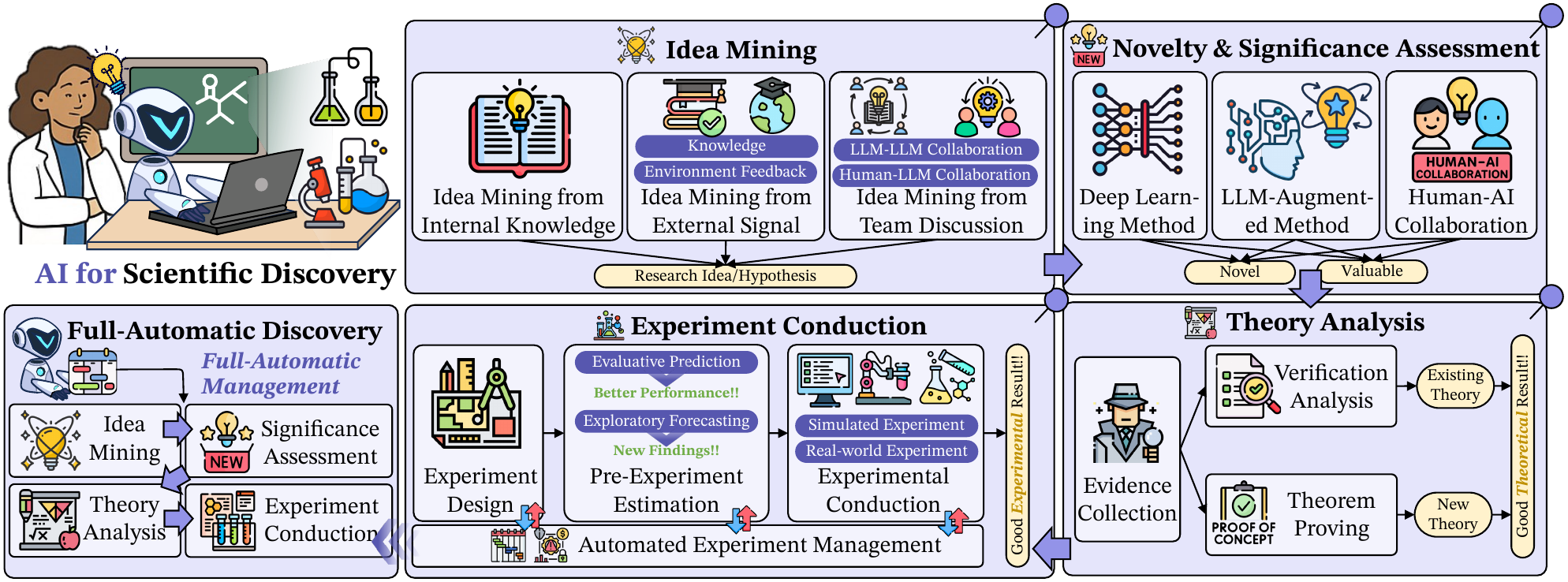}
	\caption{The AI-augmented pipeline for scientific discovery, encompassing Idea Mining, Novelty \& Significance Assessment, Theory Analysis, and Experiment Conduction. Full-Automatic Discovery integrates these elements into a cohesive, end-to-end process, supporting scientific exploration and innovation.}
	\label{fig:scientific-discovery}
\end{figure*}

\vspace{-1mm}
\label{sec:ai4scientific-discovery}
AI for Scientific Discovery~\citep{wang2023scientific,roy2025ai,raeini2025rise} leverages AI to generate novel hypotheses, theories, or ideas based on existing knowledge. Its goal is to expedite the research process by automating tasks such as idea generation, novelty and significance evaluation, theoretical analysis, and experimental design. This approach not only guides new research directions but also addresses complex scientific challenges \citep{lu2024beyond,harris2025airus,he2025reasoning}. As shown in Figure~\ref{fig:scientific-discovery}, it contains five main categories: Idea Mining ($\S$~\ref{sec:idea-mining}), Novelty \& Significance Assessment ($\S$~\ref{sec:novelty-judgment}), Theory Analysis ($\S$~\ref{sec:theory-analysis}), Experiment Conduction ($\S$~\ref{sec:scientific-experiment-conduction}) and Full-Automatic Discovery ($\S$~\ref{sec:full-automatic-discovery}).

\begin{table*}[t]
	\centering
	\setlength{\tabcolsep}{12pt}
	\resizebox{0.98\textwidth}{!}{
		\begin{tabular}{l|ccccc|c}
			\toprule
			\textbf{Model}                                        & \textbf{Fluency} & \textbf{Feasibility} & \textbf{Clarity} & \textbf{Originality} & \textbf{Flexibility} & \textbf{Average} \\

			\midrule

			DeepSeek-R1~\citep{guo2025deepseek}                   & 6.63             & \textbf{6.52}        & \textbf{8.10}    & \textbf{7.84}        & 6.83                 & \textbf{7.18}    \\
			Deepseek-R1-Distill-Qwen-32B~\citep{guo2025deepseek}  & 7.06             & 6.08                 & 7.43             & 7.13                 & 6.62                 & 6.86             \\
			Deepseek-R1-Distill-Llama-70B~\citep{guo2025deepseek} & 6.66             & 6.07                 & 7.43             & 6.98                 & 6.41                 & 6.71             \\

			\midrule
			Claude-3.7-Sonnet~\citep{anthropic2024claude3}        & \textbf{7.80}    & 5.46                 & 7.61             & 7.81                 & \textbf{6.92}        & 7.12             \\
			Claude-3.5-Sonnet~\citep{anthropic2024claude3}        & 6.90             & 5.42                 & 7.85             & 7.83                 & 6.62                 & 6.92             \\
			Claude-3.5-Haiku~\citep{anthropic2024claude3}         & 5.61             & 5.05                 & 7.40             & 7.72                 & 6.08                 & 6.37             \\
			Claude-3-Opus~\citep{anthropic2024claude3}            & 5.74             & 5.66                 & 7.72             & 6.66                 & 6.04                 & 6.36             \\

			\midrule
			Gemini-2.0-Flash-Exp~\citep{team2023gemini}           & 7.30             & 6.02                 & 7.84             & 7.37                 & 6.83                 & 7.18             \\
			Gemini-2.0-Flash-Thinking-Exp~\citep{team2023gemini}  & 7.38             & 6.05                 & 7.69             & 7.35                 & 6.83                 & 7.06             \\
			Gemini-2.0-Pro-Exp~\citep{team2023gemini}             & 6.84             & 5.88                 & 7.90             & 7.76                 & 6.75                 & 7.03             \\
			Gemini-Pro-1.5~\citep{team2024gemini}                 & 6.68             & 5.92                 & 7.75             & 7.33                 & 6.58                 & 6.85             \\

			\midrule
			GPT-4o~\citep{achiam2023gpt}                          & 6.12             & 5.58                 & 7.74             & 7.64                 & 6.38                 & 6.69             \\
			o1-mini~\citep{jaech2024openai}                       & 5.89             & 6.20                 & 7.77             & 7.09                 & 6.33                 & 6.66             \\
			o1~\citep{jaech2024openai}                            & 6.23             & 5.88                 & 7.42             & 7.23                 & 6.29                 & 6.61             \\
			o3-mini~\citep{jaech2024openai}                       & 5.57             & 5.91                 & 7.43             & 7.45                 & 6.21                 & 6.51             \\
			o3-mini-high~\citep{jaech2024openai}                  & 5.76             & 5.82                 & 7.62             & 6.95                 & 6.17                 & 6.47             \\
			GPT-4o-mini~\citep{achiam2023gpt}                     & 5.28             & 5.86                 & 7.45             & 6.67                 & 6.00                 & 6.25             \\

			\midrule
			Llama-3.1-405B-Instruct~\citep{grattafiori2024llama}  & 6.57             & 5.56                 & 7.48             & 7.18                 & 6.33                 & 6.62             \\
			Llama-3.1-70b-Instruct~\citep{grattafiori2024llama}   & 6.71             & 5.49                 & 7.34             & 7.16                 & 6.38                 & 6.62             \\

			\midrule
			QwQ-32B~\citep{qwq2025qwq}                            & 6.45             & 6.35                 & 7.98             & 7.77                 & 6.75                 & 7.06             \\
			QwQ-32B-Preview~\citep{qwen2024qwq}                   & 7.49             & 6.10                 & 7.46             & 6.87                 & 6.71                 & 6.93             \\
			Qwen-2.5-72B-Instruct~\citep{qwen25}                  & 6.17             & 5.99                 & 7.72             & 6.91                 & 6.29                 & 6.62             \\
			Qwen-2.5-7B-Instruct~\citep{qwen25}                   & 6.66             & 6.02                 & 7.17             & 6.34                 & 6.17                 & 6.47             \\

			\midrule
			Mistral-Small~\citep{jiang2024identifying}            & 7.36             & 5.97                 & 7.36             & 6.98                 & 6.62                 & 6.86             \\
			Mistral-Large~\citep{jiang2024identifying}            & 6.68             & 6.06                 & 7.69             & 7.01                 & 6.50                 & 6.79             \\

			\midrule
			Nova-Pro-v1~\citep{intelligence2024amazon}            & 6.45             & 6.19                 & 7.41             & 6.59                 & 6.25                 & 6.58             \\
			Nova-Lite-v1~\citep{intelligence2024amazon}           & 4.51             & 6.06                 & 7.38             & 6.60                 & 5.71                 & 6.05             \\

			\midrule
			Phi-4~\citep{abdin2024phi}                            & 6.58             & 5.80                 & 7.57             & 7.24                 & 6.42                 & 6.72             \\
			Gemma-2-27b-IT~\citep{team2024gemma}                  & 7.18             & 5.50                 & 7.36             & 6.86                 & 6.38                 & 6.65             \\
			Grok-2~\citep{xai2024grok2}                           & 5.76             & 5.82                 & 7.62             & 6.95                 & 6.17                 & 6.47             \\

			\bottomrule
		\end{tabular}}
	\caption{Results from the Liveideabench benchmark~\citep{ruan2024liveideabench} across five key dimensions: Fluency, Feasibility, Clarity, Originality, and Flexibility. Data is sourced from \citet{ruan2024liveideabench}. The \textbf{bolded} contents indicate the highest performance for each metric.}
	\label{tab:ai4idea-results}
\end{table*}

\vspace{-2mm}\subsection{Idea Mining}\vspace{-1mm}
\label{sec:idea-mining}
Idea mining, also known as hypothesis generation, is crucial for producing innovative, impactful research~\citep{o2025sparks,azher2025futuregen,cohrs2025large}. Recent studies show that the LLMs exhibit strong creativity and can facilitate automated scientific discovery \citep{liu2025researchbench,kumar2024can,si2024can,gu2024llms}. A comprehensive comparison of these findings is presented in Table~\ref{tab:ai4idea-results}. This suggests a future where AI agents act as independent researchers. Current efforts in this domain focus on extracting ideas from various sources to foster innovation~\citep{sanyal2025spark, sternlicht2025chimera,lin2025cognitio}. These methods can be broadly categorized into three main approaches:

\vspace{-2mm}\subsubsection{Idea Mining from Internal Knowledge}\vspace{-1mm}
Idea mining from internal knowledge leverages the latent knowledge and generative capabilities of large language models to discover novel concepts without relying on external data~\citep{girotra2023ideas,si2024can,kumar2024can}. By leveraging pretrained parameters and customized prompts, researchers can extract a variety of high-quality ideas embedded within the model~\citep{meincke2024innovation,chen2025structuring,swanson2025virtual}.
Earlier, \citet{meincke2024prompting} guided LLMs toward distinct ``idea spaces'' by adjusting decoding temperatures and applying constraint-based prompts, effectively encouraging exploration of diverse thematic trajectories.
Building on these insights, \citet{liu2025enhance} conduct an empirical study where undergraduates use an interactive tool that auto-prompted GPT-4 with business-model templates, resulting in ideas with higher novelty and feasibility, all without extensive innovation training.
Additionally, \citet{liu2025improving} demonstrate that injecting metadata into the LLM-based ideation process and applying automated validation during selection significantly increased idea feasibility and overall quality in climate-negotiation experiments.
Furthermore, \citet{chen2025ecm} model the process of inference-time learning and reasoning as a circuit, enhancing the idea-mining ability of the model through various voltage-enhancing techniques.

\vspace{-2mm}\subsubsection{Idea Mining from External Signal}\vspace{-1mm}
LLMs in research workflows can leverage not only internal parameterized knowledge, but also external signals to generate more novel, feasible, and contextually relevant hypotheses and ideas. By incorporating structured knowledge repositories or experimental feedback, these approaches extend beyond purely internal reasoning. We categorize them into two types: \vspace{-15pt}

\paragraph{Idea Mining from External Knowledge} involves supplying AI with curated academic data, such as publication metadata, citation networks, or knowledge graphs, to drive idea mining. Integrating up-to-date, domain-specific information ensures that generated hypotheses align with the latest developments in the field~\citep{liu2024literature,li2024learning,o2025sparks}.
Early efforts, limited by the capacity of earlier language models, focus on predicting relationships between concepts to generate classical ``A+B'' ideas~\citep{henry2017literature,krenn2022predicting}.
With recent advancements in language modeling~\citep{qin2024large,zhao2023survey}, attention has shifted to utilizing LLMs to explore ideas from scholarly data~\citep{majumder2024position,liu2024literature}. To support this, researchers have proposed various strategies for organizing literature to optimize knowledge extraction and mining. These mainly include ternary knowledge representations~\citep{wang2024scimon}, chained structures~\citep{li2024chain}, comprehensive databases~\citep{wang2024scipip}, and knowledge graphs~\citep{buehler2024accelerating,gu2024generation,gao2025graph}. Furthermore, efforts have focused on refining the knowledge injection process in idea mining. For example, \citet{gu2024llms} propose a framework with (1) a generalized retrieval system for cross-domain knowledge discovery and (2) a structured combinatorial process for improved idea mining.\vspace{-15pt}

\paragraph{Idea Mining from External Environment Feedback} involves treating idea mining as an interactive loop that incorporates feedback from experimental or simulated environments~\citep{baek2024researchagent, pu2025ideasynth}. Static document mining, by contrast, often overlooks the complexities of the real world, limiting its potential for innovation. These methods enable AI systems to propose experiments, receive outcome data, and refine subsequent ideas, thereby mimicking the research cycle of design, execution, and analysis~\citep{gpt-researcher, zochi2025}. In this domain, researchers primarily utilize multi-agent-based autonomous research systems, integrating idea mining agents with experiment conduction agents~\citep{schmidgall2025agent, schmidgall2025agentrxiv, carl2025}. Furthermore, researchers have successfully extended idea mining to various experimental disciplines, including chemistry~\citep{m2024augmenting}, materials science~\citep{ni2024matpilot}, biology~\citep{cui2025lumi}, medicine~\citep{swanson2025virtual, gottweis2025towards}, and machine learning~\citep{huang2023mlagentbench,ou2025automind}.

\vspace{-2mm}\subsubsection{Idea Mining from Team discussion}\vspace{-1mm}
Idea mining from team discussion encompasses approaches that simulate or facilitate collaborative brainstorming among multiple agents, either purely algorithmic or involving human participants, leveraging iterative critique, background knowledge retrieval, and facet recombination to generate richer, more diverse idea portfolios than single-agent pipelines.\vspace{-15pt}

\paragraph{AI-AI Collaboration} improves scientific ideation by refining hypotheses, critiquing proposals, and integrating external knowledge (also referred to as multi-agent collaboration)~\citep{swanson2025virtual,liu2024aigs}. We categorize current approaches into two mainstreams:
\textit{\textbf{(1) Feedback-guided Mining}}  involves agents exchanging critiques at various research stages to refine hypotheses through iterative feedback. Some studies introduce feedback loops across idea mining, experimental design, and result interpretation to optimize performance~\citep{zhou2024hypothesis,yang2023large,sinha2025can}, while others refine hypotheses using earlier outputs~\citep{hu2024nova}. These methods integrate peer review~\citep{lu2024ai}, direct critiques of hypotheses~\citep{baek2024researchagent}, and evaluations of experimental results~\citep{ma2024llm,yuan2025dolphin}.
\textit{\textbf{(2) Team-Discussion-guided Mining}} assembles multiple agents with distinct roles to simulate human research team dynamics \citep{pu2025piflow,nigam2024acceleron,ghafarollahi2024sciagents}. Specifically, \citet{su2024two} create a virtual research team (VirSci) where agents iteratively propose and critique ideas, producing more novel concepts than single-agent prompts by leveraging an expanding idea archive~\citep{zhang2023exploring,lu2024ai}. \citet{yanglarge} has developed a multi-intelligentsia framework, MOOSE-Chem, based on LLMs, specialized in scientific hypothesis discovery in chemistry, which can perform the functions of retrieving inspiration and generating hypotheses based on research contexts. Moreover, \citet{li2024chain} introduce the Chain-of-Ideas (CoI) agent, which organizes literature into a sequential chain, mirroring a topic's historical progression. This method generates outputs of similar quality to small research teams with minimal costs. \citet{lagzian2025multi} further enhance diversity and novelty via inference-time multi-view brainstorming.\vspace{-15pt}

\paragraph{Human-AI Collaboration} means the process where a human researcher guides an LLM’s exploration by selecting and curating intermediate artifacts, which the model then recombines and refines~\citep{nigam-etal-2024-interactive,ni2024matpilot}. For instance, \citet{radensky2024scideator} introduce Scideator, a system that enables researchers to select various facets, such as the problem statement, methodology, and dataset, from existing papers. The LLM subsequently recombines these facets to generate novel candidate ideas, significantly improving the idea qualities. Similarly, \citet{garikaparthi2025iris} present IRIS, an interactive research ideation system that facilitates human-AI collaboration by validating research motivations and synthesizing methodological suggestions in response to researcher queries. However, the research findings \citep{kumar2025human} show that, although LLM assistance can yield short-term boosts in creativity during supported tasks, it may inadvertently hamper users’ independent creative performance when working unassisted, thereby raising concerns about its long-term effects on human creativity and cognitive abilities.

\vspace{-2mm}\subsection{Novelty \& Significance Assessment}\vspace{-1mm}
\label{sec:novelty-judgment}
Novelty \& Significance Assessment focuses on AI methods that evaluate the originality and impact of ideas and scholarly papers~\citep{johnson2024greater,si2024can,keya2025sci}. The field predominantly employs three approaches:
\textit{\textbf{(1) Traditional Methods:}} Initially, models are trained to classify or regressively assess novelty and significance~\citep{dove2025semi,ikoma2025can,wang2025enabling}. For instance, \citet{singh2024supporting} propose SAPPhIRE that utilizes the causality ontology to quantify novelty in design problems, measuring textual similarity at multiple abstraction levels against historical works. Additionally, \citet{wang2024content} introduce ``surprise'' as an alternative measure of novelty, comparing a paper’s word distribution to a language model’s representation of scholarly discourse. This approach aligns with scientific intuition (face validity) and shows a correlation with expert judgments (construct validity).
\textit{\textbf{(2) LLM-Augmented Methods:}} With the significant development of LLMs, a series of works try to integrate LLMs for better novelty and significance assessment~\citep{wu2025sc4anm}. Typically,
\citet{feng2025grapheval} propose GraphEval, a lightweight, graph-based LLM framework for reasoning evaluation, that prompts a small-scale LLM to decompose complex reasoning processes into easily interpretable ``viewpoint'' nodes, thereby enhancing the robustness of reasoning assessment.
\textit{\textbf{(3) Human-AI Collaboration  Methods:}}
Unfortunately, purely LLM-augmented assessments of novelty may overestimate creativity~\citep{chakrabarty2024art,laverghetta2025humans} and lead to homogenization effects without human input~\citep{anderson2024homogenization,zhou2024shared}. As a result, there is growing interest in human-AI collaboration for novelty assessment, with several works~\citep{ashkinaze2024ai,liu2024ai,padmakumar2023does} advocating for the integration of human-guided ideation alongside LLM-based workflows.

\vspace{-2mm}\subsection{Theory Analysis}\vspace{-1mm}
\label{sec:theory-analysis}
Any scientific idea or hypothesis must be rigorously evaluated to confirm its validity. Theory analysis involves using AI methods to determine whether a hypothesis aligns with established scientific principles. AI applications in theory analysis can be divided into three main components:

\vspace{-2mm}\subsubsection{Scientific Claim Formalization}\vspace{-1mm}
Scientific claim formalization converts natural-language assertions into structured representations for systematic verification~\citep{rao2025nsf,kumar2025sciclaimhunt,metropolitansky2025towards}. Early approaches relied on template-based methods~\citep{goodsell2024lf}. For example, \citet{heger2024natural} describe a pipeline that converts complex hypotheses into machine-readable templates. Subsequent works focus on incorporating LLMs to refine these templates. \citet{ganguly2025grammars} propose a PCFG-based framework to address common LLM failure modes, while Valsci~\citep{edelman2025valsci} automates the conversion of natural-language claims into templated queries for LLM-driven verification. More recently, \citet{yan2025position} suggest that integrating text, images, and other modalities in multimodal LLMs provides richer structured representations, facilitating cross-domain reasoning.

\vspace{-2mm}\subsubsection{Scientific Evidence Collection}\vspace{-1mm}
Scientific evidence collection involves systematically identifying, retrieving, and curating data sources to support or challenge research claims \citep{pan2023investigating, wadden2021multivers, ke2024can}. Previous studies have focused on methods for evaluating and improving the quality of retrieved sources \citep{vladika2024comparing} and optimizing retrieval configurations \citep{vladika2024improving}. Additionally, strategies have emerged to address incomplete or faulty evidence, including techniques for detecting missing information \citep{glockner2022missing} and understanding the causes of retrieval errors \citep{wuhrl2024understanding, glockner2024grounding}.
More recent efforts have integrated LLMs with retrieval systems to enhance the accuracy of evidence retrieval and verification. For instance, SciClaims \citep{ortega2025sciclaims} combines claim extraction, evidence retrieval, and verification into a single LLM-powered pipeline, streamlining the entire process. Similarly, \citet{alvarez2024zero} and \citet{wang2025llmbased} extend retrieval-augmented generation by producing structured query representations and retrieving corroborating or refuting evidence in a single step.

\vspace{-2mm}\subsubsection{Scientific Verification Analysis}\vspace{-1mm}
Scientific verification analysis plays a critical role in AI-driven theoretical qualitative studies by assessing the logical coherence~\citep{ku2025theoremexplainagent, liang2025explainable}, factual consistency~\citep{muharram2024enhancing, kao2024magic, cao2024can, braun2024defame}, and robustness~\citep{jafari2024robust} of claims based on existing evidence.
Research underscores the importance of domain expertise for accurate and reliable verification \citep{das2023state, altuncu2023aedfact, bazaga2023unsupervised, wuhrl2024makes}. To mitigate errors and enhance interpretability, some frameworks adopt human-like, stepwise pipelines~\citep{kim2023factkg, dammu2024claimver, wu2023characterizing, atanasova2024generating}. For instance, HiSS~\citep{zhang2023towards} and ProToCo~\citep{zeng2023prompt} employ multiple cueing to validate each substatement, improving reliability.
Other methods integrate verification with experimental results to boost transparency and interpretability \citep{krishna2022proofver, pan2023fact, eldifrawi2024automated, zhang2024augmenting}. \citet{gx2025language} show that LLMs inherit reasoning heuristics from training data, leading to cognitive biases. To mitigate this, they propose an inference-time-scaling sampling procedure that reduces implicit causal assumptions and aligns the model’s reasoning with causal rigor. More recently, \citet{ku2025theoremexplainagent} introduced the task of generating coherent visual explanations and demonstrated that combining agents with Manim animations to produce long-form theorem explanation videos (over five minutes) results in more effective visual explanations.

\vspace{-2mm}\subsubsection{Theorem Proving}\vspace{-1mm}
Theorem proving involves the development of algorithms and models, often incorporating generative language models, to autonomously generate and verify formal mathematical proofs~\citep{li2024survey,frieder2024data,zhu2025deep,yang2025discovering}. Early methods~\citep{wang2023dt,lample2022hypertree} introduce dynamic tree proof search techniques and integrate retrieval algorithms with language models for theorem proving~\citep{polu2020generative}. However, retrieval algorithms tend to prioritize trivial intermediate conjectures, resulting in poor performance \citep{wang2024proving}.
To overcome this, some researchers have introduced novel approaches that replace retrieval algorithms entirely~\citep{jiang2022thor,jiang2022draft}. LEGO-Prover \citep{wang2023lego} employs Growing Libraries to enhance LLM reasoning, while \citet{zhao2023decomposing} suggest Subgoal-based Demonstration Learning for more effective theorem proving. Additionally, Lean Copilot \citep{song2024towards} and Lean-STaR \citep{lin2024lean} leverage the Lean programming language and theorem prover to enable improved human-AI collaboration in proof completion.
Recent studies have focused on fine-tuning specialized proving LLMs~\citep{first2023baldur}. For example, MUSTARD \citep{huang2024mustard} and DeepSeek-Prover \citep{xin2024deepseek} aim to generate high-quality synthetic data to fine-tune models and improve theorem proving.

\vspace{-2mm}\subsection{Scientific Experiment Conduction}\vspace{-1mm}
\label{sec:scientific-experiment-conduction}
Automatic Scientific Experiment Conduction leverages AI to design, conduct, and analyze scientific studies autonomously, aiming to automate the entire process, from hypothesis formulation to data interpretation. This automation seeks to accelerate research and improve reproducibility~\citep{chen2025ai,kon2025exp,baydin2021toward,vischia2025ai}. However, \citet{zhu2025ai} highlight a critical challenge: AI scientists currently lack the validation capabilities needed for rigorous experimentation and high-quality manuscript production. Without these essential competencies, such platforms cannot succeed.

\vspace{-2mm}\subsubsection{Experiment Design}\vspace{-1mm}
Experimental design is vital for efficiency and provides the foundation for AI-assisted experiment conduction methods~\citep{vischia2025ai,cui2025lumi,gottweis2025towards,ghafarollahi2024sciagents}. Evidence shows that systematic design plays a central role in automating and enhancing experimental processes~\citep{ni2024matpilot,m2024augmenting}.\vspace{-15pt}

\paragraph{Semi-Automatic Experiment Design} involves the creation of experimental plans through human-AI collaboration~\citep{ni2024matpilot,m2024augmenting}. \citet{arlt2024meta} present a transformer-based framework that autonomously generates quantum experiment protocols and uncovers state preparation principles. \citet{huang2024ai} integrate deep learning with multi-objective optimization to design polymer sequences with both high thermal conductivity and synthetic feasibility, validating their results through molecular dynamics. \citet{liang2024application} apply a variational autoencoder combined with reinforcement learning to enhance the design and efficiency of parameters for cultural creative products. \citet{craig-2025-human} propose a human-AI collaboration framework based on experimental design, case-based reasoning, and a note-taking system, offering scientists a structured LLM tool with transparent documentation, resulting in verifiable experimental designs and knowledge integration.\vspace{-15pt}

\paragraph{Full-Automatic Experiment Design} refers to the application of agent-centric methods for the automatic scheduling of scientific experiments~\citep{carl2025,zochi2025,ghareeb2025robin}. Platforms such as The AI Scientist~\citep{lu2024ai} and Agent Laboratory~\citep{schmidgall2025agent} continuously refine experimental protocols by incorporating new data in real-time~\citep{swanson2025virtual,baek2024researchagent,schmidgall2025agentrxiv}. In a significant development, \citet{liu2024largelanguagemodel} proposed an end-to-end generative-agent framework that enables fully autonomous planning, spanning from literature review to protocol iteration, without the need for human intervention. Additionally, \citet{roohani2024biodiscoveryagent} introduced a biodiscovery agent capable of designing, evaluating, and optimizing gene-perturbation experiments. This system has shown superior performance over traditional Bayesian methods, especially in targeting non-essential genes.

\vspace{-2mm}\subsubsection{Pre-Experiment Estimation}\vspace{-1mm}
Pre-experiment prediction leverages AI to forecast experimental outcomes, aiming to improve research efficiency and accuracy. This process can be divided into two categories:\vspace{-15pt}

\paragraph{Evaluative Prediction} predicts quantitative values or trends of experimental outcomes, such as estimating drug concentration effects, determining whether a compound affects cellular activity, and assessing protocol feasibility \citep{cui2025lumi}.
\textit{\textbf{(1) Deep-Learning  Methods:}}
With the rise of deep learning, hierarchical prediction models have emerged \citep{wu2024deepcre}. \citet{li2024physical} incorporated physical equations to predict pharmacokinetic parameters, reducing data requirements and enhancing noise robustness. More recently, \citet{li2025unimatch} proposed a dual-matching framework, combining hierarchical molecular alignment with meta-learning, which showed significant improvements in drug feature estimation.
\textit{\textbf{(2) LLM-Augmented  Methods:}}
More recently, with the advent of LLM capabilities, \citet{zhang2024massw} demonstrate successful LLM-assisted evaluative prediction, a method that has been further extended in subsequent studies. Notably, \citet{luo2025large} integrate BrainGPT into neuroscience literature retrieval, outperforming domain experts in evaluating experimental estimation. \citet{wen2025predicting} develop a system combining fine-tuned GPT-4.1 with a paper retrieval agent, which outperformed 25 human experts in evaluating experimental predictions.
\vspace{-15pt}

\paragraph{Exploratory Forecasting} utilizes AI to predict experimental outcomes, generate new compounds, design reaction pathways, and propose combinatorial schemes to drive scientific discovery~\citep{m2024augmenting,liu2025moose}. Several studies have applied deep generative models for chemical-space forecasting~\citep{gomez2018automatic,de2018molgan}. \citet{seo-2025-flavordiffusion} introduce a framework that uses graph diffusion modeling to predict ingredient-chemical molecule interactions, enabling innovative pairing exploration. Furthermore, based on a massive computation model, DeepMind's GNoME~\citep{barber2023_gnome} predicts approximately 380,000 stable material structures, demonstrating AI's potential in materials discovery. Recently, multi-turn interactive methods have also been developed to improve exploratory forecasting~\citep{gottweis2025towards}. For instance, \citet{zhang2024massw} and \citet{swanson2025virtual} present platforms that integrate multi-agent debates to better forecast experimental performance and foster idea exchange, advancing the discovery of new method variants.

\vspace{-2mm}\subsubsection{Experiment Management}\vspace{-1mm}
The integration of machine learning and robotics in AI-driven experiment management enables hypothesis generation, high-throughput experimentation, and iterative procedure refinement without human intervention~\citep{basford2024development,shahin2025agents,zhao2025artificial,angelopoulos2024transforming}. These paradigms, also named as ``self-driving laboratories''~\citep{britton2024ai,canty2025science,hatakeyama2025perspective}, promise accelerated discoveries~\citep{fehlis2025accelerating} in biology, chemistry, and materials science~\citep{kvapil2025intelligent,cao2024agents,fehlis2025uncovering}.\vspace{-15pt}

\paragraph{Open-Loop Management} involves experimental management without human oversight~\citep{wang2025autonomous}. \citet{hysmith2024future} explore human-AI collaboration, emphasizing the interoperability of robots, predictive models, and data pipelines. In bioprocessing, \citet{zournas2025machine} combine active learning with a semi-automated Design-Build-Test-Learn cycle to optimize microbial media, showing that higher NaCl levels significantly improve metabolite yield and process efficiency.
Google DeepMind and BioNTech~\citep{ft2024_deepmind_biontech} have introduced an AI-driven laboratory assistant to autonomously design protocols and predict outcomes, aiming to enhance research efficiency in the medical, energy, and educational sectors. Reports indicate that such systems could reduce the traditional 20-year, \$100 million timeline for materials discovery to just months. The U.S. government is supporting these efforts through strategic funding initiatives~\citep{axios2024_self_driving_labs}.\vspace{-15pt}

\paragraph{Close-Loop Management} entails fully autonomous experimental management without human intervention~\citep{wang2025autonomous}. The Functional Genomics Explorer~\citep{king2004functional} is a landmark in this area, being the first fully autonomous research platform that generates hypotheses, designs experiments, and validates results.
\citet{macleod2020self} describe a robotic system that formulates, deposits, and characterizes thin films using model-based optimization to enhance charge transport. AI-driven optimization algorithms are transforming experimental workflows, as seen in closed-loop Bayesian optimization methods for chemical and materials discovery~\citep{tom2024self}. \citet{knox2025self} apply multi-objective optimization to polymer nanoparticle synthesis, optimizing size, dispersity, and functionality.

\vspace{-2mm}\subsubsection{Experiment Conduction}\vspace{-1mm}
Experiment conduction refers to the application of AI techniques in executing and managing scientific experiments. This process is essential for automating workflows, ensuring experiments are carried out efficiently and accurately. The primary aim of experiment conduction is to reduce human involvement in the experimental process. It can be further divided into two categories:\vspace{-15pt}

\paragraph{Automated Machine Learning Experiment Conduction}  uses AI to streamline the design, training, and evaluation of ML models, reducing dependence on human expertise by covering the entire pipeline from preprocessing to hyperparameter optimization~\citep{zhang2023automl,tornede2023automl,grosnit2024large,zhang2025mlrc,novikov2025alphaevolve,10.1145/3579355}. Typically, \citet{wang2024opendevin} present a community-driven sandbox allowing agents to write code, browse the web, and coordinate through an event-stream API. For Kaggle challenges, \citet{li2024autokaggle} propose an iterative, collaborative multi-agent system that incorporates debugging and unit testing across the competition pipeline. AIDE \citep{AIDE2024} utilizes a tree-search loop to generate, evaluate, and refine solutions, achieving a bronze medal in Kaggle competitions. At the research level, \citet{li2024mlr} formalize a three-phase LLM agent workflow, idea generation, implementation, and execution, to automate experiments. \citet{zhao2025autoreproduce} and \citet{liu-etal-2025-variable} present a multi-agent method for extracting model variables from scientific texts, significantly improving experimental reproduction accuracy. \citet{cheng2025language} introduce a multi-agent framework for crafting novel LLMs by emulating standard research stages and leveraging scaling laws through a ``Ladder of Scales''. Designs are proposed, adversarially reviewed, implemented, and evaluated across model sizes from 14 million to 350 million parameters.
\vspace{-15pt}

\begin{table*}[t]
	\centering
	\setlength{\tabcolsep}{12pt}
	\resizebox{0.94\textwidth}{!}{
		\begin{tabular}{llcccccccc}
			\toprule
			\multirow{2}{*}{\textbf{Models}}                     & \multirow{2}{*}{} & \multicolumn{4}{c}{\textbf{Without Knowledge}} & \multicolumn{4}{c}{\textbf{With Knowledge}}                                                                                                         \\
			\cmidrule(lr){3-6}\cmidrule(lr){7-10}                &                   & SR                                             & CBS                                         & VER           & Cost $\downarrow$ & SR            & CBS           & VER           & Cost $\downarrow$ \\
			\midrule

			\rowcolor{gray!10} \multicolumn{10}{c}{\textit{Direct Prompting}}                                                                                                                                                                                                               \\
			\midrule
			Llama-3.1-Instruct-70B~\citep{grattafiori2024llama}  &                   & 5.9                                            & 81.5                                        & 29.4          & 0.001             & 4.9           & 82.1          & 27.5          & 0.001             \\
			Llama-3.1-Instruct-405B~\citep{grattafiori2024llama} &                   & 3.9                                            & 79.4                                        & 35.3          & 0.010             & 2.9           & 81.5          & 25.5          & 0.001             \\
			Mistral-Large-2~\citep{jiang2024identifying}         &                   & 13.7                                           & 82.3                                        & 47.1          & 0.009             & 16.7          & 84.7          & 39.2          & 0.001             \\
			GPT-4o~\citep{achiam2023gpt}                         &                   & 11.8                                           & 82.6                                        & \textbf{52.9} & 0.011             & 10.8          & 83.8          & 41.2          & 0.016             \\
			Claude-3.5-Sonnet~\citep{anthropic2024claude3}       &                   & \textbf{17.7}                                  & \textbf{83.6}                               & 51.0          & 0.017             & \textbf{21.6} & \textbf{85.4} & \textbf{41.2} & 0.016             \\
			o1-preview                                           &                   & 34.3                                           & 87.1                                        & 70.6          & 0.221             & 31.4          & 87.4          & 63.7          & 0.236             \\

			\midrule
			\rowcolor{gray!10} \multicolumn{10}{c}{\textit{OpenHands CodeAct}~\citep{wang2024openhands}}                                                                                                                                                                                    \\
			\midrule
			Llama-3.1-Instruct-70B~\citep{grattafiori2024llama}  &                   & 6.9                                            & 63.5                                        & 30.4          & 0.145             & 2.9           & 65.7          & 25.5          & 0.252             \\
			Llama-3.1-Instruct-405B~\citep{grattafiori2024llama} &                   & 5.9                                            & 65.3                                        & 32.0          & 0.383             & 8.3           & 71.4          & 58.0          & 0.384             \\
			Mistral-Large-2~\citep{jiang2024identifying}         &                   & 9.8                                            & 72.5                                        & 53.9          & 0.513             & 13.7          & 78.8          & 50.0          & 0.759             \\
			GPT-4o~\citep{achiam2023gpt}                         &                   & 19.6                                           & 83.4                                        & 87.5          & 0.803             & \textbf{27.5} & \textbf{86.3} & 73.5          & 1.094             \\
			Claude-3.5-Sonnet~\citep{anthropic2024claude3}       &                   & \textbf{21.6}                                  & \textbf{83.6}                               & \textbf{87.3} & 0.958             & 24.5          & 85.1          & \textbf{88.2} & 0.900             \\
			o1-preview~\citep{jaech2024openai}                   &                   & 33.4                                           & 86.2                                        & 87.0          & 0.999             & 35.3          & 88.4          & 91.5          & 0.913             \\

			\midrule
			\rowcolor{gray!10} \multicolumn{10}{c}{\textit{Self-Debug}~\citep{chen2023teaching}}                                                                                                                                                                                            \\
			\midrule
			Llama-3.1-Instruct-70B                               &                   & 13.7                                           & 82.7                                        & 40.4          & 0.007             & 16.7          & 83.5          & 73.5          & 0.005             \\
			Llama-3.1-Instruct-405B                              &                   & 16.7                                           & 80.0                                        & 35.3          & 0.006             & 23.6          & 79.4          & 40.4          & 0.004             \\
			Mistral-Large-2~\citep{jiang2024identifying}         &                   & 23.5                                           & 85.1                                        & 78.4          & 0.007             & 26.5          & 86.7          & 84.3          & 0.006             \\
			GPT-4o~\citep{achiam2023gpt}                         &                   & 22.6                                           & 84.4                                        & 84.3          & 0.024             & 33.4          & 87.1          & 86.3          & 0.037             \\
			Claude-3.5-Sonnet~\citep{anthropic2024claude3}       &                   & 32.4                                           & 86.4                                        & \textbf{92.2} & 0.026             & 34.5          & \textbf{87.1} & \textbf{86.3} & 0.015             \\
			o1-preview~\citep{jaech2024openai}                   &                   & \textbf{42.2}                                  & \textbf{88.4}                               & 92.0          & 0.636             & \textbf{41.2} & 88.9          & 91.2          & 0.713             \\

			\bottomrule
		\end{tabular}}
	\caption{Full-automatic discovery capability comparison on ScienceAgentBench~\citep{chen2024scienceagentbench}. The data presented are derived from \citet{chen2024scienceagentbench}. The \textbf{bolded} contents indicate the highest performance for each metric.}
	\label{tab:ai4discovery-results}
\end{table*}

\paragraph{Real-world Experimental Simulation \& Conduction.}
Recent advancements in the planning and reasoning capabilities of LLMs have led to their use in simulating experimental results~\citep{kambhampati2024position,zabaleta2024simulating,lee2025psyche,yue2025foam} and even direct conduct real-world experiments \citep{chen2025reinforcing,ruan2024automatic,feng2025openfoamgpt}. Real-world experimental simulation \& conduction generally employ four strategies:
\textit{\textbf{(1) Self-Improvement}}: Models iteratively refine their performance based on feedback \citep{huang2022large,shinn2023reflexion,zhang2024wrong,peng2025dlpo,yuan2025dolphin}. For example, \citet{siddiqui-etal-2025-evaluating} enhance functional approximation through iterative knowledge application. Further refinement occurs through analytical insights \citep{li2024mlr,szymanski2023autonomous,baek2024researchagent} and hyperparameter tuning \citep{ni2024matpilot,liu2024large,zhang2023mlcopilot}.
\textit{\textbf{(2) Multi-Agent Interaction}}: Models simulate collaborative research teams by assigning roles such as experimenter, analyst, or critic \citep{baek2024researchagent,ghafarollahi2024sciagents,sreedhar2025simulating,hu2025owl,solovevtowards}. For instance, MechAgent \citep{ni2024mechagents}, Researchcodeagent~\citep{gandhi2025researchcodeagent} and The AI Scientist \citep{lu2024ai,yamada2025ai} automate experiments through multi-agent collaboration, with LLMs acting as proxies in fields like computer science \citep{lu2024ai,yamada2025ai}, social science \citep{mou2024individual,manning2024automated}, and physical science \citep{zhang2025mooseagent}.
\textit{\textbf{(3) External Tool Integration}}: Researchers enhance model capabilities by linking them to databases, APIs, and other tools during experiments \citep{hao2023toolkengpt,qin2023toolllm,schick2023toolformer}. For example, \citet{boiko2023autonomous} integrate internet search, code execution, and automation into a GPT-4 system. Studies like ChemCrow \citep{m2024augmenting} and Crispr-GPT \citep{huang2024crispr} support chemistry and gene editing experiments through massive specialized toolchains.
\textit{\textbf{(4) Specific Fine-Tuning}}: A growing body of work explores the fine-tuning of specific models to improve experimental simulations. For instance, \citet{cui2024scgpt} present a transformer-based model trained on large single-cell transcriptomic datasets, achieving state-of-the-art accuracy in cell-type annotation and in silico perturbation response~\citep{liu2023training}.

\vspace{-2mm}\subsubsection{Experimental Analysis}\vspace{-1mm}
Experimental Analysis involves systematically testing hypotheses, evaluating models, or validating theoretical assumptions to draw meaningful conclusions. This process encompasses three main sub-processes:\vspace{-15pt}

\paragraph{Automated Evaluation Metrics} refer to systems like AutoML that automatically generate model learning curves, parameter sensitivity analysis graphs, and other evaluation tools to assess model performance~\citep{adriaensen2023efficient}. For instance, AutoML platforms assist researchers by automatically producing learning curves and sensitivity analysis graphs to better understand model behavior~\citep{barbudo2023eight, baratchi2024automated}.\vspace{-15pt}

\paragraph{Theoretical Consistency Analysis} ensures that the theoretical methods align with the experimental implementations~\citep{liu-etal-2025-variable}. AutoReproduce \citep{zhao2025autoreproduce} uses a large language model to create a multi-intelligent body system, enabling automatic comprehension, code reproduction, and execution verification of experiments in scientific papers. This process completes the consistency analysis between theoretical methods and experimental outcomes.\vspace{-15pt}

\paragraph{Exploratory Analysis} is essential for investigating and understanding datasets through statistical and visualization techniques to identify patterns, spot anomalies, test assumptions, or validate hypotheses~\citep{Rajdip2024, sui2024table}. This process extends the capabilities of language models for data exploration in structured formats. For example, \citet{xing2024table} utilizes a generator-validator fine-tuning approach to enable language models to specialize in parsing tabular data, improving table structure inference and summarization. Additionally, \citet{bian2023helm} developed HeLM, which facilitates high-quality natural language summarization of table content, aiding in the generation of conclusions.

\vspace{-2mm}\subsection{Full-Automatic Discovery}\vspace{-1mm}
\label{sec:full-automatic-discovery}
Full-automatic discovery refers to the ability to close the loop of the scientific process, from hypothesis generation and experimental design to autonomous execution, result analysis, and iterative feedback, powered by end-to-end artificial intelligence~\citep{mandal2024autonomous}.  A comprehensive comparison of the results is presented in Table~\ref{tab:ai4discovery-results}.
Advances in laboratory automation and closed-loop assistants are driving fully automated discovery toward greater reliability, innovation, and faster iteration through multi-agent systems~\citep{narayanan2024aviary,naumov2025dora,team2025novelseek,carl2025}.
For example, \citet{lu2024ai} and \citet{yuan2025dolphin} use literature mining to rank research topics, employ an ``anomaly-guided'' code-synthesis framework to generate and debug experimental scripts, and feed results back into the ideation module to iteratively refine hypotheses~\citep{schmidgall2025agent,schmidgall2025agentrxiv,mathur2025vision}.
\citet{kon2025curie} introduce rigor through three modules: intra-agent rigor for reliability, inter-agent rigor for systematic control, and an experimental knowledge module for interpretability, addressing issues of insufficient rigor and overstated claims. \citet{li2025autosdt} extend this approach to data-driven discovery, enhancing exploration diversity. Further, Zochi~\citep{zochi2025} is developed as an AI-driven system for end-to-end scientific discovery, demonstrating its comprehensive capabilities across the research lifecycle. Papers generated through Zochi have even been accepted by ACL 2025.

%% file: sections/writing.tex
\vspace{-2mm}\section{AI for Academic Writing}
\begin{figure*}[t]
	\centering
	\includegraphics[width=0.99\textwidth]{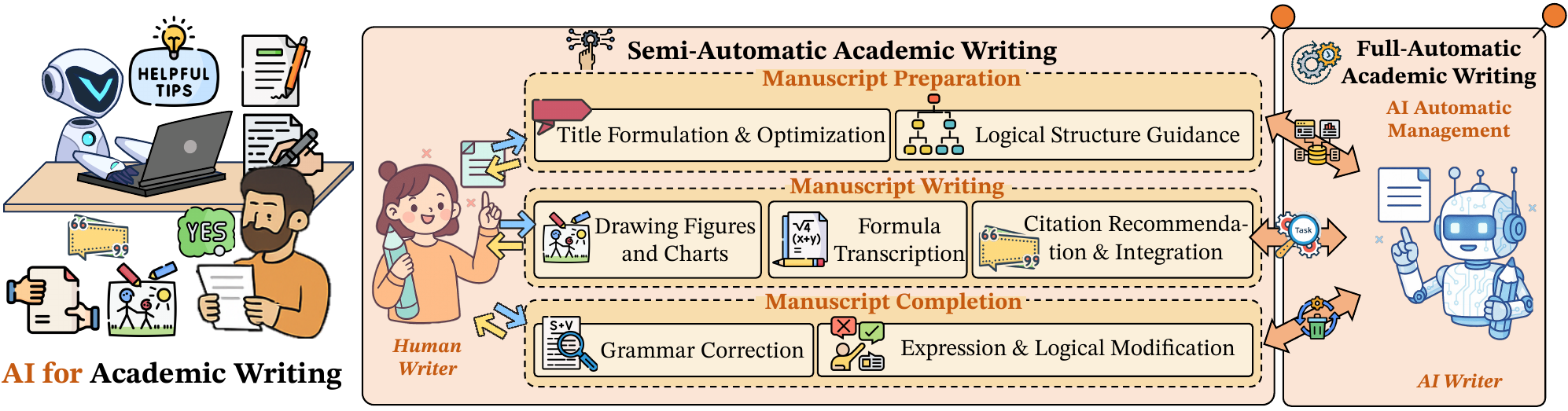}
	\caption{The main paradigms of AI for Academic Writing. It can be divided into two main categories: Semi-Automatic Academic Writing and Full-Automatic Academic Writing. Specifically, Semi-Automatic Academic Writing encompasses Manuscript Preparation, Manuscript Writing, and Manuscript Completion.}

	\label{fig:academic-writing}
\end{figure*}

\vspace{-1mm}
\label{sec:ai4writing}
AI for Academic writing involves the use of AI techniques to assist researchers or generate from scratch in drafting, editing, and formatting scientific manuscripts~\citep{khalifa2024using}. With the development of deeper interaction between human and LLMs, human and LLMs are quickly shaping each other's better writing habits~\citep{geng2025human,calderon2025and,zhou2025large}. As shown in Figure~\ref{fig:academic-writing}, it contains two main categories: Semi-Automatic Academic Writing ($\S$~\ref{sec:semi-automatic-academic-writing}) and Full-Automatic Academic Writing ($\S$~\ref{sec:full-automatic-academic-writing}).

\vspace{-2mm}\subsection{Semi-Automatic Academic Writing}\vspace{-1mm}
\label{sec:semi-automatic-academic-writing}
Semi-automatic academic writing involves the use of AI tools to assist researchers in drafting and editing scientific manuscripts, requiring human input and oversight. This approach aims to enhance the quality and efficiency of scientific writing by offering AI-generated suggestions, corrections, and formatting assistance. Semi-automatic academic writing can be categorized into three phases:

\vspace{-2mm}\subsubsection{Assistance During Manuscript Preparation}\vspace{-1mm}
Assistance During Manuscript Preparation refers to the support provided by models and tools throughout the manuscript creation process. This includes generating and refining titles, guiding the overall structure, and ensuring content coherence, all aimed at improving the clarity, quality, and readiness for submission.\vspace{-15pt}

\paragraph{Title Formulation and Optimization} involves using models to generate multiple title candidates and selecting the most suitable one~\citep{bolucu2025modest,bikku2025generating}. For example, given a research topic like ``new energy battery materials'', a model generates 5-10 titles with varying focuses, which are evaluated based on novelty, complexity, and potential impact, helping the author select the best option~\citep{bikku2025generating}. To further improve the title quality, \citet{rehman2025can} fine-tune PEGASUS-large and use GPT-3.5-turbo (zero-shot) to generate titles from abstracts. \citet{au2025personalized} enhance title quality by incorporating user preferences for more coherent and personalized titles.\vspace{-15pt}

\paragraph{Overall Logical Structure Guidance} involves providing the model with section and subsection headings, as well as paragraph outlines, to evaluate the logical flow, ensure the avoidance of repetition, correct ordering errors, and identify any missing elements~\citep{hashemi-etal-2024-llm}. \citet{sun2024lalaeval} propose a multi-stage workflow for assessing paper structure. Their rubric emphasizes the importance of section completeness and content cohesion, ensuring that headings and paragraphs adhere to established formatting standards.

\vspace{-2mm}\subsubsection{Assistance During Manuscript Writing}\vspace{-1mm}
Semi-automatic AI writing involves a collaborative process where humans create the primary content, while AI contributes supplementary elements. In this collaboration, AI assists with tasks that are secondary to the main content~\citep{lin2025divergent,cheng2025artificial,xu2025patterns,song2023enhancing,nguyen2024human}. This process can be divided into three primary tasks:\vspace{-15pt}

\paragraph{Drawing Figures and Charts} serves as an effective means of conveying experimental data and analysis~\citep{belouadi2025tikzero,chang2025sridbench}. Recent advancements in AI research on automatic scientific figure generation have led to significant breakthroughs~\citep{zhang2024scimage,hogan2024aiscivision,illustrae2025how,rashid2022text2chart}. \citet{rodriguez2023figgen} introduce the FigGen model, which maps textual descriptions to complex academic figures with high fidelity. Beyond direct image generation, several studies have explored programming scientific diagrams using Python~\citep{zhao2025chartcoder,zhang2025enhancing,cheng2023chartreader,moured2024chartformer}, SVG~\citep{rodriguez2025starvector}, or tikz~\citep{belouadi2025tikzero,belouadi2023automatikz} for better figure quality.
However, figures alone cannot fully convey results; captions are essential for ensuring that readers understand each figure~\citep{cao2024figuring,lu2024ai,yin2025understanding}. To address this, \citet{hsu2024scicapenter} develop SciCapenter, an interactive system that generates multiple caption candidates, scores them, and quality-checks each, assisting authors in selecting the most optimal phrasing. MLBCAP~\citep{kim2025multi} utilize multiple LLMs to collaborate on chart-based title generation. Additionally, frameworks like AI Scientist~\citep{lu2024ai,yamada2025ai} can identify key data from experimental logs, generate figures with captions, and integrate them into authoring tools, advancing end-to-end AI-driven scientific visualization.

\vspace{-15pt}

\paragraph{Formula Transcription} need to digitize extensive mathematical formulas and tables in academic and instructional materials, driven the development of automated tools that convert handwritten or image-based expressions into editable LaTeX. Specifically, \citet{sundararaj2024automated} employ a Vision Transformer (ViT) to transcribe these expressions into LaTeX, improving accuracy and reducing manual proofreading. \citet{vrevcar2024towards} introduce a semi-automated tool for semantic annotation, enhancing the accessibility and interoperability of mathematical symbols in LaTeX. \citet{jiang2025latte} propose the iterative refinement framework, which generates an initial LaTeX draft, compares it to the source image for feedback, and iteratively corrects errors, thereby minimizing manual verification and facilitating better transcription.\vspace{-15pt}

\paragraph{Citation Recommendation \& Integration} has emerged as a key research area in academic writing, focusing on retrieving and incorporating relevant literature into documents to enhance writing efficiency and citation accuracy~\citep{zhang2023large,maharjan2024benchmark,watson2024directed,li2025scirgc,algaba2025deep,liebling-etal-2025-towards}. Earlier, \citet{ma2021chronological} introduce a temporal preference model for ranking citations, setting the stage for subsequent research considering time factors~\citep{roy2024ilciter}. In generative models, \citet{ccelik2024citebart} develope CiteBART, which masks citation markers in context and reconstructs them, enabling zero-shot citation generation. More recently, \citet{wang2025scholarcopilot} introduce ScholarCopilot, which generates special retrieval tokens and dynamically queries literature databases to embed references in real time. \citet{he2025pasa} propose PaSa, an advanced paper-search agent powered by large language models. PaSa autonomously invokes search tools, reviews manuscripts, and selects relevant references, achieving results that surpass even Google + GPT-4o in complex scholarly queries.

\vspace{-2mm}\subsubsection{Assistance After Manuscript Completion}\vspace{-1mm}
After completing the paper, the author typically needs to refine its quality, focusing on language accuracy and logical coherence. At this stage, support tools can assist in grammar correction, expression and logical revision, ensuring the paper is clear, fluent, and logically sound.\vspace{-15pt}

\paragraph{Grammar Correction} means the model proofreads each paragraph, identifies spelling errors, improper punctuation, repetitive phrasing, and character-encoding issues, and provides corresponding revision suggestions~\citep{george2024paperpal,yang2025transforming,wang2024lm,sun2023csed}.
Specifically, \citet{wang2024improving_b} propose a synthetic data construction method based on contextual augmentation, which can ensure an efficient augmentation of the original data with a more consistent error distribution.
\citet{wang2024neural} propose an integrate automated writing evaluation system with grammatical error correction to support L2 essay writers by providing immediate feedback, offering targeted guidance to improve grammar and coherence, reducing manual grading efforts.
Further, \citet{zheng2025usage} present a Transformer-based feedback framework that generates real-time suggestions on grammar, vocabulary, sentence structure, and logical coherence for non-native English writers. Its modular design and dynamic parameter adjustment enable personalized learning paths while ensuring low-latency feedback and differential privacy.
\vspace{-15pt}

\paragraph{Expression \& Logical Revision} highlights AI systems’ role in refining scientific manuscripts post-initial draft, focusing on expression and logic~\citep{zheng2025usage}.
\textit{\textbf{(1) Self-guided Revision}} involves AI autonomously analyzing drafts and suggesting edits to improve language, cohesion, and structure~\citep{song2024enhancing,faruqui2018wikiatomicedits}. \citet{ito2019diamonds} propos sentence-level edits, adjusting or rewriting sentences based on draft content. Additionally, \citet{botha2018learning} use revision histories to segment and rewrite the text, further enhancing revision quality.
\textit{\textbf{(2) Human-guided Revision}} refers to interactive systems where users provide specific instructions or highlight sections for the AI to modify, forming a collaborative editing process~\citep{padmakumar2021machine,gero2022sparks,lee2022coauthor,mysore2025prototypical}. \citet{faltings2020text} develop an interactive editor that responds to user commands. Wordcraft \citep{coenen2021wordcraft} supports few-shot learning and dialogue for interaction. However, these methods struggle to capture the diversity and iterative nature of revision. XtraGPT~\citep{chen2025xtragpt} provides open-source LLMs for context-aware, instruction-guided revisions, addressing surface-level and section-level coherence.
\textit{\textbf{(3) Human-in-loop Revision}} emphasizes a cyclical workflow combining AI suggestions, human evaluation, and document updates through multiple optimization loops~\citep{feng2024cocoa,ifargan2025autonomous,tang2024step,nair2024closing}. \citet{du2022read} propose a human-in-the-loop system that integrates model-generated edits, user feedback, and document updates for high-quality revisions~\citep{tu2024augmenting}. \citet{lin2024techniques} shows that human-AI frameworks improve collaborative efficiency. \citet{wen2024overleafcopilot} develop OverleafCopilot, a browser extension integrating LLMs into Overleaf for real-time suggestions, automatic rewriting, translation, and prompt sharing through PromptGenius to enhance LaTeX writing.

\vspace{-2mm}\subsection{Full-Automatic Academic Writing}\vspace{-1mm}
\label{sec:full-automatic-academic-writing}
Full-automatic academic writing refers to using AI to generate complete scientific manuscripts without human intervention. This process spans drafting, formatting, and producing high-quality papers ready for submission, effectively removing the need for human input in manuscript preparation~\citep{wang2025scholawrite,xiong2025beyond}.
Recent research has primarily adopted multi-agent, modular designs with self-feedback mechanisms for iterative refinement. The AI Scientist \citep{lu2024ai} treats writing and reviewing as pipeline modules: by simulating peer review and providing score-based feedback, it refines drafts. \citet{yamada2025ai} extend this system by incorporating vision-language model feedback loops to improve both content and figure presentation. Agent Laboratory \citep{schmidgall2025agent,schmidgall2025agentrxiv} employs a paper-solver module with role-based agents that simulate lab workflows, evaluate drafts against NeurIPS criteria, and iteratively enhance them. Zochi \citep{zochi2025} uses a multi-agent architecture for initial draft generation, combining automated review with self-feedback for further polishing. Despite these successes, including some papers passing human peer review, no system has yet fully eliminated human editing, especially regarding correct citation use \citep{carl2025,zochi2025,schmidgall2025agent}.

\begin{figure*}[t]
	\centering
	\includegraphics[width=0.99\textwidth]{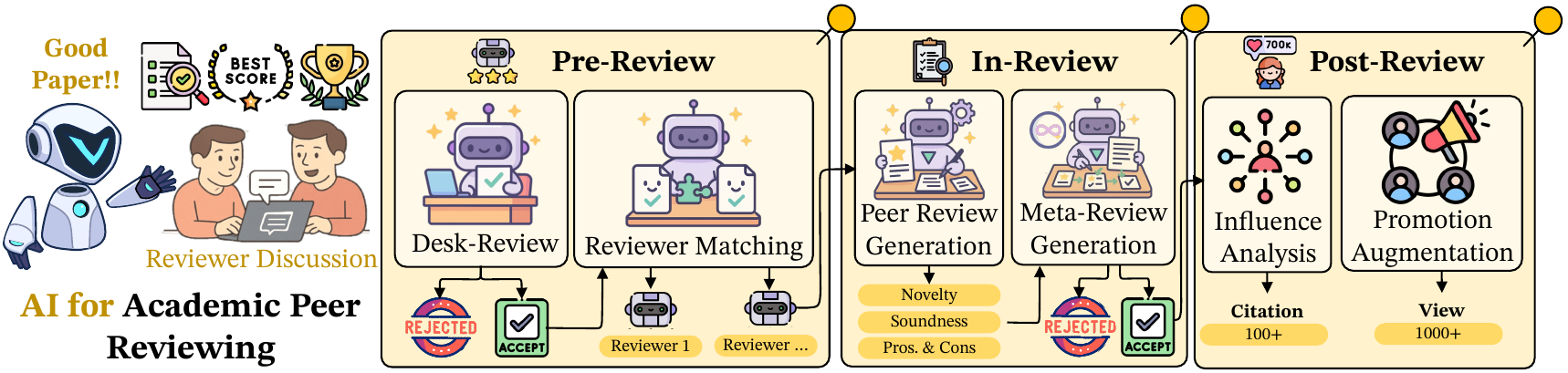}
	\caption{The primary process pipelines in AI for Academic Peer Review, encompassing three key stages: (1) Pre-Review, including Desk-Review and Reviewer Matching to ensure higher quality and more efficient evaluations; (2) In-Review, comprising Peer Review and Meta-Review, aimed at providing comprehensive scholar feedback and evaluation; and (3) Post-Review, featuring Influence Analysis and Promotion Augmentation, designed to assess the impact of the review process and improve the dissemination of scholarly work.}
	\label{fig:academic-peer-review}
\end{figure*}

%% file: sections/review.tex
\vspace{-2mm}\section{AI for Academic Peer Reviewing}\vspace{-1mm}
\label{sec:ai4reviewing}
Peer reviewing plays a crucial role in enhancing the quality of academic papers. However, it is often hindered by delays, time demands, and growing academic workloads \citep{lin2023automated,kousha2024artificial,thelwall2025evaluating,zhuang2025large}. To address these challenges and improve dissertation quality, researchers are exploring the integration of AI into the review process \citep{yuan2022can,liu2023reviewergpt,niu2023unveiling,kuznetsov2024can,machi2025framework}. As shown in Figure~\ref{fig:academic-peer-review}, it contains three main categories: Pre-Review ($\S$~\ref{sec:pre-review}), In-Review ($\S$~\ref{sec:in-review}) and Post-Review ($\S$~\ref{sec:post-review}).

\vspace{-2mm}\subsection{Pre-Review}\vspace{-1mm}
\label{sec:pre-review}
In this phase, editors or track chairs are tasked with preliminary scoring, identifying the manuscript's subject domain, and assigning appropriate reviewers to ensure review quality and prevent conflicts of interest.

\vspace{-2mm}\subsubsection{Desk-Review}\vspace{-1mm}
As manuscript submissions to academic journals increase, editorial offices face heavier workloads during the desk-review stage. To address this, major publishers have introduced AI-driven tools, such as automated keyword extraction, topic matching, and preliminary scoring, to enhance efficiency, shorten turnaround times, and reduce manual screening \citep{doskaliuk2025artificial,leyton2024matching,farber2025enhancing}. For example, Elsevier’s Evise and Editorial Manager (EM) systems use indexing and extract terms to route manuscripts to the appropriate subject areas and editorial teams \citep{tedford2015helping}. Similarly, IEEE’s Manuscript Central (built on ScholarOne) combines metadata, author-provided keywords, and an academic-network reviewer discovery tool for more accurate matching \citep{ieeecomputersociety}. Springer’s SNAPP system and Nature’s AI-assisted triage tool also demonstrate AI’s impact on desk-review workflows \citep{nature}. Additionally, \citet{diaz2024streamlining} develop AnnotateGPT, which generates annotations to help editors quickly assess a manuscript’s scope and quality, further speeding up the review process.

\vspace{-2mm}\subsubsection{Reviewer Matching}\vspace{-1mm}
Reviewer matching in peer review assigns manuscripts to experts whose knowledge aligns with the submission, aiming to maximize review quality, fairness, and workload balance~\citep{pendyala2025automated}. \citet{charlin2011framework} first formulate this as an integer program, using affinity scores to balance quality, fairness, and load. \citet{charlin2013toronto} later embed papers and reviewer profiles in a shared latent topic space, improving efficiency and accuracy in large conferences.
As submission volumes increased, automated conflict-of-interest (COI) detection became essential. \citet{wu2018pistis} introduce a semi-automated COI declaration system and a supervised ranking model to flag conflicts and ensure fairness. \citet{pradhan2020automated} further advance this with a greedy algorithm that optimizes expertise distribution and workload while maintaining COI constraints, enhancing both fairness and efficiency.
To address growing demands, \citet{leyton2024matching} develope the Large Conference Matching (LCM) algorithm, balancing expertise and load across thousands of papers. \citet{fu2025peer} and \citet{aitymbetov2025autonomous} tackle interdisciplinary submissions by forming multidisciplinary reviewer teams to improve review quality.

\vspace{-2mm}\subsection{In-Review}\vspace{-1mm}
\label{sec:in-review}
This stage involves generating or supporting review reports, either through automation or human reviewer assistance. Reviewers must assign a numerical score and provide a written evaluation. The in-review process typically involves two main stages: Peer-Review and Meta-Review.

\begin{table*}[t]
    \centering
    \resizebox{\textwidth}{!}{
        \begin{tabular}{l cccc ccc}
            \toprule
 &
            \multicolumn{4}{c}{\textbf{Focus similarity}} &
            \multicolumn{3}{c}{\textbf{Text similarity}}              \\
            \cmidrule(lr){2-5}\cmidrule(lr){6-8}
            Model & KL Divergence & Overall F1 & Strength F1  & Weakness F1
 & ROUGE-L  & BERTScore  & BLEU-4               \\ \midrule
            GPT-4o-mini~\citep{robertson2023gpt4} & 0.081 & 0.344 & 0.335 & 0.353  & 0.197 & 0.883 & 0.076 \\
            GPT-4o~\citep{robertson2023gpt4} & 0.082 & 0.348 & 0.342 & 0.354  & 0.202 & 0.885 & 0.079 \\
            o1-mini~\citep{jaech2024openai} & 0.090 & 0.359 & 0.331 & 0.385  & 0.179 & 0.878 & 0.059 \\
            o1~\citep{jaech2024openai} & 0.097 & 0.355 & 0.318 & 0.388  & 0.170 & 0.869 & 0.032 \\
            DeepSeek-R1~\citep{guo2025deepseek} & 0.120 & 0.373 & 0.341 & 0.400  & 0.156 & 0.874 & 0.045 \\
            Llama-3.1-70B~\citep{grattafiori2024llama} & 0.136 & 0.339 & 0.338 & 0.341  & 0.215 & 0.882 & 0.076 \\
            Llama-3.1-405B~\citep{grattafiori2024llama}  & 0.145 & 0.349 & 0.349 & 0.350  & 0.218 & 0.884 & 0.089 \\
            DeepSeek-V3~\citep{liu2024deepseek} & 0.151 & 0.350 & 0.330 & 0.368  & 0.199 & 0.880 & 0.069 \\
            \midrule
            GPT-4o-Finetuned~\citep{robertson2023gpt4} & 0.022 & 0.306 & 0.280 & 0.322  & 0.194 & 0.882 & 0.081 \\
            MARG~\citep{d2024marg} & 0.113 & 0.346 & \multicolumn{1}{c}{--} & 0.346  & 0.160 & 0.854 & 0.011 \\ \bottomrule
        \end{tabular}
    }
    \caption{Comparison of expert and LLM review performance sourced from \citet{shin2025mind}, where ``GPT-4o-Finetuned'' refers to GPT-4o finetuned with review data using the finetune-API. KL divergences are calculated from the average of four focus distributions (strength/target, weakness/target, strength/aspect, weakness/aspect) between expert and LLM reviews. F1 scores for overall performance, strength, and weakness are derived by comparing the (target, aspect) sets between expert and LLM reviews. Text similarity metrics are computed to assess the alignment between LLM and expert reviews. The \textbf{bolded} contents indicate the highest performance for each metric.}
    \label{tab:ai4review-results}
\end{table*}

\vspace{-2mm}\subsubsection{Peer-Review Generation}\vspace{-1mm}
Peer-review generation involves the automatic creation or assistance in the development of review comments for submitted manuscripts, including predicting quality scores and providing textual feedback. A detailed comparison of these findings is presented in Table~\ref{tab:ai4review-results}.\vspace{-15pt}

\paragraph{Score Prediction} estimates scores on criteria like innovation and clarity, assessing overall quality through multiple feature points~\citep{10.1145/3702639}. \citet{jia2021all} introduce a multi-task BERT framework that jointly detects quality features (e.g., suggestions, problem mentions) in review comments, outperforming single-task baselines. RelevAI-Reviewer \citep{couto2024relevai} treats review tasks as a classification problem to predict papers' relevance to a given call. \citet{basuki2022quality} frame score prediction as a regression task using internal paper features, excelling at distinguishing ``good'' from ``poor'' submissions. To tackle data scarcity, \citet{muangkammuen2022exploiting} improve upon this method by introducing a semi-supervised approach that fine-tunes a transformer-based model using unlabeled data, effectively utilizing contextual cues.\vspace{-15pt}

\paragraph{Comment Generation} involves generating natural-language review comments, which is the core element of manuscript evaluation~\citep{yu2024seagraph}. \citet{robertson2023gpt4} demonstrate that GPT-4 can generate plausible review comments. \citet{yuan2022kid} construct concept graphs and integrate citation mapping on a pre-trained model to generate comments. AI-Scientist \citep{lu2024ai,yamada2025ai} found that LLM-based agents approach human-level review performance~\citep{liang2024can}. MARG \citep{d2024marg} assigns paper sections to multiple LLM agents for internal discussion, improving feedback relevance. \citet{chamoun-etal-2024-automated} allocate four specialized roles to enhance specificity and comprehension. Furthermore, AgentReview \citep{jin2024agentreview} and \citet{tan2024peer} model the review process as a dynamic, multi-round dialogue.\vspace{-15pt}

\paragraph{Unified Generation} integrates textual comments and numeric scores into a single review output that mirrors real-world peer review workflows \citep{shin2025automatically,lee2025role}. There are three main paradigms for optimizing unified peer-review generation:
\textit{\textbf{(1) Single-Agent Optimization:}} A straightforward approach is to optimize a single agent through deeper analysis \citep{kang2018dataset,zhu2025deepreview,idahl2024openreviewer}. \citet{shin2025mind} observe that, by comparing the focus distributions of LLMs and human experts, off-the-shelf LLMs tend to prioritize technical validity in paper reviews while underemphasizing novelty. To address this, \citet{tyser2024ai} enhance the review system with a suite of review documents to reduce risks of misuse, score inflation, overconfident assessments, and uneven distributions. Additionally, \citet{zhu2025deepreview,zhang2025reviewing} incorporate deeper reasoning via reasoning LLMs to improve review quality.
\textit{\textbf{(2) Iterative Refinement Optimization:}} High-quality feedback is often ensured through hierarchical quality control and multi-round refinement loops \citep{bharti2021peerassist}. \citet{wu2024automated} propose an LLM-driven pipeline with hierarchical verification, producing literature surveys that match human-authored reviews. \citet{kirtani2025revieweval} introduce standardized evaluation metrics and a self-refinement cycle to align LLM-generated reviews with human accuracy and analytical depth.
\textit{\textbf{(3) Multi-Agent Optimization:}} To further enhance feedback reliability, some studies adopt multi-agent frameworks \citep{tan2024peer,huang2025papereval,gao2025reviewagents,ning2025pico,xu2023towards}. \citet{d2024marg} divide manuscripts into modules for specialized agents, leading to higher-quality feedback than single-agent systems. CycleResearcher \citep{weng2024cycleresearcher} and TreeReview \citep{chang2025treereview} apply reinforcement learning to simulate iterative review rounds and structured agent interactions, enhancing collaboration. Furthermore, \citet{taechoyotin2025remor} propose multi-objective reinforcement learning to optimize unified peer review, while \citet{taechoyotin2024mamorx} extend multi-agent scientific reviews to multimodal scenarios.

\vspace{-2mm}\subsubsection{Meta-Review Generation}\vspace{-1mm}
Meta-review generation synthesizes multiple reviewers’ opinions into a single, objective, and comprehensive critique, emphasizing the manuscript's core contributions and limitations while balancing diverse viewpoints~\citep{kumar2024towards,li2024sentiment,hossain-etal-2025-llms}.
Early studies focus on guiding the summarization process through explicit structural cues~\citep{li2023summarizing,santu2024prompting,zeng2023meta}. More recent work addresses argumentative structures and latent biases among reviewers~\citep{chen2025bridging}. Notably, PeerArg~\citep{sukpanichnant2024peerarg} introduces a Multiparty Argumentation Framework (MPAF) that combines LLMs with knowledge representation to reduce subjectivity and bias. MetaWriter~\citep{sun2024metawriter} automates the extraction of key arguments from reviewers. \citet{darrin2024glimpse} adapt the Rational Speech Act framework by creating a ``distinctiveness score'' to identify shared and unique perspectives across reviews. Moreover, \citet{kumar2023reviewers} introduce the ContraSciView corpus, which automatically detects contradictions between review pairs. Together, these efforts pave the way for more transparent and equitable meta-reviews.

\vspace{-2mm}\subsection{Post-Review}\vspace{-1mm}
\label{sec:post-review}
Post-Review refers to the suite of AI-driven methods applied after a paper has passed peer review, aiming both to assess its future scholarly impact and to broaden its dissemination. It encompasses (1) influence analysis, predicting citation trajectories and research significance from the paper’s content; and (2) promotion enhancement, automatically generating posters, lay summaries, videos, and other outreach materials to maximize visibility.
\vspace{-2mm}\subsubsection{Influence Analysis}\vspace{-1mm}

Influence analysis seeks to predict the future scholarly impact of a paper, most commonly measured by citation count, by evaluating its intrinsic characteristics~\citep{sinha2015overview,hutchins2016relative}. Early approaches predominantly rely on external metadata or handcrafted features, such as author reputation and journal impact factor~\citep{ding2011popular,tahamtan2016factors,zhu2015measuring}.
In contrast, recent methods leveraging LLMs offer the advantage of directly inferring a work’s innovativeness from its narrative. For instance, \citet{zhao2025words} frame influence prediction as a regression task, fine-tuning an LLM on titles and abstracts to generate a time- and field-normalized impact score, effectively addressing the cold-start problem. Similarly, the HLM-Cite framework \citep{hao2024hlm} adopts a two-stage approach: first, an embedding model retrieves a set of candidate citations from a large corpus, followed by a generative LLM that performs fine-grained reasoning and re-ranking to identify the most relevant references.

\vspace{-2mm}\subsubsection{Promotion Enhancement}\vspace{-1mm}
Beyond predicting impact, a parallel research strand employs generative AI to amplify a paper’s influence by producing varied, accessible promotional materials. These tools convert dense scientific manuscripts into more inviting formats, thereby broadening their reach.
For instance, \citet{sun2025p2p} present the P2P system, which automatically generates academic posters from lengthy, multimodal documents through intelligent content selection and optimized layout design.
To improve public understanding, \citet{markowitz2024complexity} leverage GPT-4 to produce lay summaries and demonstrate that these AI-generated summaries surpass human-written ones in linguistic simplicity.
More recently, \citet{park2025stealing} introduce SciTalk, a multi-agent framework that generates concise scientific videos. The rapid proliferation of such systems highlights the critical need for robust evaluation, leading to the creation of specialized metrics for assessing the quality of AI-produced scientific communications \citep{hopner2025automatic}.

%% file: sections/application.tex
\begin{figure*}[t]
	\centering
	\includegraphics[width=0.99\textwidth]{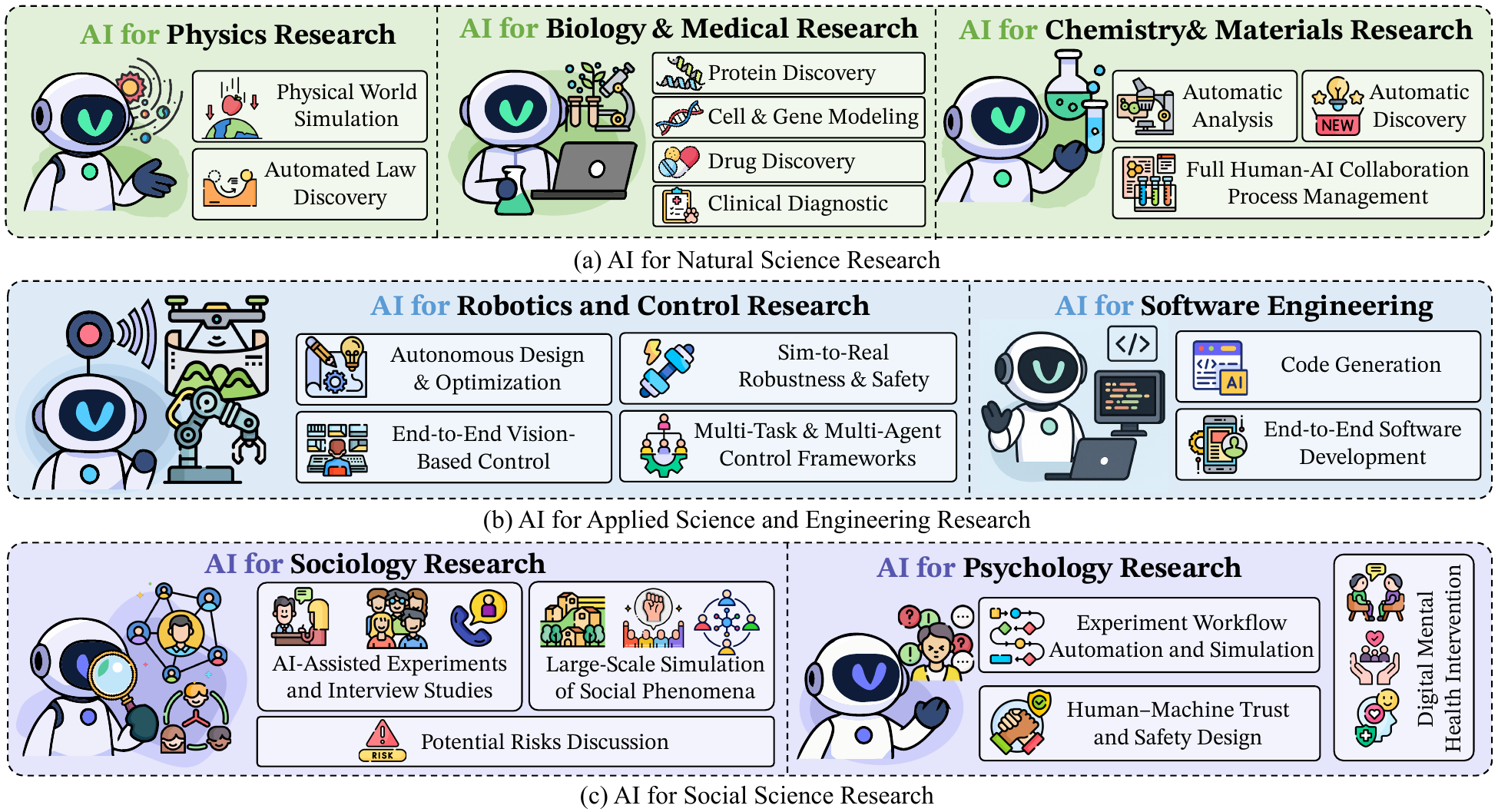}
	\caption{Multidisciplinary Applications of AI in Research. This includes three primary areas: (a) AI in Natural Sciences, covering fields such as physics, biology and medicine, and chemistry and materials science; (b) AI in Applied Sciences and Engineering, focusing on robotics and software engineering; and (c) AI in Social Sciences, encompassing disciplines such as sociology and psychology.}
	\label{fig:application}
\end{figure*}

\vspace{-2mm}\section{Application of AI for Research}\vspace{-1mm}
As shown in Figure~\ref{fig:application}, it contains three main categories: AI for Natural Science Research ($\S$~\ref{sec:ai-for-natural-science-research}), AI for Applied Science and Engineering Research ($\S$~\ref{sec:ai-for-applied-science-and-engineering-research}) and AI for Social Science Research ($\S$~\ref{sec:figure-scientific-comprehension}).

\vspace{-2mm}\subsection{AI for Natural Science Research}\vspace{-1mm}
\label{sec:ai-for-natural-science-research}
\subsubsection{AI for Physics Research}\vspace{-1mm}
In physics research, AI is now indispensable for developing new methodologies and driving discoveries~\citep{dorigo2023toward,boehnlein2022colloquium}. Its applications range from automated law discovery to physical world simulation and neural operator learning, all aimed at improving simulation accuracy, speeding up computation, and revealing hidden patterns from limited data~\citep{jiao2024ai,wang2024artificial,meng2025physics}.\vspace{-15pt}

\paragraph{Physical World Simulation} integrates physical priors with AI models to simulate complex systems while enforcing consistency with physical laws~\citep{battaglia2016interaction,de2018end,raissi2024physics}. Earlier, Physics‐Informed Neural Networks (PINNs)~\citep{raissi2019physics} embed PDE constraints in the loss function, allowing them to solve and infer nonlinear equations from sparse data. By exploiting the conserved-quantity structure of Hamiltonian mechanics, Hamiltonian Neural Networks~\citep{greydanus2019hamiltonian} exploit conserved‐quantity structures to enforce energy conservation, yielding faster convergence and drift‐free, reversible simulations. Lagrangian Neural Networks~\citep{cranmer2020lagrangian} parameterize the system’s Lagrangian directly, avoiding coordinate choices, while still preserving exact energy conservation in examples like the double pendulum.\vspace{-15pt}

\paragraph{Automated Law Discovery} leverages the reasoning power of LLMs, automated law discovery systems generate, test, and refine physical laws from noisy experimental data~\citep{shojaee2025llmsr,shojaee2025llmsrbench}. For instance, AI-Newton~\citep{fang2025ai} autonomously derives and validates physical laws, such as Newton’s laws and conservation principles, without requiring operator-provided equations. By integrating a solid knowledge base with a structured discovery workflow, AI-Newton generates interpretable models of physical phenomena. \citet{shojaee2025llmsr} propose a novel method utilizing LLMs' scientific knowledge and code-generation capabilities to discover scientific equations directly from data. The DrSR framework~\citep{wang2025drsr} enhances law discovery by analyzing structural data relationships and implementing a feedback mechanism, improving performance across various domains. LLM-Feynman~\citep{song2025llm} combines automated feature engineering, LLM-based symbolic regression, and formula interpretation to extract interpretable expressions from both empirical data and domain knowledge. Recently, \citet{li2025mllm} use enhanced visual prompting with domain expertise to uncover physical coordinates and governing equations from high-dimensional datasets more efficiently.

\vspace{-2mm}\subsubsection{AI for Biology \& Medical Research}\vspace{-1mm}
Artificial intelligence in the life sciences and medical research uses algorithms and computational models to analyze and predict across scales~\citep{lam2024large,li2025large,yang2024advancing,wang2025spatialagent,buess2025large}, from molecular structures to clinical diagnostics~\citep{wu2023can,craig-2025-human,wang2025survey}, to accelerate drug discovery~\citep{tang2024survey,othman2025advancing,guan2025ai}, optimize experimental workflows~\citep{swanson2025virtual,ifargan2025autonomous,newsham2025large}, improve diagnostic accuracy, and advance precision medicine~\citep{jiang2025fuzzy,kapitonova2024human,yang2025transforming,huang2025biomni}.\vspace{-15pt}

\paragraph{Protein Discovery.} A notable example of computational innovation is Protein Discovery and protein structure prediction, which aims to predict the three-dimensional atomic structure of proteins. This is key for understanding biological functions and guiding drug design~\citep{ding2024automating,ghafarollahi2024protagents,ghafarollahi2025sparks,pei2024leveraging}. For instance, \citet{senior2020improved} show that deep learning-based distance predictions significantly enhance de novo folding accuracy. The AlphaFold 2 system, developed by \citet{jumper2021highly}, achieves atomic-level precision and has transformed structural biology since 2021. AlphaFold 3~\citep{abramson2024accurate} builds on this by introducing a diffusion-model architecture that predicts monomeric structures and reconstructs protein-nucleic acid and protein-ligand complexes with near-experimental accuracy. Additionally, \citet{lin2025enhancing} present a dual-task LLaMA-based framework that integrates reaction and retrosynthesis into a unified recombination-fragmentation process, generating novel compounds with strong predicted protein-binding affinities through molecular docking feedback.\vspace{-15pt}

\paragraph{Cell \& Gene Modeling.} A crucial area of research is cell-level modeling and gene expression analysis, aiming to simulate cellular behavior and identify activity changes under various conditions~\citep{bunne2024build, garijo2025neurodisk, rood2024toward}. Several studies focus on pretraining models to improve cell or gene modeling~\citep{chen2024genept, bian2024general, consens2025transformers}. However, due to the scarcity of high-quality gene and cellular data, recent work has explored data augmentation techniques to enhance AI training data~\citep{afonja2024llm4grn, maleki2024efficient, agraz2024ml}. Additionally, \citet{roohani2024biodiscoveryagent} introduce an agent-based intelligent system that designs novel experiments, reasons about outcomes, and efficiently navigates hypothesis space by utilizing external tools to search the biomedical literature, analyze datasets, and engage secondary agents for evaluation, thus converging on optimal solutions. Furthermore, recent research has investigated AI-driven autonomous medical procedures, positioning AI as a collaborative tool for researchers~\citep{zou2025rise, xiao2024cellagent}. The micro-STAR system~\citep{haworth2024autonomous} integrates real-time OCT imaging with AI tissue classification to autonomously perform vascular suturing on ex vivo vessels, achieving leak-pressure performance comparable to expert surgeons, thus demonstrating the potential of AI and robotics in minimally invasive surgery.\vspace{-15pt}

\paragraph{Drug Discovery}
In recent years, artificial intelligence (AI) has made significant advancements in the field of drug discovery, driving multi-faceted innovations in drug design and showcasing immense potential and prospects~\citep{gupta2021artificial,tang2024survey,edfeldt2024data}.
\textit{\textbf{(1) Structural Prediction and Molecular Design:}} AI has made notable progress in structural prediction and molecular design. Early studies~\citep{stokes2020deep} use deep learning models to screen 23 potential antibiotic candidates from over 107 million molecules, successfully identifying a drug with antimicrobial activity. The LUMI-lab platform~\citep{cui2025lumi}, which integrates molecular models with automated experimentation, discovering ionized lipids that excel in mRNA delivery. However, challenges related to data scarcity persist in AI-driven drug discovery. To mitigate this, strategies such as multi-target drug polypharmacology, decoding drug responses, and quantum computing have been proposed to enhance model performance~\citep{gangwal2024current}.
\textit{\textbf{(2) Multi-Agent Collaborative Drug Identification:}} Multi-agent systems have proven highly effective in drug discovery, facilitating the rapid identification of new therapeutic compounds~\citep{lee2025rag}. For instance, the DrugAgent~\citep{liu2024drugagent} and  DrugPilot~\citep{li2025drugpilot} automate machine learning programming through multi-agent collaboration, achieving full automation from data acquisition to model evaluation, thus improving efficiency. \citet{solovevtowards} introduces multi-agent approach that combines LLMs with specialized generative models and validation tools to automate the end-to-end drug discovery process. \citet{lee2025rag} develop a multi-agent framework that retrieves and integrates information from biomedical knowledge bases to generate responses, avoiding the need for expensive domain-specific fine-tuning. Despite these advances, challenges remain in addressing data quality, model interpretability, and regulatory hurdles~\citep{othman2025advancing,yuan2025hallucinations}.
\textit{\textbf{(3) Drug Repurposing:}}
Drug repurposing involves the use of approved drugs for new therapeutic indications~\citep{lu2024drugclip,huang2024foundation}. In liver fibrosis research, the AI-assisted Collaborative Scientist system has successfully recommended drugs with significant anti-fibrotic activity using human liver organoid platforms, showing promise for treating liver fibrosis~\citep{guan2025ai}. By integrating knowledge graphs and diverse data sources, \citet{liu2024drugagent} and \citet{gharizadeh2024hgtdr} identify potential drug repurposing candidates and provide interpretable predictions. \citet{lee2024deep} combine subgroup analysis and treatment effect estimation, simulating clinical trials to identify drug candidates and characterize patient subgroups based on treatment effects. This approach, tested on a real-world database of more than 8 million patients, simulate over 1,000 drug trials, identifying 14 drug candidates beneficial to specific subgroups.\vspace{-15pt}

\paragraph{Clinical Diagnosis}
Clinical diagnosis advances through three converging breakthroughs. \textit{\textbf{(1) Clinical Brains:}} First, LLMs serve as clinical ``brains'', matching physician-level performance on medical licensing examinations \citep{brodsky2025generative,gottweis2025towards,khalifa2024advancing,Singhal2023,wu2023can}. Further, to enhance transparency and structure in decision-making, a human-AI note-taking framework~\citep{craig-2025-human} and Long-CoT reasoning techniques~\citep{lan2025clinicalgpt} has been proposed, leveraging case-based reasoning to guide clinical inquiry. \textit{\textbf{(2) Multi-Agent Hospital Simulation:}} Multi-agent systems such as Agent Hospital replicate AI-AI~\citep{fan2024ai,kyung2025patientsim} and human-AI~\citep{sayin2025medsyn} collaborative diagnostic and treatment workflows, effectively serving as an organizational ``nervous system'' for care coordination \citep{li2024agent,bao2024piors}. \textit{\textbf{(3) Interactive Physical Actuation:}} Robotic platforms guided by LLMs perform precise physical interventions. For example, an autonomous optical coherence tomography system delivers surgeon-level accuracy in delicate procedures such as vascular anastomosis \citep{haworth2024autonomous}.
Together, these breakthroughs demonstrate the feasibility of fully autonomous medical facilities in which artificial agents seamlessly integrate diagnostic reasoning, therapeutic planning, and procedural execution.

\vspace{-2mm}\subsubsection{AI for Chemistry \& Materials Research}\vspace{-1mm}
AI‐driven automation in chemistry~\citep{hayes2025simulating,m2024augmenting,mroz2025cross} and materials~\citep{pyzer2022accelerating,jiang2025ai4materials,choi2024accelerating} integrates machine learning, robotics, and instrumentation into a closed‐loop system for design, synthesis, and characterization, speeding decisions and experiments~\citep{chen2024nano,yuan2025empowering,wu2025literature}. \vspace{-15pt}

\paragraph{Automatic Analysis} seeks to identify optimal or novel material compositions in virtual or automated experimental setups while minimizing the number of required experiments~\citep{batista2025high,rajabi2025adaptive,seo-2025-flavordiffusion}. Specifically, \citet{chen2019graph} introduce MEGNet, demonstrating that graph neural networks can achieve density functional theory-level accuracy for both molecular and crystalline properties. \citet{li2024sequential} employ two‐stage Bayesian optimization to screen 560 organic photocatalysts using only 2.4\% of the experiment conditions, thereby obtaining significantly improved performance. \citet{ekosso2024accelerating} combine low-cost robotic platforms with high-throughput microscopy and Gaussian process models to map vesicle formation processes. More recently, \citet{szymanski2023autonomous} implement a robot-machine-learning platform that accelerate compound discovery and identify 41 novel inorganic materials within 17 days.\vspace{-15pt}

\paragraph{Automatic Discovery} is an automated experimental platform that combines robotic operations, online characterization, and real-time decision-making algorithms to autonomously execute the full experimental process, from reagent dispensing to result analysis~\citep{butler2018machine,liang2024real,darvish2025organa}. Early research laid both theoretical and practical foundations: \citet{butler2018machine} review machine learning methods that accelerate materials design and discovery~\citep{merchant2023scaling,dangayach2024machine,huang2024ai}. Furthermore, \citet{dai2024autonomous} employ a mobile robot with UPLC-MS and NMR to plan and interpret syntheses in a manner akin to human chemists. \citet{jayarathna2024experimental} leverage literature data to reduce the number of experiments in an active-learning loop, discovering new Ru-based catalysts. \citet{dai2025adaptive} introduce an AI advisor for ion-electron polymers, improving performance by 150\% over spin-coating methods. More recently, several studies have incorporated LLMs to enhance innovation and knowledge in chemistry and materials discovery~\citep{zhang5127472chatgpt,yang2024batgpt,kristiadi2024sober}.\vspace{-15pt}

\paragraph{Full Human-AI Collaboration Process Management} leverages LLMs or natural-language understanding and generation to support hypothesis formulation, experimental design, and iterative optimization, aiming to facilitate more intuitive and efficient research interactions~\citep{wang2024efficient,luo2025physics,leng2025intelligent}. The AILA framework~\citep{mandal2024autonomous} embeds LLMs within a fully automated atomic-force microscopy workflow, illustrating both the potential and current constraints of language models in guiding real-time microscopic experiments. \citet{sprueill2024chemreasoner} and \citet{feng2025agentic} integrate LLM-driven linguistic reasoning with chemical feedback in a heuristic search loop to propose novel catalysts and reaction pathways within uncertain chemical spaces. Recognizing that collective intelligence often outperforms individual reasoning, several studies~\citep{ni2024matpilot,song2025multiagent,lai2025prim} employ multi-agent architectures in which LLMs generate hypotheses, design experiments and direct iterative optimization, thereby achieving seamless human-AI collaboration. \citet{ma2025automated} introduce the first fully automated retrosynthetic planning agent tailored for LLM-driven macromolecular design, enabling comprehensive enumeration of viable multi-branch reaction routes. Meanwhile, \citet{zhu2024automated} demonstrate a robotic AI chemist that performs ore pretreatment and catalyst optimization on Martian meteorite samples.

\vspace{-2mm}\subsection{AI for Applied Science and Engineering Research}
\label{sec:ai-for-applied-science-and-engineering-research}
\vspace{-1mm}\subsubsection{AI for Robotics and Control Research}\vspace{-1mm}
AI for Robotics and Control Research applies AI methods: deep learning, reinforcement learning, and large language models, to the perception, decision‐making, and control of robots, aiming to boost adaptability, robustness, and autonomy in novel environments~\citep{lee2025generative,lange2025ai}.\vspace{-15pt}

\paragraph{Autonomous Design \& Optimization} systems integrate robotics, machine learning, and domain expertise to automate experiment planning, execution, and optimization~\citep{liang2024real}. \citet{uddin2025ai} introduce OptoMate, a system using a fine-tuned language model for optical setup design and a robotic arm to assemble spectroscopy components with submillimeter precision, enabling cloud-based optical labs. \citet{mieszczanek2024towards} employ computer vision and feedforward neural networks in a feedback optimizer to adjust 3D printing parameters in real-time, reducing data collection from days to hours and ensuring consistent part quality. \citet{angello2024closed} apply physically informed feature selection and supervised learning in a closed-loop system to enhance photostability and uncover solvent-mediated triplet-state mechanisms. \citet{bu2024closed} combine a text-conditioned video diffusion model with a feedback-driven controller to generate visual plans and iteratively refine actions, significantly improving performance.\vspace{-15pt}

\paragraph{End-to-End Vision-Based Control}
End‐to‐end vision‐based control feeds raw images or video frames directly into a neural network to generate control signals, removing separate perception, planning, and control modules. Early work combine convolutional neural networks (CNNs) with reinforcement learning or guided policy search to map camera inputs to motion commands. \citet{levine2016end} first apply Guided Policy Search to train perception and control jointly, mapping raw images to motor torques and demonstrating reliable real‐world grasping. \citet{levine2018learning} train a CNN on massive real grasp attempts to predict grasp success in real time, closing the loop on novel objects. \citet{kalashnikov2018scalable} introduce a self‐supervised, closed‐loop Q‐learning framework trained on 580,000 grasps, enabling dynamic strategy adjustment, retries, and disturbance resilience. \citet{tobin2017domain} propose domain randomization, varying simulator rendering parameters so models trained on synthetic data transfer directly to real‐world detection and grasping.\vspace{-15pt}

\paragraph{Sim-to-Real Robustness \& Safety} ensures reliable transfer of simulation-trained policies to real-world tasks while adhering to safety constraints. \citet{bochem2024improving} integrate sharpness-aware optimization into gradient-based RL to identify flat minima, enhancing transfer robustness in contact-rich tasks without compromising sample efficiency. \citet{ayabe2024robustness} assess offline RL methods on a legged robot subjected to random and adversarial torque disturbances, revealing vulnerability to sudden perturbations and emphasizing the need for real-time adaptation and safety measures. \citet{radosavovic2024real} train a Transformer-based controller using deep RL and deploy it outdoors for one week without safety scaffolding, showcasing adaptive performance amidst disturbances, rugged terrain, and varying payloads. \citet{yang2025zero} apply domain randomization for vision-based servoing of soft robots, eliminating the need for on-robot fine-tuning and enabling direct transfer of simulation-trained models to continuum manipulators. \citet{guerrier2025guided} combine control barrier functions with RL to enforce safety constraints during learning, preventing hazardous states in complex environments.\vspace{-15pt}

\paragraph{Multi-Task \& Multi-Agent Control Frameworks} facilitate concurrent task execution or enable collaboration among agents in complex workflows, thereby enhancing parallelism and automation. \citet{tahmid2025value} introduce a reinforcement learning-based framework designed to dynamically learn and compose task policies in robotic systems with redundant architectures, incorporating time-varying priority stacks to adjust task priorities. \citet{team2025novelseek} propose a unified multi-agent system capable of automatically generating hypotheses, designing and conducting experiments, and refining methods through iterative feedback, establishing a closed-loop process that accelerates interdisciplinary research.

\vspace{-2mm}\subsubsection{AI for Software Engineering}\vspace{-1mm}
AI for Software Engineering Research focuses on applying AI techniques to automate software development tasks, enhance code quality, and improve developer productivity. This includes code generation, bug detection, code review, and software testing.\vspace{-15pt}

\paragraph{Code Generation} refers to the use of AI models to automatically generate code snippets or entire programs from natural language descriptions or existing code patterns~\citep{roziere2023code, guo2024deepseek, li2023starcoder,huang2025mldebugging}. This can accelerate the development process and reduce manual coding~\citep{lozhkov2024starcoder, zhang2025seed}. For instance, \citet{chen2021evaluating} develop Codex, a GPT model fine-tuned on GitHub’s publicly available code, which supports GitHub Copilot. To democratize program synthesis, \citet{nijkamp2022codegen} train and release CodeGen, an LLM based on both natural and programming language data, alongside the open-source training library JAXFORMER. Additionally, several studies have explored advanced code capabilities and support for multiple programming languages, like Python, Java, and R~\citep{roziere2023code, guo2024deepseek, li2023starcoder}.\vspace{-15pt}

\paragraph{End-to-End Software Development} covers the entire software development lifecycle, with AI automating various stages~\citep{ozkaya2023application, fan2023large, jimenez2024swebench, le2024repoexec}. For example, \citet{phan2024hyperagent} develop HyperAgent, a generalist multi-agent system designed to handle various SE tasks across different programming languages, mimicking human developers’ workflows. \citet{qian2023experiential} introduce Experiential Co-Learning, which enables software development agents to leverage historical experiences to improve task performance. Meanwhile, \citet{qian2023chatdev} introduce ChatDev, a chat-based framework for software development, and \citet{kang2025explainable} present an explainable automated debugging framework powered by LLM-driven scientific debugging.

\vspace{-2mm}\subsection{AI for Social Science Research}\vspace{-1mm}
\label{sec:figure-scientific-comprehension}
AI has been widely utilized to automate the design, execution, and analysis of social science experiments, encompassing tasks from hypothesis generation to data collection, with minimal human intervention. In this context, we will focus on two key domains:

\vspace{-2mm}\subsubsection{AI for Sociology Research}\vspace{-1mm}
AI in sociology research refers to the use of machine learning, natural language processing, and multi-agent systems to simulate, analyze, and explore social phenomena~\citep{loffredo2025agent,xia2025reimagining,karjus2025machine}. Through AI, researchers can reconstruct macro-level patterns of collective behavior and gain deeper insights into micro-level cultural contexts and individual interactions, thereby revitalizing traditional sociological methods~\citep{li2024ethnography}.\vspace{-15pt}

\paragraph{AI-Assisted Experimental and Interview Studies.}
Controlled experiments and simulated interviews are increasingly employed by scholars to test social science hypotheses and evaluate the effects of various social mechanisms and policy interventions. \citet{manning2024automated} propose a methodology that combines structural causal models with large language models to automatically generate and empirically validate social science hypotheses in contexts such as negotiations, bail hearings, job interviews, and auctions. This approach effectively bridges the gap between theory and practice by utilizing the model both as a scientific tool for hypothesis generation and as an experimental subject for validation. \citet{liu2024step} develop MimiTalk, an automated interview system, and conduct a comparative study of AI-led and human-led interviews with 20 participants on the Prolific platform. This study demonstrates the feasibility of AI-mediated interviews and highlights their potential in experimental settings.\vspace{-15pt}

\paragraph{Large-Scale Simulation of Social Phenomena.}
This approach leverages algorithmic tools to automate the collection and analysis of extensive textual, visual, and interaction data to simulate and examine the macro-level dynamics of community practices and value evolution~\cite{hoffmann2024malinowski,sreedhar2025simulating,huang2024adasociety}. \citet{perez2024cultural} automate the extraction and analysis of large-scale text and image datasets to map cultural practices, value systems, and trends in contemporary online communities~\citep{yu2024researchtown,sheldon2024economic}. \citet{zamudio2024raise} propose a simulation framework for cultural evolution using multi-agent LLMs, enabling the manipulation of network structures, individual traits, and biases in information transmission to investigate factors driving cultural diffusion and change. \citet{chen2025predicting} develop a GPT-based three-module framework, including information extraction, variant generation, and outcome prediction, that achieved high consistency in predicting outcomes across 319 economic field experiments, while also reflecting the impact of gender, race, and social norms on performance. Additionally, \citet{bao2025language} reveal the underlying, often unspoken codes within societies.\vspace{-15pt}

\paragraph{Potential Risks Discussion.}
While LLMs demonstrate strong predictive capabilities in the natural sciences, their performance in the social sciences remains limited. \citet{manning2024automated} find that, although LLMs can predict the signs of estimated effects well when given a proposed structural causal model, they struggle to predict the magnitudes reliably. Additionally, \citet{predicting2024} highlight that LLMs face challenges in handling treatment effect heterogeneity and exhibit systematic biases when predicting social science outcomes. As such, LLMs' predictive capabilities are still underdeveloped, particularly in forecasting novel empirical patterns that could inform future experimentation \citep{lehr2024chatgpt}.

\vspace{-2mm}\subsubsection{AI for Psychology Research}\vspace{-1mm}
Research methodology focuses on the design, implementation, and validation of psychological experiments to ensure validity and reproducibility \citep{tong2024automating,li2024can,guozhen2024human}.\vspace{-15pt}

\paragraph{Experiment Workflow Automation and Simulation.}
Recent research has explored integrating AI into the management and data simulation of psychology experiments~\citep{qiu2024interactive,wang2024clientcenteredassessmentllmtherapists,Binz_2023}. \citet{zamudio2024raise} introduce the RAISE pipeline, automating the generation and validation of visual stimuli. In five experiments, AI-generated images match researcher-designed stimuli in both validity and recognizability. \citet{cingillioglu2024ai} conduct a fully automated online RCT with 1,193 participants, where AI managed recruitment, random assignment, intervention delivery, and data collection, successfully replicating eight classical hypotheses with gold-standard rigor. \citet{cui2024can,strachan2024theory,suri2023largelanguagemodelsdecision} use GPT-4 to simulate responses for 154 classical experiments, reproducing 76\% of primary effects but yielding 71.6\% unexpected significant outcomes, illustrating the promise of AI-assisted replication while emphasizing the need for cautious interpretation~\citep{chen2023emergenceeconomicrationalitygpt,DILLION2023597,goli2024can}.\vspace{-15pt}

\paragraph{Human-AI Trust and Safety Design.}
Research on human-AI trust explores the development of trust during human-AI interactions and derives strategies for ensuring safety~\citep{qin2023mmsd2}. \citet{li2024developing} introduce a three-dimensional framework encompassing the trustor, trustee, and context, identifying key factors influencing trust and offering design recommendations for improving safety. Building on this, \citet{chandra2024lived}, through interviews with 283 individuals who have mental health experiences, develop a taxonomy comprising 19 risky AI behaviors and 21 negative psychological impacts. From this, they propose a multi-path case-method framework and a set of safety guidelines aimed at mitigating these risks.\vspace{-15pt}

\paragraph{Psychological Interventions.}
Psychological interventions increasingly employ AI-driven chatbots to provide scalable and cost-effective psychological support\citep{PRAVEEN2024103975}. Earlier, \citet{Hagendorff_2023}, \citet{DILLION2023597}, and \citet{Binz_2023} discuss whether and when LLMs can replace human participants in psychological research, reviewing early evidence and proposing a theoretical framework while highlighting methodological caveats. In a randomized controlled trial, \citet{heinz2025randomized} find that the Therabot chatbot resulted in significant reductions in clinical-level symptoms compared to the control group. Similarly, \citet{spytska2025use} explore the use of the Friend chatbot for crisis support, demonstrating that its efficacy is comparable to traditional face-to-face therapy. These findings highlight the potential of generative AI to enhance accessibility and effectiveness in mental health services, particularly in settings with limited resources~\citep{dekok2025chatgpt,deiner2024llmthematic}. Recent work further demonstrates that LLMs can match or exceed human performance in generating emotionally resonant narratives~\citep{doi:10.1177/07439156241297973,rousmaniere2025large} and even pass standard Turing tests~\citep{jones2025large}, underscoring their broader psychological and communicative capabilities.

%% file: sections/resources.tex
\vspace{-2mm}\section{Resources}\vspace{-1mm}
To further advance research in this field, we will provide an expanded and more comprehensive suite of relevant resources, including tools, benchmarks, and datasets spanning all stages.
\subsection{AI for Scientific Comprehension}\vspace{-1mm}

\begin{table*}[t]
    \centering
    \setlength{\tabcolsep}{15pt}
    \resizebox{0.98\textwidth}{!}{
        \begin{tabular}{ll}
            \toprule
            \textbf{Tool}                                                          & \textbf{Description}                                                            \\
            \midrule
            \href{https://scispace.com}{SciSpace Copilot}                          & AI-powered Literature Q\&A, Annotations, Auto-Summarization, Chart Explanations \\
            \href{https://elicit.org}{Elicit}                                      & AI-powered Literature Q\&A, Auto-Summarization, Suggestions                     \\
            \href{https://jenni.ai}{Jenni AI} / \href{https://notegpt.io}{NoteGPT} & AI-powered Note-Taking, Auto-Summarization                                      \\
            \href{https://www.scholarcy.com}{Scholarcy}                            & AI-powered Auto-Summarization, Summarization Card, Analysis and Organization    \\
            \href{https://github.com/Byaidu/PDFMathTranslate}{PDFMathTranslate}    & AI-powered PDF Math-Augmented Translation                                       \\
            \bottomrule
        \end{tabular}
    }
    \caption{Representative and established AI systems and assistant tools for advancing scientific comprehension.}
    \label{tab:scientific-comprehension-research-tool}
\end{table*}

\subsubsection{Textual Scientific Comprehension}\vspace{-1mm}
To advance the evaluation of scientific question-answering systems, various benchmarks have been developed with increasing task complexity and domain specificity~\citep{rostam2024fine,yang2024graphusion,feng2024sciknoweval,he2025pasa}. Table~\ref{tab:scientific-comprehension-research-tool} presents a comprehensive overview of typical, mature AI systems and associated tools for scientific comprehension.

Datasets like ScienceQA~\citep{saikh2022scienceqa}, LitQA~\citep{lala2023paperqa}, LitQA2~\citep{skarlinski2024language}, SciQA~\citep{auer2023sciqa}, SciQAG-24D~\citep{wan2024sciqag}, and TriviaQA~\citep{kuhn2022clam} support QA for scientific content. SciBench~\citep{wang2023scibench} broadens scientific reasoning across physics, chemistry, and mathematics. Further, SciInstruct~\citep{zhang2024sciglm} broadens reasoning across formal proofs with instruction-tuned data~\citep{wang2023enabling,wadden2024sciriff}. Moreover, AutoPaperBench~\citep{kim2025autopaperbench} and SciCUEval~\citep{yu2025scicueval} are proposed for automatic paper or scientific content understanding evaluation.

Furthermore, datasets have expanded into broader domains, including biomedicine~\citep{bolton2024biomedlm,jin2019pubmedqa,krithara2023bioasq,raza2022coquad,pal2022medmcqa,lin2024biokgbench,zhao2025biomaze}, academic chemistry~\citep{chen2024scholarchemqa,peretz2023if}, materials science~\citep{luo2024generating}, physics~\citep{zheng2025scaling} and other scientific fields \citep{xu2025earthse}. TheoremQA~\citep{chen2023theoremqa} evaluates AI models' ability to apply theorems to solve challenging science problems.
Multi-task and multi-modal assessment frameworks, such as M3CoT~\citep{chen2024m}, SciFIBench~\citep{roberts2024scifibench}, MMSCI~\citep{li2024mmsci}, SPIQA \citep{pramanickspiqa}, and MultimodalArxiv~\citep{li-etal-2024-multimodal-arxiv}, further extend these evaluations. To address broader multimodal and multi-document challenges, M3SciQA~\citep{li-etal-2024-m3sciqa}, SceMQA~\citep{liang-etal-2024-scemqa} and SciDQA~\citep{singh2024scidqa} have also been introduced.

Beyond static benchmarks, dynamic and interactive evaluation frameworks have emerged. SCITOOLBENCH~\citep{ma-etal-2024-sciagent} target tool use in scientific reasoning across domains, while \citet{kuhn2022clam} propose a multi-round dialog framework to simulate user interactions, introducing metrics like adjusted accuracy. In terms of generation alignment, \citet{yu2025frame} design a system to evaluate the semantic fidelity between generated content and scientific texts, combining automated scores with human judgment.

\vspace{-2mm}\subsubsection{Table \& Chart Scientific Comprehension}\vspace{-1mm}
In the domain of reasoning based on charts and tables, a number of benchmarks have emerged to evaluate the ability of LLMs in both structural and logical comprehension. Early works, such as ChartQA~\citep{masry-etal-2022-chartqa}, CharXiv~\citep{wang2024charxiv}, ChartX~\citep{xia2024chartx}, MISSQA~\citep{zhao-etal-2025-multimodal-foundation}, and NovaChart \citep{hu2024novachart}, focus on assessing LLMs' performance in answering questions related to charts, utilizing both synthetic and real-world data. On the other hand, benchmarks like SUC~\citep{si2024table}, TableBench~\citep{wu2025tablebench}, and ToRR~\citep{ashury2025mighty} emphasize the evaluation of LLMs' structural understanding, rather than content comprehension, across various tasks such as form interpretation, numerical reasoning, and textual analysis.

\begin{table*}[t]
    \centering
    \resizebox{\textwidth}{!}{
        \begin{tabular}{ll}
            \toprule
            \textbf{Tool}                                                                                                                                                                                         & \textbf{Description}                                                     \\
            \midrule
            \href{https://scholar.google.com}{Google Scholar} / \href{https://www.webofscience.com}{Web of Science} / \href{https://www.scopus.com}{Scopus} / \href{https://www.aminer.cn/}{AMiner}               & Literature Search, Citation Tracking, Author Profiles, Citation Analysis \\
            \href{https://www.semanticscholar.org}{Semantic Scholar}                                                                                                                                              & AI-Assisted Academic Search Platform (Semantic Graph)                    \\
            \href{https://www.researchrabbit.ai}{Research Rabbit} / \href{https://www.connectedpapers.com}{Connected Papers} / \href{https://citationgecko.com}{Citation Gecko} / \href{https://iris.ai}{Iris.ai} & Visual graph of Works                                                    \\
            \href{https://scite.ai}{Scite.ai}                                                                                                                                                                     & Shows Citation Context (Supporting/Contradicting/Neutral)                \\
            \href{https://consensus.app}{Consensus.app}                                                                                                                                                           & Opinion-based Literature Search, Ideal for YES/NO Questions              \\
            \href{https://researchgpt.com}{ResearchGPT}                                                                                                                                                           & AI-generated Knowledge Graphs and Paper Structures                       \\
            \bottomrule
        \end{tabular}
    }
    \caption{Representative AI systems and assistive technologies that have been widely adopted to enhance academic surveys.}
    \label{tab:survey-tools}
\end{table*}

\vspace{-2mm}\subsection{AI for Academic Survey}\vspace{-1mm}
In the task of generating academic surveys, several representative public corpora stand out due to their distinct characteristics in terms of scale, domain, and structure. Table~\ref{tab:survey-tools} presents a comprehensive overview of typical, mature AI systems and associated tools for academic survey collection and generation.

To evaluate the capabilities of scholarly retrieval systems, several studies have focused on the scholarly deep research~\citep{wei2025browsecomp}, such as AcademicBrowse~\citep{zhou2025academicbrowse}. To facilitate section-level generation of related works, several large-scale datasets have been introduced, such as Cochrane~\cite{wallace2021generating}, MSLR 2022~\cite{wang2022overview}, MS$^2$~\cite{deyoung2021ms2}, and OARelatedWork~\citep{docekal2024oarelatedwork}, OAG-Bench~\citep{zhang2024oag}. These datasets pair comprehensive ``Related Work'' sections with their corresponding full texts, providing valuable resources for this task.
Moreover, systematic benchmarks for evaluating the quality of automatic scholarly survey generation have been developed. Examples include SciReviewGen~\cite{kasanishi2023scireviewgen}, BigSurvey~\cite{liu2023generating}, SurveySum~\cite{fernandes2024surveysum}, SurveyBench~\cite{yan2025surveyforge}, AutoSurvey~\cite{wang2024autosurvey}, and SurveyX~\cite{liang2025surveyx}. These benchmarks provide critical metrics for assessing the performance of automatic systems in generating academic surveys.
For fine-grained manual annotation, SurveyEval~\cite{wang2025llm} offers a hierarchical title tree that includes a vast number of reviews and citations. It is accompanied by hierarchical consistency and citation-chapter alignment metrics, which serve as essential tools for evaluating the distribution of synopsis generation and citation accuracy.

\vspace{-2mm}\subsection{AI for Scientific Discovery}\vspace{-1mm}
To facilitate scientific discovery, a variety of AI datasets and tools have been developed to assist researchers in different stages of the research process. These tools range from experiment design and management to full-automatic discovery, as shown in Table~\ref{tab:discovery-tools}.

\paragraph{Idea Mining} has seen the introduction of several key resources that significantly contribute to scientific discovery tasks~\citep{chen2025structuring}. Notably, LiveIdeaBench~\citep{ruan2024liveideabench}, ResearchBench~\citep{liu2025researchbench}, Genome-Bench~\citep{yin2025genome}, AIIdeaBench2025~\citep{qiu2025ai}, the AP-FRI Corpus~\citep{kumar2024can}, HypoGen~\citep{o2025sparks}, CLIMATEDATABANK~\citep{liu2025improving}, CHIMERA~\citep{sternlicht2025chimera} and OMATO-Chem~\citep{yanglarge} provide structured datasets for idea mining and hypothesis generation, enabling systematic training and evaluation of LLMs. Furthermore, OAG-Bench~\citep{zhang2024oag} offers a comprehensive, fine-grained benchmark for academic graph mining, spanning 10 tasks, 20 datasets, and over 70 baseline methods, all curated by human experts. This resource fosters systematic evaluation and encourages community-driven research. Additionally, SPARK~\citep{sanyal2025spark} and the ICLR-NeurIPS Ideas Dataset~\citep{li2024learning} introduce curated datasets of idea-centered abstract-review pairs from OpenReview submissions, supporting supervised and reinforcement learning for research idea generation with multi-dimensional idea quality control, including novelty, feasibility, and effectiveness.\vspace{-15pt}

\begin{table*}[t]
    \centering
    \resizebox{\textwidth}{!}{
        \begin{tabular}{ll}
            \toprule
            \textbf{Tool}                                                                                                                                                                 & \textbf{Description}                                                              \\
            \midrule
            \rowcolor{gray!10} \multicolumn{2}{c}{\textit{Experiment Design}}                                                                                                                                                                                                 \\
            \midrule
            \href{https://www.snapgene.com}{SnapGene}                                                                                                                                     & Molecular Cloning and DNA Visualization                                           \\
            \href{https://elicit.org}{Elicit}                                                                                                                                             & AI-driven Experiments Design                                                      \\
            \midrule
            \rowcolor{gray!10} \multicolumn{2}{c}{\textit{Experiment Management}}                                                                                                                                                                                             \\
            \midrule
            \href{https://www.notion.so}{Notion} / \href{https://asana.com}{Asana} / \href{https://clickup.com}{ClickUp} / \href{https://www.atlassian.com/software/rovo}{Atlassian Rovo} & Project Management, Task Tracking, Collaboration                                  \\
            \href{https://trello.com}{Trello} / \href{https://www.wrike.com}{Wrike}                                                                                                       & Generate Content, Help Brainstorming, Conceive Product                            \\
            \href{https://gitmind.com}{GitMind}                                                                                                                                           & Mind Mapping and Brainstorming Tool for Project Planning                          \\
            \href{https://www.forecast.app}{Forecast}                                                                                                                                     & Project Risk and Status Management                                                \\
            \href{https://www.tableau.com}{Tableau} / \href{https://powerbi.microsoft.com}{Power BI}                                                                                      & Interactive Dashboards and Reports                                                \\
            \midrule
            \rowcolor{gray!10} \multicolumn{2}{c}{\textit{Experiment Conduction}}                                                                                                                                                                                             \\
            \midrule
            \href{https://github.com/features/copilot}{Copilot} / \href{https://www.cursor.so}{Cursor} / \href{https://www.tabnine.com}{Tabnine} / \href{https://www.qodo.ai}{Qodo}       & AI-powered Code Completion, Generation, Review, Documentation                     \\
            \href{https://ai.google.dev}{Gemini CLI}                                                                                                                                      & AI-powered Open Source Command Line Tool                                          \\
            \href{https://www.diffblue.com}{Diffblue}                                                                                                                                     & AI-powered Unit Test Generation for Java Code                                     \\
            \href{https://mlflow.org}{MLflow} / \href{https://wandb.ai}{Weights \& Biases} / \href{https://www.tensorflow.org/tensorboard}{TensorBoard}                                   & AI Experiment Tracking And Visualization                                          \\
            \href{https://paperswithcode.com}{Papers with Code}                                                                                                                           & Paper-Code Pairs for Easier Reproducibility                                       \\
            \href{https://github.com/Project-MONAI/MONAI}{MONAI}                                                                                                                          & AI-powered Medical Imaging Framework for Reproducibility                          \\
            \href{https://www.taskade.com}{Taskade}                                                                                                                                       & Generate Code Snippets and Debugging to Facilitate Collaboration Among Developers \\
            \midrule
            \rowcolor{gray!10} \multicolumn{2}{c}{\textit{Full-Automatic Discovery}}                                                                                                                                                                                          \\
            \midrule
            \href{https://chat.openai.com}{ChatGPT} / \href{https://claude.ai}{Claude} / \href{https://gemini.google.com}{Gemini}                                                         & Problem Solving, Code Assistance, Writing Polishing, Full-Lifecycle Management    \\
            \href{https://researchgpt.com}{ResearchGPT}                                                                                                                                   & AI-generated Knowledge Graphs and Paper Structures                                \\
            \href{https://github.com/Torantulino/Auto-GPT}{AutoGPT} / \href{https://github.com/OpenDevin/OpenDevin}{OpenDevin}                                                            & Multi-step Research Automation                                                    \\
            \href{https://www.agentlabs.dev}{AgentLabs}                                                                                                                                   & Multi-Agent AI Platform for Research Automation                                   \\
            \href{https://github.com/SakanaAI/AI-Scientist}{AI-Scientist} / \href{https://www.intology.ai}{Zochi}                                                                         & AI-powered Research Assistant for Scientific Discovery                            \\
            \bottomrule
        \end{tabular}
    }
    \caption{Representative AI tools and their role in facilitating scientific discovery, with a particular focus on experiment conduction.}
    \label{tab:discovery-tools}
\end{table*}

\paragraph{Novelty \& Significant Assesment}
In the development of automated scientific research evaluation, the academic community has primarily focused on the dual criteria of ``novelty and significance'', systematically exploring the ability of language models to assess the innovation within scientific research~\citep{tan2025hierarchical}. The SchNovel framework~\citep{lin-etal-2025-evaluating} and NoveltyDetection~\citep{liu2025harnessing} are introduced to evaluate AI systems' capacity to assess scholarly novelty across multiple scientific disciplines sampled from arXiv, aiming to facilitate the automated evaluation of research originality in scientific workflows. Building upon this, \citet{gu2024blade} introduce BLADE, a system that integrates 12 expert-labeled datasets with multiple automated scoring methods, enabling the model to explore diverse inference strategies in open, data-driven scientific analysis. HypoBench~\citep{liu2025hypobench}, \citet{dasguptaempowering} and \citet{linllms} are designed to evaluate LLMs and hypothesis generation methods across multiple aspects, including practical utility, generalizability, hypothesis, novelty \citep{dasguptaempowering} and rigor \citep{linllms} discovery rate.\vspace{-15pt}

\paragraph{Theory Analysis} requires the collection of scientific evidence, theoretical verification, and theorem proving. Specifically, FV-Generalization Benchmark~\citep{pan2023investigating}, SCitance \citep{alvarez2024zero}, and MissciPlus \citep{glockner2024grounding} are designed to complete scientific evidence collection. TheoremExplainBench \citep{ku2025theoremexplainagent} , XClaimCheck \citep{kao2024magic}, SciNews \citep{cao2024can}, ClaimReview2024+ \citep{braun2024defame}, FactKG \citep{kim2023factkg}, TrendFact \citep{zhang2024augmenting}, and SciVer~\citep{wang-etal-2025-sciver} provide datasets and benchmarks for scientific verification analysis. MiniF2F~\citep{zheng2021minif2f}, FIMO~\citep{liu2023fimo}, MUSTARDSAUCE \citep{huang2024mustard} are used to fine-tune and evaluate LLMs on scientific theorem proving tasks. Furthermore, datasets and benchmarks have expanded into broader domains, including biomedicine \citep{wang2025biodsa}.\vspace{-15pt}

\paragraph{Experiment Design}
In terms of experimental design, \citet{tian2024benchmarking} propose evaluation frameworks for zero-shot and few-shot scenarios within virtual screening and lead compound optimization, establishing a comprehensive set of metrics tailored for AI-driven drug discovery. Concurrently, \citet{feng2024bioactivity} leverage 1.6 million bioactivity measurements to train a universal model using pairwise meta-learning, which facilitates rapid adaptation and robust generalization to new biological systems. For biological protocol understanding and reasoning, BioProBench~\citep{liu2025bioprobench} is the first large-scale, multi-task benchmark. In order to provide valuable insights for the safe and effective deployment of LLMs in medical domains, \citet{zhang2025llmeval} develop LLMEval-Med, a real-world clinical benchmark for medical LLMs with physician validation.\vspace{-15pt}

\paragraph{Experiment Conduction}
To evaluate model performance in realistic research environments, MLAgentBench~\citep{huang2023mlagentbench}, Exp-Bench~\citep{kon2025exp},  MLRC-Bench~\citep{zhang2025mlrc}, MLE-Bench \citep{chan2024mle}, DS-Bench~\citep{jing2024dsbench}, ScienceBoard~\citep{sun2025scienceboard}, ScienceArena~\citep{zhao2025sciarena}, AutoReproduce~\citep{zhao2025autoreproduce}, SciReplicate-Bench~\citep{xiang2025scireplicate}, DO Challenge~\citep{smbatyan2025can}, and MLR-Bench~\citep{chen2025mlr}  assess AI agents' abilities to perform typical research tasks, such as optimizing CIFAR-10 classifiers and tuning BabyLM. In a similar vein, \citet{hu2024infiagent} develop InfiAgent-DABench, which is based on real-world CSV datasets and evaluates models' ability to interact with tools in end-to-end data analysis tasks. MLGym-Bench~\citep{nathani2025mlgym} is the first Gym environment for machine learning tasks, enabling research on reinforcement learning algorithms for training such agents. ResearchCodeBench \citep{hua2025researchcodebench} enables continuous understanding and advancement of LLM-driven innovation in research code generation. AutoBio~\citep{lan2025autobio} is designed to evaluate robotic automation in biology laboratory environments.\vspace{-15pt}

\paragraph{Experimental Analysis}
Experimental Analysis involves systematically testing hypotheses, evaluating models, or validating theoretical assumptions to draw meaningful conclusions. MicroVQA~\citep{burgess2025microvqa} is proposed  to assess three reasoning capabilities vital in research workflows: expert image understanding, hypothesis generation, and experiment proposal.\vspace{-15pt}

\paragraph{Full Automatic Discovery}
In recent years, benchmark suites have been developed to assess AI-driven research agents. These suites offer standardized datasets, predefined tasks, and evaluation metrics, thereby facilitating systematic advances in algorithmic optimization. They encompass multi-domain scenarios spanning chemical synthesis, materials discovery, and biological experimentation~\citep{gu2024blade,guo2024ds,liu2025vision}. Notable examples include ScienceAgentBench~\citep{chen2024scienceagentbench}, BaisBench~\citep{luo2025benchmarking}, Curie~\citep{kon2025curie}, which are designed to evaluate AI scientists’ abilities to generate novel discoveries in different disciplines through data analysis and reasoning with external knowledge~\citep{smbatyan2025can,shojaee2025llm,kao2025towards}. And DiscoveryWorld \citep{jansen2024discoveryworld} evaluates end-to-end scientific discovery agents, while DiscoveryBench~\citep{majumder2024discoverybench} challenges large language models with 264 real-world and 903 synthetic tasks across six domains, using structured protocols to measure multi-step, data-driven discovery and to elucidate both capabilities and failure modes.

\begin{table*}[t]
    \centering
    \resizebox{0.96\textwidth}{!}{
        \begin{tabular}{ll}
            \toprule
            \textbf{Tool}                                                                                                                                                                                                                                                               & \textbf{Description}                                      \\
            \midrule
            \href{https://endnote.com}{EndNote} / \href{https://www.mendeley.com}{Mendeley plugins}                                                                                                                                                                                     & Reference Insertion and Auto-Formatting                   \\
            \href{https://mathpix.com}{Mathpix Snip} / \href{https://mathhandwrit.ing}{MathHandwriting}                                                                                                                                                                                 & AI-powered Math Equation Recognition and LaTeX Conversion \\
            \href{https://github.com/eseckel/ai-for-grant-writing}{AI for Grant Writing}                                                                                                                                                                                                & AI-powered Grant Writing Assistance                       \\
            \href{https://writefull.com}{Writefull} / \href{https://www.trinka.ai}{Trinka} / \href{https://www.grammarly.com}{Grammarly AI} / \href{https://www.paperpal.com}{Paperpal} / \href{https://www.overleaf.com}{Overleaf Copilot} / \href{https://www.wordtune.com}{Wordtune} & AI-powered Scientific English Polishing Tools             \\
            \href{https://typeset.io/copilot}{SciSpace Copilot} / \href{https://jenni.ai}{Jenni AI}                                                                                                                                                                                     & AI Writing Assistants for Editing and Suggestions         \\
            \href{https://chat.openai.com}{ChatGPT} / \href{https://claude.ai}{Claude} / \href{https://gemini.google.com}{Gemini}                                                                                                                                                       & Writing Inspiration, Summarization, Editing               \\
            \href{https://chat.openai.com}{GPT-4o} / \href{https://www.vizcom.ai/}{Vizcom}  / \href{https://illustrae.co}{Illustrae} / \href{https://openart.ai}{OpenArt}                                                                                                               & AI-powered Figure Generation and Illustration Tools       \\

            \bottomrule
        \end{tabular}
    }
    \caption{An overview of representative AI tools and their contributions to enhancing academic writing.}
    \label{tab:writing-tools}
\end{table*}

\vspace{-2mm}\subsection{AI for Academic Writing}\vspace{-1mm}
The field of AI in academic writing is supported by a comprehensive array of meticulously curated datasets that address various aspects of the academic writing process. Detailed tools are summarized in Table~\ref{tab:writing-tools}, which provides an overview of representative AI systems and their contributions to enhancing academic writing.

\subsubsection{Semi-Automatic Academic Writing}\vspace{-1mm}
Recent advancements in semi-automatic academic writing have led to the development of several datasets designed to assist researchers in different stages of manuscript preparation, writing, and editing.\vspace{-15pt}

\paragraph{Assistance During Manuscript Preparation.}
In the early stages of manuscript preparation, recent datasets such as MoDeST~\citep{BOLUCU2025113557} and LLM-Rubric~\citep{hashemi-etal-2024-llm} offer valuable tools for generating multi-domain scientific titles and assessing the scientific idea generation capabilities of LLMs.\vspace{-15pt}

\paragraph{Assistance During Manuscript Writing}
Several datasets, including FigGen~\citep{rodriguez2023figgen}, SridBench~\citep{chang2025sridbench}, Figuring out Figures~\citep{cao2024figuring}, SciCapenter~\citep{hsu2024scicapenter}, and TikZero~\citep{belouadi2025tikzero}, support figure and formula generation, from text-to-figure creation to automated TikZ code generation. For citation management, datasets like CITEWORTH~\citep{wright2021citeworth}, CiteBART~\citep{ccelik2024citebart}, and ScholarCopilot~\citep{wang2025scholarcopilot} enhance context-aware automatic citation generation~\citep{chen2025xtragpt}. Additionally, FutureGen~\citep{azher2025futuregen} extracts future work statements from thousands of papers, using LLMs to identify and validate forward-looking scientific content.\vspace{-15pt}

\paragraph{Assistance After Manuscript Completion.}
Once a manuscript is completed, further enhancement can be achieved through grammar correction and expression optimization. To support this process, datasets from the Automated Writing Evaluation (AWE) system~\citep{wang2024neural}, and AAAR-1~\citep{lou2024aaar} provide valuable resources. Additionally, the transformer-based Feedback Dataset~\citep{zheng2025usage} offers comprehensive support for multi-dimensional writing quality assessment~\citep{ito2019diamonds,faruqui2018wikiatomicedits}.
Moreover, datasets such as Wikipedia Revision Histories~\citep{botha2018learning}, which track real-world editing histories, play an important role in refining language and improving overall clarity. \citet{pang2025paper2poster} introduce the first benchmark and metric suite for poster generation for visual quality, coherence-language fluency, and the ability to convey core paper content.

\vspace{-2mm}\subsection{AI for Academic Peer Reviewing}\vspace{-1mm}
Research on AI for Academic Peer Reviewing is grounded in diverse datasets, addressing tasks from AI text detection to review generation, quality assessment, and decision support~\citep{zhou2024llm,farber2024enhancing,rao2024withdrarxiv,chu2024pre,chu2024automatic,staudinger2024analysis,baumgartner2025peerqa}. To simulate realistic peer review interactions, datasets such as ReviewCritique~\citep{du-etal-2024-llms}, PeerRead~\citep{kang2018dataset}, SPOT~\citep{son2025ai}, NLPeer~\citep{dycke-etal-2023-nlpeer}, ReviewMT~\citep{tan2024peer}, MOPRD~\citep{lin2023moprd}, OpenReviewer~\citep{idahl2024openreviewer}, MASSW~\citep{zhang2024massw}, COMPARE~\citep{singh2021compare} , PeerArg~\citep{sukpanichnant2024peerarg}, Re$^2$~\citep{zhang2025re}, ReviewEval~\citep{kirtani2025revieweval}, AAAR-1~\citep{lou2024aaar}, Papereval~\citep{huang2025papereval}, ORB~\citep{szumega2023open} and ORSUM~\citep{zeng2024scientific} collect extensive paper and review data from leading conferences or journals, enabling the training and evaluation of LLMs in multi-turn, long-context, or role-based peer reviews~\citep{plank2019citetracked}. Additionally, LLMart~\citep{liang2024can} offers a toolkit for evaluating LLM robustness through adversarial testing and prompt optimization, ensuring AI reliability in sensitive academic contexts. To assess LLM review quality more precisely, \citet{shin2025mind} and \citet{couto2024relevai} analyze the quality of peer review content across multiple predefined aspects, highlighting discrepancies between LLM and human review focus.

In the field of review quality detection, researchers have investigated diverse quality features~\citep{ghosal2022peer}. \citet{purkayastha2025lazyreview}, and PolitePEER~\citep{bharti2024politepeer} assess AI systems' ability to identify instances of ``lazy thinking'' or politeness in peer reviews. Furthermore, both the AI-Peer-Review-Detection-Benchmark~\citep{yu2024your} and TRIED~\citep{liu2023reviewergpt}  include thousands of AI-generated peer reviews alongside human-authored reviews from the ICLR and NeurIPS conferences. These datasets provide standard corpora essential for evaluating methods designed to detect AI-generated peer reviews.

%% file: sections/future.tex
\vspace{-2mm}\section{Frontiers \& Future Direction}\vspace{-1mm}
\begin{figure*}[t]
	\centering
	\includegraphics[width=0.99\textwidth]{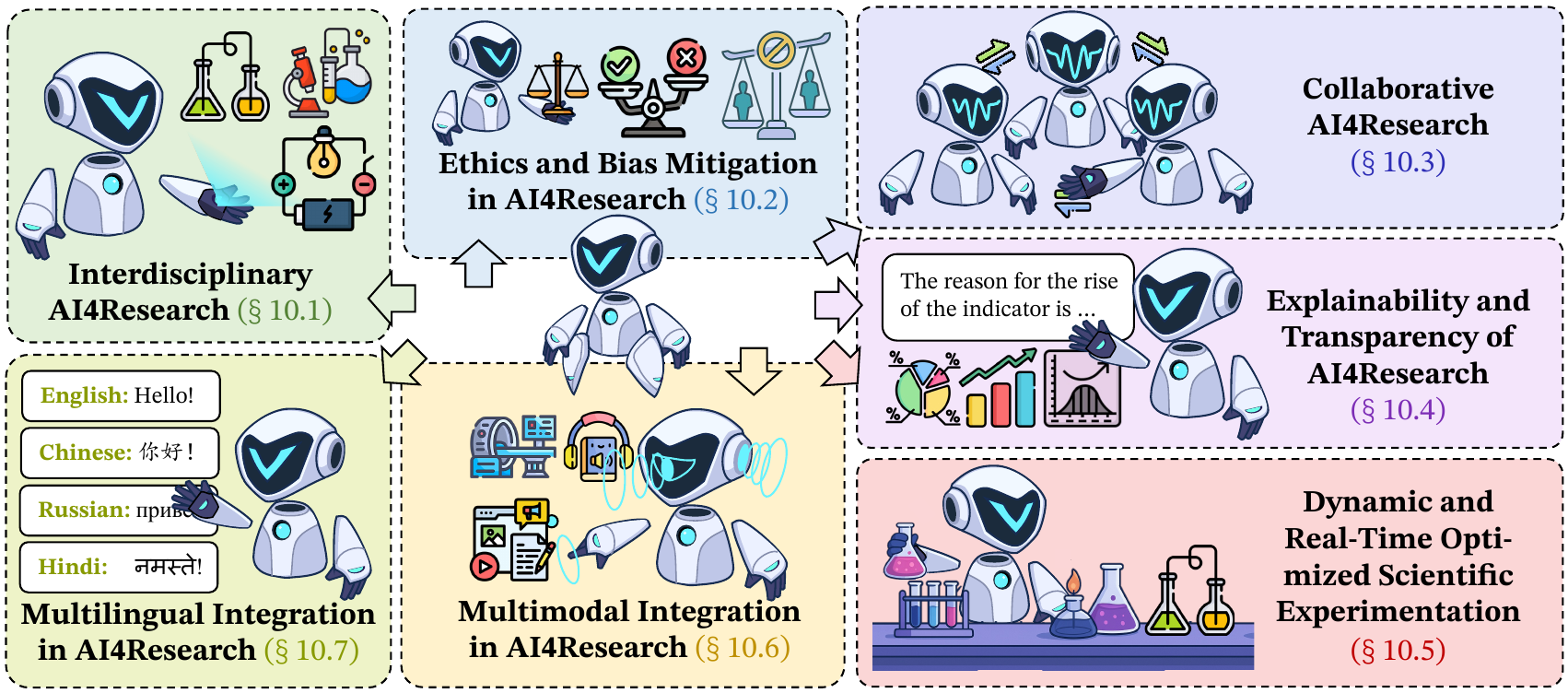}
	\caption{Frontiers and Future Directions of Artificial Intelligence in Research: This includes (1) Interdisciplinary AI models, (2) Ethics and Safety in AI4Research, (3) AI for Collaborative Research, (4) Explainability and Transparency of AI4Research, (5) Dynamic and Real-Time Optimized Scientific Experimentation, (6) Multimodal Integration in AI4Research, and (7) Multilingual Integration in AI4Research.
	}
	\label{fig:future-work}
\end{figure*}
\subsection{Interdisciplinary AI Models}\vspace{-1mm}
As AI advances across research domains, we need models that integrate knowledge from multiple fields. Future work should develop general-purpose AI systems able to understand and generate insights in biology, physics, social sciences, and beyond.
The primary research directions are:
\textit{\textbf{(1) Foundation Models.}} This paradigm has become the cornerstone of cross-domain AI. These models are pretrained self-supervised on vast unlabeled or weakly labeled datasets, then fine-tuned on new tasks with minimal data. They have driven performance gains in medical imaging, natural language processing, and robotics~\citep{huang2025foundation,khan2025comprehensive}.
\textit{\textbf{(2) Graph Models.}} Graph methods naturally handle relational data by propagating information along nodes and edges. This enables cross-field knowledge flow, e.g., integrating ontologies and neural graphs in medical text classification for precise concept capture and efficient inference~\citep{fan2022generating,buehler2024accelerating,gu2024generation,lin2024biokgbench,lan2023contrastive}.

The greatest challenges at present are:
\textit{\textbf{(1) Heterogeneous Interdisciplinary Data.}} Interdisciplinary research involves diverse modalities, from high-dimensional sensor signals to categorical labels and unstructured text. These sources vary in scale, noise characteristics, and missing-data patterns, hindering unified preprocessing and feature fusion~\citep{putrama2024heterogeneous,ye2023heterogeneous,cirillo2021artificial}.
\textit{\textbf{(2) Cross-Domain Knowledge Transfer.}} Transferring knowledge across domains requires extracting and adapting relevant information for new tasks. Techniques such as policy transfer, domain-adversarial training, and semantic alignment can narrow some gaps, but negative transfer persists in highly heterogeneous settings~\citep{niu2024comprehensive,serrano2024knowledge}. Moreover, preserving reliability and interpretability during transfer, to ensure more applicable and trustworthy application in novel contexts, remains an urgent open problem~\citep{zhu2024survey}.

\vspace{-2mm}\subsection{Ethics and Safety in AI4Research}\vspace{-1mm}
As AI assumes a central role in scientific research, a range of ethical, safety, fairness, and bias concerns has emerged~\citep{ferrara2023fairness,ye2024we,glickman2025human,huang2025survey}, making mitigation essential~\citep{alvarez2024policy,gonzalez2024mitigating,xiong2025toward,pan2025hidden,cheng2025ai}. Early work by \citet{farber2024enhancing} shows that, while AI improves reviewer matching and response rates, it disadvantages authors in low-resource languages or on niche topics. Worse, text-similarity matching can be abused by collusive rings to manipulate peer review, underscoring the need for built-in anti-collusion measures~\citep{raghunathan2024vulnerability,hsieh2024automated}. Moreover, \citet{mcshane2025artificial} find that AI-assisted statistical interpreters fall prey to ``dichotomous mania'', reducing results to simply significant or not, a flaw that prompt-engineering alone has not resolved~\citep{rao2025detecting}.
There are two main mitigation strategies:
\textit{\textbf{(1) Fairness-Aware Training:}} Integrate fairness constraints into the loss function to balance accuracy and equity across groups~\citep{ferrara2023fairness,hanna2025ethical}. Causal-inference methods then detect and adjust for hidden biases, enabling counterfactual fairness interventions~\citep{cheng2021causal,mensah2023artificial}.
\textit{\textbf{(2) Training-free Debiasing:}} Without retraining, apply unsupervised pruning and reweighting to model outputs at regular intervals, correcting biases in large language models by leveraging their pretrained behaviors~\citep{ebrahimi2024axolotl,rosbach2025automation,cheng2025biasfilter}.
\textit{\textbf{(3) Establishing Ethical Framework:}} Some studies are establishing benchmarks for professional and broad ethical frameworks to regulate AI-generated content in a controlled area through security risk and ethical issue monitoring \citep{zhu2025safescientist,spotte2025considering}.

Nonetheless, these endeavors confront two core challenges: \textit{\textbf{(1) Balancing performance and fairness:}} The inherent tension between maximizing predictive accuracy and enforcing fairness constraints is difficult to reconcile and typically demands meticulous, application‐specific tuning to avoid degrading model utility~\citep{herron2024scitrust,sun2025openreview}.
\textit{\textbf{(2) Avoiding AI Plagiarism:}} 
A major ethical concern in AI-driven scientific research is plagiarism~\citep{lin2024beyond, ganguly2025generative,wheeler2025responsible}. Large-scale text generation by LLMs could lead to a ``plagiarism singularity'', where text originality is diminished, raising concerns about the ethical and copyright risks of AI-generated content~\citep{ranga2025plagiarism}. Studies have also revealed significant instances of intelligent plagiarism in LLM-generated scientific literature~\citep{gupta2025all}.

\vspace{-2mm}\subsection{AI for Collaborative Research}\vspace{-1mm}
As interdisciplinary research advances, the diversity of team members’ backgrounds can impede information flow and decision coordination. AI techniques can automatically extract and synchronize cross-document and cross-domain information, thereby narrowing the information gap among collaborators~\citep{blaurock2024designing,schleiger2024collaborative}. Simultaneously, AI-driven arbitrators within real-time collaboration platforms can adjust task allocation dynamically based on project progress and member expertise, improving both efficiency and the quality of innovative outcomes~\citep{lu2024ai,fehlis2025accelerating,hosseini2025role}.
The main research directions can be broadly divided into two categories:
\textit{\textbf{(1) Collaborative Agents and Cooperative Intelligent Systems.}}
Collaborative agents are AI systems endowed with decision-making, autonomous execution, and communication capabilities. They simulate and augment human collaborators by participating in complex project management and research workflows through task assignment and autonomous role switching~\citep{hosseini2025role,hu2025text2world}. Through semantic retrieval, reasoning validation, and context awareness, multi-agent frameworks are creating a new paradigm of human-AI collective intelligence, enabling automated hypothesis generation, experimental design planning, and preliminary results analysis to accelerate scientific discovery~\citep{zimmermann202534,li2025drugpilot,sreedhar2025simulating}. These advances support efficient human–AI hybrid teams and suggest fertile directions for further work on collaborative agents and distributed modeling.
\textit{\textbf{(2) Federated Learning and Distributed Modeling Mechanisms.}}
Because sensitive data across institutions cannot be fully shared, recent research has adopted federated learning as a privacy-preserving distributed modeling approach. By training models collectively while keeping data local, federated learning mitigates data silos among institutions and specialist teams~\citep{li2020review,zhang2021survey,kuo2025distributed}. To enhance both performance and privacy guarantees, differential privacy, and homomorphic encryption are being integrated with federated optimization algorithms, offering scalability and regulatory compliance for large-scale, multi-scenario collaborative research~\citep{truex2019hybrid}.

Current challenges in this field include:
\textit{\textbf{(1) Interaction Complexity.}}
Repeated task reassignments, control handovers, and heterogeneous communication modalities can lead to misunderstandings, inefficiencies, and compounded coordination errors~\citep{gomez2025human,holter2024deconstructing}. Addressing this issue requires adaptive collaboration mechanisms that allow AI systems to adjust their behavior dynamically to match human collaborators’ working styles and decision-making preferences. Multi-intelligence relationships are also critical, with three failure modes of miscoordination, conflict, and collusion \citep{hammond2025multi}.
\textit{\textbf{(2) Tension between Data Privacy and Accessibility.}}
A fundamental tension exists between data privacy and accessibility: stringent anonymization or legal restrictions often reduce the quality and diversity of training data. Although anonymization techniques and compliance with regulations  protect privacy, they can diminish data utility and hinder AI models from capturing representative features, thereby affecting the accuracy and credibility of interdisciplinary research~\citep{myakala2024federated}. Moreover, differences in data access permissions, network bandwidth, and legal frameworks across institutions can cause communication delays and inconsistent model updates during distributed training, undermining the efficiency and stability of federated learning~\citep{guendouzi2023systematic}.

\vspace{-2mm}\subsection{Explainability and Transparency of AI4Research}\vspace{-1mm}
As AI models increasingly drive scientific discovery, ensuring their trustworthiness, transparency and explainability is essential. Future work should strengthen model interpretability so that researchers can trace how conclusions and recommendations are generated, particularly in high-stakes scientific applications~\citep{frasca2024explainable,ehsan2024explainable,wang2025medcite}.
Efforts to improve explainability fall into two main categories: \textit{\textbf{(1) White-box Analysis:}} This approach investigates the model’s internal structure by linking specific network ``circuits'' to conceptual representations. It has attracted considerable interest from both the security and transparency communities~\citep{bereska2024mechanistic,zhao2024towards,rai2024practical}. \textit{\textbf{(2) Black-box Analysis:}} More recent work focuses on interpreting models without direct access to internal parameters. By examining reasoning trajectories and aggregate behavior, black-box methods provide insights into a model’s knowledge representation and enable more reliable control over its outputs~\citep{cai2023gradient,hassija2024interpreting,chen2024unlocking,chen2025rbf++,chen2025ecm}.

Despite these advances, two principal challenges remain: \textit{\textbf{(1) Lack of Standardized Frameworks:}}  Explanation techniques and metrics vary widely across the AI4Research community. Such absence  can produce conflicting results and undermine user confidence.\textit{\textbf{(2) Transparency–Performance Trade-off:}} Highly capable black-box models often sacrifice interpretability, whereas intrinsically transparent models may lag in performance. This tension complicates scientific adoption and raises uncertainty about whether novel outputs represent genuine discoveries or the recombination of existing data \citep{lin2024beyond}.

\vspace{-2mm}\subsection{AI for Dynamic and Real‑Time Optimized Scientific Experimentation}\vspace{-1mm}
Real-time AI models can automatically adjust experimental protocols in response to unforeseen variables or shifting conditions, while performing immediate data analysis to substantially enhance research efficiency and innovative potential. Two prominent research directions have emerged:
\textit{\textbf{(1) Agentic Real-Time AI:}} This approach advances AI beyond passive data analysis, transforming it into an autonomous research optimizing agent endowed with reasoning, planning, and decision-making capabilities based on real-time experimental feedback. Such agents can systematically survey the literature, generate hypotheses, design experiments, and iteratively refine workflows based on experimental feedback \citep{liang2024real,desai2025autoscilab,dai2025adaptive}.
\textit{\textbf{(2) Coordination in self-driving laboratories:}} These systems integrate robotic platforms, analytical instruments, and AI models into closed-loop frameworks that manage every stage, from experimental planning and execution to data processing. They support applications such as compound screening and novel materials discovery based on real-time signals with minimal human intervention \citep{tom2024self,canty2025science,lo2024review}.

Despite these advances, two core challenges must be addressed before dynamic, real-time AI experiments become routine:
\textit{\textbf{(1) Reliable integration of heterogeneous devices and AI systems:}} Laboratory environments comprise diverse instruments and robotic platforms requiring precise, real-time control and feedback. Systems must ensure compatibility, robustness, and low latency to avoid deviations or downtime caused by integration failures or timing mismatches.
\textit{\textbf{(2) Low-latency decision-making and dynamic optimization:}} AI-driven experiments must continuously ingest multisensor and instrument data on the millisecond to second timescales, update model parameters in real-time, and adjust protocols dynamically to maintain workflow continuity and efficiency. Simultaneously, they must uphold robustness and safety to prevent interruptions or hazards due to network jitter or computational bottlenecks~\citep{hu2023tree,hu2024hiagent}.

\vspace{-2mm}\subsection{Multimodal Integration in AI4Research}\vspace{-1mm}
As scientific data become more diverse, encompassing text, figures, tables, code snippets, and experimental signals, effective multimodal integration has emerged as a linchpin for AI‐driven discovery~\citep{chen2024m,wang2023t,wang2024s3,rodriguez2025bigdocs,cheng2025comt}. Early work \citep{gomez2019look,chen2020uniter,maruf2024vlm4bio,shi2024every} show that jointly embedding text and figures can substantially boost deep analysis and literature-based discovery, yet this approach often falters when aligning highly specialized diagrams with their textual descriptions~\citep{cheng2025visual,qin2024factors}.
There are two main integration strategies:
\textit{\textbf{(1) Rigorous Multi-Source Data Ingestion:}} Scientific datasets span manuscripts, high-resolution images, time-series signals, code artifacts, and structured tables. Each modality requires tailored preprocessing, such as OCR for figures, noise filtering for sensor data, syntax checking for code, to preserve integrity and alignment with domain ontologies \citep{gokdemir2025hiperrag,leong2025mermaid}.
\textit{\textbf{(2) Interactive Human-in-the-Loop Refinement:}} Unlike general-purpose systems, research workflows integrate expert feedback at multiple stages. Interactive interfaces enable domain scientists to validate figure captions, correct table alignments, or adjust the experimental setting based on the multi-modal signals, creating an iterative loop that refines model outputs and builds trust~\citep{singh2023figcaps,wang2024model,zhao2025interfeedback}.

Nonetheless, multimodal integration in AI4Research faces two core challenges:
\textit{\textbf{(1) Scarcity of cross-modal data and annotation bottleneck:}}
High-quality aligned annotations are exceedingly scarce, particularly in specialized scientific domains where expert involvement is required for fine-grained pairing, leading to a dramatic escalation of training and evaluation costs.
\textit{\textbf{(2) Quantification of inter-modal uncertainty:}}
Data originating from diverse sources contain heterogeneous noise; how to uniformly quantify and propagate this uncertainty to support reliable scientific decision-making remains an open challenge.

\vspace{-2mm}\subsection{Multilingual Integration in AI4Research}\vspace{-1mm}
Scientific research transcends linguistic and geographic borders. Global initiatives, such as COVID-19 containment and climate modeling, depend on integrating literature, datasets, and expert insights across diverse languages efficiently. If AI tools favor only English or other high-resource languages, research sharing suffers, reinforcing ``information silos'' and the ``knowledge divide''~\citep{amano2021ten,amano2016languages}.
Most researchers’ native languages lie in the ``long tail'' of AI systems. Neglecting low-resource languages limits discoverability and citation of high-quality studies and sidelines region-specific topics (e.g., tropical agriculture, minority health). Multilingual pre-training and data augmentation can generate accurate summaries, retrievals, and translations in low-resource languages, breaking down academic barriers~\citep{conneau2019unsupervised,hangya2022improving,chua2024crosslingual,qin2025survey,pradier2025smack}.
There are two principal integration strategies:
\textit{\textbf{(1) Alignment of Scientific Terminology:}} Reproducibility demands consistent terms and semantic fidelity. Multilingual terminology alignment and contextual‐fidelity techniques ensure accurate translation of experiments and publications, so researchers worldwide build on a common knowledge base~\citep{sabet2020simalign,zheng2023hit,frontull2024rule}.
\textit{\textbf{(2) Equilibrating Multilingual Performance:}} Data imbalances between high- and low-resource languages hinder cross-lingual transfer. Equalizing performance across languages enhances zero-shot and few-shot capabilities in research applications~\citep{chua2024crosslingual,qin2025survey,wang2025x,zhang2024autocap}.

Nonetheless, multilingual integration in AI4Research faces two core challenges:
\textit{\textbf{(1) Balancing Capacity and Coverage:}} Under finite computational and parameter budgets, striking the right balance between supporting core research capabilities and maintaining broad multilingual performance is critical to prevent ``language breadth'' from sacrificing ``research depth''. This requires fine‐grained architectural pruning and resource allocation tailored to specific domains and language pairs.
\textit{\textbf{(2) Analysis of Cross-Lingual Academic Rhetorical Fidelity:}} Ensuring that conceptual meanings remain consistent across different languages, preserving the logical integrity of academic argumentation in translation, and addressing language‐specific academic conventions constitute important directions for future research.

%% file: sections/related_work.tex
\vspace{-2mm}\section{Related work}\vspace{-1mm}

Recent years have seen increasing interest in AI-assisted or autonomous research across multiple research communities~\citep{barman2025large}. The empirical use of large language models (LLMs) in research workflows indicates that most researchers are incorporating these models into their processes~\citep{liao2024llms}. Additionally, \citet{yu2025unlocking} survey and predict the rise in AI4Science publications, suggesting strategies to empower AI researchers.
Early survey~\citep{agrawal2024artificial,reddy2025towards,zhang2024comprehensive,trifonov2025ai,zhang2025advancing} summarize how LLMs are transforming scientific discovery~\citep{eger2025transforming, zheng2025automation, gridach2025agentic,zhu2025ai}. \citet{li2025review} focus more on the ideation developments for LLM-assisted ideation, while \citet{kulkarni2025scientific} and \citet{ren2025towards} summarize the architectures and benchmarks for LLM-driven discovery methods. \citet{chen2025ai} propose the Science-of-Science framework, which surveys the AI4Science in multi-agent simulation perspective.
Meanwhile, \citet{huang2025towards} describe the AI-driven scientific discovery process from the perspective of the hypothesis lifecycle~\citep{ismayilzada2024creativity}. In particular, \citet{zhou2024hypothesis} and \citet{luo2025llm4sr} develop a three-stage taxonomy to systematically review assistance role in each phase. Building on this framework, \citet{alkan2025survey} and \citet{bazgir2025agentichypothesis} offer a comprehensive classification of LLM-based hypothesis generation methods.
In response to the peer-review crisis, \citet{kim2025position} focus more on the bidirectional feedback system with certified reviewers, while \citet{bolanos2024artificial} and \citet{zhuang2025large} review the rise of automated scientific paper reviews, which coexist with human oversight.

Although significant advancements have been made in AI4Research, much of the existing survey has focused primarily on scientific discovery and academic writing, often under the umbrella of AI4Science or the limited research stages. However, these discussions typically overlook the broader research lifecycle, including scientific comprehension, academic survey, and peer review. Additionally, they tend to neglect AI applications across these stages. This paper introduces the AI4Research framework and offers a systematic survey of key factors and recent developments driving AI-enabled research. Our goal is to provide the research community with streamlined access to essential resources and insights, thereby facilitating innovative breakthroughs.

%% file: sections/conclusion.tex
\vspace{-2mm}\section{Conclusion}\vspace{-1mm}
In conclusion, rapid advancements in artificial intelligence, particularly large language models like OpenAI-o1 and DeepSeek-R1, have demonstrated substantial potential in areas such as logical reasoning and experimental coding. These developments have sparked increasing interest in applying AI to scientific research. However, despite the growing potential of AI in this domain, there is a lack of comprehensive surveys that consolidate current knowledge, hindering further progress. This paper addresses this gap by providing a detailed survey and unified framework for AI4Research. Our contributions include a systematic taxonomy for classifying AI4Research tasks, identification of key research gaps and future directions, and a compilation of open-source resources to support the community. We believe this work will enhance our understanding of AI’s role in research and serve as a catalyst for future advancements in the field.

%% file: main.bbl
\begin{thebibliography}{960}
\providecommand{\natexlab}[1]{#1}
\providecommand{\url}[1]{\texttt{#1}}
\expandafter\ifx\csname urlstyle\endcsname\relax
  \providecommand{\doi}[1]{doi: #1}\else
  \providecommand{\doi}{doi: \begingroup \urlstyle{rm}\Url}\fi

\bibitem[Abdin et~al.(2024)Abdin, Aneja, Behl, Bubeck, Eldan, Gunasekar, Harrison, Hewett, Javaheripi, Kauffmann, et~al.]{abdin2024phi}
Marah Abdin, Jyoti Aneja, Harkirat Behl, S{\'e}bastien Bubeck, Ronen Eldan, Suriya Gunasekar, Michael Harrison, Russell~J Hewett, Mojan Javaheripi, Piero Kauffmann, et~al.
\newblock Phi-4 technical report.
\newblock \emph{arXiv preprint arXiv:2412.08905}, 2024.

\bibitem[Abramson et~al.(2024)Abramson, Adler, Dunger, Evans, Green, Pritzel, Ronneberger, Willmore, Ballard, Bambrick, et~al.]{abramson2024accurate}
Josh Abramson, Jonas Adler, Jack Dunger, Richard Evans, Tim Green, Alexander Pritzel, Olaf Ronneberger, Lindsay Willmore, Andrew~J Ballard, Joshua Bambrick, et~al.
\newblock Accurate structure prediction of biomolecular interactions with alphafold 3.
\newblock \emph{Nature}, 630\penalty0 (8016):\penalty0 493--500, May 2024.

\bibitem[Achiam et~al.(2023)Achiam, Adler, Agarwal, Ahmad, Akkaya, Aleman, Almeida, Altenschmidt, Altman, Anadkat, et~al.]{achiam2023gpt}
Josh Achiam, Steven Adler, Sandhini Agarwal, Lama Ahmad, Ilge Akkaya, Florencia~Leoni Aleman, Diogo Almeida, Janko Altenschmidt, Sam Altman, Shyamal Anadkat, et~al.
\newblock Gpt-4 technical report.
\newblock \emph{arXiv preprint arXiv:2303.08774}, 2023.

\bibitem[Achkar et~al.(2025)Achkar, Gollub, and Potthast]{achkar2025ask}
Pierre Achkar, Tim Gollub, and Martin Potthast.
\newblock Ask, retrieve, summarize: A modular pipeline for scientific literature summarization.
\newblock \emph{arXiv preprint arXiv:2505.16349}, 2025.

\bibitem[Adriaensen et~al.(2023)Adriaensen, Rakotoarison, M{\"u}ller, and Hutter]{adriaensen2023efficient}
Steven Adriaensen, Herilalaina Rakotoarison, Samuel M{\"u}ller, and Frank Hutter.
\newblock Efficient bayesian learning curve extrapolation using prior-data fitted networks.
\newblock \emph{Advances in Neural Information Processing Systems}, 36:\penalty0 19858--19886, Dec 2023.

\bibitem[Afonja et~al.(2024)Afonja, Sheth, Binkyte, Hanif, Ulas, Becker, and Fritz]{afonja2024llm4grn}
Tejumade Afonja, Ivaxi Sheth, Ruta Binkyte, Waqar Hanif, Thomas Ulas, Matthias Becker, and Mario Fritz.
\newblock Llm4grn: Discovering causal gene regulatory networks with llms--evaluation through synthetic data generation.
\newblock \emph{arXiv preprint arXiv:2410.15828}, 2024.

\bibitem[Agarwal et~al.(2024)Agarwal, Sahu, Puri, Laradji, Dvijotham, Stanley, Charlin, and Pal]{agarwal2024litllm}
Shubham Agarwal, Gaurav Sahu, Abhay Puri, Issam~H Laradji, Krishnamurthy~DJ Dvijotham, Jason Stanley, Laurent Charlin, and Christopher Pal.
\newblock Litllm: A toolkit for scientific literature review.
\newblock \emph{arXiv preprint arXiv:2402.01788}, 2024.

\bibitem[Agarwal et~al.(2024{\natexlab{2}})Agarwal, Sahu, Puri, Laradji, Dvijotham, Stanley, Charlin, and Pal]{agarwal2024llms}
Shubham Agarwal, Gaurav Sahu, Abhay Puri, Issam~H Laradji, Krishnamurthy~DJ Dvijotham, Jason Stanley, Laurent Charlin, and Christopher Pal.
\newblock Llms for literature review: Are we there yet?
\newblock \emph{arXiv preprint arXiv:2412.15249}, 2024{\natexlab{2}}.

\bibitem[Agarwal et~al.(2025)Agarwal, Sahu, Puri, Laradji, Dvijotham, Stanley, Charlin, and Pal]{agarwal2025litllms}
Shubham Agarwal, Gaurav Sahu, Abhay Puri, Issam~H. Laradji, Krishnamurthy~Dj Dvijotham, Jason Stanley, Laurent Charlin, and Christopher Pal.
\newblock Lit{LLM}s, {LLM}s for literature review: Are we there yet?
\newblock \emph{Transactions on Machine Learning Research}, Apr 2025.
\newblock ISSN 2835-8856.
\newblock URL \url{https://openreview.net/forum?id=heeJqQXKg7}.

\bibitem[Agrawal et~al.(2024)Agrawal, McHale, and Oettl]{agrawal2024artificial}
Ajay Agrawal, John McHale, and Alexander Oettl.
\newblock Artificial intelligence and scientific discovery: A model of prioritized search.
\newblock \emph{Research Policy}, 53\penalty0 (5):\penalty0 104989, Jun 2024.

\bibitem[Agraz et~al.(2024)Agraz, Goksuluk, Zhang, Choi, Clements, Choudhary, and Karniadakis]{agraz2024ml}
Melih Agraz, Dincer Goksuluk, Peng Zhang, Bum-Rak Choi, Richard~T Clements, Gaurav Choudhary, and George~Em Karniadakis.
\newblock Ml-gap: machine learning-enhanced genomic analysis pipeline using autoencoders and data augmentation.
\newblock \emph{Frontiers in Genetics}, 15:\penalty0 1442759, Sep 2024.

\bibitem[AI(2025)]{zochi2025}
Intology AI.
\newblock Zochi technical report, Mar 2025.
\newblock URL \url{https://github.com/IntologyAI/Zochi/blob/main/Zochi_Technical_Report.pdf}.
\newblock Zochi Technical Report.

\bibitem[Aitymbetov and Zorbas(2025)]{aitymbetov2025autonomous}
Nurmukhammed Aitymbetov and Dimitrios Zorbas.
\newblock Autonomous machine learning-based peer reviewer selection system.
\newblock In \emph{Proceedings of the 31st International Conference on Computational Linguistics: System Demonstrations}, pages 199--207, Jan 2025.

\bibitem[Algaba et~al.(2025)Algaba, Holst, Tori, Mobini, Verbeken, Wenmackers, and Ginis]{algaba2025deep}
Andres Algaba, Vincent Holst, Floriano Tori, Melika Mobini, Brecht Verbeken, Sylvia Wenmackers, and Vincent Ginis.
\newblock How deep do large language models internalize scientific literature and citation practices?
\newblock \emph{arXiv preprint arXiv:2504.02767}, 2025.

\bibitem[Alkan et~al.(2025)Alkan, Sourav, Jablonska, Astarita, Chakrabarty, Garuda, Khetarpal, Pi{\'o}ro, Tanoglidis, Iyer, et~al.]{alkan2025survey}
Atilla~Kaan Alkan, Shashwat Sourav, Maja Jablonska, Simone Astarita, Rishabh Chakrabarty, Nikhil Garuda, Pranav Khetarpal, Maciej Pi{\'o}ro, Dimitrios Tanoglidis, Kartheik~G Iyer, et~al.
\newblock A survey on hypothesis generation for scientific discovery in the era of large language models.
\newblock \emph{arXiv preprint arXiv:2504.05496}, 2025.

\bibitem[Altuncu et~al.(2023)Altuncu, Nurse, Bagriacik, Kaleba, Yuan, Bonheme, and Li]{altuncu2023aedfact}
Enes Altuncu, Jason~RC Nurse, Meryem Bagriacik, Sophie Kaleba, Haiyue Yuan, Lisa Bonheme, and Shujun Li.
\newblock aedfact: Scientific fact-checking made easier via semi-automatic discovery of relevant expert opinions.
\newblock \emph{arXiv preprint arXiv:2305.07796}, 2023.

\bibitem[Alvarez et~al.(2024)Alvarez, Bennett, and Wang]{alvarez2024zero}
Carlos Alvarez, Maxwell Bennett, and Lucy~Lu Wang.
\newblock Zero-shot scientific claim verification using llms and citation text.
\newblock In \emph{Proceedings of the Fourth Workshop on Scholarly Document Processing (SDP 2024)}, pages 269--276, Aug 2024.

\bibitem[Alvarez et~al.(2024{\natexlab{2}})Alvarez, Colmenarejo, Elobaid, Fabbrizzi, Fahimi, Ferrara, Ghodsi, Mougan, Papageorgiou, Reyero, et~al.]{alvarez2024policy}
Jose~M Alvarez, Alejandra~Bringas Colmenarejo, Alaa Elobaid, Simone Fabbrizzi, Miriam Fahimi, Antonio Ferrara, Siamak Ghodsi, Carlos Mougan, Ioanna Papageorgiou, Paula Reyero, et~al.
\newblock Policy advice and best practices on bias and fairness in ai.
\newblock \emph{Ethics and Information Technology}, 26\penalty0 (2):\penalty0 31, Apr 2024{\natexlab{2}}.

\bibitem[Amano et~al.(2016)Amano, Gonz{\'a}lez-Varo, and Sutherland]{amano2016languages}
Tatsuya Amano, Juan~P Gonz{\'a}lez-Varo, and William~J Sutherland.
\newblock Languages are still a major barrier to global science.
\newblock \emph{PLoS biology}, 14\penalty0 (12):\penalty0 e2000933, Dec 2016.

\bibitem[Amano et~al.(2021)Amano, Rios~Rojas, Boum~II, Calvo, and Misra]{amano2021ten}
Tatsuya Amano, Clarissa Rios~Rojas, Yap Boum~II, Margarita Calvo, and Biswapriya~B Misra.
\newblock Ten tips for overcoming language barriers in science.
\newblock \emph{Nature Human Behaviour}, 5\penalty0 (9):\penalty0 1119--1122, Jul 2021.

\bibitem[Anderson et~al.(2024)Anderson, Shah, and Kreminski]{anderson2024homogenization}
Barrett~R Anderson, Jash~Hemant Shah, and Max Kreminski.
\newblock Homogenization effects of large language models on human creative ideation.
\newblock In \emph{Proceedings of the 16th conference on creativity \& cognition}, pages 413--425, Jun 2024.

\bibitem[Angello et~al.(2024)Angello, Friday, Hwang, Yi, Cheng, Torres-Flores, Jira, Wang, Aspuru-Guzik, Burke, et~al.]{angello2024closed}
Nicholas~H Angello, David~M Friday, Changhyun Hwang, Seungjoo Yi, Austin~H Cheng, Tiara~C Torres-Flores, Edward~R Jira, Wesley Wang, Al{\'a}n Aspuru-Guzik, Martin~D Burke, et~al.
\newblock Closed-loop transfer enables artificial intelligence to yield chemical knowledge.
\newblock \emph{Nature}, 633\penalty0 (8029):\penalty0 351--358, 2024.

\bibitem[Angelopoulos et~al.(2024)Angelopoulos, Cahoon, and Alterovitz]{angelopoulos2024transforming}
Angelos Angelopoulos, James~F Cahoon, and Ron Alterovitz.
\newblock Transforming science labs into automated factories of discovery.
\newblock \emph{Science Robotics}, 9\penalty0 (95):\penalty0 eadm6991, 2024.

\bibitem[Anthropic(2024)]{anthropic2024claude3}
AI~Anthropic.
\newblock The claude 3 model family: Opus, sonnet, haiku.
\newblock Claude-3 Model Card, Mar 2024.

\bibitem[Arlt et~al.(2024)Arlt, Duan, Li, Xie, Wu, and Krenn]{arlt2024meta}
S{\"o}ren Arlt, Haonan Duan, Felix Li, Sang~Michael Xie, Yuhuai Wu, and Mario Krenn.
\newblock Meta-designing quantum experiments with language models.
\newblock \emph{arXiv preprint arXiv:2406.02470}, 2024.

\bibitem[Asai et~al.(2024)Asai, He, Shao, Shi, Singh, Chang, Lo, Soldaini, Feldman, D'arcy, et~al.]{asai2024openscholar}
Akari Asai, Jacqueline He, Rulin Shao, Weijia Shi, Amanpreet Singh, Joseph~Chee Chang, Kyle Lo, Luca Soldaini, Sergey Feldman, Mike D'arcy, et~al.
\newblock Openscholar: Synthesizing scientific literature with retrieval-augmented lms.
\newblock \emph{arXiv preprint arXiv:2411.14199}, 2024.

\bibitem[Ashkinaze et~al.(2024)Ashkinaze, Mendelsohn, Qiwei, Budak, and Gilbert]{ashkinaze2024ai}
Joshua Ashkinaze, Julia Mendelsohn, Li~Qiwei, Ceren Budak, and Eric Gilbert.
\newblock How ai ideas affect the creativity, diversity, and evolution of human ideas: evidence from a large, dynamic experiment.
\newblock \emph{arXiv preprint arXiv:2401.13481}, 2024.

\bibitem[Ashury-Tahan et~al.(2025)Ashury-Tahan, Mai, Gera, Perlitz, Yehudai, Bandel, Choshen, Shnarch, Liang, Shmueli-Scheuer, et~al.]{ashury2025mighty}
Shir Ashury-Tahan, Yifan Mai, Ariel Gera, Yotam Perlitz, Asaf Yehudai, Elron Bandel, Leshem Choshen, Eyal Shnarch, Percy Liang, Michal Shmueli-Scheuer, et~al.
\newblock The mighty torr: A benchmark for table reasoning and robustness.
\newblock \emph{arXiv preprint arXiv:2502.19412}, 2025.

\bibitem[Assafelovic(2023)]{gpt-researcher}
Assafelovic.
\newblock gpt-researcher, May 2023.
\newblock URL \url{https://github.com/assafelovic/gpt-researcher}.
\newblock gpt-researcher.

\bibitem[Atanasova(2024)]{atanasova2024generating}
Pepa Atanasova.
\newblock Generating fact checking explanations.
\newblock In \emph{Accountable and Explainable Methods for Complex Reasoning over Text}, pages 83--103. Springer, Apr 2024.

\bibitem[Au et~al.(2025)Au, Dimacali, Pedirappagari, Park, Dernoncourt, Wang, Kanakaris, Deilamsalehy, Rossi, and Ahmed]{au2025personalized}
Steven Au, Cameron~J Dimacali, Ojasmitha Pedirappagari, Namyong Park, Franck Dernoncourt, Yu~Wang, Nikos Kanakaris, Hanieh Deilamsalehy, Ryan~A Rossi, and Nesreen~K Ahmed.
\newblock Personalized graph-based retrieval for large language models.
\newblock \emph{arXiv preprint arXiv:2501.02157}, 2025.

\bibitem[Auer et~al.(2023)Auer, Barone, Bartz, Cortes, Jaradeh, Karras, Koubarakis, Mouromtsev, Pliukhin, Radyush, et~al.]{auer2023sciqa}
S{\"o}ren Auer, Dante~AC Barone, Cassiano Bartz, Eduardo~G Cortes, Mohamad~Yaser Jaradeh, Oliver Karras, Manolis Koubarakis, Dmitry Mouromtsev, Dmitrii Pliukhin, Daniil Radyush, et~al.
\newblock The sciqa scientific question answering benchmark for scholarly knowledge.
\newblock \emph{Scientific Reports}, 13\penalty0 (1):\penalty0 7240, May 2023.

\bibitem[{Axios}(2024)]{axios2024_self_driving_labs}
{Axios}.
\newblock Self-driving labs are the new ai asset.
\newblock \emph{Axios}, Aug 2024.
\newblock URL \url{https://www.axios.com/2024/08/09/ai-self-driving-science-labs-research}.

\bibitem[Ayabe et~al.(2024)Ayabe, Otomo, Kera, and Kawamoto]{ayabe2024robustness}
Shingo Ayabe, Takuto Otomo, Hiroshi Kera, and Kazuhiko Kawamoto.
\newblock Robustness evaluation of offline reinforcement learning for robot control against action perturbations.
\newblock \emph{arXiv preprint arXiv:2412.18781}, 2024.

\bibitem[Azher et~al.(2025)Azher, Mokarrama, Guo, Choudhury, and Alhoori]{azher2025futuregen}
Ibrahim~Al Azher, Miftahul~Jannat Mokarrama, Zhishuai Guo, Sagnik~Ray Choudhury, and Hamed Alhoori.
\newblock Futuregen: Llm-rag approach to generate the future work of scientific article.
\newblock \emph{arXiv preprint arXiv:2503.16561}, 2025.

\bibitem[Baek et~al.(2024)Baek, Jauhar, Cucerzan, and Hwang]{baek2024researchagent}
Jinheon Baek, Sujay~Kumar Jauhar, Silviu Cucerzan, and Sung~Ju Hwang.
\newblock Researchagent: Iterative research idea generation over scientific literature with large language models.
\newblock \emph{arXiv preprint arXiv:2404.07738}, 2024.

\bibitem[Bai et~al.(2019)Bai, Wang, Lee, Yang, Kong, and Xia]{bai2019scientific}
Xiaomei Bai, Mengyang Wang, Ivan Lee, Zhuo Yang, Xiangjie Kong, and Feng Xia.
\newblock Scientific paper recommendation: A survey.
\newblock \emph{Ieee Access}, 7:\penalty0 9324--9339, Jan 2019.

\bibitem[Bao et~al.(2025)Bao, Wu, Choi, Mao, and Evans]{bao2025language}
Honglin Bao, Siyang Wu, Jiwoong Choi, Yingrong Mao, and James~A Evans.
\newblock Language models surface the unwritten code of science and society.
\newblock \emph{arXiv preprint arXiv:2505.18942}, 2025.

\bibitem[Bao et~al.(2024)Bao, Liu, Guo, Ye, Shen, Xie, Peng, Huang, and Wei]{bao2024piors}
Zhijie Bao, Qingyun Liu, Ying Guo, Zhengqiang Ye, Jun Shen, Shirong Xie, Jiajie Peng, Xuanjing Huang, and Zhongyu Wei.
\newblock Piors: Personalized intelligent outpatient reception based on large language model with multi-agents medical scenario simulation.
\newblock \emph{arXiv preprint arXiv:2411.13902}, 2024.

\bibitem[Baratchi et~al.(2024)Baratchi, Wang, Limmer, van Rijn, Hoos, B{\"a}ck, and Olhofer]{baratchi2024automated}
Mitra Baratchi, Can Wang, Steffen Limmer, Jan~N van Rijn, Holger Hoos, Thomas B{\"a}ck, and Markus Olhofer.
\newblock Automated machine learning: past, present and future.
\newblock \emph{Artificial intelligence review}, 57\penalty0 (5):\penalty0 122, Apr 2024.

\bibitem[Barber(2023)]{barber2023_gnome}
Gregory Barber.
\newblock Google deepmind's ai dreamed up 380,000 new materials. the next challenge is making them.
\newblock \emph{WIRED}, Nov 2023.
\newblock URL \url{https://www.wired.com/story/an-ai-dreamed-up-380000-new-materials-the-next-challenge-is-making-them/}.

\bibitem[Barbudo et~al.(2023)Barbudo, Ventura, and Romero]{barbudo2023eight}
Rafael Barbudo, Sebasti{\'a}n Ventura, and Jos{\'e}~Ra{\'u}l Romero.
\newblock Eight years of automl: categorisation, review and trends.
\newblock \emph{Knowledge and Information Systems}, 65\penalty0 (12):\penalty0 5097--5149, Aug 2023.

\bibitem[Barman et~al.(2025)Barman, Caron, Sullivan, de~Regt, de~Austri, Boon, F{\"a}rber, Fr{\"o}se, Hasibi, Ipp, et~al.]{barman2025large}
Kristian~G Barman, Sascha Caron, Emily Sullivan, Henk~W de~Regt, Roberto~Ruiz de~Austri, Mieke Boon, Michael F{\"a}rber, Stefan Fr{\"o}se, Faegheh Hasibi, Andreas Ipp, et~al.
\newblock Large physics models: Towards a collaborative approach with large language models and foundation models.
\newblock \emph{arXiv preprint arXiv:2501.05382}, 2025.

\bibitem[Basford et~al.(2024)Basford, Bernardino, Teeuwen, Egleston, Humphreys, Jelfs, Nitschke, Riddell, and Greenaway]{basford2024development}
Annabel~R Basford, Aaron~H Bernardino, Paula~CP Teeuwen, Benjamin~D Egleston, Joshua Humphreys, Kim~E Jelfs, Jonathan~R Nitschke, Imogen~A Riddell, and Rebecca~L Greenaway.
\newblock Development of an automated workflow for screening the assembly and host--guest behavior of metal-organic cages towards accelerated discovery.
\newblock \emph{Angewandte Chemie International Edition}, page e202424270, Apr 2024.

\bibitem[Basuki and Tsuchiya(2022)]{basuki2022quality}
Setio Basuki and Masatoshi Tsuchiya.
\newblock The quality assist: A technology-assisted peer review based on citation functions to predict the paper quality.
\newblock \emph{IEEE Access}, 10:\penalty0 126815--126831, Dec 2022.

\bibitem[Batista et~al.(2025)Batista, Amrutha, Yan, Dangi, and Steinbock]{batista2025high}
Bruno~C Batista, SV~Amrutha, Jie Yan, Beni~B Dangi, and Oliver Steinbock.
\newblock High-throughput robotic collection, imaging, and machine learning analysis of salt patterns: composition and concentration from dried droplet photos.
\newblock \emph{Digital Discovery}, 4\penalty0 (4):\penalty0 1030--1041, Feb 2025.

\bibitem[Battaglia et~al.(2016)Battaglia, Pascanu, Lai, Jimenez~Rezende, et~al.]{battaglia2016interaction}
Peter Battaglia, Razvan Pascanu, Matthew Lai, Danilo Jimenez~Rezende, et~al.
\newblock Interaction networks for learning about objects, relations and physics.
\newblock \emph{Advances in Neural Information Processing Systems}, 29, Dec 2016.

\bibitem[Baumg{\"a}rtner et~al.(2025)Baumg{\"a}rtner, Briscoe, and Gurevych]{baumgartner2025peerqa}
Tim Baumg{\"a}rtner, Ted Briscoe, and Iryna Gurevych.
\newblock Peerqa: A scientific question answering dataset from peer reviews.
\newblock In \emph{Proceedings of the 2025 Conference of the Nations of the Americas Chapter of the Association for Computational Linguistics: Human Language Technologies (Volume 1: Long Papers)}, pages 508--544, Feb 2025.

\bibitem[Baydin et~al.(2021)Baydin, Cranmer, Manzano, Delaere, Derkach, Donini, Dorigo, Giammanco, Kieseler, Layer, et~al.]{baydin2021toward}
At{\i}l{\i}m~G{\"u}ne{\c{s}} Baydin, Kyle Cranmer, Pablo de~Castro Manzano, Christophe Delaere, Denis Derkach, Julien Donini, Tommaso Dorigo, Andrea Giammanco, Jan Kieseler, Lukas Layer, et~al.
\newblock Toward machine learning optimization of experimental design.
\newblock \emph{Nuclear Physics News}, 31\penalty0 (1):\penalty0 25--28, Feb 2021.

\bibitem[Bazaga et~al.(2023)Bazaga, Lio, and Micklem]{bazaga2023unsupervised}
Adri{\'a}n Bazaga, Pietro Lio, and Gos Micklem.
\newblock Unsupervised pretraining for fact verification by language model distillation.
\newblock \emph{arXiv preprint arXiv:2309.16540}, 2023.

\bibitem[Bazgir et~al.(2025)Bazgir, Zhang, et~al.]{bazgir2025agentichypothesis}
Adib Bazgir, Yuwen Zhang, et~al.
\newblock Agentichypothesis: A survey on hypothesis generation using llm systems.
\newblock \emph{Towards Agentic AI for Science: Hypothesis Generation, Comprehension, Quantification, and Validation}, Mar 2025.

\bibitem[Beel et~al.(2016)Beel, Gipp, Langer, and Breitinger]{beel2016paper}
Joeran Beel, Bela Gipp, Stefan Langer, and Corinna Breitinger.
\newblock Paper recommender systems: a literature survey.
\newblock \emph{International Journal on Digital Libraries}, 17\penalty0 (4):\penalty0 305--338, Jul 2016.

\bibitem[Beel et~al.(2025)Beel, Kan, and Baumgart]{beel2025evaluating}
Joeran Beel, Min-Yen Kan, and Moritz Baumgart.
\newblock Evaluating sakana's ai scientist for autonomous research: Wishful thinking or an emerging reality towards' artificial research intelligence'(ari)?
\newblock \emph{arXiv preprint arXiv:2502.14297}, 2025.

\bibitem[Belouadi et~al.(2023)Belouadi, Lauscher, and Eger]{belouadi2023automatikz}
Jonas Belouadi, Anne Lauscher, and Steffen Eger.
\newblock Automatikz: Text-guided synthesis of scientific vector graphics with tikz.
\newblock \emph{arXiv preprint arXiv:2310.00367}, 2023.

\bibitem[Belouadi et~al.(2025)Belouadi, Ilg, Keuper, Tanaka, Utiyama, Dabre, Eger, and Ponzetto]{belouadi2025tikzero}
Jonas Belouadi, Eddy Ilg, Margret Keuper, Hideki Tanaka, Masao Utiyama, Raj Dabre, Steffen Eger, and Simone~Paolo Ponzetto.
\newblock Tikzero: Zero-shot text-guided graphics program synthesis.
\newblock \emph{arXiv preprint arXiv:2503.11509}, 2025.

\bibitem[Beltagy et~al.(2019)Beltagy, Lo, and Cohan]{beltagy-etal-2019-scibert}
Iz~Beltagy, Kyle Lo, and Arman Cohan.
\newblock {S}ci{BERT}: A pretrained language model for scientific text.
\newblock In Kentaro Inui, Jing Jiang, Vincent Ng, and Xiaojun Wan, editors, \emph{Proceedings of the 2019 Conference on Empirical Methods in Natural Language Processing and the 9th International Joint Conference on Natural Language Processing (EMNLP-IJCNLP)}, pages 3615--3620, Hong Kong, China, November 2019. Association for Computational Linguistics.
\newblock \doi{10.18653/v1/D19-1371}.
\newblock URL \url{https://aclanthology.org/D19-1371/}.

\bibitem[Bereska and Gavves(2024)]{bereska2024mechanistic}
Leonard Bereska and Efstratios Gavves.
\newblock Mechanistic interpretability for ai safety--a review.
\newblock \emph{arXiv preprint arXiv:2404.14082}, 2024.

\bibitem[Bevilacqua et~al.(2025)Bevilacqua, Oketch, Qin, Stamey, Zhang, Gan, Yang, and Abbasi]{10.1145/3702639}
Marialena Bevilacqua, Kezia Oketch, Ruiyang Qin, Will Stamey, Xinyuan Zhang, Yi~Gan, Kai Yang, and Ahmed Abbasi.
\newblock When automated assessment meets automated content generation: Examining text quality in the era of gpts.
\newblock \emph{ACM Trans. Inf. Syst.}, 43\penalty0 (2), January 2025.
\newblock ISSN 1046-8188.
\newblock \doi{10.1145/3702639}.
\newblock URL \url{https://doi.org/10.1145/3702639}.

\bibitem[Bharti et~al.(2021)Bharti, Ranjan, Ghosal, Agrawal, and Ekbal]{bharti2021peerassist}
Prabhat~Kumar Bharti, Shashi Ranjan, Tirthankar Ghosal, Mayank Agrawal, and Asif Ekbal.
\newblock Peerassist: leveraging on paper-review interactions to predict peer review decisions.
\newblock In \emph{Towards Open and Trustworthy Digital Societies: 23rd International Conference on Asia-Pacific Digital Libraries, ICADL 2021, Virtual Event, December 1--3, 2021, Proceedings 23}, pages 421--435. Springer, Nov 2021.

\bibitem[Bharti et~al.(2024)Bharti, Navlakha, Agarwal, and Ekbal]{bharti2024politepeer}
Prabhat~Kumar Bharti, Meith Navlakha, Mayank Agarwal, and Asif Ekbal.
\newblock Politepeer: does peer review hurt? a dataset to gauge politeness intensity in the peer reviews.
\newblock \emph{Language Resources and Evaluation}, 58\penalty0 (4):\penalty0 1291--1313, May 2024.

\bibitem[Bian et~al.(2024)Bian, Chen, Luo, Wu, Hao, Wei, and Zhang]{bian2024general}
Haiyang Bian, Yixin Chen, Erpai Luo, Xinze Wu, Minsheng Hao, Lei Wei, and Xuegong Zhang.
\newblock General-purpose pre-trained large cellular models for single-cell transcriptomics.
\newblock \emph{National Science Review}, 11\penalty0 (11):\penalty0 nwae340, Sep 2024.

\bibitem[Bian et~al.(2023)Bian, Qin, Zou, Huang, Luo, Zhang, and Zhang]{bian2023helm}
Junyi Bian, Xiaolei Qin, Wuhe Zou, Mengzuo Huang, Congyi Luo, Ke~Zhang, and Weidong Zhang.
\newblock Helm: Highlighted evidence augmented language model for enhanced table-to-text generation.
\newblock \emph{arXiv preprint arXiv:2311.08896}, 2023.

\bibitem[Bikku et~al.(2025)Bikku, Narimalla, Konda, Nakkala, Yarlagadda, and Sachuthananthan]{bikku2025generating}
Thulasi Bikku, Nirmala~Rani Narimalla, Keerthi Konda, Anusha Nakkala, Avanti Yarlagadda, and B~Sachuthananthan.
\newblock Generating accurate and engaging research paper titles using nlp techniques.
\newblock In \emph{International Conference on Innovations in Bio-Inspired Computing and Applications}, pages 428--437. Springer, May 2025.

\bibitem[Binz and Schulz(2023)]{Binz_2023}
Marcel Binz and Eric Schulz.
\newblock Using cognitive psychology to understand gpt-3.
\newblock \emph{Proceedings of the National Academy of Sciences}, 120\penalty0 (6), February 2023.
\newblock ISSN 1091-6490.
\newblock \doi{10.1073/pnas.2218523120}.
\newblock URL \url{http://dx.doi.org/10.1073/pnas.2218523120}.

\bibitem[Blaurock et~al.(2024)Blaurock, B{\"u}ttgen, and Schepers]{blaurock2024designing}
Marah Blaurock, Marion B{\"u}ttgen, and Jeroen Schepers.
\newblock Designing collaborative intelligence systems for employee-ai service co-production.
\newblock \emph{Journal of Service Research}, page 10946705241238751, Mar 2024.

\bibitem[Bochem et~al.(2024)Bochem, Gonzalez-Sanchez, Bicker, and Fadini]{bochem2024improving}
Severin Bochem, Eduardo Gonzalez-Sanchez, Yves Bicker, and Gabriele Fadini.
\newblock Improving generalization of robot locomotion policies via sharpness-aware reinforcement learning.
\newblock \emph{arXiv preprint arXiv:2411.19732}, 2024.

\bibitem[Boehnlein et~al.(2022)Boehnlein, Diefenthaler, Sato, Schram, Ziegler, Fanelli, Hjorth-Jensen, Horn, Kuchera, Lee, et~al.]{boehnlein2022colloquium}
Amber Boehnlein, Markus Diefenthaler, Nobuo Sato, Malachi Schram, Veronique Ziegler, Cristiano Fanelli, Morten Hjorth-Jensen, Tanja Horn, Michelle~P Kuchera, Dean Lee, et~al.
\newblock Colloquium: Machine learning in nuclear physics.
\newblock \emph{Reviews of modern physics}, 94\penalty0 (3):\penalty0 031003, Sep 2022.

\bibitem[Boiko et~al.(2023)Boiko, MacKnight, Kline, and Gomes]{boiko2023autonomous}
Daniil~A Boiko, Robert MacKnight, Ben Kline, and Gabe Gomes.
\newblock Autonomous chemical research with large language models.
\newblock \emph{Nature}, 624\penalty0 (7992):\penalty0 570--578, Dec 2023.

\bibitem[Bolanos et~al.(2024)Bolanos, Salatino, Osborne, and Motta]{bolanos2024artificial}
Francisco Bolanos, Angelo Salatino, Francesco Osborne, and Enrico Motta.
\newblock Artificial intelligence for literature reviews: Opportunities and challenges.
\newblock \emph{Artificial Intelligence Review}, 57\penalty0 (10):\penalty0 259, Aug 2024.

\bibitem[Bolotova et~al.(2022)Bolotova, Blinov, Scholer, Croft, and Sanderson]{bolotova2022non}
Valeriia Bolotova, Vladislav Blinov, Falk Scholer, W~Bruce Croft, and Mark Sanderson.
\newblock A non-factoid question-answering taxonomy.
\newblock In \emph{Proceedings of the 45th International ACM SIGIR Conference on Research and Development in Information Retrieval}, pages 1196--1207, Jul 2022.

\bibitem[Bolton et~al.(2024)Bolton, Venigalla, Yasunaga, Hall, Xiong, Lee, Daneshjou, Frankle, Liang, Carbin, et~al.]{bolton2024biomedlm}
Elliot Bolton, Abhinav Venigalla, Michihiro Yasunaga, David Hall, Betty Xiong, Tony Lee, Roxana Daneshjou, Jonathan Frankle, Percy Liang, Michael Carbin, et~al.
\newblock Biomedlm: A 2.7 b parameter language model trained on biomedical text.
\newblock \emph{arXiv preprint arXiv:2403.18421}, 2024.

\bibitem[B{\"o}l{\"u}c{\"u} et~al.(2025)B{\"o}l{\"u}c{\"u}, Bilge, {\c{C}}etinta{\c{s}}, and Y{\"u}cel]{bolucu2025modest}
Necva B{\"o}l{\"u}c{\"u}, Yunus~Can Bilge, Dilber {\c{C}}etinta{\c{s}}, and Zehra Y{\"u}cel.
\newblock Modest: A dataset for multi domain scientific title generation.
\newblock \emph{Knowledge-Based Systems}, page 113557, Jun 2025.

\bibitem[Botha et~al.(2018)Botha, Faruqui, Alex, Baldridge, and Das]{botha2018learning}
Jan~A Botha, Manaal Faruqui, John Alex, Jason Baldridge, and Dipanjan Das.
\newblock Learning to split and rephrase from wikipedia edit history.
\newblock \emph{arXiv preprint arXiv:1808.09468}, 2018.

\bibitem[Boylan et~al.(2024)Boylan, Mangla, Thorn, Ghalandari, Ghaffari, and Hokamp]{boylan2024kgvalidator}
Jack Boylan, Shashank Mangla, Dominic Thorn, Demian~Gholipour Ghalandari, Parsa Ghaffari, and Chris Hokamp.
\newblock Kgvalidator: A framework for automatic validation of knowledge graph construction.
\newblock \emph{arXiv preprint arXiv:2404.15923}, 2024.

\bibitem[Braun et~al.(2024)Braun, Rothermel, Rohrbach, and Rohrbach]{braun2024defame}
Tobias Braun, Mark Rothermel, Marcus Rohrbach, and Anna Rohrbach.
\newblock Defame: Dynamic evidence-based fact-checking with multimodal experts.
\newblock \emph{arXiv preprint arXiv:2412.10510}, 2024.

\bibitem[Britton et~al.(2024)Britton, Bedwell, Chawhan, Crowe, Jarvis, Jeske, Kalra, Lawrence, and McSpadden]{britton2024ai}
Thomas Britton, Cullan Bedwell, Abhijeet Chawhan, Julie Crowe, Naomi Jarvis, Torri Jeske, Nikhil Kalra, David Lawrence, and Diana McSpadden.
\newblock Ai driven experiment calibration and control.
\newblock In \emph{EPJ Web of Conferences}, volume 295, page 02003. EDP Sciences, May 2024.

\bibitem[Brodsky et~al.(2025)Brodsky, Ullah, Bychkov, Song, Walk, Louis, Rasool, Singh, Mahmood, Bui, et~al.]{brodsky2025generative}
Victor Brodsky, Ehsan Ullah, Andrey Bychkov, Andrew~H Song, Eric~E Walk, Peter Louis, Ghulam Rasool, Rajendra~S Singh, Faisal Mahmood, Marilyn~M Bui, et~al.
\newblock Generative artificial intelligence in anatomic pathology.
\newblock \emph{Archives of Pathology \& Laboratory Medicine}, Apr 2025.

\bibitem[Bu et~al.(2024)Bu, Zeng, Chen, Yang, Zhou, Yan, Luo, Cui, Ma, and Li]{bu2024closed}
Qingwen Bu, Jia Zeng, Li~Chen, Yanchao Yang, Guyue Zhou, Junchi Yan, Ping Luo, Heming Cui, Yi~Ma, and Hongyang Li.
\newblock Closed-loop visuomotor control with generative expectation for robotic manipulation.
\newblock In \emph{The Thirty-eighth Annual Conference on Neural Information Processing Systems}, 2024.

\bibitem[Buehler(2024)]{buehler2024accelerating}
Markus~J Buehler.
\newblock Accelerating scientific discovery with generative knowledge extraction, graph-based representation, and multimodal intelligent graph reasoning.
\newblock \emph{Machine Learning: Science and Technology}, 5\penalty0 (3):\penalty0 035083, Sep 2024.

\bibitem[Buess et~al.(2025)Buess, Keicher, Navab, Maier, and Arasteh]{buess2025large}
Lukas Buess, Matthias Keicher, Nassir Navab, Andreas Maier, and Soroosh~Tayebi Arasteh.
\newblock From large language models to multimodal ai: A scoping review on the potential of generative ai in medicine.
\newblock \emph{arXiv preprint arXiv:2502.09242}, 2025.

\bibitem[Bunne et~al.(2024)Bunne, Roohani, Rosen, Gupta, Zhang, Roed, Alexandrov, AlQuraishi, Brennan, Burkhardt, et~al.]{bunne2024build}
Charlotte Bunne, Yusuf Roohani, Yanay Rosen, Ankit Gupta, Xikun Zhang, Marcel Roed, Theo Alexandrov, Mohammed AlQuraishi, Patricia Brennan, Daniel~B Burkhardt, et~al.
\newblock How to build the virtual cell with artificial intelligence: Priorities and opportunities.
\newblock \emph{Cell}, 187\penalty0 (25):\penalty0 7045--7063, Dec 2024.

\bibitem[Burgess et~al.(2025)Burgess, Nirschl, Bravo-S{\'a}nchez, Lozano, Gupte, Galaz-Montoya, Zhang, Su, Bhowmik, Coman, et~al.]{burgess2025microvqa}
James Burgess, Jeffrey~J Nirschl, Laura Bravo-S{\'a}nchez, Alejandro Lozano, Sanket~Rajan Gupte, Jesus~G Galaz-Montoya, Yuhui Zhang, Yuchang Su, Disha Bhowmik, Zachary Coman, et~al.
\newblock Microvqa: A multimodal reasoning benchmark for microscopy-based scientific research.
\newblock In \emph{Proceedings of the Computer Vision and Pattern Recognition Conference}, pages 19552--19564, Mar 2025.

\bibitem[Butler et~al.(2018)Butler, Davies, Cartwright, Isayev, and Walsh]{butler2018machine}
Keith~T Butler, Daniel~W Davies, Hugh Cartwright, Olexandr Isayev, and Aron Walsh.
\newblock Machine learning for molecular and materials science.
\newblock \emph{Nature}, 559\penalty0 (7715):\penalty0 547--555, Jul 2018.

\bibitem[Bölücü et~al.(2025)Bölücü, Bilge, Çetintaş, and Yücel]{BOLUCU2025113557}
Necva Bölücü, Yunus~Can Bilge, Dilber Çetintaş, and Zehra Yücel.
\newblock Modest: A dataset for multi domain scientific title generation.
\newblock \emph{Knowledge-Based Systems}, 321:\penalty0 113557, Jun 2025.
\newblock ISSN 0950-7051.
\newblock \doi{https://doi.org/10.1016/j.knosys.2025.113557}.
\newblock URL \url{https://www.sciencedirect.com/science/article/pii/S0950705125006033}.

\bibitem[Cai and Wunder(2023)]{cai2023gradient}
Yi~Cai and Gerhard Wunder.
\newblock On gradient-like explanation under a black-box setting: when black-box explanations become as good as white-box.
\newblock \emph{arXiv preprint arXiv:2308.09381}, 2023.

\bibitem[Calderon and Herrera(2025)]{calderon2025and}
Reyes Calderon and Francisco Herrera.
\newblock And plato met chatgpt: an ethical reflection on the use of chatbots in scientific research writing, with a particular focus on the social sciences.
\newblock \emph{Humanities and Social Sciences Communications}, 12\penalty0 (1):\penalty0 1--13, May 2025.

\bibitem[Canty et~al.(2025)Canty, Bennett, Brown, Buonassisi, Kalinin, Kitchin, Maruyama, Moore, Schrier, Seifrid, et~al.]{canty2025science}
Richard~B Canty, Jeffrey~A Bennett, Keith~A Brown, Tonio Buonassisi, Sergei~V Kalinin, John~R Kitchin, Benji Maruyama, Robert~G Moore, Joshua Schrier, Martin Seifrid, et~al.
\newblock Science acceleration and accessibility with self-driving labs.
\newblock \emph{Nature Communications}, 16\penalty0 (1):\penalty0 3856, Apr 2025.

\bibitem[Cao and Liu(2025)]{cao2025tablemaster}
Lang Cao and Hanbing Liu.
\newblock Tablemaster: A recipe to advance table understanding with language models.
\newblock \emph{arXiv preprint arXiv:2501.19378}, 2025.

\bibitem[Cao et~al.(2024)Cao, Zhang, Alghadeer, Fasciati, Piscitelli, Bakr, Leek, and Aspuru-Guzik]{cao2024agents}
Shuxiang Cao, Zijian Zhang, Mohammed Alghadeer, Simone~D Fasciati, Michele Piscitelli, Mustafa Bakr, Peter Leek, and Al{\'a}n Aspuru-Guzik.
\newblock Agents for self-driving laboratories applied to quantum computing.
\newblock \emph{arXiv preprint arXiv:2412.07978}, 2024.

\bibitem[Cao and Liu(2024)]{cao2024figuring}
Stanley Cao and Kevin Liu.
\newblock Figuring out figures: Using textual references to caption scientific figures.
\newblock \emph{arXiv preprint arXiv:2407.11008}, 2024.

\bibitem[Cao et~al.(2024{\natexlab{2}})Cao, Nair, Eyimife, Soofi, Subbalakshmi, Wullert~II, Basu, and Shallcross]{cao2024can}
Yupeng Cao, Aishwarya~Muralidharan Nair, Elyon Eyimife, Nastaran~Jamalipour Soofi, KP~Subbalakshmi, John~R Wullert~II, Chumki Basu, and David Shallcross.
\newblock Can large language models detect misinformation in scientific news reporting?
\newblock \emph{arXiv preprint arXiv:2402.14268}, 2024{\natexlab{2}}.

\bibitem[{\c{C}}elik and Tekir(2024)]{ccelik2024citebart}
Ege~Yi{\u{g}}it {\c{C}}elik and Selma Tekir.
\newblock Citebart: Learning to generate citations for local citation recommendation.
\newblock \emph{arXiv preprint arXiv:2412.17534}, 2024.

\bibitem[Chakrabarty et~al.(2024)Chakrabarty, Laban, Agarwal, Muresan, and Wu]{chakrabarty2024art}
Tuhin Chakrabarty, Philippe Laban, Divyansh Agarwal, Smaranda Muresan, and Chien-Sheng Wu.
\newblock Art or artifice? large language models and the false promise of creativity.
\newblock In \emph{Proceedings of the 2024 CHI Conference on Human Factors in Computing Systems}, pages 1--34, May 2024.

\bibitem[Chamoun et~al.(2024)Chamoun, Schlichtkrull, and Vlachos]{chamoun-etal-2024-automated}
Eric Chamoun, Michael Schlichtkrull, and Andreas Vlachos.
\newblock Automated focused feedback generation for scientific writing assistance.
\newblock In Lun-Wei Ku, Andre Martins, and Vivek Srikumar, editors, \emph{Findings of the Association for Computational Linguistics: ACL 2024}, pages 9742--9763, Bangkok, Thailand, August 2024. Association for Computational Linguistics.
\newblock \doi{10.18653/v1/2024.findings-acl.580}.
\newblock URL \url{https://aclanthology.org/2024.findings-acl.580/}.

\bibitem[Chan et~al.(2024)Chan, Chowdhury, Jaffe, Aung, Sherburn, Mays, Starace, Liu, Maksin, Patwardhan, et~al.]{chan2024mle}
Jun~Shern Chan, Neil Chowdhury, Oliver Jaffe, James Aung, Dane Sherburn, Evan Mays, Giulio Starace, Kevin Liu, Leon Maksin, Tejal Patwardhan, et~al.
\newblock Mle-bench: Evaluating machine learning agents on machine learning engineering.
\newblock \emph{arXiv preprint arXiv:2410.07095}, 2024.

\bibitem[Chandra et~al.(2024)Chandra, Naik, Ford, Okoli, De~Choudhury, Ershadi, Ramos, Hernandez, Bhattacharjee, Warreth, et~al.]{chandra2024lived}
Mohit Chandra, Suchismita Naik, Denae Ford, Ebele Okoli, Munmun De~Choudhury, Mahsa Ershadi, Gonzalo Ramos, Javier Hernandez, Ananya Bhattacharjee, Shahed Warreth, et~al.
\newblock From lived experience to insight: Unpacking the psychological risks of using ai conversational agents.
\newblock \emph{arXiv preprint arXiv:2412.07951}, 2024.

\bibitem[Chang et~al.(2025)Chang, Feng, Sun, Ai, Li, Zhou, and Zhang]{chang2025sridbench}
Yifan Chang, Yukang Feng, Jianwen Sun, Jiaxin Ai, Chuanhao Li, S~Kevin Zhou, and Kaipeng Zhang.
\newblock Sridbench: Benchmark of scientific research illustration drawing of image generation model.
\newblock \emph{arXiv preprint arXiv:2505.22126}, 2025.

\bibitem[Chang et~al.(2025{\natexlab{2}})Chang, Li, Zhang, Kong, Wu, Guo, and Wong]{chang2025treereview}
Yuan Chang, Ziyue Li, Hengyuan Zhang, Yuanbo Kong, Yanru Wu, Zhijiang Guo, and Ngai Wong.
\newblock Treereview: A dynamic tree of questions framework for deep and efficient llm-based scientific peer review.
\newblock \emph{arXiv preprint arXiv:2506.07642}, 2025{\natexlab{2}}.

\bibitem[Charlin and Zemel(2013)]{charlin2013toronto}
Laurent Charlin and Richard Zemel.
\newblock The toronto paper matching system: an automated paper-reviewer assignment system.
\newblock May 2013.

\bibitem[Charlin et~al.(2011)Charlin, Zemel, and Boutilier]{charlin2011framework}
Laurent Charlin, Richard Zemel, and Craig Boutilier.
\newblock A framework for optimizing paper matching.
\newblock In \emph{Proceedings of the Twenty-Seventh Conference on Uncertainty in Artificial Intelligence}, pages 86--95, Jul 2011.

\bibitem[Chaudhuri(2024)]{Rajdip2024}
Rajdip Chaudhuri.
\newblock Llm based exploratory data analysis using bigquery data canvas, Oct 2024.
\newblock URL \url{https://medium.com/google-cloud/llm-based-exploratory-data-analysis-using-bigquery-data-canvas-42fbecb9f009}.
\newblock LLM Based Exploratory Data Analysis Using BigQuery Data Canvas.

\bibitem[Chen et~al.(2019)Chen, Ye, Zuo, Zheng, and Ong]{chen2019graph}
Chi Chen, Weike Ye, Yunxing Zuo, Chen Zheng, and Shyue~Ping Ong.
\newblock Graph networks as a universal machine learning framework for molecules and crystals.
\newblock \emph{Chemistry of Materials}, 31\penalty0 (9):\penalty0 3564--3572, Apr 2019.

\bibitem[Chen et~al.(2025)Chen, Xiong, Lu, Han, Deng, He, Wu, Li, Liu, and Hooi]{chen2025mlr}
Hui Chen, Miao Xiong, Yujie Lu, Wei Han, Ailin Deng, Yufei He, Jiaying Wu, Yibo Li, Yue Liu, and Bryan Hooi.
\newblock Mlr-bench: Evaluating ai agents on open-ended machine learning research.
\newblock \emph{arXiv preprint arXiv:2505.19955}, 2025.

\bibitem[Chen and Zhuge(2019)]{chen2019automatic}
Jingqiang Chen and Hai Zhuge.
\newblock Automatic generation of related work through summarizing citations.
\newblock \emph{Concurrency and Computation: Practice and Experience}, 31\penalty0 (3):\penalty0 e4261, Sep 2019.

\bibitem[Chen et~al.(2025{\natexlab{2}})Chen, Zhang, Li, Feng, Zhang, and Deng]{chen2025structuring}
Junlan Chen, Kexin Zhang, Daifeng Li, Yangyang Feng, Yuxuan Zhang, and Bowen Deng.
\newblock Structuring scientific innovation: A framework for modeling and discovering impactful knowledge combinations.
\newblock \emph{arXiv preprint arXiv:2503.18865}, 2025{\natexlab{2}}.

\bibitem[Chen et~al.(2021)Chen, Tworek, Jun, Yuan, Pinto, Kaplan, Edwards, Burda, Joseph, Brockman, et~al.]{chen2021evaluating}
Mark Chen, Jerry Tworek, Heewoo Jun, Qiming Yuan, Henrique Ponde De~Oliveira Pinto, Jared Kaplan, Harri Edwards, Yuri Burda, Nicholas Joseph, Greg Brockman, et~al.
\newblock Evaluating large language models trained on code.
\newblock \emph{arXiv preprint arXiv:2107.03374}, 2021.

\bibitem[Chen et~al.(2025{\natexlab{3}})Chen, HuiKai, Wu, Hou, Zhang, Wang, Wang, and He]{chen2025xtragpt}
Nuo Chen, Andre~Lin HuiKai, Jiaying Wu, Junyi Hou, Zining Zhang, Qian Wang, Xidong Wang, and Bingsheng He.
\newblock Xtragpt: Llms for human-ai collaboration on controllable academic paper revision.
\newblock \emph{arXiv preprint arXiv:2505.11336}, 2025{\natexlab{3}}.

\bibitem[Chen et~al.(2024)Chen, Qin, Wang, Zhou, and Che]{chen2024unlocking}
Qiguang Chen, Libo Qin, Jiaqi Wang, Jingxuan Zhou, and Wanxiang Che.
\newblock Unlocking the capabilities of thought: A reasoning boundary framework to quantify and optimize chain-of-thought.
\newblock \emph{Advances in Neural Information Processing Systems}, 37:\penalty0 54872--54904, Sep 2024.
\newblock URL \url{https://proceedings.neurips.cc/paper_files/paper/2024/hash/62ab1c2cb4b03e717005479efb211841-Abstract-Conference.html}.

\bibitem[Chen et~al.(2024{\natexlab{2}})Chen, Qin, Zhang, Chen, Xu, and Che]{chen2024m}
Qiguang Chen, Libo Qin, Jin Zhang, Zhi Chen, Xiao Xu, and Wanxiang Che.
\newblock {M}$^3${C}o{T}: A novel benchmark for multi-domain multi-step multi-modal chain-of-thought.
\newblock pages 8199--8221, August 2024{\natexlab{2}}.
\newblock \doi{10.18653/v1/2024.acl-long.446}.
\newblock URL \url{https://aclanthology.org/2024.acl-long.446/}.

\bibitem[Chen et~al.(2025{\natexlab{4}})Chen, Qin, Liu, Liao, Wang, Zhou, and Che]{chen2025rbf++}
Qiguang Chen, Libo Qin, Jinhao Liu, Yue Liao, Jiaqi Wang, Jingxuan Zhou, and Wanxiang Che.
\newblock Rbf++: Quantifying and optimizing reasoning boundaries across measurable and unmeasurable capabilities for chain-of-thought reasoning.
\newblock \emph{arXiv preprint arXiv:2505.13307}, 2025{\natexlab{4}}.

\bibitem[Chen et~al.(2025{\natexlab{5}})Chen, Qin, Liu, Peng, Guan, Wang, Hu, Zhou, Gao, and Che]{chen2025towards}
Qiguang Chen, Libo Qin, Jinhao Liu, Dengyun Peng, Jiannan Guan, Peng Wang, Mengkang Hu, Yuhang Zhou, Te~Gao, and Wangxiang Che.
\newblock Towards reasoning era: A survey of long chain-of-thought for reasoning large language models.
\newblock \emph{arXiv preprint arXiv:2503.09567}, 2025{\natexlab{5}}.

\bibitem[Chen et~al.(2025{\natexlab{6}})Chen, Qin, Liu, Peng, Wang, Hu, Chen, Che, and Liu]{chen2025ecm}
Qiguang Chen, Libo Qin, Jinhao Liu, Dengyun Peng, Jiaqi Wang, Mengkang Hu, Zhi Chen, Wanxiang Che, and Ting Liu.
\newblock Ecm: A unified electronic circuit model for explaining the emergence of in-context learning and chain-of-thought in large language model.
\newblock \emph{arXiv preprint arXiv:2502.03325}, 2025{\natexlab{6}}.

\bibitem[Chen et~al.(2025{\natexlab{7}})Chen, Su, Tang, Yin, Wu, Li, Sun, Dong, Ouyang, and Torr]{chen2025ai}
Renqi Chen, Haoyang Su, Shixiang Tang, Zhenfei Yin, Qi~Wu, Hui Li, Ye~Sun, Nanqing Dong, Wanli Ouyang, and Philip Torr.
\newblock Ai-driven automation can become the foundation of next-era science of science research.
\newblock \emph{arXiv preprint arXiv:2505.12039}, 2025{\natexlab{7}}.

\bibitem[Chen et~al.(2025{\natexlab{8}})Chen, Ding, Yuan, Zhang, Wang, and Zhuang]{chen2025bridging}
Wei Chen, Han Ding, Meng Yuan, Zhao Zhang, Deqing Wang, and Fuzhen Zhuang.
\newblock Bridging social psychology and llm reasoning: Conflict-aware meta-review generation via cognitive alignment.
\newblock \emph{arXiv preprint arXiv:2503.13879}, 2025{\natexlab{8}}.

\bibitem[Chen et~al.(2023)Chen, Yin, Ku, Lu, Wan, Ma, Xu, Wang, and Xia]{chen2023theoremqa}
Wenhu Chen, Ming Yin, Max Ku, Pan Lu, Yixin Wan, Xueguang Ma, Jianyu Xu, Xinyi Wang, and Tony Xia.
\newblock Theoremqa: A theorem-driven question answering dataset.
\newblock \emph{arXiv preprint arXiv:2305.12524}, 2023.

\bibitem[Chen et~al.(2024{\natexlab{3}})Chen, Buriak, Salanne, and Xin]{chen2024nano}
Xiaodong Chen, Jillian~M Buriak, Mathieu Salanne, and Huolin Xin.
\newblock Nano \& ai: A nobel partnership, Nov 2024{\natexlab{3}}.

\bibitem[Chen et~al.(2023{\natexlab{2}})Chen, Lin, Sch{\"a}rli, and Zhou]{chen2023teaching}
Xinyun Chen, Maxwell Lin, Nathanael Sch{\"a}rli, and Denny Zhou.
\newblock Teaching large language models to self-debug.
\newblock \emph{arXiv preprint arXiv:2304.05128}, 2023{\natexlab{2}}.

\bibitem[Chen et~al.(2021{\natexlab{2}})Chen, Alamro, Li, Gao, Zhang, Zhao, and Yan]{chen2021capturing}
Xiuying Chen, Hind Alamro, Mingzhe Li, Shen Gao, Xiangliang Zhang, Dongyan Zhao, and Rui Yan.
\newblock Capturing relations between scientific papers: An abstractive model for related work section generation.
\newblock In Chengqing Zong, Fei Xia, Wenjie Li, and Roberto Navigli, editors, \emph{Proceedings of the 59th Annual Meeting of the Association for Computational Linguistics and the 11th International Joint Conference on Natural Language Processing (Volume 1: Long Papers)}, pages 6068--6077, Online, August 2021{\natexlab{2}}. Association for Computational Linguistics.
\newblock \doi{10.18653/v1/2021.acl-long.473}.
\newblock URL \url{https://aclanthology.org/2021.acl-long.473/}.

\bibitem[Chen et~al.(2022)Chen, Alamro, Li, Gao, Yan, Gao, and Zhang]{chen2022target}
Xiuying Chen, Hind Alamro, Mingzhe Li, Shen Gao, Rui Yan, Xin Gao, and Xiangliang Zhang.
\newblock Target-aware abstractive related work generation with contrastive learning.
\newblock In \emph{Proceedings of the 45th international ACM SIGIR conference on research and development in information retrieval}, pages 373--383, Jul 2022.

\bibitem[Chen et~al.(2024{\natexlab{4}})Chen, Wang, Guo, Guo, Zhou, Li, Zhuge, Schmidhuber, Gao, and Zhang]{chen2024scholarchemqa}
Xiuying Chen, Tairan Wang, Taicheng Guo, Kehan Guo, Juexiao Zhou, Haoyang Li, Mingchen Zhuge, J{\"u}rgen Schmidhuber, Xin Gao, and Xiangliang Zhang.
\newblock Scholarchemqa: Unveiling the power of language models in chemical research question answering.
\newblock \emph{arXiv preprint arXiv:2407.16931}, 2024{\natexlab{4}}.

\bibitem[Chen et~al.(2025{\natexlab{9}})Chen, Hu, and Lu]{chen2025predicting}
Yaoyu Chen, Yuheng Hu, and Yingda Lu.
\newblock Predicting field experiments with large language models.
\newblock \emph{arXiv preprint arXiv:2504.01167}, 2025{\natexlab{9}}.

\bibitem[Chen et~al.(2020)Chen, Li, Yu, El~Kholy, Ahmed, Gan, Cheng, and Liu]{chen2020uniter}
Yen-Chun Chen, Linjie Li, Licheng Yu, Ahmed El~Kholy, Faisal Ahmed, Zhe Gan, Yu~Cheng, and Jingjing Liu.
\newblock Uniter: Universal image-text representation learning.
\newblock In \emph{European conference on computer vision}, pages 104--120. Springer, Sep 2020.

\bibitem[Chen et~al.(2025{\natexlab{10}})Chen, Albarqawi, and Chen]{chen2025reinforcing}
Ying-Jung Chen, Ahmad Albarqawi, and Chi-Sheng Chen.
\newblock Reinforcing clinical decision support through multi-agent systems and ethical ai governance.
\newblock \emph{arXiv preprint arXiv:2504.03699}, 2025{\natexlab{10}}.

\bibitem[Chen and Zou(2024)]{chen2024genept}
Yiqun Chen and James Zou.
\newblock Genept: a simple but effective foundation model for genes and cells built from chatgpt.
\newblock \emph{bioRxiv}, pages 2023--10, Mar 2024.

\bibitem[Chen et~al.(2023{\natexlab{3}})Chen, Liu, Shan, and Zhong]{chen2023emergenceeconomicrationalitygpt}
Yiting Chen, Tracy~Xiao Liu, You Shan, and Songfa Zhong.
\newblock The emergence of economic rationality of gpt.
\newblock \emph{Proceedings of the National Academy of Sciences}, 120\penalty0 (51):\penalty0 e2316205120, Dec 2023{\natexlab{3}}.

\bibitem[Chen et~al.(2024{\natexlab{5}})Chen, Chen, Qin, Guo, Lv, Zou, Che, Yan, Chen, and Lin]{chen2024essential}
Zhi Chen, Qiguang Chen, Libo Qin, Qipeng Guo, Haijun Lv, Yicheng Zou, Wanxiang Che, Hang Yan, Kai Chen, and Dahua Lin.
\newblock What are the essential factors in crafting effective long context multi-hop instruction datasets? insights and best practices.
\newblock \emph{arXiv preprint arXiv:2409.01893}, 2024{\natexlab{5}}.

\bibitem[Chen et~al.(2024{\natexlab{6}})Chen, Chen, Ning, Zhang, Wang, Yu, Li, Liao, Wei, Lu, et~al.]{chen2024scienceagentbench}
Ziru Chen, Shijie Chen, Yuting Ning, Qianheng Zhang, Boshi Wang, Botao Yu, Yifei Li, Zeyi Liao, Chen Wei, Zitong Lu, et~al.
\newblock Scienceagentbench: Toward rigorous assessment of language agents for data-driven scientific discovery.
\newblock \emph{arXiv preprint arXiv:2410.05080}, 2024{\natexlab{6}}.

\bibitem[Cheng et~al.(2025)Cheng, Calhoun, and Reedy]{cheng2025artificial}
Adam Cheng, Aaron Calhoun, and Gabriel Reedy.
\newblock Artificial intelligence-assisted academic writing: recommendations for ethical use.
\newblock \emph{Advances in Simulation}, 10\penalty0 (1):\penalty0 22, Apr 2025.

\bibitem[Cheng et~al.(2025{\natexlab{2}})Cheng, Clark, and Richardson]{cheng2025language}
Junyan Cheng, Peter Clark, and Kyle Richardson.
\newblock Language modeling by language models.
\newblock \emph{arXiv preprint arXiv:2506.20249}, 2025{\natexlab{2}}.

\bibitem[Cheng et~al.(2021)Cheng, Mosallanezhad, Sheth, and Liu]{cheng2021causal}
Lu~Cheng, Ahmadreza Mosallanezhad, Paras Sheth, and Huan Liu.
\newblock Causal learning for socially responsible ai.
\newblock \emph{arXiv preprint arXiv:2104.12278}, 2021.

\bibitem[Cheng et~al.(2025{\natexlab{3}})Cheng, Chen, Zan, Jia, and Peng]{cheng2025biasfilter}
Xiaoqing Cheng, Ruizhe Chen, Hongying Zan, Yuxiang Jia, and Min Peng.
\newblock Biasfilter: An inference-time debiasing framework for large language models.
\newblock \emph{arXiv preprint arXiv:2505.23829}, 2025{\natexlab{3}}.

\bibitem[Cheng and Zhang(2025)]{cheng2025ai}
Xusen Cheng and Lulu Zhang.
\newblock Ai-generated literature reviews threaten scientific progress.
\newblock \emph{Nature}, 641\penalty0 (8064):\penalty0 852--852, 2025.

\bibitem[Cheng et~al.(2023)Cheng, Dai, and Hauptmann]{cheng2023chartreader}
Zhi-Qi Cheng, Qi~Dai, and Alexander~G Hauptmann.
\newblock Chartreader: A unified framework for chart derendering and comprehension without heuristic rules.
\newblock In \emph{Proceedings of the IEEE/CVF International Conference on Computer Vision}, pages 22202--22213, Apr 2023.

\bibitem[Cheng et~al.(2025{\natexlab{4}})Cheng, Chen, Xu, Wang, Wang, Fei, Wang, Wang, Chen, Che, et~al.]{cheng2025visual}
Zihui Cheng, Qiguang Chen, Xiao Xu, Jiaqi Wang, Weiyun Wang, Hao Fei, Yidong Wang, Alex~Jinpeng Wang, Zhi Chen, Wanxiang Che, et~al.
\newblock Visual thoughts: A unified perspective of understanding multimodal chain-of-thought.
\newblock \emph{arXiv preprint arXiv:2505.15510}, 2025{\natexlab{4}}.

\bibitem[Cheng et~al.(2025{\natexlab{5}})Cheng, Chen, Zhang, Fei, Feng, Che, Li, and Qin]{cheng2025comt}
Zihui Cheng, Qiguang Chen, Jin Zhang, Hao Fei, Xiaocheng Feng, Wanxiang Che, Min Li, and Libo Qin.
\newblock Comt: A novel benchmark for chain of multi-modal thought on large vision-language models.
\newblock In \emph{Proceedings of the AAAI Conference on Artificial Intelligence}, volume~39, pages 23678--23686, Apr 2025{\natexlab{5}}.

\bibitem[Choi and Lee(2024)]{choi2024accelerating}
Jaewoong Choi and Byungju Lee.
\newblock Accelerating materials language processing with large language models.
\newblock \emph{Communications Materials}, 5\penalty0 (1):\penalty0 13, Feb 2024.

\bibitem[Chu et~al.(2025)Chu, Fan, Chen, Wang, Yang, Liang, Wang, Li, Tang, Liu, et~al.]{chu2025self}
Zheng Chu, Huiming Fan, Jingchang Chen, Qianyu Wang, Mingda Yang, Jiafeng Liang, Zhongjie Wang, Hao Li, Guo Tang, Ming Liu, et~al.
\newblock Self-critique guided iterative reasoning for multi-hop question answering.
\newblock \emph{arXiv preprint arXiv:2505.19112}, 2025.

\bibitem[Chu et~al.(2024)Chu, Ai, Tu, Li, and Liu]{chu2024automatic}
Zhumin Chu, Qingyao Ai, Yiteng Tu, Haitao Li, and Yiqun Liu.
\newblock Automatic large language model evaluation via peer review.
\newblock In \emph{Proceedings of the 33rd ACM International Conference on Information and Knowledge Management}, pages 384--393, Oct 2024.

\bibitem[Chu et~al.(2024{\natexlab{2}})Chu, Ai, Tu, Li, and Liu]{chu2024pre}
Zhumin Chu, Qingyao Ai, Yiteng Tu, Haitao Li, and Yiqun Liu.
\newblock Pre: A peer review based large language model evaluator.
\newblock \emph{arXiv preprint arXiv:2401.15641}, 2024{\natexlab{2}}.

\bibitem[Chua et~al.(2024)Chua, Ghazi, Huang, Kamath, Kumar, Manurangsi, Sinha, Xie, and Zhang]{chua2024crosslingual}
Lynn Chua, Badih Ghazi, Yangsibo Huang, Pritish Kamath, Ravi Kumar, Pasin Manurangsi, Amer Sinha, Chulin Xie, and Chiyuan Zhang.
\newblock Crosslingual capabilities and knowledge barriers in multilingual large language models.
\newblock \emph{arXiv preprint arXiv:2406.16135}, 2024.

\bibitem[Cingillioglu et~al.(2024)Cingillioglu, Gal, and Prokhorov]{cingillioglu2024ai}
Ilker Cingillioglu, Uri Gal, and Artem Prokhorov.
\newblock Ai-experiments in education: An ai-driven randomized controlled trial for higher education research.
\newblock \emph{Education and Information Technologies}, 29\penalty0 (15):\penalty0 19649--19677, 2024.

\bibitem[Circi et~al.(2024)Circi, Khalighinejad, Chen, Dhingra, and Brinson]{circi2024well}
Defne Circi, Ghazal Khalighinejad, Anlan Chen, Bhuwan Dhingra, and L~Catherine Brinson.
\newblock How well do large language models understand tables in materials science?
\newblock \emph{Integrating Materials and Manufacturing Innovation}, 13\penalty0 (3):\penalty0 669--687, Jul 2024.
\newblock URL \url{https://link.springer.com/article/10.1007/s40192-024-00362-6}.

\bibitem[Cirillo et~al.(2021)Cirillo, N{\'u}{\~n}ez-Carpintero, and Valencia]{cirillo2021artificial}
Davide Cirillo, Iker N{\'u}{\~n}ez-Carpintero, and Alfonso Valencia.
\newblock Artificial intelligence in cancer research: learning at different levels of data granularity.
\newblock \emph{Molecular oncology}, 15\penalty0 (4):\penalty0 817--829, Apr 2021.

\bibitem[Clark et~al.(2019)Clark, Lee, Chang, Kwiatkowski, Collins, and Toutanova]{clark2019boolq}
Christopher Clark, Kenton Lee, Ming-Wei Chang, Tom Kwiatkowski, Michael Collins, and Kristina Toutanova.
\newblock Boolq: Exploring the surprising difficulty of natural yes/no questions.
\newblock \emph{arXiv preprint arXiv:1905.10044}, 2019.

\bibitem[Coenen et~al.(2021)Coenen, Davis, Ippolito, Reif, and Yuan]{coenen2021wordcraft}
Andy Coenen, Luke Davis, Daphne Ippolito, Emily Reif, and Ann Yuan.
\newblock Wordcraft: A human-ai collaborative editor for story writing.
\newblock \emph{arXiv preprint arXiv:2107.07430}, 2021.

\bibitem[Cohrs et~al.(2025)Cohrs, Diaz, Sitokonstantinou, Varando, and Camps-Valls]{cohrs2025large}
Kai-Hendrik Cohrs, Emiliano Diaz, Vasileios Sitokonstantinou, Gherardo Varando, and Gustau Camps-Valls.
\newblock Large language models for causal hypothesis generation in science.
\newblock \emph{Machine Learning: Science and Technology}, 6\penalty0 (1):\penalty0 013001, Jan 2025.

\bibitem[Conneau et~al.(2019)Conneau, Khandelwal, Goyal, Chaudhary, Wenzek, Guzm{\'a}n, Grave, Ott, Zettlemoyer, and Stoyanov]{conneau2019unsupervised}
Alexis Conneau, Kartikay Khandelwal, Naman Goyal, Vishrav Chaudhary, Guillaume Wenzek, Francisco Guzm{\'a}n, Edouard Grave, Myle Ott, Luke Zettlemoyer, and Veselin Stoyanov.
\newblock Unsupervised cross-lingual representation learning at scale.
\newblock \emph{arXiv preprint arXiv:1911.02116}, 2019.

\bibitem[Consens et~al.(2025)Consens, Dufault, Wainberg, Forster, Karimzadeh, Goodarzi, Theis, Moses, and Wang]{consens2025transformers}
Micaela~E Consens, Cameron Dufault, Michael Wainberg, Duncan Forster, Mehran Karimzadeh, Hani Goodarzi, Fabian~J Theis, Alan Moses, and Bo~Wang.
\newblock Transformers and genome language models.
\newblock \emph{Nature Machine Intelligence}, pages 1--17, Mar 2025.

\bibitem[Couto et~al.(2024)Couto, Ho, Kumari, Rachmat, Khuong, Ullah, and Sun-Hosoya]{couto2024relevai}
Paulo~Henrique Couto, Quang~Phuoc Ho, Nageeta Kumari, Benedictus~Kent Rachmat, Thanh Gia~Hieu Khuong, Ihsan Ullah, and Lisheng Sun-Hosoya.
\newblock Relevai-reviewer: A benchmark on ai reviewers for survey paper relevance.
\newblock \emph{arXiv preprint arXiv:2406.10294}, 2024.

\bibitem[Craig(2025)]{craig-2025-human}
Douglas~B Craig.
\newblock A human-{LLM} note-taking system with case-based reasoning as framework for scientific discovery.
\newblock In Peter Jansen, Bhavana Dalvi~Mishra, Harsh Trivedi, Bodhisattwa Prasad~Majumder, Tom Hope, Tushar Khot, Doug Downey, and Eric Horvitz, editors, \emph{Proceedings of the 1st Workshop on AI and Scientific Discovery: Directions and Opportunities}, pages 22--30, Albuquerque, New Mexico, USA, May 2025. Association for Computational Linguistics.
\newblock ISBN 979-8-89176-224-4.
\newblock \doi{10.18653/v1/2025.aisd-main.3}.
\newblock URL \url{https://aclanthology.org/2025.aisd-main.3/}.

\bibitem[Cranmer et~al.(2020)Cranmer, Greydanus, Hoyer, Battaglia, Spergel, and Ho]{cranmer2020lagrangian}
Miles Cranmer, Sam Greydanus, Stephan Hoyer, Peter Battaglia, David Spergel, and Shirley Ho.
\newblock Lagrangian neural networks.
\newblock \emph{arXiv preprint arXiv:2003.04630}, 2020.

\bibitem[Cui et~al.(2024)Cui, Wang, Maan, Pang, Luo, Duan, and Wang]{cui2024scgpt}
Haotian Cui, Chloe Wang, Hassaan Maan, Kuan Pang, Fengning Luo, Nan Duan, and Bo~Wang.
\newblock scgpt: toward building a foundation model for single-cell multi-omics using generative ai.
\newblock \emph{Nature Methods}, 21\penalty0 (8):\penalty0 1470--1480, Feb 2024.

\bibitem[Cui et~al.(2025)Cui, Xu, Pang, Li, Gong, Wang, and Li]{cui2025lumi}
Haotian Cui, Yue Xu, Kuan Pang, Gen Li, Fanglin Gong, Bo~Wang, and Bowen Li.
\newblock Lumi-lab: a foundation model-driven autonomous platform enabling discovery of new ionizable lipid designs for mrna delivery.
\newblock \emph{BioRxiv}, pages 2025--02, Feb 2025.

\bibitem[Cui et~al.(2024{\natexlab{2}})Cui, Li, and Zhou]{cui2024can}
Ziyan Cui, Ning Li, and Huaikang Zhou.
\newblock Can ai replace human subjects? a large-scale replication of psychological experiments with llms.
\newblock \emph{arXiv preprint arXiv:2409.00128}, 2024{\natexlab{2}}.

\bibitem[Dai et~al.(2024)Dai, Vijayakrishnan, Szczypi{\'n}ski, Ayme, Simaei, Fellowes, Clowes, Kotopanov, Shields, Zhou, et~al.]{dai2024autonomous}
Tianwei Dai, Sriram Vijayakrishnan, Filip~T Szczypi{\'n}ski, Jean-Fran{\c{c}}ois Ayme, Ehsan Simaei, Thomas Fellowes, Rob Clowes, Lyubomir Kotopanov, Caitlin~E Shields, Zhengxue Zhou, et~al.
\newblock Autonomous mobile robots for exploratory synthetic chemistry.
\newblock \emph{Nature}, pages 1--8, Nov 2024.

\bibitem[Dai et~al.(2025)Dai, Chan, Vriza, Kim, Wang, Liu, Shan, Xu, Weires, Wu, et~al.]{dai2025adaptive}
Yahao Dai, Henry Chan, Aikaterini Vriza, Fredrick Kim, Yunfei Wang, Wei Liu, Naisong Shan, Jing Xu, Max Weires, Yukun Wu, et~al.
\newblock Adaptive ai decision interface for autonomous electronic material discovery.
\newblock \emph{arXiv preprint arXiv:2504.13344}, 2025.

\bibitem[Dammu et~al.(2024)Dammu, Naidu, Dewan, Kim, Roosta, Chadha, and Shah]{dammu2024claimver}
Preetam Prabhu~Srikar Dammu, Himanshu Naidu, Mouly Dewan, YoungMin Kim, Tanya Roosta, Aman Chadha, and Chirag Shah.
\newblock Claimver: Explainable claim-level verification and evidence attribution of text through knowledge graphs.
\newblock \emph{arXiv preprint arXiv:2403.09724}, 2024.

\bibitem[Dangayach et~al.(2024)Dangayach, Jeong, Demirel, Uzal, Fung, and Chen]{dangayach2024machine}
Raghav Dangayach, Nohyeong Jeong, Elif Demirel, Nigmet Uzal, Victor Fung, and Yongsheng Chen.
\newblock Machine learning-aided inverse design and discovery of novel polymeric materials for membrane separation.
\newblock \emph{Environmental Science \& Technology}, 59\penalty0 (2):\penalty0 993--1012, Dec 2024.

\bibitem[D'Arcy et~al.(2024)D'Arcy, Hope, Birnbaum, and Downey]{d2024marg}
Mike D'Arcy, Tom Hope, Larry Birnbaum, and Doug Downey.
\newblock Marg: Multi-agent review generation for scientific papers.
\newblock \emph{arXiv preprint arXiv:2401.04259}, 2024.

\bibitem[Darrin et~al.(2024)Darrin, Arous, Piantanida, and Cheung]{darrin2024glimpse}
Maxime Darrin, Ines Arous, Pablo Piantanida, and Jackie~CK Cheung.
\newblock Glimpse: Pragmatically informative multi-document summarization for scholarly reviews.
\newblock \emph{arXiv preprint arXiv:2406.07359}, 2024.

\bibitem[Darvish et~al.(2025)Darvish, Skreta, Zhao, Yoshikawa, Som, Bogdanovic, Cao, Hao, Xu, Aspuru-Guzik, et~al.]{darvish2025organa}
Kourosh Darvish, Marta Skreta, Yuchi Zhao, Naruki Yoshikawa, Sagnik Som, Miroslav Bogdanovic, Yang Cao, Han Hao, Haoping Xu, Al{\'a}n Aspuru-Guzik, et~al.
\newblock Organa: a robotic assistant for automated chemistry experimentation and characterization.
\newblock \emph{Matter}, 8\penalty0 (2), Feb 2025.

\bibitem[Das et~al.(2023)Das, Liu, Kovatchev, and Lease]{das2023state}
Anubrata Das, Houjiang Liu, Venelin Kovatchev, and Matthew Lease.
\newblock The state of human-centered nlp technology for fact-checking.
\newblock \emph{Information processing \& management}, 60\penalty0 (2):\penalty0 103219, Mar 2023.

\bibitem[Dasgupta et~al.(2024)Dasgupta, Mondal, and Chakrabarti]{dasguptaempowering}
Debajyoti Dasgupta, Arijit Mondal, and Partha~Pratim Chakrabarti.
\newblock Empowering ai as autonomous researchers: Evaluating llms in generating novel research ideas through automated metrics.
\newblock In \emph{2nd AI4Research Workshop: Towards a Knowledge-grounded Scientific Research Lifecycle}, Dec 2024.

\bibitem[de~Avila Belbute-Peres et~al.(2018)de~Avila Belbute-Peres, Smith, Allen, Tenenbaum, and Kolter]{de2018end}
Filipe de~Avila Belbute-Peres, Kevin Smith, Kelsey Allen, Josh Tenenbaum, and J~Zico Kolter.
\newblock End-to-end differentiable physics for learning and control.
\newblock \emph{Advances in Neural Information Processing Systems}, 31, Dec 2018.

\bibitem[De~Cao and Kipf(2018)]{de2018molgan}
Nicola De~Cao and Thomas Kipf.
\newblock Molgan: An implicit generative model for small molecular graphs.
\newblock \emph{arXiv preprint arXiv:1805.11973}, 2018.

\bibitem[de~Kok(2025)]{dekok2025chatgpt}
Ties de~Kok.
\newblock Chatgpt for textual analysis? how to use generative llms in accounting research.
\newblock \emph{Management Science}, Jan 2025.
\newblock \doi{10.1287/mnsc.2023.03253}.
\newblock URL \url{https://doi.org/10.1287/mnsc.2023.03253}.
\newblock Published online in Articles in Advance, 13 Jan 2025.

\bibitem[Deiner et~al.(2024)Deiner, Honcharov, Li, Mackey, Porco, and Sarkar]{deiner2024llmthematic}
Michael~S. Deiner, Vlad Honcharov, Jiawei Li, Tim~K. Mackey, Travis~C. Porco, and Urmimala Sarkar.
\newblock Large language models can enable inductive thematic analysis of a social media corpus in a single prompt: Human validation study.
\newblock \emph{JMIR Infodemiology}, 4:\penalty0 e59641, August 2024.
\newblock \doi{10.2196/59641}.
\newblock URL \url{https://doi.org/10.2196/59641}.

\bibitem[Deng et~al.(2021)Deng, Zeng, Gu, Ji, and Hua]{deng2021automatic}
Zekun Deng, Zixin Zeng, Weiye Gu, Jiawen Ji, and Bolin Hua.
\newblock Automatic related work section generation by sentence extraction and reordering.
\newblock In \emph{AII@ iConference}, pages 101--110, Jan 2021.

\bibitem[Desai et~al.(2025)Desai, Addamane, Tsao, Brener, Swiler, Dingreville, and Iyer]{desai2025autoscilab}
Saaketh Desai, Sadhvikas Addamane, Jeffrey~Y Tsao, Igal Brener, Laura~P Swiler, Remi Dingreville, and Prasad~P Iyer.
\newblock Autoscilab: A self-driving laboratory for interpretable scientific discovery.
\newblock In \emph{Proceedings of the AAAI Conference on Artificial Intelligence}, volume~39, pages 146--154, Apr 2025.

\bibitem[DeYoung et~al.(2021)DeYoung, Beltagy, van Zuylen, Kuehl, and Wang]{deyoung2021ms2}
Jay DeYoung, Iz~Beltagy, Madeleine van Zuylen, Bailey Kuehl, and Lucy~Lu Wang.
\newblock Ms2: Multi-document summarization of medical studies.
\newblock \emph{arXiv preprint arXiv:2104.06486}, 2021.

\bibitem[D{\'\i}az et~al.(2024)D{\'\i}az, Garmendia, and Pereira]{diaz2024streamlining}
Oscar D{\'\i}az, Xabier Garmendia, and Juanan Pereira.
\newblock Streamlining the review process: Ai-generated annotations in research manuscripts.
\newblock \emph{arXiv preprint arXiv:2412.00281}, 2024.

\bibitem[Dillion et~al.(2023)Dillion, Tandon, Gu, and Gray]{DILLION2023597}
Danica Dillion, Niket Tandon, Yuling Gu, and Kurt Gray.
\newblock Can ai language models replace human participants?
\newblock \emph{Trends in Cognitive Sciences}, 27\penalty0 (7):\penalty0 597--600, Jul 2023.
\newblock ISSN 1364-6613.
\newblock \doi{https://doi.org/10.1016/j.tics.2023.04.008}.
\newblock URL \url{https://www.sciencedirect.com/science/article/pii/S1364661323000980}.

\bibitem[Ding et~al.(2024)Ding, Qu, Xie, Li, Liu, Zhang, Xiong, Zuo, Chen, Hua, et~al.]{ding2024automating}
Ning Ding, Shang Qu, Linhai Xie, Yifei Li, Zaoqu Liu, Kaiyan Zhang, Yibai Xiong, Yuxin Zuo, Zhangren Chen, Ermo Hua, et~al.
\newblock Automating exploratory proteomics research via language models.
\newblock \emph{arXiv preprint arXiv:2411.03743}, 2024.

\bibitem[Ding and Cronin(2011)]{ding2011popular}
Ying Ding and Blaise Cronin.
\newblock Popular and/or prestigious? measures of scholarly esteem.
\newblock \emph{Information processing \& management}, 47\penalty0 (1):\penalty0 80--96, Jan 2011.

\bibitem[Docekal et~al.(2024)Docekal, Fajcik, and Smrz]{docekal2024oarelatedwork}
Martin Docekal, Martin Fajcik, and Pavel Smrz.
\newblock Oarelatedwork: A large-scale dataset of related work sections with full-texts from open access sources.
\newblock \emph{arXiv preprint arXiv:2405.01930}, 2024.

\bibitem[Dominik et~al.(2023)Dominik, Zhengyao, and Yuxiang]{AIDE2024}
Schmidt Dominik, Jiang Zhengyao, and Wu~Yuxiang.
\newblock Aide: Human-level performance on data science competitions, Apr 2023.
\newblock URL \url{https://www.weco.ai/blog/technical-report}.
\newblock AIDE.

\bibitem[Dorigo et~al.(2023)Dorigo, Giammanco, Vischia, Aehle, Bawaj, Boldyrev, de~Castro~Manzano, Derkach, Donini, Edelen, et~al.]{dorigo2023toward}
Tommaso Dorigo, Andrea Giammanco, Pietro Vischia, Max Aehle, Mateusz Bawaj, Alexey Boldyrev, Pablo de~Castro~Manzano, Denis Derkach, Julien Donini, Auralee Edelen, et~al.
\newblock Toward the end-to-end optimization of particle physics instruments with differentiable programming.
\newblock \emph{Reviews in Physics}, 10:\penalty0 100085, Jun 2023.

\bibitem[Doskaliuk et~al.(2025)Doskaliuk, Zimba, Yessirkepov, Klishch, and Yatsyshyn]{doskaliuk2025artificial}
Bohdana Doskaliuk, Olena Zimba, Marlen Yessirkepov, Iryna Klishch, and Roman Yatsyshyn.
\newblock Artificial intelligence in peer review: enhancing efficiency while preserving integrity.
\newblock \emph{Journal of Korean medical science}, 40\penalty0 (7), Feb 2025.

\bibitem[Dove et~al.(2025)Dove, Hartfield, and Carpenter]{dove2025semi}
Ryan~S Dove, Roy~J Hartfield, and Mark Carpenter.
\newblock Semi-supervised classification with novelty detection using support vector machines and linear discriminant analysis.
\newblock In \emph{AIAA SCITECH 2025 Forum}, page 0705, Jan 2025.

\bibitem[Du et~al.(2024)Du, Wang, Zhao, Deng, Liu, Lou, Zou, Narayanan~Venkit, Zhang, Srinath, Zhang, Gupta, Li, Li, Wang, Liu, Liu, Gao, Xia, Xing, Jiayang, Wang, Su, Shah, Guo, Gu, Li, Wei, Wang, Cheng, Ranathunga, Fang, Fu, Liu, Huang, Blanco, Cao, Zhang, Yu, and Yin]{du-etal-2024-llms}
Jiangshu Du, Yibo Wang, Wenting Zhao, Zhongfen Deng, Shuaiqi Liu, Renze Lou, Henry~Peng Zou, Pranav Narayanan~Venkit, Nan Zhang, Mukund Srinath, Haoran~Ranran Zhang, Vipul Gupta, Yinghui Li, Tao Li, Fei Wang, Qin Liu, Tianlin Liu, Pengzhi Gao, Congying Xia, Chen Xing, Cheng Jiayang, Zhaowei Wang, Ying Su, Raj~Sanjay Shah, Ruohao Guo, Jing Gu, Haoran Li, Kangda Wei, Zihao Wang, Lu~Cheng, Surangika Ranathunga, Meng Fang, Jie Fu, Fei Liu, Ruihong Huang, Eduardo Blanco, Yixin Cao, Rui Zhang, Philip~S. Yu, and Wenpeng Yin.
\newblock {LLM}s assist {NLP} researchers: Critique paper (meta-)reviewing.
\newblock In Yaser Al-Onaizan, Mohit Bansal, and Yun-Nung Chen, editors, \emph{Proceedings of the 2024 Conference on Empirical Methods in Natural Language Processing}, pages 5081--5099, Miami, Florida, USA, November 2024. Association for Computational Linguistics.
\newblock \doi{10.18653/v1/2024.emnlp-main.292}.
\newblock URL \url{https://aclanthology.org/2024.emnlp-main.292/}.

\bibitem[Du et~al.(2025)Du, Xu, Zhu, Wang, and Mao]{du2025deepresearch}
Mingxuan Du, Benfeng Xu, Chiwei Zhu, Xiaorui Wang, and Zhendong Mao.
\newblock Deepresearch bench: A comprehensive benchmark for deep research agents.
\newblock \emph{arXiv preprint arXiv:2506.11763}, 2025.

\bibitem[Du et~al.(2022)Du, Kim, Raheja, Kumar, and Kang]{du2022read}
Wanyu Du, Zae~Myung Kim, Vipul Raheja, Dhruv Kumar, and Dongyeop Kang.
\newblock Read, revise, repeat: A system demonstration for human-in-the-loop iterative text revision.
\newblock \emph{arXiv preprint arXiv:2204.03685}, 2022.

\bibitem[Dycke et~al.(2023)Dycke, Kuznetsov, and Gurevych]{dycke-etal-2023-nlpeer}
Nils Dycke, Ilia Kuznetsov, and Iryna Gurevych.
\newblock {NLP}eer: A unified resource for the computational study of peer review.
\newblock In Anna Rogers, Jordan Boyd-Graber, and Naoaki Okazaki, editors, \emph{Proceedings of the 61st Annual Meeting of the Association for Computational Linguistics (Volume 1: Long Papers)}, pages 5049--5073, Toronto, Canada, July 2023. Association for Computational Linguistics.
\newblock \doi{10.18653/v1/2023.acl-long.277}.
\newblock URL \url{https://aclanthology.org/2023.acl-long.277/}.

\bibitem[Ebrahimi et~al.(2024)Ebrahimi, Chen, Asudeh, Das, and Koudas]{ebrahimi2024axolotl}
Sana Ebrahimi, Kaiwen Chen, Abolfazl Asudeh, Gautam Das, and Nick Koudas.
\newblock Axolotl: fairness through assisted self-debiasing of large language model outputs.
\newblock \emph{arXiv preprint arXiv:2403.00198}, 2024.

\bibitem[Edelman and Skolnick(2025)]{edelman2025valsci}
Brice Edelman and Jeffrey Skolnick.
\newblock Valsci: an open-source, self-hostable literature review utility for automated large-batch scientific claim verification using large language models.
\newblock \emph{BMC bioinformatics}, 26\penalty0 (1):\penalty0 1--25, May 2025.

\bibitem[Edfeldt et~al.(2024)Edfeldt, Edwards, Engkvist, G{\"u}nther, Hartley, Hulcoop, Leach, Marsden, Menge, Misquitta, et~al.]{edfeldt2024data}
Kristina Edfeldt, Aled~M Edwards, Ola Engkvist, Judith G{\"u}nther, Matthew Hartley, David~G Hulcoop, Andrew~R Leach, Brian~D Marsden, Amelie Menge, Leonie Misquitta, et~al.
\newblock A data science roadmap for open science organizations engaged in early-stage drug discovery.
\newblock \emph{Nature Communications}, 15\penalty0 (1):\penalty0 5640, Jul 2024.

\bibitem[Eger et~al.(2025)Eger, Cao, D'Souza, Geiger, Greisinger, Gross, Hou, Krenn, Lauscher, Li, et~al.]{eger2025transforming}
Steffen Eger, Yong Cao, Jennifer D'Souza, Andreas Geiger, Christian Greisinger, Stephanie Gross, Yufang Hou, Brigitte Krenn, Anne Lauscher, Yizhi Li, et~al.
\newblock Transforming science with large language models: A survey on ai-assisted scientific discovery, experimentation, content generation, and evaluation.
\newblock \emph{arXiv preprint arXiv:2502.05151}, 2025.

\bibitem[Ehsan and Riedl(2024)]{ehsan2024explainable}
Upol Ehsan and Mark Riedl.
\newblock Explainable ai reloaded: Challenging the xai status quo in the era of large language models.
\newblock In \emph{Proceedings of the Halfway to the Future Symposium}, pages 1--8, Oct 2024.

\bibitem[Ekosso et~al.(2024)Ekosso, Liu, Glagovich, Nguyen, Maurer, and Schrier]{ekosso2024accelerating}
Christelle Ekosso, Hao Liu, Avery Glagovich, Dustin Nguyen, Sarah Maurer, and Joshua Schrier.
\newblock Accelerating the discovery of abiotic vesicles with ai-guided automated experimentation.
\newblock \emph{Langmuir}, 41\penalty0 (1):\penalty0 858--867, 2024.

\bibitem[Eldifrawi et~al.(2024)Eldifrawi, Wang, and Trabelsi]{eldifrawi2024automated}
Islam Eldifrawi, Shengrui Wang, and Amine Trabelsi.
\newblock Automated justification production for claim veracity in fact checking: A survey on architectures and approaches.
\newblock \emph{arXiv preprint arXiv:2407.12853}, 2024.

\bibitem[Faltings et~al.(2020)Faltings, Galley, Hintz, Brockett, Quirk, Gao, and Dolan]{faltings2020text}
Felix Faltings, Michel Galley, Gerold Hintz, Chris Brockett, Chris Quirk, Jianfeng Gao, and Bill Dolan.
\newblock Text editing by command.
\newblock \emph{arXiv preprint arXiv:2010.12826}, 2020.

\bibitem[Fan and Gardent(2022)]{fan2022generating}
Angela Fan and Claire Gardent.
\newblock Generating full length wikipedia biographies: The impact of gender bias on the retrieval-based generation of women biographies.
\newblock \emph{arXiv preprint arXiv:2204.05879}, 2022.

\bibitem[Fan et~al.(2023)Fan, Gokkaya, Harman, Lyubarskiy, Sengupta, Yoo, and Zhang]{fan2023large}
Angela Fan, Beliz Gokkaya, Mark Harman, Mitya Lyubarskiy, Shubho Sengupta, Shin Yoo, and Jie~M Zhang.
\newblock Large language models for software engineering: Survey and open problems.
\newblock In \emph{2023 IEEE/ACM International Conference on Software Engineering: Future of Software Engineering (ICSE-FoSE)}, pages 31--53. IEEE, Oct 2023.

\bibitem[Fan et~al.(2024)Fan, Ding, Ning, Wang, Li, Yin, Chua, and Li]{fan2024survey}
Wenqi Fan, Yujuan Ding, Liangbo Ning, Shijie Wang, Hengyun Li, Dawei Yin, Tat-Seng Chua, and Qing Li.
\newblock A survey on rag meeting llms: Towards retrieval-augmented large language models.
\newblock In \emph{Proceedings of the 30th ACM SIGKDD Conference on Knowledge Discovery and Data Mining}, pages 6491--6501, Aug 2024.

\bibitem[Fan et~al.(2024{\natexlab{2}})Fan, Tang, Chen, Wang, Wei, Xi, Huang, and Zhou]{fan2024ai}
Zhihao Fan, Jialong Tang, Wei Chen, Siyuan Wang, Zhongyu Wei, Jun Xi, Fei Huang, and Jingren Zhou.
\newblock Ai hospital: Benchmarking large language models in a multi-agent medical interaction simulator.
\newblock \emph{arXiv preprint arXiv:2402.09742}, 2024{\natexlab{2}}.

\bibitem[Fang et~al.(2025)Fang, Jian, Li, and Ma]{fang2025ai}
You-Le Fang, Dong-Shan Jian, Xiang Li, and Yan-Qing Ma.
\newblock Ai-newton: A concept-driven physical law discovery system without prior physical knowledge.
\newblock \emph{arXiv preprint arXiv:2504.01538}, 2025.

\bibitem[Farber(2024)]{farber2024enhancing}
Shai Farber.
\newblock Enhancing peer review efficiency: A mixed-methods analysis of artificial intelligence-assisted reviewer selection across academic disciplines.
\newblock \emph{Learned Publishing}, 37\penalty0 (4):\penalty0 e1638, Oct 2024.

\bibitem[Farber(2025)]{farber2025enhancing}
Shai Farber.
\newblock Enhancing academic decision-making: A pilot study of ai-supported journal selection in higher education.
\newblock \emph{Innovative Higher Education}, pages 1--19, Feb 2025.

\bibitem[Faruqui et~al.(2018)Faruqui, Pavlick, Tenney, and Das]{faruqui2018wikiatomicedits}
Manaal Faruqui, Ellie Pavlick, Ian Tenney, and Dipanjan Das.
\newblock Wikiatomicedits: A multilingual corpus of wikipedia edits for modeling language and discourse.
\newblock \emph{arXiv preprint arXiv:1808.09422}, 2018.

\bibitem[Fehlis(2025)]{fehlis2025uncovering}
Yao Fehlis.
\newblock Uncovering bottlenecks and optimizing scientific lab workflows with cycle time reduction agents.
\newblock \emph{arXiv preprint arXiv:2505.21534}, 2025.

\bibitem[Fehlis et~al.(2025)Fehlis, Mandel, Crain, Liu, and Fuller]{fehlis2025accelerating}
Yao Fehlis, Paul Mandel, Charles Crain, Betty Liu, and David Fuller.
\newblock Accelerating drug discovery with artificial: a whole-lab orchestration and scheduling system for self-driving labs.
\newblock \emph{arXiv preprint arXiv:2504.00986}, 2025.

\bibitem[Feng et~al.(2024)Feng, Liu, Huang, Xiao, Zhang, Mirzoyan, Xu, Hao, Xu, Zhang, et~al.]{feng2024bioactivity}
Bin Feng, Zequn Liu, Nanlan Huang, Zhiping Xiao, Haomiao Zhang, Srbuhi Mirzoyan, Hanwen Xu, Jiaran Hao, Yinghui Xu, Ming Zhang, et~al.
\newblock A bioactivity foundation model using pairwise meta-learning.
\newblock \emph{Nature Machine Intelligence}, 6\penalty0 (8):\penalty0 962--974, Aug 2024.

\bibitem[Feng et~al.(2025)Feng, Xu, and Chu]{feng2025openfoamgpt}
Jingsen Feng, Ran Xu, and Xu~Chu.
\newblock Openfoamgpt 2.0: end-to-end, trustworthy automation for computational fluid dynamics.
\newblock \emph{arXiv preprint arXiv:2504.19338}, 2025.

\bibitem[Feng et~al.(2024{\natexlab{2}})Feng, Ding, Wang, Zhuang, Wang, Qin, Zhao, Yao, Zhang, and Chen]{feng2024sciknoweval}
Kehua Feng, Keyan Ding, Weijie Wang, Xiang Zhuang, Zeyuan Wang, Ming Qin, Yu~Zhao, Jianhua Yao, Qiang Zhang, and Huajun Chen.
\newblock Sciknoweval: Evaluating multi-level scientific knowledge of large language models.
\newblock \emph{arXiv preprint arXiv:2406.09098}, 2024{\natexlab{2}}.

\bibitem[Feng et~al.(2024{\natexlab{3}})Feng, Pu, Latzke, August, Siangliulue, Bragg, Weld, Zhang, and Chang]{feng2024cocoa}
KJ~Feng, Kevin Pu, Matt Latzke, Tal August, Pao Siangliulue, Jonathan Bragg, Daniel~S Weld, Amy~X Zhang, and Joseph~Chee Chang.
\newblock Cocoa: Co-planning and co-execution with ai agents.
\newblock \emph{arXiv preprint arXiv:2412.10999}, 2024{\natexlab{3}}.

\bibitem[Feng et~al.(2025{\natexlab{2}})Feng, Liang, Yin, Gao, and Wang]{feng2025agentic}
Ruozhu Feng, Yangang Liang, Tianzhixi Yin, Peiyuan Gao, and Wei Wang.
\newblock Agentic assistant for material scientists.
\newblock Apr 2025{\natexlab{2}}.

\bibitem[Feng et~al.(2025{\natexlab{3}})Feng, Sun, and You]{feng2025grapheval}
Tao Feng, Yihang Sun, and Jiaxuan You.
\newblock Grapheval: A lightweight graph-based llm framework for idea evaluation.
\newblock \emph{arXiv preprint arXiv:2503.12600}, 2025{\natexlab{3}}.

\bibitem[Fernandes et~al.(2024)Fernandes, Guedes, Laitz, Almeida, Nogueira, Lotufo, and Pereira]{fernandes2024surveysum}
Leandro~Car{\'\i}sio Fernandes, Gustavo~Bartz Guedes, Thiago~Soares Laitz, Thales~Sales Almeida, Rodrigo Nogueira, Roberto Lotufo, and Jayr Pereira.
\newblock Surveysum: A dataset for summarizing multiple scientific articles into a survey section.
\newblock In \emph{Brazilian Conference on Intelligent Systems}, pages 431--444. Springer, Jan 2024.

\bibitem[Ferrara(2023)]{ferrara2023fairness}
Emilio Ferrara.
\newblock Fairness and bias in artificial intelligence: A brief survey of sources, impacts, and mitigation strategies.
\newblock \emph{Sci}, 6\penalty0 (1):\penalty0 3, Dec 2023.

\bibitem[{Financial Times}(2024)]{ft2024_deepmind_biontech}
{Financial Times}.
\newblock Deepmind and biontech build ai lab assistants for scientific research.
\newblock \emph{Financial Times}, Oct 2024.
\newblock URL \url{https://www.ft.com/content/64b1bb33-095e-4cc5-a911-50df76fa3d1d}.

\bibitem[First et~al.(2023)First, Rabe, Ringer, and Brun]{first2023baldur}
Emily First, Markus~N Rabe, Talia Ringer, and Yuriy Brun.
\newblock Baldur: Whole-proof generation and repair with large language models.
\newblock In \emph{Proceedings of the 31st ACM Joint European Software Engineering Conference and Symposium on the Foundations of Software Engineering}, pages 1229--1241, Nov 2023.

\bibitem[Fonseca and Cohen(2024)]{fonseca-cohen-2024-large-language}
Marcio Fonseca and Shay Cohen.
\newblock Can large language model summarizers adapt to diverse scientific communication goals?
\newblock In Lun-Wei Ku, Andre Martins, and Vivek Srikumar, editors, \emph{Findings of the Association for Computational Linguistics: ACL 2024}, pages 8599--8618, Bangkok, Thailand, August 2024. Association for Computational Linguistics.
\newblock \doi{10.18653/v1/2024.findings-acl.508}.
\newblock URL \url{https://aclanthology.org/2024.findings-acl.508/}.

\bibitem[Formal et~al.(2021)Formal, Lassance, Piwowarski, and Clinchant]{formal2021splade}
Thibault Formal, Carlos Lassance, Benjamin Piwowarski, and St{\'e}phane Clinchant.
\newblock Splade v2: Sparse lexical and expansion model for information retrieval.
\newblock \emph{arXiv preprint arXiv:2109.10086}, 2021.

\bibitem[Frasca et~al.(2024)Frasca, La~Torre, Pravettoni, and Cutica]{frasca2024explainable}
Maria Frasca, Davide La~Torre, Gabriella Pravettoni, and Ilaria Cutica.
\newblock Explainable and interpretable artificial intelligence in medicine: a systematic bibliometric review.
\newblock \emph{Discover Artificial Intelligence}, 4\penalty0 (1):\penalty0 15, Feb 2024.

\bibitem[Frieder et~al.(2024)Frieder, Bayer, Collins, Berner, Loader, Juh{\'a}sz, Ruehle, Welleck, Poesia, Griffiths, et~al.]{frieder2024data}
Simon Frieder, Jonas Bayer, Katherine~M Collins, Julius Berner, Jacob Loader, Andr{\'a}s Juh{\'a}sz, Fabian Ruehle, Sean Welleck, Gabriel Poesia, Ryan-Rhys Griffiths, et~al.
\newblock Data for mathematical copilots: Better ways of presenting proofs for machine learning.
\newblock \emph{arXiv preprint arXiv:2412.15184}, 2024.

\bibitem[Frontull and Moser(2024)]{frontull2024rule}
Samuel Frontull and Georg Moser.
\newblock Rule-based, neural and llm back-translation: Comparative insights from a variant of ladin.
\newblock \emph{arXiv preprint arXiv:2407.08819}, 2024.

\bibitem[Fu et~al.(2025)Fu, Luo, Nan, and Li]{fu2025peer}
Yongfan Fu, Jian Luo, Guofang Nan, and Dahui Li.
\newblock Peer review expert group recommendation: A multi-subject coverage-based approach.
\newblock \emph{Expert Systems with Applications}, 264:\penalty0 125971, Mar 2025.

\bibitem[Galli et~al.(2024)Galli, Moretti, and Calciolari]{galli2024intelligent}
Carlo Galli, Chiara Moretti, and Elena Calciolari.
\newblock Intelligent summaries: Will artificial intelligence mark the finale for biomedical literature reviews?
\newblock \emph{Learned Publishing}, Dec 2024.

\bibitem[Gandhi et~al.(2025)Gandhi, Shah, Patwardhan, Vig, and Shroff]{gandhi2025researchcodeagent}
Shubham Gandhi, Dhruv Shah, Manasi Patwardhan, Lovekesh Vig, and Gautam Shroff.
\newblock Researchcodeagent: An llm multi-agent system for automated codification of research methodologies.
\newblock \emph{arXiv preprint arXiv:2504.20117}, 2025.

\bibitem[Ganguly et~al.(2025)Ganguly, Johri, Ali, and McDonald]{ganguly2025generative}
Amrita Ganguly, Aditya Johri, Areej Ali, and Nora McDonald.
\newblock Generative artificial intelligence for academic research: evidence from guidance issued for researchers by higher education institutions in the united states.
\newblock \emph{AI and Ethics}, pages 1--17, Mar 2025.

\bibitem[Ganguly et~al.(2025{\natexlab{2}})Ganguly, Singh, Sankar, Zhang, Zhang, Iyengar, Han, Sharma, Kalyanaraman, and Chaudhary]{ganguly2025grammars}
Debargha Ganguly, Vikash Singh, Sreehari Sankar, Biyao Zhang, Xuecen Zhang, Srinivasan Iyengar, Xiaotian Han, Amit Sharma, Shivkumar Kalyanaraman, and Vipin Chaudhary.
\newblock Grammars of formal uncertainty: When to trust llms in automated reasoning tasks.
\newblock \emph{arXiv preprint arXiv:2505.20047}, 2025{\natexlab{2}}.

\bibitem[Gangwal et~al.(2024)Gangwal, Ansari, Ahmad, Azad, and Sulaiman]{gangwal2024current}
Amit Gangwal, Azim Ansari, Iqrar Ahmad, Abul~Kalam Azad, and Wan Mohd Azizi~Wan Sulaiman.
\newblock Current strategies to address data scarcity in artificial intelligence-based drug discovery: A comprehensive review.
\newblock \emph{Computers in Biology and Medicine}, 179:\penalty0 108734, Sep 2024.

\bibitem[Gao et~al.(2025)Gao, Ruan, Gao, Liu, and Fu]{gao2025reviewagents}
Xian Gao, Jiacheng Ruan, Jingsheng Gao, Ting Liu, and Yuzhuo Fu.
\newblock Reviewagents: Bridging the gap between human and ai-generated paper reviews.
\newblock \emph{arXiv preprint arXiv:2503.08506}, 2025.

\bibitem[Gao et~al.(2025{\natexlab{2}})Gao, Zhang, Xie, Liu, and Fu]{gao2025graph}
Xian Gao, Zongyun Zhang, Mingye Xie, Ting Liu, and Yuzhuo Fu.
\newblock Graph of ai ideas: Leveraging knowledge graphs and llms for ai research idea generation.
\newblock \emph{arXiv preprint arXiv:2503.08549}, 2025{\natexlab{2}}.

\bibitem[Garcia-Silva et~al.(2022)Garcia-Silva, Berrio, Gomez-Perez, Mart{\'\i}nez-Heras, Donati, and Roma]{garcia2022spaceqa}
Andres Garcia-Silva, Cristian Berrio, Jose~Manuel Gomez-Perez, Jose~Antonio Mart{\'\i}nez-Heras, Alessandro Donati, and Ilaria Roma.
\newblock Spaceqa: Answering questions about the design of space missions and space craft concepts.
\newblock In \emph{Proceedings of the 45th International ACM SIGIR Conference on Research and Development in Information Retrieval}, pages 3306--3311, Jul 2022.

\bibitem[Garijo et~al.(2025)Garijo, Yang, Vargas, Gadewar, Low, Ratnakar, Osorio, Zhu, McMahon, Gil, et~al.]{garijo2025neurodisk}
Daniel Garijo, Qifan Yang, Hern{\'a}n Vargas, Shruti~P Gadewar, Kevin Low, Varun Ratnakar, Maximiliano Osorio, Alyssa~H Zhu, Agnes McMahon, Yolanda Gil, et~al.
\newblock Neurodisk: An ai approach to automate continuous inquiry-driven discoveries in neuroimaging genetics.
\newblock \emph{bioRxiv}, Feb 2025.

\bibitem[Garikaparthi et~al.(2025)Garikaparthi, Patwardhan, Kanade, Hassan, Vig, and Cohan]{garikaparthi2025mir}
Aniketh Garikaparthi, Manasi Patwardhan, Aditya~Sanjiv Kanade, Aman Hassan, Lovekesh Vig, and Arman Cohan.
\newblock Mir: Methodology inspiration retrieval for scientific research problems.
\newblock \emph{arXiv preprint arXiv:2506.00249}, 2025.

\bibitem[Garikaparthi et~al.(2025{\natexlab{2}})Garikaparthi, Patwardhan, Vig, and Cohan]{garikaparthi2025iris}
Aniketh Garikaparthi, Manasi Patwardhan, Lovekesh Vig, and Arman Cohan.
\newblock Iris: Interactive research ideation system for accelerating scientific discovery.
\newblock \emph{arXiv preprint arXiv:2504.16728}, 2025{\natexlab{2}}.

\bibitem[Ge et~al.(2021)Ge, Dinh, Liu, Su, Lu, Wang, and Diesner]{ge2021baco}
Yubin Ge, Ly~Dinh, Xiaofeng Liu, Jinsong Su, Ziyao Lu, Ante Wang, and Jana Diesner.
\newblock Baco: A background knowledge-and content-based framework for citing sentence generation.
\newblock In \emph{Proceedings of the 59th Annual Meeting of the Association for Computational Linguistics and the 11th International Joint Conference on Natural Language Processing (Volume 1: Long Papers)}, pages 1466--1478, Aug 2021.

\bibitem[Geng and Trotta(2025)]{geng2025human}
Mingmeng Geng and Roberto Trotta.
\newblock Human-llm coevolution: Evidence from academic writing.
\newblock \emph{arXiv preprint arXiv:2502.09606}, 2025.

\bibitem[George(2024)]{george2024paperpal}
Elizabeth~Oommen George.
\newblock How paperpal enhances english writing quality and improves productivity for japanese academics.
\newblock Aug 2024.

\bibitem[Gero et~al.(2022)Gero, Liu, and Chilton]{gero2022sparks}
Katy~Ilonka Gero, Vivian Liu, and Lydia Chilton.
\newblock Sparks: Inspiration for science writing using language models.
\newblock In \emph{Proceedings of the 2022 ACM Designing Interactive Systems Conference}, pages 1002--1019, Jun 2022.

\bibitem[Ghafarollahi and Buehler(2024)]{ghafarollahi2024protagents}
Alireza Ghafarollahi and Markus~J Buehler.
\newblock Protagents: protein discovery via large language model multi-agent collaborations combining physics and machine learning.
\newblock \emph{Digital Discovery}, 3\penalty0 (7):\penalty0 1389--1409, May 2024.

\bibitem[Ghafarollahi and Buehler(2024{\natexlab{2}})]{ghafarollahi2024sciagents}
Alireza Ghafarollahi and Markus~J Buehler.
\newblock Sciagents: Automating scientific discovery through multi-agent intelligent graph reasoning.
\newblock \emph{arXiv preprint arXiv:2409.05556}, 2024{\natexlab{2}}.

\bibitem[Ghafarollahi and Buehler(2025)]{ghafarollahi2025sparks}
Alireza Ghafarollahi and Markus~J Buehler.
\newblock Sparks: Multi-agent artificial intelligence model discovers protein design principles.
\newblock \emph{arXiv preprint arXiv:2504.19017}, 2025.

\bibitem[Ghareeb et~al.()Ghareeb, Chang, Mitchener, Yiu, Szostkiewicz, Laurent, Razzak, White, Hinks, and Rodriques]{ghareeb2025robin}
Ali~Essam Ghareeb, Benjamin Chang, Ludovico Mitchener, Angela Yiu, Caralyn~J Szostkiewicz, Jon~M Laurent, Muhammed~T Razzak, Andrew~D White, Michaela~M Hinks, and Samuel~G Rodriques.
\newblock Robin: A multi-agent system for automating scientific discovery.
\newblock \emph{arXiv preprint arXiv:2505.13400}.

\bibitem[Gharizadeh et~al.(2024)Gharizadeh, Abbasi, Ghareyazi, Mofrad, and Rabiee]{gharizadeh2024hgtdr}
Ali Gharizadeh, Karim Abbasi, Amin Ghareyazi, Mohammad~RK Mofrad, and Hamid~R Rabiee.
\newblock Hgtdr: Advancing drug repurposing with heterogeneous graph transformers.
\newblock \emph{Bioinformatics}, 40\penalty0 (7):\penalty0 btae349, Jul 2024.

\bibitem[Ghosal et~al.(2022)Ghosal, Kumar, Bharti, and Ekbal]{ghosal2022peer}
Tirthankar Ghosal, Sandeep Kumar, Prabhat~Kumar Bharti, and Asif Ekbal.
\newblock Peer review analyze: A novel benchmark resource for computational analysis of peer reviews.
\newblock \emph{Plos one}, 17\penalty0 (1):\penalty0 e0259238, Jan 2022.

\bibitem[Ghoshal et~al.(2022)Ghoshal, Iyer, Paranjape, Lakhotia, Yih, and Mehdad]{ghoshal2022quaser}
Asish Ghoshal, Srinivasan Iyer, Bhargavi Paranjape, Kushal Lakhotia, Scott Wen-tau Yih, and Yashar Mehdad.
\newblock Quaser: Question answering with scalable extractive rationalization.
\newblock In \emph{Proceedings of the 45th International ACM SIGIR Conference on Research and Development in Information Retrieval}, pages 1208--1218, Jul 2022.

\bibitem[Girotra et~al.(2023)Girotra, Meincke, Terwiesch, and Ulrich]{girotra2023ideas}
Karan Girotra, Lennart Meincke, Christian Terwiesch, and Karl~T Ulrich.
\newblock Ideas are dimes a dozen: Large language models for idea generation in innovation.
\newblock \emph{The Wharton School Research Paper Forthcoming}, Jul 2023.

\bibitem[Glickman and Sharot(2025)]{glickman2025human}
Moshe Glickman and Tali Sharot.
\newblock How human--ai feedback loops alter human perceptual, emotional and social judgements.
\newblock \emph{Nature Human Behaviour}, 9\penalty0 (2):\penalty0 345--359, 2025.

\bibitem[Glockner et~al.(2022)Glockner, Hou, and Gurevych]{glockner2022missing}
Max Glockner, Yufang Hou, and Iryna Gurevych.
\newblock Missing counter-evidence renders nlp fact-checking unrealistic for misinformation.
\newblock \emph{arXiv preprint arXiv:2210.13865}, 2022.

\bibitem[Glockner et~al.(2024)Glockner, Hou, Nakov, and Gurevych]{glockner2024grounding}
Max Glockner, Yufang Hou, Preslav Nakov, and Iryna Gurevych.
\newblock Grounding fallacies misrepresenting scientific publications in evidence.
\newblock \emph{arXiv preprint arXiv:2408.12812}, 2024.

\bibitem[Gokdemir et~al.(2025)Gokdemir, Siebenschuh, Brace, Wells, Hsu, Hippe, Setty, Ajith, Pauloski, Sastry, et~al.]{gokdemir2025hiperrag}
Ozan Gokdemir, Carlo Siebenschuh, Alexander Brace, Azton Wells, Brian Hsu, Kyle Hippe, Priyanka~V Setty, Aswathy Ajith, J~Gregory Pauloski, Varuni Sastry, et~al.
\newblock Hiperrag: High-performance retrieval augmented generation for scientific insights.
\newblock \emph{arXiv preprint arXiv:2505.04846}, Jun 2025.

\bibitem[Goli and Singh(2024)]{goli2024can}
Ali Goli and Amandeep Singh.
\newblock Frontiers: Can large language models capture human preferences?
\newblock \emph{Marketing Science}, 43\penalty0 (4):\penalty0 709--722, Apr 2024.
\newblock \doi{10.1287/mksc.2023.0306}.
\newblock URL \url{https://doi.org/10.1287/mksc.2023.0306}.

\bibitem[Gomez et~al.(2025)Gomez, Cho, Ke, Huang, and Unberath]{gomez2025human}
Catalina Gomez, Sue~Min Cho, Shichang Ke, Chien-Ming Huang, and Mathias Unberath.
\newblock Human-ai collaboration is not very collaborative yet: A taxonomy of interaction patterns in ai-assisted decision making from a systematic review.
\newblock \emph{Frontiers in Computer Science}, 6:\penalty0 1521066, Jan 2025.

\bibitem[G{\'o}mez-Bombarelli et~al.(2018)G{\'o}mez-Bombarelli, Wei, Duvenaud, Hern{\'a}ndez-Lobato, S{\'a}nchez-Lengeling, Sheberla, Aguilera-Iparraguirre, Hirzel, Adams, and Aspuru-Guzik]{gomez2018automatic}
Rafael G{\'o}mez-Bombarelli, Jennifer~N Wei, David Duvenaud, Jos{\'e}~Miguel Hern{\'a}ndez-Lobato, Benjam{\'\i}n S{\'a}nchez-Lengeling, Dennis Sheberla, Jorge Aguilera-Iparraguirre, Timothy~D Hirzel, Ryan~P Adams, and Al{\'a}n Aspuru-Guzik.
\newblock Automatic chemical design using a data-driven continuous representation of molecules.
\newblock \emph{ACS central science}, 4\penalty0 (2):\penalty0 268--276, Jan 2018.

\bibitem[Gomez-Perez and Ortega(2019)]{gomez2019look}
Jose~Manuel Gomez-Perez and Raul Ortega.
\newblock Look, read and enrich-learning from scientific figures and their captions.
\newblock In \emph{Proceedings of the 10th International Conference on Knowledge Capture}, pages 101--108, Sep 2019.

\bibitem[Gonz{\'a}lez-Sendino et~al.(2024)Gonz{\'a}lez-Sendino, Serrano, and Bajo]{gonzalez2024mitigating}
Rub{\'e}n Gonz{\'a}lez-Sendino, Emilio Serrano, and Javier Bajo.
\newblock Mitigating bias in artificial intelligence: Fair data generation via causal models for transparent and explainable decision-making.
\newblock \emph{Future Generation Computer Systems}, 155:\penalty0 384--401, Jun 2024.

\bibitem[Goodsell and Yli-Vakkuri(2024)]{goodsell2024lf}
Zachary Goodsell and Juhani Yli-Vakkuri.
\newblock Lf: a foundational higher-order-logic.
\newblock \emph{arXiv preprint arXiv:2401.11050}, 2024.

\bibitem[Gosmar and Dahl(2025)]{gosmar2025hallucination}
Diego Gosmar and Deborah~A Dahl.
\newblock Hallucination mitigation using agentic ai natural language-based frameworks.
\newblock \emph{arXiv preprint arXiv:2501.13946}, 2025.

\bibitem[Gottweis et~al.(2025)Gottweis, Weng, Daryin, Tu, Palepu, Sirkovic, Myaskovsky, Weissenberger, Rong, Tanno, et~al.]{gottweis2025towards}
Juraj Gottweis, Wei-Hung Weng, Alexander Daryin, Tao Tu, Anil Palepu, Petar Sirkovic, Artiom Myaskovsky, Felix Weissenberger, Keran Rong, Ryutaro Tanno, et~al.
\newblock Towards an ai co-scientist.
\newblock \emph{arXiv preprint arXiv:2502.18864}, 2025.

\bibitem[Grattafiori et~al.(2024)Grattafiori, Dubey, Jauhri, Pandey, Kadian, Al-Dahle, Letman, Mathur, Schelten, Vaughan, et~al.]{grattafiori2024llama}
Aaron Grattafiori, Abhimanyu Dubey, Abhinav Jauhri, Abhinav Pandey, Abhishek Kadian, Ahmad Al-Dahle, Aiesha Letman, Akhil Mathur, Alan Schelten, Alex Vaughan, et~al.
\newblock The llama 3 herd of models.
\newblock \emph{arXiv preprint arXiv:2407.21783}, 2024.

\bibitem[Greydanus et~al.(2019)Greydanus, Dzamba, and Yosinski]{greydanus2019hamiltonian}
Samuel Greydanus, Misko Dzamba, and Jason Yosinski.
\newblock Hamiltonian neural networks.
\newblock \emph{Advances in Neural Information Processing Systems}, 32, Jul 2019.

\bibitem[Gridach et~al.(2025)Gridach, Nanavati, Abidine, Mendes, and Mack]{gridach2025agentic}
Mourad Gridach, Jay Nanavati, Khaldoun Zine~El Abidine, Lenon Mendes, and Christina Mack.
\newblock Agentic ai for scientific discovery: A survey of progress, challenges, and future directions.
\newblock \emph{arXiv preprint arXiv:2503.08979}, 2025.

\bibitem[Grosnit et~al.(2024)Grosnit, Maraval, Doran, Paolo, Thomas, Beevi, Gonzalez, Khandelwal, Iacobacci, Benechehab, et~al.]{grosnit2024large}
Antoine Grosnit, Alexandre Maraval, James Doran, Giuseppe Paolo, Albert Thomas, Refinath Shahul Hameed~Nabeezath Beevi, Jonas Gonzalez, Khyati Khandelwal, Ignacio Iacobacci, Abdelhakim Benechehab, et~al.
\newblock Large language models orchestrating structured reasoning achieve kaggle grandmaster level.
\newblock \emph{arXiv preprint arXiv:2411.03562}, 2024.

\bibitem[Gu et~al.(2024)Gu, Shang, Jiang, Kuang, Lin, Lyu, Mao, Pan, Wu, Yu, et~al.]{gu2024blade}
Ken Gu, Ruoxi Shang, Ruien Jiang, Keying Kuang, Richard-John Lin, Donghe Lyu, Yue Mao, Youran Pan, Teng Wu, Jiaqian Yu, et~al.
\newblock Blade: Benchmarking language model agents for data-driven science.
\newblock \emph{arXiv preprint arXiv:2408.09667}, 2024.

\bibitem[Gu and Hahnloser(2022)]{gu2022controllable}
Nianlong Gu and Richard~HR Hahnloser.
\newblock Controllable citation sentence generation with language models.
\newblock \emph{arXiv preprint arXiv:2211.07066}, 2022.

\bibitem[Gu et~al.(2024{\natexlab{2}})Gu, Wang, Zhang, and Li]{gu2024llms}
Tianyang Gu, Jingjin Wang, Zhihao Zhang, and HaoHong Li.
\newblock Llms can realize combinatorial creativity: generating creative ideas via llms for scientific research.
\newblock \emph{arXiv preprint arXiv:2412.14141}, 2024{\natexlab{2}}.

\bibitem[Gu and Krenn(2024)]{gu2024generation}
Xuemei Gu and Mario Krenn.
\newblock Generation and human-expert evaluation of interesting research ideas using knowledge graphs and large language models.
\newblock \emph{arXiv preprint arXiv:2405.17044}, 2024.

\bibitem[Guan et~al.(2025)Guan, Inchai, Fang, Law, Brito, Pawlosky, GottWeis, Daryin, Myaskovsky, Palepu, et~al.]{guan2025ai}
Yuan Guan, Jakkapong Inchai, Zhuoqing Fang, Jacky Law, Alberto Alonzo~Garcia Brito, Annalisa Pawlosky, Juraj GottWeis, Alexander Daryin, Artiom Myaskovsky, Anil Palepu, et~al.
\newblock Ai-assisted drug re-purposing for human liver fibrosis.
\newblock \emph{bioRxiv}, pages 2025--04, May 2025.

\bibitem[Guendouzi et~al.(2023)Guendouzi, Ouchani, Assaad, and Zaher]{guendouzi2023systematic}
Badra~Souhila Guendouzi, Samir Ouchani, Hiba~EL Assaad, and Madeleine~EL Zaher.
\newblock A systematic review of federated learning: Challenges, aggregation methods, and development tools.
\newblock \emph{Journal of Network and Computer Applications}, 220:\penalty0 103714, Nov 2023.

\bibitem[Guerrier et~al.(2025)Guerrier, Soma, Fouad, and Beltrame]{guerrier2025guided}
Maeva Guerrier, Karthik Soma, Hassan Fouad, and Giovanni Beltrame.
\newblock Guided by guardrails: Control barrier functions as safety instructors for robotic learning.
\newblock \emph{arXiv preprint arXiv:2505.18858}, 2025.

\bibitem[Guo et~al.(2024)Guo, Zhu, Yang, Xie, Dong, Zhang, Chen, Bi, Wu, Li, et~al.]{guo2024deepseek}
Daya Guo, Qihao Zhu, Dejian Yang, Zhenda Xie, Kai Dong, Wentao Zhang, Guanting Chen, Xiao Bi, Yu~Wu, YK~Li, et~al.
\newblock Deepseek-coder: When the large language model meets programming--the rise of code intelligence.
\newblock \emph{arXiv preprint arXiv:2401.14196}, 2024.

\bibitem[Guo et~al.(2025)Guo, Yang, Zhang, Song, Zhang, Xu, Zhu, Ma, Wang, Bi, et~al.]{guo2025deepseek}
Daya Guo, Dejian Yang, Haowei Zhang, Junxiao Song, Ruoyu Zhang, Runxin Xu, Qihao Zhu, Shirong Ma, Peiyi Wang, Xiao Bi, et~al.
\newblock Deepseek-r1: Incentivizing reasoning capability in llms via reinforcement learning.
\newblock \emph{arXiv preprint arXiv:2501.12948}, 2025.

\bibitem[Guo et~al.(2024{\natexlab{2}})Guo, Deng, Wen, Chen, Chang, and Wang]{guo2024ds}
Siyuan Guo, Cheng Deng, Ying Wen, Hechang Chen, Yi~Chang, and Jun Wang.
\newblock Ds-agent: Automated data science by empowering large language models with case-based reasoning.
\newblock \emph{arXiv preprint arXiv:2402.17453}, 2024{\natexlab{2}}.

\bibitem[Guo et~al.(2021)Guo, Qiu, Wang, and Cohen]{guo2021automated}
Yue Guo, Wei Qiu, Yizhong Wang, and Trevor Cohen.
\newblock Automated lay language summarization of biomedical scientific reviews.
\newblock In \emph{Proceedings of the AAAI Conference on Artificial Intelligence}, volume~35, pages 160--168, May 2021.

\bibitem[Guo et~al.(2025{\natexlab{2}})Guo, Zhang, Chen, Gao, Jiang, Wang, and Heng]{guo2025sciverse}
Ziyu Guo, Ray Zhang, Hao Chen, Jialin Gao, Dongzhi Jiang, Jiaze Wang, and Pheng-Ann Heng.
\newblock Sciverse: Unveiling the knowledge comprehension and visual reasoning of lmms on multi-modal scientific problems.
\newblock \emph{arXiv preprint arXiv:2503.10627}, 2025{\natexlab{2}}.

\bibitem[Gupta et~al.(2021)Gupta, Srivastava, Sahu, Tiwari, Ambasta, and Kumar]{gupta2021artificial}
Rohan Gupta, Devesh Srivastava, Mehar Sahu, Swati Tiwari, Rashmi~K Ambasta, and Pravir Kumar.
\newblock Artificial intelligence to deep learning: machine intelligence approach for drug discovery.
\newblock \emph{Molecular diversity}, 25:\penalty0 1315--1360, Apr 2021.

\bibitem[Gupta and Pruthi(2025)]{gupta2025all}
Tarun Gupta and Danish Pruthi.
\newblock All that glitters is not novel: Plagiarism in ai generated research.
\newblock \emph{arXiv preprint arXiv:2502.16487}, 2025.

\bibitem[GX-Chen et~al.(2025)GX-Chen, Lin, Samiei, Precup, Richards, Fergus, and Marino]{gx2025language}
Anthony GX-Chen, Dongyan Lin, Mandana Samiei, Doina Precup, Blake~A Richards, Rob Fergus, and Kenneth Marino.
\newblock Language agents mirror human causal reasoning biases. how can we help them think like scientists?
\newblock \emph{arXiv preprint arXiv:2505.09614}, 2025.

\bibitem[Haarmann(2025)]{liu2025enhance}
Hendrik Haarmann.
\newblock Enhance innovation by boosting idea generation with large language models.
\newblock \emph{INFORMS Journal on Computing}, Jul 2025.

\bibitem[Hagendorff et~al.(2023)Hagendorff, Fabi, and Kosinski]{Hagendorff_2023}
Thilo Hagendorff, Sarah Fabi, and Michal Kosinski.
\newblock Human-like intuitive behavior and reasoning biases emerged in large language models but disappeared in chatgpt.
\newblock \emph{Nature Computational Science}, 3\penalty0 (10):\penalty0 833--838, October 2023.
\newblock ISSN 2662-8457.
\newblock \doi{10.1038/s43588-023-00527-x}.
\newblock URL \url{http://dx.doi.org/10.1038/s43588-023-00527-x}.

\bibitem[Hammond et~al.(2025)Hammond, Chan, Clifton, Hoelscher-Obermaier, Khan, McLean, Smith, Barfuss, Foerster, Gaven{\v{c}}iak, et~al.]{hammond2025multi}
Lewis Hammond, Alan Chan, Jesse Clifton, Jason Hoelscher-Obermaier, Akbir Khan, Euan McLean, Chandler Smith, Wolfram Barfuss, Jakob Foerster, Tom{\'a}{\v{s}} Gaven{\v{c}}iak, et~al.
\newblock Multi-agent risks from advanced ai.
\newblock \emph{arXiv preprint arXiv:2502.14143}, 2025.

\bibitem[Hangya et~al.(2022)Hangya, Saadi, and Fraser]{hangya2022improving}
Viktor Hangya, Hossain~Shaikh Saadi, and Alexander Fraser.
\newblock Improving low-resource languages in pre-trained multilingual language models.
\newblock In \emph{Proceedings of the 2022 Conference on Empirical Methods in Natural Language Processing}, pages 11993--12006, Jan 2022.

\bibitem[Hanna et~al.(2025)Hanna, Pantanowitz, Jackson, Palmer, Visweswaran, Pantanowitz, Deebajah, and Rashidi]{hanna2025ethical}
Matthew~G Hanna, Liron Pantanowitz, Brian Jackson, Octavia Palmer, Shyam Visweswaran, Joshua Pantanowitz, Mustafa Deebajah, and Hooman~H Rashidi.
\newblock Ethical and bias considerations in artificial intelligence/machine learning.
\newblock \emph{Modern Pathology}, 38\penalty0 (3):\penalty0 100686, Mar 2025.

\bibitem[Hao et~al.(2024)Hao, Fan, Xu, Yuan, and Li]{hao2024hlm}
Qianyue Hao, Jingyang Fan, Fengli Xu, Jian Yuan, and Yong Li.
\newblock Hlm-cite: Hybrid language model workflow for text-based scientific citation prediction.
\newblock \emph{arXiv preprint arXiv:2410.09112}, 2024.

\bibitem[Hao et~al.(2023)Hao, Liu, Wang, and Hu]{hao2023toolkengpt}
Shibo Hao, Tianyang Liu, Zhen Wang, and Zhiting Hu.
\newblock Toolkengpt: Augmenting frozen language models with massive tools via tool embeddings.
\newblock \emph{Advances in Neural Information Processing Systems}, 36:\penalty0 45870--45894, Dec 2023.

\bibitem[Harris(2025)]{harris2025airus}
Kenneth~D Harris.
\newblock Airus: a simple workflow for ai-assisted exploration of scientific data.
\newblock \emph{bioRxiv}, pages 2025--02, Feb 2025.

\bibitem[Hartley(2024)]{hartley2024efficacy}
Russell Hartley.
\newblock Efficacy analysis of online artificial intelligence fact-checking tools.
\newblock \emph{The International Review of Information Ethics}, 33\penalty0 (1), Apr 2024.

\bibitem[Hashemi et~al.(2024)Hashemi, Eisner, Rosset, Van~Durme, and Kedzie]{hashemi-etal-2024-llm}
Helia Hashemi, Jason Eisner, Corby Rosset, Benjamin Van~Durme, and Chris Kedzie.
\newblock {LLM}-rubric: A multidimensional, calibrated approach to automated evaluation of natural language texts.
\newblock In Lun-Wei Ku, Andre Martins, and Vivek Srikumar, editors, \emph{Proceedings of the 62nd Annual Meeting of the Association for Computational Linguistics (Volume 1: Long Papers)}, pages 13806--13834, Bangkok, Thailand, August 2024. Association for Computational Linguistics.
\newblock \doi{10.18653/v1/2024.acl-long.745}.
\newblock URL \url{https://aclanthology.org/2024.acl-long.745/}.

\bibitem[Hassija et~al.(2024)Hassija, Chamola, Mahapatra, Singal, Goel, Huang, Scardapane, Spinelli, Mahmud, and Hussain]{hassija2024interpreting}
Vikas Hassija, Vinay Chamola, Atmesh Mahapatra, Abhinandan Singal, Divyansh Goel, Kaizhu Huang, Simone Scardapane, Indro Spinelli, Mufti Mahmud, and Amir Hussain.
\newblock Interpreting black-box models: a review on explainable artificial intelligence.
\newblock \emph{Cognitive Computation}, 16\penalty0 (1):\penalty0 45--74, Aug 2024.

\bibitem[Hatakeyama-Sato et~al.(2025)Hatakeyama-Sato, Nishida, Kitamura, Ushiku, Takahashi, Nabae, and Hayakawa]{hatakeyama2025perspective}
Kan Hatakeyama-Sato, Toshihiko Nishida, Kenta Kitamura, Yoshitaka Ushiku, Koichi Takahashi, Yuta Nabae, and Teruaki Hayakawa.
\newblock Perspective on utilizing foundation models for laboratory automation in materials research.
\newblock \emph{arXiv preprint arXiv:2506.12312}, 2025.

\bibitem[Haworth et~al.(2024)Haworth, Biswas, Opfermann, Kam, Wang, Pantalone, Creighton, Yang, Kang, and Krieger]{haworth2024autonomous}
Jesse Haworth, Rishi Biswas, Justin Opfermann, Michael Kam, Yaning Wang, Desire Pantalone, Francis~X Creighton, Robin Yang, Jin~U Kang, and Axel Krieger.
\newblock Autonomous robotic system with optical coherence tomography guidance for vascular anastomosis.
\newblock \emph{arXiv preprint arXiv:2410.07493}, 2024.

\bibitem[Hayes et~al.(2025)Hayes, Rao, Akin, Sofroniew, Oktay, Lin, Verkuil, Tran, Deaton, Wiggert, et~al.]{hayes2025simulating}
Thomas Hayes, Roshan Rao, Halil Akin, Nicholas~J Sofroniew, Deniz Oktay, Zeming Lin, Robert Verkuil, Vincent~Q Tran, Jonathan Deaton, Marius Wiggert, et~al.
\newblock Simulating 500 million years of evolution with a language model.
\newblock \emph{Science}, page eads0018, Jan 2025.

\bibitem[He and Chen(2025)]{he2025reasoning}
Kaiyu He and Zhiyu Chen.
\newblock From reasoning to learning: A survey on hypothesis discovery and rule learning with large language models.
\newblock \emph{arXiv preprint arXiv:2505.21935}, 2025.

\bibitem[He et~al.(2025)He, Huang, Feng, Lin, Zhang, Li, et~al.]{he2025pasa}
Yichen He, Guanhua Huang, Peiyuan Feng, Yuan Lin, Yuchen Zhang, Hang Li, et~al.
\newblock Pasa: An llm agent for comprehensive academic paper search.
\newblock \emph{arXiv preprint arXiv:2501.10120}, 2025.

\bibitem[Heger et~al.(2024)Heger, Algergawy, Brinner, Jeschke, K{\"o}nig-Ries, Mietchen, and Zarrie{\ss}]{heger2024natural}
Tina Heger, Alsayed Algergawy, Marc Brinner, Jonathan~M Jeschke, Birgitta K{\"o}nig-Ries, Daniel Mietchen, and Sina Zarrie{\ss}.
\newblock Natural language hypotheses in scientific papers and how to tame them: Suggested steps for formalizing complex scientific claims.
\newblock In \emph{Conference on Advances in Robust Argumentation Machines}, pages 3--19. Springer Nature Switzerland Cham, Jun 2024.

\bibitem[Heinz et~al.(2025)Heinz, Mackin, Trudeau, Bhattacharya, Wang, Banta, Jewett, Salzhauer, Griffin, and Jacobson]{heinz2025randomized}
Michael~V Heinz, Daniel~M Mackin, Brianna~M Trudeau, Sukanya Bhattacharya, Yinzhou Wang, Haley~A Banta, Abi~D Jewett, Abigail~J Salzhauer, Tess~Z Griffin, and Nicholas~C Jacobson.
\newblock Randomized trial of a generative ai chatbot for mental health treatment.
\newblock \emph{Nejm Ai}, 2\penalty0 (4):\penalty0 AIoa2400802, Mar 2025.

\bibitem[Henry and McInnes(2017)]{henry2017literature}
Sam Henry and Bridget~T McInnes.
\newblock Literature based discovery: models, methods, and trends.
\newblock \emph{Journal of biomedical informatics}, 74:\penalty0 20--32, Oct 2017.

\bibitem[Herron et~al.(2024)Herron, Yin, and Wang]{herron2024scitrust}
Emily Herron, Junqi Yin, and Feiyi Wang.
\newblock Scitrust: Evaluating the trustworthiness of large language models for science.
\newblock In \emph{SC24-W: Workshops of the International Conference for High Performance Computing, Networking, Storage and Analysis}, pages 72--78. IEEE, Nov 2024.

\bibitem[Hilgert et~al.(2024)Hilgert, Liu, and Niehues]{hilgert-etal-2024-evaluating}
Lukas Hilgert, Danni Liu, and Jan Niehues.
\newblock Evaluating and training long-context large language models for question answering on scientific papers.
\newblock In Sachin Kumar, Vidhisha Balachandran, Chan~Young Park, Weijia Shi, Shirley~Anugrah Hayati, Yulia Tsvetkov, Noah Smith, Hannaneh Hajishirzi, Dongyeop Kang, and David Jurgens, editors, \emph{Proceedings of the 1st Workshop on Customizable NLP: Progress and Challenges in Customizing NLP for a Domain, Application, Group, or Individual (CustomNLP4U)}, pages 220--236, Miami, Florida, USA, November 2024. Association for Computational Linguistics.
\newblock \doi{10.18653/v1/2024.customnlp4u-1.17}.
\newblock URL \url{https://aclanthology.org/2024.customnlp4u-1.17/}.

\bibitem[Hoang and Kan(2010)]{hoang2010towards}
Cong Duy~Vu Hoang and Min-Yen Kan.
\newblock Towards automated related work summarization.
\newblock In \emph{Coling 2010: Posters}, pages 427--435, Aug 2010.

\bibitem[Hoffmann et~al.(2024)Hoffmann, Fillies, and Paschke]{hoffmann2024malinowski}
Michael~Peter Hoffmann, Jan Fillies, and Adrian Paschke.
\newblock Malinowski in the age of ai: Can large language models create a text game based on an anthropological classic?
\newblock \emph{arXiv preprint arXiv:2410.20536}, 2024.

\bibitem[Hogan et~al.(2024)Hogan, Kabra, Pacheco, Greenstreet, Fan, Ferber, Ummus, Brito, Graham, Aoki, et~al.]{hogan2024aiscivision}
Brendan Hogan, Anmol Kabra, Felipe~Siqueira Pacheco, Laura Greenstreet, Joshua Fan, Aaron Ferber, Marta Ummus, Alecsander Brito, Olivia Graham, Lillian Aoki, et~al.
\newblock Aiscivision: A framework for specializing large multimodal models in scientific image classification.
\newblock \emph{arXiv preprint arXiv:2410.21480}, 2024.

\bibitem[Holter and El-Assady(2024)]{holter2024deconstructing}
Steffen Holter and Mennatallah El-Assady.
\newblock Deconstructing human-ai collaboration: Agency, interaction, and adaptation.
\newblock In \emph{Computer Graphics forum}, volume~43, page e15107. Wiley Online Library, Jun 2024.

\bibitem[H{\"o}pner et~al.(2025)H{\"o}pner, Eshuijs, Alivanistos, Zamprogno, and Tiddi]{hopner2025automatic}
Niklas H{\"o}pner, Leon Eshuijs, Dimitrios Alivanistos, Giacomo Zamprogno, and Ilaria Tiddi.
\newblock Automatic evaluation metrics for artificially generated scientific research.
\newblock \emph{arXiv preprint arXiv:2503.05712}, 2025.

\bibitem[Hossain et~al.(2025)Hossain, Sinha, Bansal, Knipper, Sarkar, Salvador, Mahajan, Guttikonda, Akter, Hassan, Freestone, Jr., Feng, and Karmaker]{hossain-etal-2025-llms}
Eftekhar Hossain, Sanjeev~Kumar Sinha, Naman Bansal, R.~Alexander Knipper, Souvika Sarkar, John Salvador, Yash Mahajan, Sri Ram Pavan~Kumar Guttikonda, Mousumi Akter, Md.~Mahadi Hassan, Matthew Freestone, Matthew C.~Williams Jr., Dongji Feng, and Santu Karmaker.
\newblock {LLM}s as meta-reviewers' assistants: A case study.
\newblock In Luis Chiruzzo, Alan Ritter, and Lu~Wang, editors, \emph{Proceedings of the 2025 Conference of the Nations of the Americas Chapter of the Association for Computational Linguistics: Human Language Technologies (Volume 1: Long Papers)}, pages 7763--7803, Albuquerque, New Mexico, April 2025. Association for Computational Linguistics.
\newblock ISBN 979-8-89176-189-6.
\newblock \doi{10.18653/v1/2025.naacl-long.395}.
\newblock URL \url{https://aclanthology.org/2025.naacl-long.395/}.

\bibitem[Hosseini and Seilani(2025)]{hosseini2025role}
Soodeh Hosseini and Hossein Seilani.
\newblock The role of agentic ai in shaping a smart future: A systematic review.
\newblock \emph{Array}, page 100399, Jul 2025.

\bibitem[Hsieh(2024)]{hsieh2024automated}
Jhih-Yi Hsieh.
\newblock \emph{Automated Peer-Reviewer Assignment can be Manipulated to Secure Reviews from Colluders}.
\newblock PhD thesis, Carnegie Mellon University Pittsburgh, PA, May 2024.

\bibitem[Hsu et~al.(2024)Hsu, Bransom, Sparks, Kuehl, Tan, Wadden, Wang, and Naik]{hsu2024chime}
Chao-Chun Hsu, Erin Bransom, Jenna Sparks, Bailey Kuehl, Chenhao Tan, David Wadden, Lucy~Lu Wang, and Aakanksha Naik.
\newblock Chime: Llm-assisted hierarchical organization of scientific studies for literature review support.
\newblock \emph{arXiv preprint arXiv:2407.16148}, 2024.

\bibitem[Hsu et~al.(2024{\natexlab{2}})Hsu, Huang, Huang, Rossi, Kim, Yu, Giles, and Huang]{hsu2024scicapenter}
Ting-Yao Hsu, Chieh-Yang Huang, Shih-Hong Huang, Ryan Rossi, Sungchul Kim, Tong Yu, C~Lee Giles, and Ting-Hao~Kenneth Huang.
\newblock Scicapenter: Supporting caption composition for scientific figures with machine-generated captions and ratings.
\newblock In \emph{Extended Abstracts of the CHI Conference on Human Factors in Computing Systems}, pages 1--9, May 2024{\natexlab{2}}.

\bibitem[Hu et~al.(2024)Hu, Wang, Pan, Yu, Shao, Feng, and Nie]{hu2024novachart}
Linmei Hu, Duokang Wang, Yiming Pan, Jifan Yu, Yingxia Shao, Chong Feng, and Liqiang Nie.
\newblock Novachart: A large-scale dataset towards chart understanding and generation of multimodal large language models.
\newblock In \emph{Proceedings of the 32nd ACM International Conference on Multimedia}, pages 3917--3925, Oct 2024.
\newblock URL \url{https://openreview.net/forum?id=PTYL6011vp}.

\bibitem[Hu et~al.(2023)Hu, Mu, Yu, Ding, Wu, Shao, Chen, Wang, Qiao, and Luo]{hu2023tree}
Mengkang Hu, Yao Mu, Xinmiao Yu, Mingyu Ding, Shiguang Wu, Wenqi Shao, Qiguang Chen, Bin Wang, Yu~Qiao, and Ping Luo.
\newblock Tree-planner: Efficient close-loop task planning with large language models.
\newblock \emph{arXiv preprint arXiv:2310.08582}, 2023.

\bibitem[Hu et~al.(2024{\natexlab{2}})Hu, Chen, Chen, Mu, Shao, and Luo]{hu2024hiagent}
Mengkang Hu, Tianxing Chen, Qiguang Chen, Yao Mu, Wenqi Shao, and Ping Luo.
\newblock Hiagent: Hierarchical working memory management for solving long-horizon agent tasks with large language model.
\newblock \emph{arXiv preprint arXiv:2408.09559}, 2024{\natexlab{2}}.

\bibitem[Hu et~al.(2025)Hu, Chen, Zou, Lei, Chen, Li, Mu, Zhang, Shao, and Luo]{hu2025text2world}
Mengkang Hu, Tianxing Chen, Yude Zou, Yuheng Lei, Qiguang Chen, Ming Li, Yao Mu, Hongyuan Zhang, Wenqi Shao, and Ping Luo.
\newblock Text2world: Benchmarking large language models for symbolic world model generation.
\newblock \emph{arXiv preprint arXiv:2502.13092}, 2025.

\bibitem[Hu et~al.(2025{\natexlab{2}})Hu, Zhou, Fan, Nie, Xia, Sun, Ye, Jin, Li, Chen, et~al.]{hu2025owl}
Mengkang Hu, Yuhang Zhou, Wendong Fan, Yuzhou Nie, Bowei Xia, Tao Sun, Ziyu Ye, Zhaoxuan Jin, Yingru Li, Qiguang Chen, et~al.
\newblock Owl: Optimized workforce learning for general multi-agent assistance in real-world task automation.
\newblock \emph{arXiv preprint arXiv:2505.23885}, 2025{\natexlab{2}}.

\bibitem[Hu et~al.(2024{\natexlab{3}})Hu, Fu, Wang, Wang, Li, Xu, Lu, Jin, Pan, and Lan]{hu2024nova}
Xiang Hu, Hongyu Fu, Jinge Wang, Yifeng Wang, Zhikun Li, Renjun Xu, Yu~Lu, Yaochu Jin, Lili Pan, and Zhenzhong Lan.
\newblock Nova: An iterative planning and search approach to enhance novelty and diversity of llm generated ideas.
\newblock \emph{arXiv preprint arXiv:2410.14255}, 2024{\natexlab{3}}.

\bibitem[Hu et~al.(2024{\natexlab{4}})Hu, Zhao, Wei, Chai, Ma, Wang, Wang, Su, Xu, Zhu, et~al.]{hu2024infiagent}
Xueyu Hu, Ziyu Zhao, Shuang Wei, Ziwei Chai, Qianli Ma, Guoyin Wang, Xuwu Wang, Jing Su, Jingjing Xu, Ming Zhu, et~al.
\newblock Infiagent-dabench: Evaluating agents on data analysis tasks.
\newblock \emph{arXiv preprint arXiv:2401.05507}, 2024{\natexlab{4}}.

\bibitem[Hu and Wan(2014)]{hu2014automatic}
Yue Hu and Xiaojun Wan.
\newblock Automatic generation of related work sections in scientific papers: an optimization approach.
\newblock In \emph{Proceedings of the 2014 Conference on Empirical Methods in Natural Language Processing (EMNLP)}, pages 1624--1633, Oct 2014.

\bibitem[Hu et~al.(2024{\natexlab{5}})Hu, Li, Zhang, Ling, Kanjiani, Zhao, and Zhao]{hu2024taxonomy}
Yuntong Hu, Zhuofeng Li, Zheng Zhang, Chen Ling, Raasikh Kanjiani, Boxin Zhao, and Liang Zhao.
\newblock Taxonomy tree generation from citation graph.
\newblock \emph{arXiv preprint arXiv:2410.03761}, 2024{\natexlab{5}}.

\bibitem[Hua et~al.(2025)Hua, Hua, Xiang, Klieger, Truong, Liang, Sun, and Haber]{hua2025researchcodebench}
Tianyu Hua, Harper Hua, Violet Xiang, Benjamin Klieger, Sang~T Truong, Weixin Liang, Fan-Yun Sun, and Nick Haber.
\newblock Researchcodebench: Benchmarking llms on implementing novel machine learning research code.
\newblock \emph{arXiv preprint arXiv:2506.02314}, 2025.

\bibitem[Huang et~al.(2022)Huang, Gu, Hou, Wu, Wang, Yu, and Han]{huang2022large}
Jiaxin Huang, Shixiang~Shane Gu, Le~Hou, Yuexin Wu, Xuezhi Wang, Hongkun Yu, and Jiawei Han.
\newblock Large language models can self-improve.
\newblock \emph{arXiv preprint arXiv:2210.11610}, 2022.

\bibitem[Huang et~al.(2025)Huang, Xu, Wang, Wang, Liang, Wang, Zhang, Wei, Zhang, Huang, et~al.]{huang2025foundation}
Jincai Huang, Yongjun Xu, Qi~Wang, Qi~Cheems Wang, Xingxing Liang, Fei Wang, Zhao Zhang, Wei Wei, Boxuan Zhang, Libo Huang, et~al.
\newblock Foundation models and intelligent decision-making: Progress, challenges, and perspectives.
\newblock \emph{The Innovation}, Jun 2025.

\bibitem[Huang et~al.(2025{\natexlab{2}})Huang, Feng, Chen, Zhao, Cheng, Bai, Zhou, Li, and Qin]{huang2025mldebugging}
Jinyang Huang, Xiachong Feng, Qiguang Chen, Hanjie Zhao, Zihui Cheng, Jiesong Bai, Jingxuan Zhou, Min Li, and Libo Qin.
\newblock Mldebugging: Towards benchmarking code debugging across multi-library scenarios.
\newblock \emph{arXiv preprint arXiv:2506.13824}, 2025{\natexlab{2}}.

\bibitem[Huang et~al.(2024)Huang, Qu, Cousins, Johnson, Yin, Shah, Zhou, Altman, Wang, and Cong]{huang2024crispr}
Kaixuan Huang, Yuanhao Qu, Henry Cousins, William~A Johnson, Di~Yin, Mihir Shah, Denny Zhou, Russ Altman, Mengdi Wang, and Le~Cong.
\newblock Crispr-gpt: An llm agent for automated design of gene-editing experiments.
\newblock \emph{arXiv preprint arXiv:2404.18021}, 2024.

\bibitem[Huang et~al.(2024{\natexlab{2}})Huang, Chandak, Wang, Havaldar, Vaid, Leskovec, Nadkarni, Glicksberg, Gehlenborg, and Zitnik]{huang2024foundation}
Kexin Huang, Payal Chandak, Qianwen Wang, Shreyas Havaldar, Akhil Vaid, Jure Leskovec, Girish~N Nadkarni, Benjamin~S Glicksberg, Nils Gehlenborg, and Marinka Zitnik.
\newblock A foundation model for clinician-centered drug repurposing.
\newblock \emph{Nature Medicine}, 30\penalty0 (12):\penalty0 3601--3613, Sep 2024{\natexlab{2}}.

\bibitem[Huang et~al.(2025{\natexlab{3}})Huang, Zhang, Wang, Qu, Lu, Roohani, Li, Qiu, Zhang, Di, et~al.]{huang2025biomni}
Kexin Huang, Serena Zhang, Hanchen Wang, Yuanhao Qu, Yingzhou Lu, Yusuf Roohani, Ryan Li, Lin Qiu, Junze Zhang, Yin Di, et~al.
\newblock Biomni: A general-purpose biomedical ai agent.
\newblock \emph{bioRxiv}, pages 2025--05, Jun 2025{\natexlab{3}}.

\bibitem[Huang et~al.(2025{\natexlab{4}})Huang, Yu, Ma, Zhong, Feng, Wang, Chen, Peng, Feng, Qin, et~al.]{huang2025survey}
Lei Huang, Weijiang Yu, Weitao Ma, Weihong Zhong, Zhangyin Feng, Haotian Wang, Qianglong Chen, Weihua Peng, Xiaocheng Feng, Bing Qin, et~al.
\newblock A survey on hallucination in large language models: Principles, taxonomy, challenges, and open questions.
\newblock \emph{ACM Transactions on Information Systems}, 43\penalty0 (2):\penalty0 1--55, Jan 2025{\natexlab{4}}.

\bibitem[Huang et~al.(2025{\natexlab{5}})Huang, Koutra, Kulkarni, Prioleau, Wu, Yan, Yang, Zou, and Zhou]{huang2025towards}
Lifu Huang, Danai Koutra, Adithya Kulkarni, Temiloluwa Prioleau, Qingyun Wu, Yujun Yan, Yaoqing Yang, James Zou, and Dawei Zhou.
\newblock Towards agentic ai for science: Hypothesis generation, comprehension, quantification, and validation.
\newblock In \emph{Companion Proceedings of the ACM on Web Conference 2025}, pages 1639--1642, May 2025{\natexlab{5}}.

\bibitem[Huang et~al.(2025{\natexlab{6}})Huang, Zhang, Ma, Lai, Xu, Li, Wu, Wu, and Liu]{huang2025chartsketcher}
Muye Huang, Lingling Zhang, Jie Ma, Han Lai, Fangzhi Xu, Yifei Li, Wenjun Wu, Yaqiang Wu, and Jun Liu.
\newblock Chartsketcher: Reasoning with multimodal feedback and reflection for chart understanding.
\newblock \emph{arXiv preprint arXiv:2505.19076}, 2025{\natexlab{6}}.

\bibitem[Huang et~al.(2023)Huang, Vora, Liang, and Leskovec]{huang2023mlagentbench}
Qian Huang, Jian Vora, Percy Liang, and Jure Leskovec.
\newblock Mlagentbench: Evaluating language agents on machine learning experimentation.
\newblock \emph{arXiv preprint arXiv:2310.03302}, 2023.

\bibitem[Huang et~al.(2025{\natexlab{7}})Huang, Wang, Lu, Liu, Xu, and Huang]{huang2025papereval}
Shengzhi Huang, Qicong Wang, Wei Lu, Lingyu Liu, Zhenzhen Xu, and Yong Huang.
\newblock Papereval: A universal, quantitative, and explainable paper evaluation method powered by a multi-agent system.
\newblock \emph{Information Processing \& Management}, 62\penalty0 (6):\penalty0 104225, Nov 2025{\natexlab{7}}.

\bibitem[Huang et~al.(2024{\natexlab{3}})Huang, Zhao, Wang, and Ju]{huang2024ai}
Xiang Huang, CY~Zhao, Hong Wang, and Shenghong Ju.
\newblock Ai-assisted inverse design of sequence-ordered high intrinsic thermal conductivity polymers.
\newblock \emph{Materials Today Physics}, 44:\penalty0 101438, May 2024{\natexlab{3}}.

\bibitem[Huang et~al.(2024{\natexlab{4}})Huang, Lin, Liu, Cao, Xin, Wang, Li, Song, and Liang]{huang2024mustard}
Yinya Huang, Xiaohan Lin, Zhengying Liu, Qingxing Cao, Huajian Xin, Haiming Wang, Zhenguo Li, Linqi Song, and Xiaodan Liang.
\newblock Mustard: Mastering uniform synthesis of theorem and proof data.
\newblock \emph{arXiv preprint arXiv:2402.08957}, 2024{\natexlab{4}}.

\bibitem[Huang et~al.(2024{\natexlab{5}})Huang, Wang, Liu, Kong, Qin, Tang, Wang, Zhu, Bi, Qi, et~al.]{huang2024adasociety}
Yizhe Huang, Xingbo Wang, Hao Liu, Fanqi Kong, Aoyang Qin, Min Tang, Xiaoxi Wang, Song-Chun Zhu, Mingjie Bi, Siyuan Qi, et~al.
\newblock Adasociety: An adaptive environment with social structures for multi-agent decision-making.
\newblock \emph{arXiv preprint arXiv:2411.03865}, 2024{\natexlab{5}}.

\bibitem[Hutchins et~al.(2016)Hutchins, Yuan, Anderson, and Santangelo]{hutchins2016relative}
B~Ian Hutchins, Xin Yuan, James~M Anderson, and George~M Santangelo.
\newblock Relative citation ratio (rcr): a new metric that uses citation rates to measure influence at the article level.
\newblock \emph{PLoS biology}, 14\penalty0 (9):\penalty0 e1002541, Sep 2016.

\bibitem[Hysmith et~al.(2024)Hysmith, Foadian, Padhy, Kalinin, Moore, Ovchinnikova, and Ahmadi]{hysmith2024future}
Holland Hysmith, Elham Foadian, Shakti~P Padhy, Sergei~V Kalinin, Rob~G Moore, Olga~S Ovchinnikova, and Mahshid Ahmadi.
\newblock The future of self-driving laboratories: from human in the loop interactive ai to gamification.
\newblock \emph{Digital Discovery}, 3\penalty0 (4):\penalty0 621--636, Mar 2024.

\bibitem[Idahl and Ahmadi(2024)]{idahl2024openreviewer}
Maximilian Idahl and Zahra Ahmadi.
\newblock Openreviewer: A specialized large language model for generating critical scientific paper reviews.
\newblock \emph{arXiv preprint arXiv:2412.11948}, 2024.

\bibitem[Ifargan et~al.(2025)Ifargan, Hafner, Kern, Alcalay, and Kishony]{ifargan2025autonomous}
Tal Ifargan, Lukas Hafner, Maor Kern, Ori Alcalay, and Roy Kishony.
\newblock Autonomous llm-driven research—from data to human-verifiable research papers.
\newblock \emph{NEJM AI}, 2\penalty0 (1):\penalty0 AIoa2400555, Dec 2025.

\bibitem[Ikoma and Mitamura(2025)]{ikoma2025can}
Hayato Ikoma and Teruko Mitamura.
\newblock Can ai examine novelty of patents?: Novelty evaluation based on the correspondence between patent claim and prior art.
\newblock \emph{arXiv preprint arXiv:2502.06316}, 2025.

\bibitem[Institute(2025)]{carl2025}
Autoscience Institute.
\newblock Carl technical report, Mar 2025.
\newblock URL \url{https://drive.google.com/file/d/1iVedOdZDuEdjS4lcm9Z7i8oEDGWfzVJq/view}.
\newblock Carl Technical Report.

\bibitem[Intelligence(2025)]{intelligence2024amazon}
Amazon Artificial~General Intelligence.
\newblock The amazon nova family of models: Technical report and model card.
\newblock Jun 2025.

\bibitem[Ismayilzada et~al.(2024)Ismayilzada, Paul, Bosselut, and van~der Plas]{ismayilzada2024creativity}
Mete Ismayilzada, Debjit Paul, Antoine Bosselut, and Lonneke van~der Plas.
\newblock Creativity in ai: Progresses and challenges.
\newblock \emph{arXiv preprint arXiv:2410.17218}, 2024.

\bibitem[Ito et~al.(2019)Ito, Kuribayashi, Kobayashi, Brassard, Hagiwara, Suzuki, and Inui]{ito2019diamonds}
Takumi Ito, Tatsuki Kuribayashi, Hayato Kobayashi, Ana Brassard, Masato Hagiwara, Jun Suzuki, and Kentaro Inui.
\newblock Diamonds in the rough: Generating fluent sentences from early-stage drafts for academic writing assistance.
\newblock \emph{arXiv preprint arXiv:1910.09180}, 2019.

\bibitem[Jaech et~al.(2024)Jaech, Kalai, Lerer, Richardson, El-Kishky, Low, Helyar, Madry, Beutel, Carney, et~al.]{jaech2024openai}
Aaron Jaech, Adam Kalai, Adam Lerer, Adam Richardson, Ahmed El-Kishky, Aiden Low, Alec Helyar, Aleksander Madry, Alex Beutel, Alex Carney, et~al.
\newblock Openai o1 system card.
\newblock \emph{arXiv preprint arXiv:2412.16720}, 2024.

\bibitem[Jafari and Allan(2024)]{jafari2024robust}
Nazanin Jafari and James Allan.
\newblock Robust claim verification through fact detection.
\newblock \emph{arXiv preprint arXiv:2407.18367}, 2024.

\bibitem[Jansen et~al.(2024)Jansen, C{\^o}t{\'e}, Khot, Bransom, Dalvi~Mishra, Majumder, Tafjord, and Clark]{jansen2024discoveryworld}
Peter Jansen, Marc-Alexandre C{\^o}t{\'e}, Tushar Khot, Erin Bransom, Bhavana Dalvi~Mishra, Bodhisattwa~Prasad Majumder, Oyvind Tafjord, and Peter Clark.
\newblock Discoveryworld: A virtual environment for developing and evaluating automated scientific discovery agents.
\newblock \emph{Advances in Neural Information Processing Systems}, 37:\penalty0 10088--10116, Dec 2024.

\bibitem[Jayarathna et~al.(2024)Jayarathna, Onsree, Drummond, Naglic, and Lauterbach]{jayarathna2024experimental}
Rasika Jayarathna, Thossaporn Onsree, Samuel Drummond, Jennifer Naglic, and Jochen Lauterbach.
\newblock Experimental discovery of novel ammonia synthesis catalysts via active learning.
\newblock \emph{Journal of Materials Chemistry A}, 12\penalty0 (5):\penalty0 3046--3060, Feb 2024.

\bibitem[Ji et~al.(2024)Ji, Zhu, Gao, Xu, Lu, Ye, and Zhao]{ji2024tree}
Deyi Ji, Lanyun Zhu, Siqi Gao, Peng Xu, Hongtao Lu, Jieping Ye, and Feng Zhao.
\newblock Tree-of-table: Unleashing the power of llms for enhanced large-scale table understanding.
\newblock \emph{arXiv preprint arXiv:2411.08516}, 2024.

\bibitem[Jia et~al.(2021)Jia, Cui, Xiao, Liu, Rashid, and Gehringer]{jia2021all}
Qinjin Jia, Jialin Cui, Yunkai Xiao, Chengyuan Liu, Parvez Rashid, and Edward Gehringer.
\newblock All-in-one: Multi-task learning bert models for evaluating peer assessments.
\newblock \emph{International Educational Data Mining Society}, Oct 2021.

\bibitem[Jiang et~al.(2022)Jiang, Welleck, Zhou, Li, Liu, Jamnik, Lacroix, Wu, and Lample]{jiang2022draft}
Albert~Q Jiang, Sean Welleck, Jin~Peng Zhou, Wenda Li, Jiacheng Liu, Mateja Jamnik, Timoth{\'e}e Lacroix, Yuhuai Wu, and Guillaume Lample.
\newblock Draft, sketch, and prove: Guiding formal theorem provers with informal proofs.
\newblock \emph{arXiv preprint arXiv:2210.12283}, 2022.

\bibitem[Jiang et~al.(2022{\natexlab{2}})Jiang, Li, Tworkowski, Czechowski, Odrzyg{\'o}{\'z}d{\'z}, Mi{\l}o{\'s}, Wu, and Jamnik]{jiang2022thor}
Albert~Qiaochu Jiang, Wenda Li, Szymon Tworkowski, Konrad Czechowski, Tomasz Odrzyg{\'o}{\'z}d{\'z}, Piotr Mi{\l}o{\'s}, Yuhuai Wu, and Mateja Jamnik.
\newblock Thor: Wielding hammers to integrate language models and automated theorem provers.
\newblock \emph{Advances in Neural Information Processing Systems}, 35:\penalty0 8360--8373, Nov 2022{\natexlab{2}}.

\bibitem[Jiang(2024)]{jiang2024identifying}
Fengqing Jiang.
\newblock Identifying and mitigating vulnerabilities in llm-integrated applications.
\newblock Master's thesis, University of Washington, Jul 2024.

\bibitem[Jiang et~al.(2025)Jiang, Liang, Wang, Wang, and Tan]{jiang2025latte}
Nan Jiang, Shanchao Liang, Chengxiao Wang, Jiannan Wang, and Lin Tan.
\newblock Latte: Improving latex recognition for tables and formulae with iterative refinement.
\newblock In \emph{Proceedings of the AAAI Conference on Artificial Intelligence}, volume~39, pages 4030--4038, Apr 2025.

\bibitem[Jiang et~al.(2025{\natexlab{2}})Jiang, Ou, Chen, Ao, Chang, Do, and Lin]{jiang2025fuzzy}
Xiaowei Jiang, Liang Ou, Yanan Chen, Na~Ao, Yu-Cheng Chang, Thomas Do, and Chin-Teng Lin.
\newblock A fuzzy logic-based approach to predict human interaction by functional near-infrared spectroscopy.
\newblock \emph{IEEE Transactions on Fuzzy Systems}, Jan 2025{\natexlab{2}}.

\bibitem[Jiang et~al.(2025{\natexlab{3}})Jiang, Xue, Wang, Liu, Yang, Su, et~al.]{jiang2025ai4materials}
Xue Jiang, Dezhen Xue, William~Yi Wang, Jianjun Liu, Mingli Yang, Yanjing Su, et~al.
\newblock Ai4materials: Transforming the landscape of materials science and enigneering.
\newblock \emph{Review of Materials Research}, page 100010, Jan 2025{\natexlab{3}}.

\bibitem[Jiao et~al.(2024)Jiao, Song, You, Liu, Li, Chen, Tang, Feng, Liu, Guo, et~al.]{jiao2024ai}
Licheng Jiao, Xue Song, Chao You, Xu~Liu, Lingling Li, Puhua Chen, Xu~Tang, Zhixi Feng, Fang Liu, Yuwei Guo, et~al.
\newblock Ai meets physics: a comprehensive survey.
\newblock \emph{Artificial Intelligence Review}, 57\penalty0 (9):\penalty0 256, Aug 2024.

\bibitem[Jimenez et~al.(2024)Jimenez, Yang, Wettig, Yao, Pei, Press, and Narasimhan]{jimenez2024swebench}
Carlos~E Jimenez, John Yang, Alexander Wettig, Shunyu Yao, Kexin Pei, Ofir Press, and Karthik~R Narasimhan.
\newblock {SWE}-bench: Can language models resolve real-world github issues?
\newblock In \emph{The Twelfth International Conference on Learning Representations}, Jan 2024.
\newblock URL \url{https://openreview.net/forum?id=VTF8yNQM66}.

\bibitem[Jin et~al.(2019)Jin, Dhingra, Liu, Cohen, and Lu]{jin2019pubmedqa}
Qiao Jin, Bhuwan Dhingra, Zhengping Liu, William~W Cohen, and Xinghua Lu.
\newblock Pubmedqa: A dataset for biomedical research question answering.
\newblock \emph{arXiv preprint arXiv:1909.06146}, 2019.

\bibitem[Jin et~al.(2024)Jin, Zhao, Wang, Chen, Zhu, Xiao, and Wang]{jin2024agentreview}
Yiqiao Jin, Qinlin Zhao, Yiyang Wang, Hao Chen, Kaijie Zhu, Yijia Xiao, and Jindong Wang.
\newblock Agentreview: Exploring peer review dynamics with llm agents.
\newblock \emph{arXiv preprint arXiv:2406.12708}, 2024.

\bibitem[Jing et~al.(2024)Jing, Huang, Wang, Yao, Yu, Ma, Zhang, Du, and Yu]{jing2024dsbench}
Liqiang Jing, Zhehui Huang, Xiaoyang Wang, Wenlin Yao, Wenhao Yu, Kaixin Ma, Hongming Zhang, Xinya Du, and Dong Yu.
\newblock Dsbench: How far are data science agents to becoming data science experts?
\newblock \emph{arXiv preprint arXiv:2409.07703}, 2024.

\bibitem[Johnson and Proudfoot(2024)]{johnson2024greater}
Wayne Johnson and Devon Proudfoot.
\newblock Greater variability in judgements of the value of novel ideas.
\newblock \emph{Nature Human Behaviour}, 8\penalty0 (3):\penalty0 471--479, Jan 2024.

\bibitem[Jones and Bergen(2025)]{jones2025large}
Cameron~R Jones and Benjamin~K Bergen.
\newblock Large language models pass the turing test.
\newblock \emph{arXiv preprint arXiv:2503.23674}, 2025.

\bibitem[Jumper et~al.(2021)Jumper, Evans, Pritzel, Green, Figurnov, Ronneberger, Tunyasuvunakool, Bates, {\v{Z}}{\'\i}dek, Potapenko, et~al.]{jumper2021highly}
John Jumper, Richard Evans, Alexander Pritzel, Tim Green, Michael Figurnov, Olaf Ronneberger, Kathryn Tunyasuvunakool, Russ Bates, Augustin {\v{Z}}{\'\i}dek, Anna Potapenko, et~al.
\newblock Highly accurate protein structure prediction with alphafold.
\newblock \emph{nature}, 596\penalty0 (7873):\penalty0 583--589, Jul 2021.

\bibitem[Juneja and Mitra(2022)]{juneja2022human}
Prerna Juneja and Tanushree Mitra.
\newblock Human and technological infrastructures of fact-checking.
\newblock \emph{Proceedings of the ACM on Human-Computer Interaction}, 6\penalty0 (CSCW2):\penalty0 1--36, Nov 2022.

\bibitem[Kaiser et~al.(2025)Kaiser, Sadr, Yuen, Krenz, Chin-Chin, Sheela, Ransford, and Kobetz]{kaiser2025376}
Rebecca~E Kaiser, Farshad Sadr, Trevor Yuen, Till Krenz, Lee Chin-Chin, C~Dominguez Sheela, Daru~LL Ransford, and Erin Kobetz.
\newblock 376 using a large language model to create lay summaries of clinical study descriptions.
\newblock \emph{Journal of Clinical and Translational Science}, 9\penalty0 (s1):\penalty0 116--116, Apr 2025.

\bibitem[Kalashnikov et~al.(2018)Kalashnikov, Irpan, Pastor, Ibarz, Herzog, Jang, Quillen, Holly, Kalakrishnan, Vanhoucke, et~al.]{kalashnikov2018scalable}
Dmitry Kalashnikov, Alex Irpan, Peter Pastor, Julian Ibarz, Alexander Herzog, Eric Jang, Deirdre Quillen, Ethan Holly, Mrinal Kalakrishnan, Vincent Vanhoucke, et~al.
\newblock Scalable deep reinforcement learning for vision-based robotic manipulation.
\newblock In \emph{Conference on robot learning}, pages 651--673. PMLR, Oct 2018.

\bibitem[Kambhampati et~al.(2024)Kambhampati, Valmeekam, Guan, Verma, Stechly, Bhambri, Saldyt, and Murthy]{kambhampati2024position}
Subbarao Kambhampati, Karthik Valmeekam, Lin Guan, Mudit Verma, Kaya Stechly, Siddhant Bhambri, Lucas~Paul Saldyt, and Anil~B Murthy.
\newblock Position: Llms can’t plan, but can help planning in llm-modulo frameworks.
\newblock In \emph{Forty-first International Conference on Machine Learning}, May 2024.

\bibitem[Kang et~al.(2018)Kang, Ammar, Dalvi, van Zuylen, Kohlmeier, Hovy, and Schwartz]{kang2018dataset}
Dongyeop Kang, Waleed Ammar, Bhavana Dalvi, Madeleine van Zuylen, Sebastian Kohlmeier, Eduard Hovy, and Roy Schwartz.
\newblock A dataset of peer reviews (peerread): Collection, insights and nlp applications.
\newblock In \emph{Proceedings of the 2018 Conference of the North American Chapter of the Association for Computational Linguistics: Human Language Technologies, Volume 1 (Long Papers)}, pages 1647--1661, Apr 2018.

\bibitem[Kang et~al.(2022)Kang, Kocielnik, Head, Yang, Latzke, Kittur, Weld, Downey, and Bragg]{kang2022you}
Hyeonsu~B Kang, Rafal Kocielnik, Andrew Head, Jiangjiang Yang, Matt Latzke, Aniket Kittur, Daniel~S Weld, Doug Downey, and Jonathan Bragg.
\newblock From who you know to what you read: Augmenting scientific recommendations with implicit social networks.
\newblock In \emph{Proceedings of the 2022 CHI Conference on Human Factors in Computing Systems}, pages 1--23, Apr 2022.

\bibitem[Kang et~al.(2023)Kang, Soliman, Latzke, Chang, and Bragg]{kang2023comlittee}
Hyeonsu~B Kang, Nouran Soliman, Matt Latzke, Joseph~Chee Chang, and Jonathan Bragg.
\newblock Comlittee: Literature discovery with personal elected author committees.
\newblock In \emph{Proceedings of the 2023 CHI Conference on Human Factors in Computing Systems}, pages 1--20, Apr 2023.

\bibitem[Kang et~al.(2025)Kang, Chen, Yoo, and Lou]{kang2025explainable}
Sungmin Kang, Bei Chen, Shin Yoo, and Jian-Guang Lou.
\newblock Explainable automated debugging via large language model-driven scientific debugging.
\newblock \emph{Empirical Software Engineering}, 30\penalty0 (2):\penalty0 1--28, Mar 2025.

\bibitem[Kao et~al.(2025)Kao, Zhao, Revankar, Speas, Bhagat, Datta, Phoo, Mall, Vondrick, Bala, et~al.]{kao2025towards}
Chia~Hsiang Kao, Wenting Zhao, Shreelekha Revankar, Samuel Speas, Snehal Bhagat, Rajeev Datta, Cheng~Perng Phoo, Utkarsh Mall, Carl Vondrick, Kavita Bala, et~al.
\newblock Towards llm agents for earth observation.
\newblock \emph{arXiv preprint arXiv:2504.12110}, 2025.

\bibitem[Kao and Yen(2024)]{kao2024magic}
Wei-Yu Kao and An-Zi Yen.
\newblock Magic: Multi-argument generation with self-refinement for domain generalization in automatic fact-checking.
\newblock In \emph{Proceedings of the 2024 Joint International Conference on Computational Linguistics, Language Resources and Evaluation (LREC-COLING 2024)}, pages 10891--10902, May 2024.

\bibitem[Kapitonova and Ball(2024)]{kapitonova2024human}
Maryna Kapitonova and Tonio Ball.
\newblock Human-ai teaming using large language models: Boosting brain-computer interfacing (bci) and brain research.
\newblock \emph{arXiv preprint arXiv:2501.01451}, 2024.

\bibitem[Karjus(2025)]{karjus2025machine}
Andres Karjus.
\newblock Machine-assisted quantitizing designs: augmenting humanities and social sciences with artificial intelligence.
\newblock \emph{Humanities and Social Sciences Communications}, 12\penalty0 (1):\penalty0 1--18, Feb 2025.

\bibitem[Kasanishi et~al.(2023)Kasanishi, Isonuma, Mori, and Sakata]{kasanishi2023scireviewgen}
Tetsu Kasanishi, Masaru Isonuma, Junichiro Mori, and Ichiro Sakata.
\newblock Scireviewgen: a large-scale dataset for automatic literature review generation.
\newblock \emph{arXiv preprint arXiv:2305.15186}, 2023.

\bibitem[Katz et~al.(2024)Katz, Levy, and Goldberg]{katz2024knowledge}
Uri Katz, Mosh Levy, and Yoav Goldberg.
\newblock Knowledge navigator: Llm-guided browsing framework for exploratory search in scientific literature.
\newblock \emph{arXiv preprint arXiv:2408.15836}, 2024.

\bibitem[Ke et~al.(2024)Ke, Sawyer, Soyer, Engelcke, Reichert, Hudson, Reid, Lerchner, Rezende, Lillicrap, et~al.]{ke2024can}
Nan~Rosemary Ke, Danny~P Sawyer, Hubert Soyer, Martin Engelcke, David~P Reichert, Drew~A Hudson, John Reid, Alexander Lerchner, Danilo~Jimenez Rezende, Timothy~P Lillicrap, et~al.
\newblock Can foundation models actively gather information in interactive environments to test hypotheses?
\newblock \emph{arXiv preprint arXiv:2412.06438}, 2024.

\bibitem[Kerwer et~al.(2021)Kerwer, Chasiotis, Stricker, G{\"u}nther, and Rosman]{kerwer2021straight}
Martin Kerwer, Anita Chasiotis, Johannes Stricker, Armin G{\"u}nther, and Tom Rosman.
\newblock Straight from the scientist's mouth—plain language summaries promote laypeople's comprehension and knowledge acquisition when reading about individual research findings in psychology.
\newblock \emph{Collabra: Psychology}, 7\penalty0 (1), Feb 2021.

\bibitem[Keya et~al.(2025)Keya, Rabby, Mitra, Vahdati, Auer, and Jaradeh]{keya2025sci}
Farhana Keya, Gollam Rabby, Prasenjit Mitra, Sahar Vahdati, S{\"o}ren Auer, and Yaser Jaradeh.
\newblock Sci-idea: Context-aware scientific ideation using token and sentence embeddings.
\newblock \emph{arXiv preprint arXiv:2503.19257}, 2025.

\bibitem[Khalifa and Albadawy(2024)]{khalifa2024using}
Mohamed Khalifa and Mona Albadawy.
\newblock Using artificial intelligence in academic writing and research: An essential productivity tool.
\newblock \emph{Computer Methods and Programs in Biomedicine Update}, page 100145, Mar 2024.

\bibitem[Khalifa et~al.(2024)Khalifa, Albadawy, and Iqbal]{khalifa2024advancing}
Mohamed Khalifa, Mona Albadawy, and Usman Iqbal.
\newblock Advancing clinical decision support: The role of artificial intelligence across six domains.
\newblock \emph{Computer Methods and Programs in Biomedicine Update}, 5:\penalty0 100142, Feb 2024.

\bibitem[Khan et~al.(2025)Khan, Leem, See, Wong, Zhang, and Fang]{khan2025comprehensive}
Wasif Khan, Seowung Leem, Kyle~B See, Joshua~K Wong, Shaoting Zhang, and Ruogu Fang.
\newblock A comprehensive survey of foundation models in medicine.
\newblock \emph{IEEE Reviews in Biomedical Engineering}, Jan 2025.

\bibitem[Kim et~al.(2025)Kim, Lee, and Lee]{kim2025position}
Jaeho Kim, Yunseok Lee, and Seulki Lee.
\newblock Position: The ai conference peer review crisis demands author feedback and reviewer rewards.
\newblock \emph{arXiv preprint arXiv:2505.04966}, 2025.

\bibitem[Kim et~al.(2025{\natexlab{2}})Kim, Lee, Choi, Hsu, Huang, Kim, Rossi, Yu, Giles, Huang, et~al.]{kim2025multi}
Jaeyoung Kim, Jongho Lee, Hong-Jun Choi, Ting-Yao Hsu, Chieh-Yang Huang, Sungchul Kim, Ryan Rossi, Tong Yu, Clyde~Lee Giles, Ting-Hao'Kenneth' Huang, et~al.
\newblock Multi-llm collaborative caption generation in scientific documents.
\newblock \emph{arXiv preprint arXiv:2501.02552}, 2025{\natexlab{2}}.

\bibitem[Kim et~al.(2023)Kim, Park, Kwon, Jo, Thorne, and Choi]{kim2023factkg}
Jiho Kim, Sungjin Park, Yeonsu Kwon, Yohan Jo, James Thorne, and Edward Choi.
\newblock Factkg: Fact verification via reasoning on knowledge graphs.
\newblock \emph{arXiv preprint arXiv:2305.06590}, 2023.

\bibitem[Kim et~al.(2025{\natexlab{3}})Kim, Park, Ahn, Park, Jeon, Lee, Lee, and Choi]{kim2025autopaperbench}
Min-Woo Kim, Hyo-Bin Park, Hee-Jin Ahn, Woo-Ram Park, Jae-Wan Jeon, Kyong-Ha Lee, Ryong Lee, and Dong-Geol Choi.
\newblock Autopaperbench: An mllm-based framework for automatic generation of paper understanding evaluation benchmarks.
\newblock \emph{Electronics}, 14\penalty0 (6):\penalty0 1175, Mar 2025{\natexlab{3}}.

\bibitem[Kim(2025)]{kim2025medbiolm}
Seonok Kim.
\newblock Medbiolm: Optimizing medical and biological qa with fine-tuned large language models and retrieval-augmented generation.
\newblock \emph{arXiv preprint arXiv:2502.03004}, 2025.

\bibitem[King et~al.(2004)King, Whelan, Jones, Reiser, Bryant, Muggleton, Kell, and Oliver]{king2004functional}
Ross~D King, Kenneth~E Whelan, Ffion~M Jones, Philip~GK Reiser, Christopher~H Bryant, Stephen~H Muggleton, Douglas~B Kell, and Stephen~G Oliver.
\newblock Functional genomic hypothesis generation and experimentation by a robot scientist.
\newblock \emph{Nature}, 427\penalty0 (6971):\penalty0 247--252, Jan 2004.

\bibitem[Kirtani et~al.(2025)Kirtani, Garg, Prasad, Singhal, Mandal, and Kumar]{kirtani2025revieweval}
Chhavi Kirtani, Madhav~Krishan Garg, Tejash Prasad, Tanmay Singhal, Murari Mandal, and Dhruv Kumar.
\newblock Revieweval: An evaluation framework for ai-generated reviews.
\newblock \emph{arXiv preprint arXiv:2502.11736}, 2025.

\bibitem[Knox et~al.(2025)Knox, Wu, Islam, O'Connell, Pittaway, Chingono, Oyekan, Panoutsos, Chamberlain, Bourne, et~al.]{knox2025self}
Stephen~T Knox, Kai~E Wu, Nazrul Islam, Roisin O'Connell, Peter~M Pittaway, Kudakwashe~E Chingono, John Oyekan, George Panoutsos, Thomas~W Chamberlain, Richard~A Bourne, et~al.
\newblock Self-driving laboratory platform for many-objective self-optimisation of polymer nanoparticle synthesis with cloud-integrated machine learning and orthogonal online analytics.
\newblock \emph{Polymer Chemistry}, 16\penalty0 (12):\penalty0 1355--1364, Feb 2025.

\bibitem[Kon et~al.(2025)Kon, Liu, Ding, Qiu, Yang, Huang, Srinivasa, Lee, Chowdhury, and Chen]{kon2025curie}
Patrick Tser~Jern Kon, Jiachen Liu, Qiuyi Ding, Yiming Qiu, Zhenning Yang, Yibo Huang, Jayanth Srinivasa, Myungjin Lee, Mosharaf Chowdhury, and Ang Chen.
\newblock Curie: Toward rigorous and automated scientific experimentation with ai agents.
\newblock \emph{arXiv preprint arXiv:2502.16069}, 2025.

\bibitem[Kon et~al.(2025{\natexlab{2}})Kon, Liu, Zhu, Ding, Peng, Xing, Huang, Qiu, Srinivasa, Lee, et~al.]{kon2025exp}
Patrick Tser~Jern Kon, Jiachen Liu, Xinyi Zhu, Qiuyi Ding, Jingjia Peng, Jiarong Xing, Yibo Huang, Yiming Qiu, Jayanth Srinivasa, Myungjin Lee, et~al.
\newblock Exp-bench: Can ai conduct ai research experiments?
\newblock \emph{arXiv preprint arXiv:2505.24785}, 2025{\natexlab{2}}.

\bibitem[Kousha and Thelwall(2024)]{kousha2024artificial}
Kayvan Kousha and Mike Thelwall.
\newblock Artificial intelligence to support publishing and peer review: A summary and review.
\newblock \emph{Learned Publishing}, 37\penalty0 (1):\penalty0 4--12, Aug 2024.

\bibitem[Krenn et~al.(2022)Krenn, Buffoni, Coutinho, Eppel, Foster, Gritsevskiy, Lee, Lu, Moutinho, Sanjabi, et~al.]{krenn2022predicting}
Mario Krenn, Lorenzo Buffoni, Bruno Coutinho, Sagi Eppel, Jacob~Gates Foster, Andrew Gritsevskiy, Harlin Lee, Yichao Lu, Joao~P Moutinho, Nima Sanjabi, et~al.
\newblock Predicting the future of ai with ai: High-quality link prediction in an exponentially growing knowledge network.
\newblock \emph{arXiv preprint arXiv:2210.00881}, 2022.

\bibitem[Kreutz and Schenkel(2022)]{kreutz2022scientific}
Christin~Katharina Kreutz and Ralf Schenkel.
\newblock Scientific paper recommendation systems: a literature review of recent publications.
\newblock \emph{International journal on digital libraries}, 23\penalty0 (4):\penalty0 335--369, Oct 2022.

\bibitem[Krishna et~al.(2022)Krishna, Riedel, and Vlachos]{krishna2022proofver}
Amrith Krishna, Sebastian Riedel, and Andreas Vlachos.
\newblock Proofver: Natural logic theorem proving for fact verification.
\newblock \emph{Transactions of the Association for Computational Linguistics}, 10:\penalty0 1013--1030, Sep 2022.

\bibitem[Kristiadi et~al.(2024)Kristiadi, Strieth-Kalthoff, Skreta, Poupart, Aspuru-Guzik, and Pleiss]{kristiadi2024sober}
Agustinus Kristiadi, Felix Strieth-Kalthoff, Marta Skreta, Pascal Poupart, Al{\'a}n Aspuru-Guzik, and Geoff Pleiss.
\newblock A sober look at llms for material discovery: Are they actually good for bayesian optimization over molecules?
\newblock \emph{arXiv preprint arXiv:2402.05015}, 2024.

\bibitem[Krithara et~al.(2023)Krithara, Nentidis, Bougiatiotis, and Paliouras]{krithara2023bioasq}
Anastasia Krithara, Anastasios Nentidis, Konstantinos Bougiatiotis, and Georgios Paliouras.
\newblock Bioasq-qa: A manually curated corpus for biomedical question answering.
\newblock \emph{Scientific Data}, 10\penalty0 (1):\penalty0 170, Mar 2023.

\bibitem[Ku et~al.(2025)Ku, Chong, Leung, Shah, Yu, and Chen]{ku2025theoremexplainagent}
Max Ku, Thomas Chong, Jonathan Leung, Krish Shah, Alvin Yu, and Wenhu Chen.
\newblock Theoremexplainagent: Towards video-based multimodal explanations for llm theorem understanding.
\newblock \emph{arXiv preprint arXiv:2502.19400}, 2025.

\bibitem[Kuhn et~al.(2022)Kuhn, Gal, and Farquhar]{kuhn2022clam}
Lorenz Kuhn, Yarin Gal, and Sebastian Farquhar.
\newblock Clam: Selective clarification for ambiguous questions with generative language models.
\newblock \emph{arXiv preprint arXiv:2212.07769}, 2022.

\bibitem[Kulkarni et~al.(2025)Kulkarni, Alotaibi, Zeng, Wu, Zeng, Yao, Liu, Zhang, Huang, and Zhou]{kulkarni2025scientific}
Adithya Kulkarni, Fatimah Alotaibi, Xinyue Zeng, Longfeng Wu, Tong Zeng, Barry~Menglong Yao, Minqian Liu, Shuaicheng Zhang, Lifu Huang, and Dawei Zhou.
\newblock Scientific hypothesis generation and validation: Methods, datasets, and future directions.
\newblock \emph{arXiv preprint arXiv:2505.04651}, 2025.

\bibitem[Kumar et~al.(2024)Kumar, Ghosal, Bhattacharjee, and Ekbal]{kumar2024towards}
Asheesh Kumar, Tirthankar Ghosal, Saprativa Bhattacharjee, and Asif Ekbal.
\newblock Towards automated meta-review generation via an nlp/ml pipeline in different stages of the scholarly peer review process.
\newblock \emph{International Journal on Digital Libraries}, 25\penalty0 (3):\penalty0 493--504, Apr 2024.

\bibitem[Kumar et~al.(2025)Kumar, Vincentius, Jordan, and Anderson]{kumar2025human}
Harsh Kumar, Jonathan Vincentius, Ewan Jordan, and Ashton Anderson.
\newblock Human creativity in the age of llms: Randomized experiments on divergent and convergent thinking.
\newblock In \emph{Proceedings of the 2025 CHI Conference on Human Factors in Computing Systems}, pages 1--18, Apr 2025.

\bibitem[Kumar et~al.(2023)Kumar, Ghosal, and Ekbal]{kumar2023reviewers}
Sandeep Kumar, Tirthankar Ghosal, and Asif Ekbal.
\newblock When reviewers lock horn: Finding disagreement in scientific peer reviews.
\newblock \emph{arXiv preprint arXiv:2310.18685}, 2023.

\bibitem[Kumar et~al.(2024{\natexlab{2}})Kumar, Ghosal, Goyal, and Ekbal]{kumar2024can}
Sandeep Kumar, Tirthankar Ghosal, Vinayak Goyal, and Asif Ekbal.
\newblock Can large language models unlock novel scientific research ideas?
\newblock \emph{arXiv preprint arXiv:2409.06185}, 2024{\natexlab{2}}.

\bibitem[Kumar et~al.(2025{\natexlab{2}})Kumar, Sharma, Khincha, Shroff, Singh, and Mishra]{kumar2025sciclaimhunt}
Sujit Kumar, Anshul Sharma, Siddharth~Hemant Khincha, Gargi Shroff, Sanasam~Ranbir Singh, and Rahul Mishra.
\newblock Sciclaimhunt: A large dataset for evidence-based scientific claim verification.
\newblock \emph{arXiv preprint arXiv:2502.10003}, 2025{\natexlab{2}}.

\bibitem[Kuo et~al.(2025)Kuo, Gabriel, Koola, Schooley, and Ohno-Machado]{kuo2025distributed}
Tsung-Ting Kuo, Rodney~A Gabriel, Jejo Koola, Robert~T Schooley, and Lucila Ohno-Machado.
\newblock Distributed cross-learning for equitable federated models-privacy-preserving prediction on data from five california hospitals.
\newblock \emph{Nature Communications}, 16\penalty0 (1):\penalty0 1371, Feb 2025.

\bibitem[Kuznetsov et~al.(2024)Kuznetsov, Afzal, Dercksen, Dycke, Goldberg, Hope, Hovy, Kummerfeld, Lauscher, Leyton-Brown, et~al.]{kuznetsov2024can}
Ilia Kuznetsov, Osama~Mohammed Afzal, Koen Dercksen, Nils Dycke, Alexander Goldberg, Tom Hope, Dirk Hovy, Jonathan~K Kummerfeld, Anne Lauscher, Kevin Leyton-Brown, et~al.
\newblock What can natural language processing do for peer review?
\newblock \emph{arXiv preprint arXiv:2405.06563}, 2024.

\bibitem[Kvapil et~al.(2025)Kvapil, Borca-Tasciuc, Bossi, Chen, Chen, Morales, Da~Costa, Da~Silva, Dean, Durham, et~al.]{kvapil2025intelligent}
J~Kvapil, G~Borca-Tasciuc, H~Bossi, K~Chen, Y~Chen, Y~Corrales Morales, H~Da~Costa, C~Da~Silva, C~Dean, J~Durham, et~al.
\newblock Intelligent experiments through real-time ai: Fast data processing and autonomous detector control for sphenix and future eic detectors.
\newblock \emph{arXiv preprint arXiv:2501.04845}, 2025.

\bibitem[Kyung et~al.(2025)Kyung, Chung, Bae, Kim, Sohn, Kim, Kim, and Choi]{kyung2025patientsim}
Daeun Kyung, Hyunseung Chung, Seongsu Bae, Jiho Kim, Jae~Ho Sohn, Taerim Kim, Soo~Kyung Kim, and Edward Choi.
\newblock Patientsim: A persona-driven simulator for realistic doctor-patient interactions.
\newblock \emph{arXiv preprint arXiv:2505.17818}, 2025.

\bibitem[Lagzian et~al.(2025)Lagzian, Anumasa, and Liu]{lagzian2025multi}
Arash Lagzian, Srinivas Anumasa, and Dianbo Liu.
\newblock Multi-novelty: Improve the diversity and novelty of contents generated by large language models via inference-time multi-views brainstorming.
\newblock \emph{arXiv preprint arXiv:2502.12700}, 2025.

\bibitem[Lai et~al.(2024)Lai, Wu, Wang, Hu, and Zheng]{lai2024instruct}
Yuxuan Lai, Yupeng Wu, Yidan Wang, Wenpeng Hu, and Chen Zheng.
\newblock Instruct large language models to generate scientific literature survey step by step.
\newblock In \emph{CCF International Conference on Natural Language Processing and Chinese Computing}, pages 484--496. Springer, Nov 2024.

\bibitem[Lai and Pu(2025)]{lai2025prim}
Zheyuan Lai and Yingming Pu.
\newblock Prim: Principle-inspired material discovery through multi-agent collaboration.
\newblock \emph{arXiv preprint arXiv:2504.08810}, 2025.

\bibitem[L{\'a}la et~al.(2023)L{\'a}la, O'Donoghue, Shtedritski, Cox, Rodriques, and White]{lala2023paperqa}
Jakub L{\'a}la, Odhran O'Donoghue, Aleksandar Shtedritski, Sam Cox, Samuel~G Rodriques, and Andrew~D White.
\newblock Paperqa: Retrieval-augmented generative agent for scientific research.
\newblock \emph{arXiv preprint arXiv:2312.07559}, 2023.

\bibitem[Lam et~al.(2024)Lam, Ong, and Mutwil]{lam2024large}
Hilbert Yuen~In Lam, Xing~Er Ong, and Marek Mutwil.
\newblock Large language models in plant biology.
\newblock \emph{Trends in Plant Science}, Oct 2024.

\bibitem[Lample et~al.(2022)Lample, Lacroix, Lachaux, Rodriguez, Hayat, Lavril, Ebner, and Martinet]{lample2022hypertree}
Guillaume Lample, Timothee Lacroix, Marie-Anne Lachaux, Aurelien Rodriguez, Amaury Hayat, Thibaut Lavril, Gabriel Ebner, and Xavier Martinet.
\newblock Hypertree proof search for neural theorem proving.
\newblock \emph{Advances in Neural Information Processing Systems}, 35:\penalty0 26337--26349, Nov 2022.

\bibitem[Lan et~al.(2023)Lan, Hu, Li, and Zhang]{lan2023contrastive}
Ge~Lan, Mengting Hu, Ye~Li, and Yuzhi Zhang.
\newblock Contrastive knowledge integrated graph neural networks for chinese medical text classification.
\newblock \emph{Engineering Applications of Artificial Intelligence}, 122:\penalty0 106057, Jun 2023.

\bibitem[Lan et~al.(2025)Lan, Wang, Ji, Yang, Zhang, Liu, Wu, and Wang]{lan2025clinicalgpt}
Wuyang Lan, Wenzheng Wang, Changwei Ji, Guoxing Yang, Yongbo Zhang, Xiaohong Liu, Song Wu, and Guangyu Wang.
\newblock Clinicalgpt-r1: Pushing reasoning capability of generalist disease diagnosis with large language model.
\newblock \emph{arXiv preprint arXiv:2504.09421}, 2025.

\bibitem[Lan et~al.(2025{\natexlab{2}})Lan, Jiang, Wang, Xie, Zhang, Zhu, Li, Yang, Chen, Gao, et~al.]{lan2025autobio}
Zhiqian Lan, Yuxuan Jiang, Ruiqi Wang, Xuanbing Xie, Rongkui Zhang, Yicheng Zhu, Peihang Li, Tianshuo Yang, Tianxing Chen, Haoyu Gao, et~al.
\newblock Autobio: A simulation and benchmark for robotic automation in digital biology laboratory.
\newblock \emph{arXiv preprint arXiv:2505.14030}, 2025{\natexlab{2}}.

\bibitem[Lange et~al.(2025)Lange, Prasad, Sun, Faldor, Tang, and Ha]{lange2025ai}
Robert~Tjarko Lange, Aaditya Prasad, Qi~Sun, Maxence Faldor, Yujin Tang, and David Ha.
\newblock The ai cuda engineer: Agentic cuda kernel discovery, optimization and composition.
\newblock Technical report, Technical report, Sakana AI, 02 2025, Feb 2025.

\bibitem[Laverghetta~Jr et~al.(2025)Laverghetta~Jr, Chakrabarty, Hope, Pronchick, Bhawsar, and Beaty]{laverghetta2025humans}
Antonio Laverghetta~Jr, Tuhin Chakrabarty, Tom Hope, Jimmy Pronchick, Krupa Bhawsar, and Roger~E Beaty.
\newblock How do humans and language models reason about creativity? a comparative analysis.
\newblock \emph{arXiv preprint arXiv:2502.03253}, 2025.

\bibitem[Le~Hai et~al.(2024)Le~Hai, Nguyen, and Bui]{le2024repoexec}
Nam Le~Hai, Dung~Manh Nguyen, and Nghi~DQ Bui.
\newblock Repoexec: Evaluate code generation with a repository-level executable benchmark.
\newblock \emph{arXiv e-prints}, pages arXiv--2406, 2024.

\bibitem[Lee et~al.(2025)Lee, Lim, Jung, and Kim]{lee2025psyche}
Jingoo Lee, Kyungho Lim, Young-Chul Jung, and Byung-Hoon Kim.
\newblock Psyche: A multi-faceted patient simulation framework for evaluation of psychiatric assessment conversational agents.
\newblock \emph{arXiv preprint arXiv:2501.01594}, 2025.

\bibitem[Lee et~al.(2025{\natexlab{2}})Lee, Lee, and Yoo]{lee2025role}
Jisoo Lee, Jieun Lee, and Jeong-Ju Yoo.
\newblock The role of large language models in the peer-review process: opportunities and challenges for medical journal reviewers and editors.
\newblock \emph{Journal of Educational Evaluation for Health Professions}, 22, Jan 2025{\natexlab{2}}.

\bibitem[Lee et~al.(2022)Lee, Liang, and Yang]{lee2022coauthor}
Mina Lee, Percy Liang, and Qian Yang.
\newblock Coauthor: Designing a human-ai collaborative writing dataset for exploring language model capabilities.
\newblock In \emph{Proceedings of the 2022 CHI conference on human factors in computing systems}, pages 1--19, Apr 2022.

\bibitem[Lee et~al.(2025{\natexlab{3}})Lee, De~Brouwer, Hajiramezanali, Biancalani, Park, and Scalia]{lee2025rag}
Namkyeong Lee, Edward De~Brouwer, Ehsan Hajiramezanali, Tommaso Biancalani, Chanyoung Park, and Gabriele Scalia.
\newblock Rag-enhanced collaborative llm agents for drug discovery.
\newblock \emph{arXiv preprint arXiv:2502.17506}, 2025{\natexlab{3}}.

\bibitem[Lee et~al.(2024)Lee, Liu, Cheng, and Zhang]{lee2024deep}
Seungyeon Lee, Ruoqi Liu, Feixiong Cheng, and Ping Zhang.
\newblock A deep subgrouping framework for precision drug repurposing via emulating clinical trials on real-world patient data.
\newblock \emph{arXiv preprint arXiv:2412.20373}, 2024.

\bibitem[Lee and Ko(2025)]{lee2025generative}
Suk~Ki Lee and Hyunwoong Ko.
\newblock Generative machine learning in adaptive control of dynamic manufacturing processes: A review.
\newblock \emph{arXiv preprint arXiv:2505.00210}, 2025.

\bibitem[Lee et~al.(2024{\natexlab{2}})Lee, Kang, Latzke, Kim, Bragg, Chang, and Siangliulue]{lee2024paperweaver}
Yoonjoo Lee, Hyeonsu~B Kang, Matt Latzke, Juho Kim, Jonathan Bragg, Joseph~Chee Chang, and Pao Siangliulue.
\newblock Paperweaver: Enriching topical paper alerts by contextualizing recommended papers with user-collected papers.
\newblock In \emph{Proceedings of the 2024 CHI Conference on Human Factors in Computing Systems}, pages 1--19, May 2024{\natexlab{2}}.

\bibitem[Lehr et~al.(2024)Lehr, Caliskan, Liyanage, and Banaji]{lehr2024chatgpt}
Steven~A Lehr, Aylin Caliskan, Suneragiri Liyanage, and Mahzarin~R Banaji.
\newblock Chatgpt as research scientist: probing gpt’s capabilities as a research librarian, research ethicist, data generator, and data predictor.
\newblock \emph{Proceedings of the National Academy of Sciences}, 121\penalty0 (35):\penalty0 e2404328121, Jul 2024.

\bibitem[Leng et~al.(2025)Leng, Zhong, Gu, Li, Cui, Li, Liu, and Wan]{leng2025intelligent}
Yan Leng, Yi~Zhong, Zhi Gu, Peiyi Li, Haoting Cui, Xing Li, Yang Liu, and Jiayu Wan.
\newblock Intelligent, personalized scientific assistant via large language models for solid-state battery research.
\newblock \emph{ACS Materials Letters}, 7\penalty0 (5):\penalty0 1807--1816, Apr 2025.

\bibitem[Leong et~al.(2025)Leong, Pablo-Garc{\'\i}a, Wong, and Aspuru-Guzik]{leong2025mermaid}
Shi~Xuan Leong, Sergio Pablo-Garc{\'\i}a, Brandon Wong, and Al{\'a}n Aspuru-Guzik.
\newblock Mermaid: Universal multimodal mining of chemical reactions from pdfs using vision-language models.
\newblock Mar 2025.

\bibitem[Levine et~al.(2016)Levine, Finn, Darrell, and Abbeel]{levine2016end}
Sergey Levine, Chelsea Finn, Trevor Darrell, and Pieter Abbeel.
\newblock End-to-end training of deep visuomotor policies.
\newblock \emph{Journal of Machine Learning Research}, 17\penalty0 (39):\penalty0 1--40, Apr 2016.

\bibitem[Levine et~al.(2018)Levine, Pastor, Krizhevsky, Ibarz, and Quillen]{levine2018learning}
Sergey Levine, Peter Pastor, Alex Krizhevsky, Julian Ibarz, and Deirdre Quillen.
\newblock Learning hand-eye coordination for robotic grasping with deep learning and large-scale data collection.
\newblock \emph{The International journal of robotics research}, 37\penalty0 (4-5):\penalty0 421--436, Jun 2018.

\bibitem[Leyton-Brown et~al.(2024)Leyton-Brown, Nandwani, Zarkoob, Cameron, Newman, Raghu, et~al.]{leyton2024matching}
Kevin Leyton-Brown, Yatin Nandwani, Hedayat Zarkoob, Chris Cameron, Neil Newman, Dinesh Raghu, et~al.
\newblock Matching papers and reviewers at large conferences.
\newblock \emph{Artificial Intelligence}, 331:\penalty0 104119, Jun 2024.

\bibitem[Li et~al.(2024)Li, Guan, Dou, Feng, Wang, Xu, Wang, Chen, Wang, Xu, et~al.]{li2024can}
Bohan Li, Jiannan Guan, Longxu Dou, Yunlong Feng, Dingzirui Wang, Yang Xu, Enbo Wang, Qiguang Chen, Bichen Wang, Xiao Xu, et~al.
\newblock Can large language models understand you better? an mbti personality detection dataset aligned with population traits.
\newblock \emph{arXiv preprint arXiv:2412.12510}, 2024.

\bibitem[Li et~al.(2024{\natexlab{2}})Li, Shangguan, Zhao, Li, Liu, and Cohan]{li-etal-2024-m3sciqa}
Chuhan Li, Ziyao Shangguan, Yilun Zhao, Deyuan Li, Yixin Liu, and Arman Cohan.
\newblock {M}3{S}ci{QA}: A multi-modal multi-document scientific {QA} benchmark for evaluating foundation models.
\newblock In Yaser Al-Onaizan, Mohit Bansal, and Yun-Nung Chen, editors, \emph{Findings of the Association for Computational Linguistics: EMNLP 2024}, pages 15419--15446, Miami, Florida, USA, November 2024{\natexlab{2}}. Association for Computational Linguistics.
\newblock \doi{10.18653/v1/2024.findings-emnlp.904}.
\newblock URL \url{https://aclanthology.org/2024.findings-emnlp.904/}.

\bibitem[Li et~al.(2024{\natexlab{3}})Li, Lai, Li, Ren, Zhang, Kang, Wang, Li, Zhang, Ma, et~al.]{li2024agent}
Junkai Li, Yunghwei Lai, Weitao Li, Jingyi Ren, Meng Zhang, Xinhui Kang, Siyu Wang, Peng Li, Ya-Qin Zhang, Weizhi Ma, et~al.
\newblock Agent hospital: A simulacrum of hospital with evolvable medical agents.
\newblock \emph{arXiv preprint arXiv:2405.02957}, 2024{\natexlab{3}}.

\bibitem[Li et~al.(2025)Li, Wu, Wang, and Hu]{li2025drugpilot}
Kun Li, Zhennan Wu, Shoupeng Wang, and Wenbin Hu.
\newblock Drugpilot: Llm-based parameterized reasoning agent for drug discovery.
\newblock \emph{arXiv preprint arXiv:2505.13940}, 2025.

\bibitem[Li et~al.(2024{\natexlab{4}})Li, Wang, Xu, Wang, Feng, Kong, and Liu]{li-etal-2024-multimodal-arxiv}
Lei Li, Yuqi Wang, Runxin Xu, Peiyi Wang, Xiachong Feng, Lingpeng Kong, and Qi~Liu.
\newblock Multimodal {A}r{X}iv: A dataset for improving scientific comprehension of large vision-language models.
\newblock In Lun-Wei Ku, Andre Martins, and Vivek Srikumar, editors, \emph{Proceedings of the 62nd Annual Meeting of the Association for Computational Linguistics (Volume 1: Long Papers)}, pages 14369--14387, Bangkok, Thailand, August 2024{\natexlab{4}}. Association for Computational Linguistics.
\newblock \doi{10.18653/v1/2024.acl-long.775}.
\newblock URL \url{https://aclanthology.org/2024.acl-long.775/}.

\bibitem[Li et~al.(2020)Li, Fan, Tse, and Lin]{li2020review}
Li~Li, Yuxi Fan, Mike Tse, and Kuo-Yi Lin.
\newblock A review of applications in federated learning.
\newblock \emph{Computers \& Industrial Engineering}, 149:\penalty0 106854, Nov 2020.

\bibitem[Li et~al.(2024{\natexlab{5}})Li, Xu, Guo, Zhao, Li, Yuan, Zhang, Jiang, Xin, Dang, et~al.]{li2024chain}
Long Li, Weiwen Xu, Jiayan Guo, Ruochen Zhao, Xingxuan Li, Yuqian Yuan, Boqiang Zhang, Yuming Jiang, Yifei Xin, Ronghao Dang, et~al.
\newblock Chain of ideas: Revolutionizing research via novel idea development with llm agents.
\newblock \emph{arXiv preprint arXiv:2410.13185}, 2024{\natexlab{5}}.

\bibitem[Li et~al.(2023)Li, Hovy, and Lau]{li2023summarizing}
Miao Li, Eduard Hovy, and Jey~Han Lau.
\newblock Summarizing multiple documents with conversational structure for meta-review generation.
\newblock \emph{arXiv preprint arXiv:2305.01498}, 2023.

\bibitem[Li et~al.(2024{\natexlab{6}})Li, Lau, and Hovy]{li2024sentiment}
Miao Li, Jey~Han Lau, and Eduard Hovy.
\newblock A sentiment consolidation framework for meta-review generation.
\newblock \emph{arXiv preprint arXiv:2402.18005}, 2024{\natexlab{6}}.

\bibitem[Li et~al.(2023{\natexlab{2}})Li, Allal, Zi, Muennighoff, Kocetkov, Mou, Marone, Akiki, Li, Chim, et~al.]{li2023starcoder}
Raymond Li, Loubna~Ben Allal, Yangtian Zi, Niklas Muennighoff, Denis Kocetkov, Chenghao Mou, Marc Marone, Christopher Akiki, Jia Li, Jenny Chim, et~al.
\newblock Starcoder: may the source be with you!
\newblock \emph{arXiv preprint arXiv:2305.06161}, 2023{\natexlab{2}}.

\bibitem[Li et~al.(2024{\natexlab{7}})Li, Zhou, Shen, Zhang, Su, Li, Chen, Chen, Zhang, Zhang, et~al.]{li2024physical}
Ruifeng Li, Dongzhan Zhou, Ancheng Shen, Ao~Zhang, Mao Su, Mingqian Li, Hongyang Chen, Gang Chen, Yin Zhang, Shufei Zhang, et~al.
\newblock Physical formula enhanced multi-task learning for pharmacokinetics prediction.
\newblock \emph{arXiv preprint arXiv:2404.10354}, 2024{\natexlab{7}}.

\bibitem[Li et~al.(2025{\natexlab{2}})Li, Li, Liu, Zhou, Zhou, Yao, Zhang, and Chen]{li2025unimatch}
Ruifeng Li, Mingqian Li, Wei Liu, Yuhua Zhou, Xiangxin Zhou, Yuan Yao, Qiang Zhang, and Hongyang Chen.
\newblock Unimatch: Universal matching from atom to task for few-shot drug discovery.
\newblock \emph{arXiv preprint arXiv:2502.12453}, 2025{\natexlab{2}}.

\bibitem[Li et~al.(2025{\natexlab{3}})Li, Lu, Tang, Qi, and Ouyang]{li2025mllm}
Ruikun Li, Yan Lu, Shixiang Tang, Biqing Qi, and Wanli Ouyang.
\newblock Mllm-based discovery of intrinsic coordinates and governing equations from high-dimensional data.
\newblock \emph{arXiv preprint arXiv:2505.11940}, 2025{\natexlab{3}}.

\bibitem[Li et~al.(2024{\natexlab{8}})Li, Jing, Han, Zhou, and Du]{li2024learning}
Ruochen Li, Liqiang Jing, Chi Han, Jiawei Zhou, and Xinya Du.
\newblock Learning to generate research idea with dynamic control.
\newblock \emph{arXiv preprint arXiv:2412.14626}, 2024{\natexlab{8}}.

\bibitem[Li et~al.(2024{\natexlab{9}})Li, Patel, Wang, and Du]{li2024mlr}
Ruochen Li, Teerth Patel, Qingyun Wang, and Xinya Du.
\newblock Mlr-copilot: Autonomous machine learning research based on large language models agents.
\newblock \emph{arXiv preprint arXiv:2408.14033}, 2024{\natexlab{9}}.

\bibitem[Li et~al.(2024{\natexlab{10}})Li, Li, Wang, and Du]{li2024iqa}
Ruosen Li, Ruochen Li, Barry Wang, and Xinya Du.
\newblock Iqa-eval: Automatic evaluation of human-model interactive question answering.
\newblock \emph{Advances in Neural Information Processing Systems}, 37:\penalty0 109894--109921, Dec 2024{\natexlab{10}}.

\bibitem[Li et~al.(2025{\natexlab{4}})Li, Padilla, Bras, Dong, and Chantler]{li2025review}
Sitong Li, Stefano Padilla, Pierre~Le Bras, Junyu Dong, and Mike Chantler.
\newblock A review of llm-assisted ideation.
\newblock \emph{arXiv preprint arXiv:2503.00946}, 2025{\natexlab{4}}.

\bibitem[Li et~al.(2025{\natexlab{5}})Li, Mao, Xiao, Liao, Koffas, Chen, Ma, and Tang]{li2025large}
Wenyu Li, Zhitao Mao, Zhengyang Xiao, Xiaoping Liao, Mattheos Koffas, Yixin Chen, Hongwu Ma, and Yinjie~J Tang.
\newblock Large language model for knowledge synthesis and ai-enhanced biomanufacturing.
\newblock \emph{Trends in Biotechnology}, Mar 2025{\natexlab{5}}.

\bibitem[Li and Ouyang(2024)]{li-ouyang-2024-related}
Xiangci Li and Jessica Ouyang.
\newblock Related work and citation text generation: A survey.
\newblock In Yaser Al-Onaizan, Mohit Bansal, and Yun-Nung Chen, editors, \emph{Proceedings of the 2024 Conference on Empirical Methods in Natural Language Processing}, pages 13846--13864, Miami, Florida, USA, November 2024. Association for Computational Linguistics.
\newblock \doi{10.18653/v1/2024.emnlp-main.767}.
\newblock URL \url{https://aclanthology.org/2024.emnlp-main.767/}.

\bibitem[Li and Ouyang(2024{\natexlab{2}})]{li2024explaining}
Xiangci Li and Jessica Ouyang.
\newblock Explaining relationships among research papers.
\newblock \emph{arXiv preprint arXiv:2402.13426}, 2024{\natexlab{2}}.

\bibitem[Li and Ouyang(2024{\natexlab{3}})]{li2024related}
Xiangci Li and Jessica Ouyang.
\newblock Related work and citation text generation: A survey.
\newblock \emph{arXiv preprint arXiv:2404.11588}, 2024{\natexlab{3}}.

\bibitem[Li et~al.(2023{\natexlab{3}})Li, Lee, and Ouyang]{li2023cited}
Xiangci Li, Yi-Hui Lee, and Jessica Ouyang.
\newblock Cited text spans for citation text generation.
\newblock \emph{arXiv preprint arXiv:2309.06365}, 2023{\natexlab{3}}.

\bibitem[Li and Chen(2025)]{li2025scirgc}
Xiangyu Li and Jingqiang Chen.
\newblock Scirgc: Multi-granularity citation recommendation and citation sentence preference alignment.
\newblock \emph{arXiv preprint arXiv:2505.20103}, 2025.

\bibitem[Li et~al.(2024{\natexlab{11}})Li, Che, Chen, Liu, Wang, Liu, Yang, Pyzer-Knapp, and Cooper]{li2024sequential}
Xiaobo Li, Yu~Che, Linjiang Chen, Tao Liu, Kewei Wang, Lunjie Liu, Haofan Yang, Edward~O Pyzer-Knapp, and Andrew~I Cooper.
\newblock Sequential closed-loop bayesian optimization as a guide for organic molecular metallophotocatalyst formulation discovery.
\newblock \emph{Nature Chemistry}, 16\penalty0 (8):\penalty0 1286--1294, Jun 2024{\natexlab{11}}.

\bibitem[Li et~al.(2025{\natexlab{6}})Li, Moussa, Chen, Chen, Yu, Xue, Burns, Chiu, Dey, Lu, et~al.]{li2025autosdt}
Yifei Li, Hanane~Nour Moussa, Ziru Chen, Shijie Chen, Botao Yu, Mingyi Xue, Benjamin Burns, Tzu-Yao Chiu, Vishal Dey, Zitong Lu, et~al.
\newblock Autosdt: Scaling data-driven discovery tasks toward open co-scientists.
\newblock \emph{arXiv preprint arXiv:2506.08140}, 2025{\natexlab{6}}.

\bibitem[Li et~al.(2024{\natexlab{12}})Li, Wu, Huang, and Luan]{li2024developing}
Yugang Li, Baizhou Wu, Yuqi Huang, and Shenghua Luan.
\newblock Developing trustworthy artificial intelligence: insights from research on interpersonal, human-automation, and human-ai trust.
\newblock \emph{Frontiers in Psychology}, 15:\penalty0 1382693, Apr 2024{\natexlab{12}}.

\bibitem[Li et~al.(2024{\natexlab{13}})Li, Yang, Choi, Zhu, Hsieh, Kim, Lim, Ji, Lee, Yan, et~al.]{li2024mmsci}
Zekun Li, Xianjun Yang, Kyuri Choi, Wanrong Zhu, Ryan Hsieh, HyeonJung Kim, Jin~Hyuk Lim, Sungyoung Ji, Byungju Lee, Xifeng Yan, et~al.
\newblock Mmsci: A dataset for graduate-level multi-discipline multimodal scientific understanding.
\newblock \emph{arXiv preprint arXiv:2407.04903}, 2024{\natexlab{13}}.

\bibitem[Li et~al.(2024{\natexlab{14}})Li, Sun, Murphy, Su, Li, Zhang, Yang, and Si]{li2024survey}
Zhaoyu Li, Jialiang Sun, Logan Murphy, Qidong Su, Zenan Li, Xian Zhang, Kaiyu Yang, and Xujie Si.
\newblock A survey on deep learning for theorem proving.
\newblock \emph{arXiv preprint arXiv:2404.09939}, 2024{\natexlab{14}}.

\bibitem[Li and Zou(2019)]{li2019review}
Zhi Li and Xiaozhu Zou.
\newblock A review on personalized academic paper recommendation.
\newblock \emph{Comput. Inf. Sci.}, 12\penalty0 (1):\penalty0 33--43, Jan 2019.

\bibitem[Li and Abramson(2024)]{li2024ethnography}
Zhuofan Li and Corey~M Abramson.
\newblock Ethnography and machine learning: Synergies and new directions.
\newblock \emph{arXiv preprint arXiv:2412.06087}, 2024.

\bibitem[Li et~al.(2024{\natexlab{15}})Li, Zang, Ma, Guo, Zheng, Liu, Niu, Wang, Yang, Liu, et~al.]{li2024autokaggle}
Ziming Li, Qianbo Zang, David Ma, Jiawei Guo, Tuney Zheng, Minghao Liu, Xinyao Niu, Yue Wang, Jian Yang, Jiaheng Liu, et~al.
\newblock Autokaggle: A multi-agent framework for autonomous data science competitions.
\newblock \emph{arXiv preprint arXiv:2410.20424}, 2024{\natexlab{15}}.

\bibitem[Liang et~al.(2024)Liang, Wang, Yu, Kirsch, Pant, McDannald, Kusne, Zhao, and Takeuchi]{liang2024real}
Haotong Liang, Chuangye Wang, Heshan Yu, Dylan Kirsch, Rohit Pant, Austin McDannald, A~Gilad Kusne, Ji-Cheng Zhao, and Ichiro Takeuchi.
\newblock Real-time experiment-theory closed-loop interaction for autonomous materials science.
\newblock \emph{arXiv preprint arXiv:2410.17430}, 2024.

\bibitem[Liang(2024)]{liang2024application}
Jing Liang.
\newblock The application of artificial intelligence-assisted technology in cultural and creative product design.
\newblock \emph{Scientific Reports}, 14\penalty0 (1):\penalty0 31069, Dec 2024.

\bibitem[Liang and Sonntag(2025)]{liang2025explainable}
Siting Liang and Daniel Sonntag.
\newblock Explainable biomedical claim verification with large language models.
\newblock \emph{arXiv preprint arXiv:2502.21014}, 2025.

\bibitem[Liang et~al.(2024{\natexlab{2}})Liang, Zhang, Cao, Wang, Ding, Yang, Vodrahalli, He, Smith, Yin, et~al.]{liang2024can}
Weixin Liang, Yuhui Zhang, Hancheng Cao, Binglu Wang, Daisy~Yi Ding, Xinyu Yang, Kailas Vodrahalli, Siyu He, Daniel~Scott Smith, Yian Yin, et~al.
\newblock Can large language models provide useful feedback on research papers? a large-scale empirical analysis.
\newblock \emph{NEJM AI}, 1\penalty0 (8):\penalty0 AIoa2400196, Jul 2024{\natexlab{2}}.

\bibitem[Liang et~al.(2025)Liang, Yang, Wang, Tang, Zheng, Song, Lin, Yang, Niu, Wang, et~al.]{liang2025surveyx}
Xun Liang, Jiawei Yang, Yezhaohui Wang, Chen Tang, Zifan Zheng, Shichao Song, Zehao Lin, Yebin Yang, Simin Niu, Hanyu Wang, et~al.
\newblock Surveyx: Academic survey automation via large language models.
\newblock \emph{arXiv preprint arXiv:2502.14776}, 2025.

\bibitem[Liang et~al.(2024{\natexlab{3}})Liang, Guo, Liu, Guo, Zhou, Yang, Jiao, Pi, Zhang, and Zhang]{liang-etal-2024-scemqa}
Zhenwen Liang, Kehan Guo, Gang Liu, Taicheng Guo, Yujun Zhou, Tianyu Yang, Jiajun Jiao, Renjie Pi, Jipeng Zhang, and Xiangliang Zhang.
\newblock {S}ce{MQA}: A scientific college entrance level multimodal question answering benchmark.
\newblock In Lun-Wei Ku, Andre Martins, and Vivek Srikumar, editors, \emph{Proceedings of the 62nd Annual Meeting of the Association for Computational Linguistics (Volume 2: Short Papers)}, pages 109--119, Bangkok, Thailand, August 2024{\natexlab{3}}. Association for Computational Linguistics.
\newblock \doi{10.18653/v1/2024.acl-short.11}.
\newblock URL \url{https://aclanthology.org/2024.acl-short.11/}.

\bibitem[Liao et~al.(2024)Liao, Antoniak, Cheong, Cheng, Lee, Lo, Chang, and Zhang]{liao2024llms}
Zhehui Liao, Maria Antoniak, Inyoung Cheong, Evie Yu-Yen Cheng, Ai-Heng Lee, Kyle Lo, Joseph~Chee Chang, and Amy~X Zhang.
\newblock Llms as research tools: A large scale survey of researchers' usage and perceptions.
\newblock \emph{arXiv preprint arXiv:2411.05025}, 2024.

\bibitem[Liebling et~al.(2025)Liebling, Kane, Grunde-McLaughlin, Lang, Venugopalan, and Brenner]{liebling-etal-2025-towards}
Daniel~J. Liebling, Malcolm Kane, Madeleine Grunde-McLaughlin, Ian Lang, Subhashini Venugopalan, and Michael Brenner.
\newblock Towards {AI}-assisted academic writing.
\newblock In Peter Jansen, Bhavana Dalvi~Mishra, Harsh Trivedi, Bodhisattwa Prasad~Majumder, Tom Hope, Tushar Khot, Doug Downey, and Eric Horvitz, editors, \emph{Proceedings of the 1st Workshop on AI and Scientific Discovery: Directions and Opportunities}, pages 31--45, Albuquerque, New Mexico, USA, May 2025. Association for Computational Linguistics.
\newblock ISBN 979-8-89176-224-4.
\newblock URL \url{https://aclanthology.org/2025.aisd-main.4/}.

\bibitem[Lin and Zhu(2025)]{lin2025divergent}
Cong~William Lin and Wu~Zhu.
\newblock Divergent llm adoption and heterogeneous convergence paths in research writing.
\newblock \emph{arXiv preprint arXiv:2504.13629}, 2025.

\bibitem[Lin et~al.(2025)Lin, Peng, and Fang]{lin-etal-2025-evaluating}
Ethan Lin, Zhiyuan Peng, and Yi~Fang.
\newblock Evaluating and enhancing large language models for novelty assessment in scholarly publications.
\newblock In Peter Jansen, Bhavana Dalvi~Mishra, Harsh Trivedi, Bodhisattwa Prasad~Majumder, Tom Hope, Tushar Khot, Doug Downey, and Eric Horvitz, editors, \emph{Proceedings of the 1st Workshop on AI and Scientific Discovery: Directions and Opportunities}, pages 46--57, Albuquerque, New Mexico, USA, May 2025. Association for Computational Linguistics.
\newblock ISBN 979-8-89176-224-4.
\newblock URL \url{https://aclanthology.org/2025.aisd-main.5/}.

\bibitem[Lin et~al.(2024)Lin, Sun, Welleck, and Yang]{lin2024lean}
Haohan Lin, Zhiqing Sun, Sean Welleck, and Yiming Yang.
\newblock Lean-star: Learning to interleave thinking and proving.
\newblock \emph{arXiv preprint arXiv:2407.10040}, 2024.

\bibitem[Lin et~al.(2023)Lin, Song, Zhou, Chen, and Shi]{lin2023automated}
Jialiang Lin, Jiaxin Song, Zhangping Zhou, Yidong Chen, and Xiaodong Shi.
\newblock Automated scholarly paper review: Concepts, technologies, and challenges.
\newblock \emph{Information fusion}, 98:\penalty0 101830, Oct 2023.

\bibitem[Lin et~al.(2023{\natexlab{2}})Lin, Song, Zhou, Chen, and Shi]{lin2023moprd}
Jialiang Lin, Jiaxin Song, Zhangping Zhou, Yidong Chen, and Xiaodong Shi.
\newblock Moprd: A multidisciplinary open peer review dataset.
\newblock \emph{Neural Computing and Applications}, 35\penalty0 (34):\penalty0 24191--24206, Sep 2023{\natexlab{2}}.

\bibitem[Lin et~al.(2024{\natexlab{2}})Lin, Yang, ul~Abdeen, Sangiovanni-Vincentelli, Huang, and Jin]{linllms}
Tung-Wei Lin, Runing Yang, Zain ul~Abdeen, Alberto Sangiovanni-Vincentelli, Haibo Huang, and Ming Jin.
\newblock Llms tackle meta-analysis: Automating scientific hypothesis generation with statistical rigor.
\newblock In \emph{2nd AI4Research Workshop: Towards a Knowledge-grounded Scientific Research Lifecycle}, Dec 2024{\natexlab{2}}.

\bibitem[Lin et~al.(2024{\natexlab{3}})Lin, Ma, Shan, Zhang, Hu, Guo, Li, and Yu]{lin2024biokgbench}
Xinna Lin, Siqi Ma, Junjie Shan, Xiaojing Zhang, Shell~Xu Hu, Tiannan Guo, Stan~Z Li, and Kaicheng Yu.
\newblock Biokgbench: A knowledge graph checking benchmark of ai agent for biomedical science.
\newblock \emph{arXiv preprint arXiv:2407.00466}, 2024{\natexlab{3}}.

\bibitem[Lin et~al.(2025{\natexlab{2}})Lin, Liu, Xiang, Zeng, and Zeng]{lin2025enhancing}
Xuan Lin, Qingrui Liu, Hongxin Xiang, Daojian Zeng, and Xiangxiang Zeng.
\newblock Enhancing chemical reaction and retrosynthesis prediction with large language model and dual-task learning.
\newblock \emph{arXiv preprint arXiv:2505.02639}, 2025{\natexlab{2}}.

\bibitem[Lin(2025)]{lin2025cognitio}
Xule Lin.
\newblock Cognitio emergens: Agency, dimensions, and dynamics in human-ai knowledge co-creation.
\newblock \emph{arXiv preprint arXiv:2505.03105}, 2025.

\bibitem[Lin(2024)]{lin2024beyond}
Zhicheng Lin.
\newblock Beyond principlism: practical strategies for ethical ai use in research practices.
\newblock \emph{AI and Ethics}, pages 1--13, Oct 2024.

\bibitem[Lin(2024{\natexlab{2}})]{lin2024techniques}
Zhicheng Lin.
\newblock Techniques for supercharging academic writing with generative ai.
\newblock \emph{Nature Biomedical Engineering}, pages 1--6, Mar 2024{\natexlab{2}}.

\bibitem[Liu et~al.(2024)Liu, Feng, Xue, Wang, Wu, Lu, Zhao, Deng, Zhang, Ruan, et~al.]{liu2024deepseek}
Aixin Liu, Bei Feng, Bing Xue, Bingxuan Wang, Bochao Wu, Chengda Lu, Chenggang Zhao, Chengqi Deng, Chenyu Zhang, Chong Ruan, et~al.
\newblock Deepseek-v3 technical report.
\newblock \emph{arXiv preprint arXiv:2412.19437}, 2024.

\bibitem[Liu et~al.(2025)Liu, Li, Zhang, Wang, He, Hong, Liu, Zhang, Song, Zhu, et~al.]{liu2025advances}
Bang Liu, Xinfeng Li, Jiayi Zhang, Jinlin Wang, Tanjin He, Sirui Hong, Hongzhang Liu, Shaokun Zhang, Kaitao Song, Kunlun Zhu, et~al.
\newblock Advances and challenges in foundation agents: From brain-inspired intelligence to evolutionary, collaborative, and safe systems.
\newblock \emph{arXiv preprint arXiv:2504.01990}, 2025.

\bibitem[Liu et~al.(2025{\natexlab{2}})Liu, Wang, Cao, Ge, Wang, Zhang, Cheng, Zhao, Li, Jia, et~al.]{liu2025vision}
Chengwei Liu, Chong Wang, Jiayue Cao, Jingquan Ge, Kun Wang, Lvye Zhang, Ming-Ming Cheng, Penghai Zhao, Tianlin Li, Xiaojun Jia, et~al.
\newblock A vision for auto research with llm agents.
\newblock \emph{arXiv preprint arXiv:2504.18765}, 2025{\natexlab{2}}.

\bibitem[Liu et~al.(2023)Liu, Shen, Xin, Liu, Yuan, Wang, Ju, Zheng, Yin, Li, et~al.]{liu2023fimo}
Chengwu Liu, Jianhao Shen, Huajian Xin, Zhengying Liu, Ye~Yuan, Haiming Wang, Wei Ju, Chuanyang Zheng, Yichun Yin, Lin Li, et~al.
\newblock Fimo: A challenge formal dataset for automated theorem proving.
\newblock \emph{arXiv preprint arXiv:2309.04295}, 2023.

\bibitem[Liu et~al.(2025{\natexlab{3}})Liu, Noriega-Atala, Pyarelal, Morrison, and Cafarella]{liu-etal-2025-variable}
Chunwei Liu, Enrique Noriega-Atala, Adarsh Pyarelal, Clayton~T Morrison, and Mike Cafarella.
\newblock Variable extraction for model recovery in scientific literature.
\newblock In Peter Jansen, Bhavana Dalvi~Mishra, Harsh Trivedi, Bodhisattwa Prasad~Majumder, Tom Hope, Tushar Khot, Doug Downey, and Eric Horvitz, editors, \emph{Proceedings of the 1st Workshop on AI and Scientific Discovery: Directions and Opportunities}, pages 1--12, Albuquerque, New Mexico, USA, May 2025{\natexlab{3}}. Association for Computational Linguistics.
\newblock ISBN 979-8-89176-224-4.
\newblock \doi{10.18653/v1/2025.aisd-main.1}.
\newblock URL \url{https://aclanthology.org/2025.aisd-main.1/}.

\bibitem[Liu and Yu(2024)]{liu2024step}
Fengming Liu and Shubin Yu.
\newblock Step further towards automated social science: An ai-powered interview platform.
\newblock \emph{Available at SSRN}, Apr 2024.

\bibitem[Liu et~al.(2024{\natexlab{2}})Liu, Zhou, Li, Yuan, and Tan]{liu2024literature}
Haokun Liu, Yangqiaoyu Zhou, Mingxuan Li, Chenfei Yuan, and Chenhao Tan.
\newblock Literature meets data: A synergistic approach to hypothesis generation.
\newblock \emph{arXiv preprint arXiv:2410.17309}, 2024{\natexlab{2}}.

\bibitem[Liu et~al.(2025{\natexlab{4}})Liu, Huang, Hu, Zhou, and Tan]{liu2025hypobench}
Haokun Liu, Sicong Huang, Jingyu Hu, Yangqiaoyu Zhou, and Chenhao Tan.
\newblock Hypobench: Towards systematic and principled benchmarking for hypothesis generation.
\newblock \emph{arXiv preprint arXiv:2504.11524}, 2025{\natexlab{4}}.

\bibitem[Liu et~al.(2024{\natexlab{3}})Liu, Tang, Zhang, Zheng, and Zhu]{liu2024largelanguagemodel}
Haoyu Liu, Yifu Tang, Zizhao Zhang, Zeyu Zheng, and Tingyu Zhu.
\newblock Large language model assisted experiment design with generative human-behavior agents.
\newblock In \emph{2024 Winter Simulation Conference (WSC)}, pages 2751--2762. IEEE, Dec 2024{\natexlab{3}}.

\bibitem[Liu et~al.(2023{\natexlab{2}})Liu, Zhang, Shi, Naseem, Wang, and Tsang]{liu2023causal}
Jiachang Liu, Qi~Zhang, Chongyang Shi, Usman Naseem, Shoujin Wang, and Ivor Tsang.
\newblock Causal intervention for abstractive related work generation.
\newblock \emph{arXiv preprint arXiv:2305.13685}, 2023{\natexlab{2}}.

\bibitem[Liu et~al.(2025{\natexlab{5}})Liu, Zhu, Bai, He, Liao, Que, Wang, Zhang, Zhang, Zhang, et~al.]{liu2025comprehensive}
Jiaheng Liu, Dawei Zhu, Zhiqi Bai, Yancheng He, Huanxuan Liao, Haoran Que, Zekun Wang, Chenchen Zhang, Ge~Zhang, Jiebin Zhang, et~al.
\newblock A comprehensive survey on long context language modeling.
\newblock \emph{arXiv preprint arXiv:2503.17407}, 2025{\natexlab{5}}.

\bibitem[Liu et~al.(2024{\natexlab{4}})Liu, Li, Chen, Chen, Bao, and Li]{liu2024synchart}
Mengchen Liu, Qixiu Li, Dongdong Chen, Dong Chen, Jianmin Bao, and Yunsheng Li.
\newblock Synchart: Synthesizing charts from language models.
\newblock \emph{arXiv preprint arXiv:2409.16517}, 2024{\natexlab{4}}.

\bibitem[Liu et~al.(2023{\natexlab{3}})Liu, Yang, Jia, Zhang, Zhou, Dai, Yang, and Vosoughi]{liu2023training}
Ruibo Liu, Ruixin Yang, Chenyan Jia, Ge~Zhang, Denny Zhou, Andrew~M Dai, Diyi Yang, and Soroush Vosoughi.
\newblock Training socially aligned language models in simulated human society.
\newblock \emph{arXiv preprint arXiv:2305.16960}, 2, 2023{\natexlab{3}}.

\bibitem[Liu and Shah(2023)]{liu2023reviewergpt}
Ryan Liu and Nihar~B Shah.
\newblock Reviewergpt? an exploratory study on using large language models for paper reviewing.
\newblock \emph{arXiv preprint arXiv:2306.00622}, 2023.

\bibitem[Liu et~al.(2023{\natexlab{4}})Liu, Cao, Yang, and Wen]{liu2023generating}
Shuaiqi Liu, Jiannong Cao, Ruosong Yang, and Zhiyuan Wen.
\newblock Generating a structured summary of numerous academic papers: Dataset and method.
\newblock \emph{arXiv preprint arXiv:2302.04580}, 2023{\natexlab{4}}.

\bibitem[Liu et~al.(2024{\natexlab{5}})Liu, Gao, and Li]{liu2024large}
Siyi Liu, Chen Gao, and Yong Li.
\newblock Large language model agent for hyper-parameter optimization.
\newblock \emph{arXiv preprint arXiv:2402.01881}, 2024{\natexlab{5}}.

\bibitem[Liu et~al.(2024{\natexlab{6}})Liu, Lu, Chen, Hu, Zhao, Lu, and Zhao]{liu2024drugagent}
Sizhe Liu, Yizhou Lu, Siyu Chen, Xiyang Hu, Jieyu Zhao, Yingzhou Lu, and Yue Zhao.
\newblock Drugagent: Automating ai-aided drug discovery programming through llm multi-agent collaboration.
\newblock \emph{arXiv preprint arXiv:2411.15692}, 2024{\natexlab{6}}.

\bibitem[Liu et~al.(2025{\natexlab{6}})Liu, Yang, Wang, Bing, Zhang, Zhou, Li, Li, Cambria, and Ouyang]{liu2025moose}
Wanhao Liu, Zonglin Yang, Jue Wang, Lidong Bing, Di~Zhang, Dongzhan Zhou, Yuqiang Li, Houqiang Li, Erik Cambria, and Wanli Ouyang.
\newblock Moose-chem3: Toward experiment-guided hypothesis ranking via simulated experimental feedback.
\newblock \emph{arXiv preprint arXiv:2505.17873}, 2025{\natexlab{6}}.

\bibitem[Liu et~al.(2025{\natexlab{7}})Liu, Dong, Gao, Feng, and Pang]{liu2025improving}
Xiao Liu, Xinyi Dong, Xinyang Gao, Yansong Feng, and Xun Pang.
\newblock Improving research idea generation through data: An empirical investigation in social science.
\newblock \emph{arXiv preprint arXiv:2505.21396}, 2025{\natexlab{7}}.

\bibitem[Liu et~al.(2025{\natexlab{8}})Liu, Song, Wang, and Chen]{liu2025select}
Xiaochuan Liu, Ruihua Song, Xiting Wang, and Xu~Chen.
\newblock Select, read, and write: A multi-agent framework of full-text-based related work generation.
\newblock \emph{arXiv preprint arXiv:2505.19647}, 2025{\natexlab{8}}.

\bibitem[Liu et~al.(2025{\natexlab{9}})Liu, Yang, Poria, Nguyen, and Cambria]{liu2025harnessing}
Yan Liu, Zonglin Yang, Soujanya Poria, Thanh-Son Nguyen, and Erik Cambria.
\newblock Harnessing large language models for scientific novelty detection.
\newblock \emph{arXiv preprint arXiv:2505.24615}, 2025{\natexlab{9}}.

\bibitem[Liu et~al.(2025{\natexlab{10}})Liu, Yu, Xu, Li, and Zhu]{liu2025survey}
Yijun Liu, Jinzheng Yu, Yang Xu, Zhongyang Li, and Qingfu Zhu.
\newblock A survey on transformer context extension: Approaches and evaluation.
\newblock \emph{arXiv preprint arXiv:2503.13299}, 2025{\natexlab{10}}.

\bibitem[Liu et~al.(2024{\natexlab{7}})Liu, Chen, Cheng, Yu, Ran, Mo, Tang, and Huang]{liu2024ai}
Yiren Liu, Si~Chen, Haocong Cheng, Mengxia Yu, Xiao Ran, Andrew Mo, Yiliu Tang, and Yun Huang.
\newblock How ai processing delays foster creativity: Exploring research question co-creation with an llm-based agent.
\newblock In \emph{Proceedings of the 2024 CHI Conference on Human Factors in Computing Systems}, pages 1--25, May 2024{\natexlab{7}}.

\bibitem[Liu et~al.(2025{\natexlab{11}})Liu, Yang, Xie, Ni, Gao, Li, Tang, Ouyang, Cambria, and Zhou]{liu2025researchbench}
Yujie Liu, Zonglin Yang, Tong Xie, Jinjie Ni, Ben Gao, Yuqiang Li, Shixiang Tang, Wanli Ouyang, Erik Cambria, and Dongzhan Zhou.
\newblock Researchbench: Benchmarking llms in scientific discovery via inspiration-based task decomposition.
\newblock \emph{arXiv preprint arXiv:2503.21248}, 2025{\natexlab{11}}.

\bibitem[Liu et~al.(2025{\natexlab{12}})Liu, Lv, Zhang, Yuan, and Tian]{liu2025bioprobench}
Yuyang Liu, Liuzhenghao Lv, Xiancheng Zhang, Li~Yuan, and Yonghong Tian.
\newblock Bioprobench: Comprehensive dataset and benchmark in biological protocol understanding and reasoning.
\newblock \emph{arXiv preprint arXiv:2505.07889}, 2025{\natexlab{12}}.

\bibitem[Liu et~al.(2024{\natexlab{8}})Liu, Liu, Zhu, Lei, Yang, Zhang, Li, and Liu]{liu2024aigs}
Zijun Liu, Kaiming Liu, Yiqi Zhu, Xuanyu Lei, Zonghan Yang, Zhenhe Zhang, Peng Li, and Yang Liu.
\newblock Aigs: Generating science from ai-powered automated falsification.
\newblock \emph{arXiv preprint arXiv:2411.11910}, 2024{\natexlab{8}}.

\bibitem[Lo et~al.(2024)Lo, Baird, Schrier, Blaiszik, Carson, Foster, Aguilar-Granda, Kalinin, Maruyama, Politi, et~al.]{lo2024review}
Stanley Lo, Sterling~G Baird, Joshua Schrier, Ben Blaiszik, Nessa Carson, Ian Foster, Andr{\'e}s Aguilar-Granda, Sergei~V Kalinin, Benji Maruyama, Maria Politi, et~al.
\newblock Review of low-cost self-driving laboratories in chemistry and materials science: the “frugal twin” concept.
\newblock \emph{Digital Discovery}, 3\penalty0 (5):\penalty0 842--868, Feb 2024.

\bibitem[Loffredo and Yun(2025)]{loffredo2025agent}
Joseph~R Loffredo and Suyeol Yun.
\newblock Agent-enhanced large language models for researching political institutions.
\newblock \emph{arXiv preprint arXiv:2503.13524}, 2025.

\bibitem[Lou et~al.(2024)Lou, Xu, Wang, Du, Kamoi, Lu, Xie, Sun, Zhang, Ahn, et~al.]{lou2024aaar}
Renze Lou, Hanzi Xu, Sijia Wang, Jiangshu Du, Ryo Kamoi, Xiaoxin Lu, Jian Xie, Yuxuan Sun, Yusen Zhang, Jihyun~Janice Ahn, et~al.
\newblock Aaar-1.0: Assessing ai's potential to assist research.
\newblock \emph{arXiv preprint arXiv:2410.22394}, 2024.

\bibitem[Lozhkov et~al.(2024)Lozhkov, Li, Allal, Cassano, Lamy-Poirier, Tazi, Tang, Pykhtar, Liu, Wei, et~al.]{lozhkov2024starcoder}
Anton Lozhkov, Raymond Li, Loubna~Ben Allal, Federico Cassano, Joel Lamy-Poirier, Nouamane Tazi, Ao~Tang, Dmytro Pykhtar, Jiawei Liu, Yuxiang Wei, et~al.
\newblock Starcoder 2 and the stack v2: The next generation.
\newblock \emph{arXiv preprint arXiv:2402.19173}, 2024.

\bibitem[Lu et~al.(2024)Lu, Lu, Lange, Foerster, Clune, and Ha]{lu2024ai}
Chris Lu, Cong Lu, Robert~Tjarko Lange, Jakob Foerster, Jeff Clune, and David Ha.
\newblock The ai scientist: Towards fully automated open-ended scientific discovery.
\newblock \emph{arXiv preprint arXiv:2408.06292}, 2024.

\bibitem[Lu et~al.(2024{\natexlab{2}})Lu, Hu, and Clune]{lu2024beyond}
Cong Lu, Shengran Hu, and Jeff Clune.
\newblock Beyond benchmarking: Automated capability discovery via model self-exploration.
\newblock In \emph{Language Gamification-NeurIPS 2024 Workshop}, Oct 2024{\natexlab{2}}.

\bibitem[Lu et~al.(2024{\natexlab{3}})Lu, Hu, and Li]{lu2024drugclip}
Yingzhou Lu, Yaojun Hu, and Chenhao Li.
\newblock Drugclip: Contrastive drug-disease interaction for drug repurposing.
\newblock \emph{arXiv preprint arXiv:2407.02265}, 2024{\natexlab{3}}.

\bibitem[Luke et~al.(2024)Luke, Ashwini, Isaias, and Robb]{predicting2024}
Hewitt Luke, Ashokkumar Ashwini, Ghezae Isaias, and Willer Robb.
\newblock Predicting results of social science experiments using large language models, Aug 2024.
\newblock URL \url{https://samim.io/dl/Predicting%20results%20of%20social%20science%20experiments%20using%20large%20language%20models.pdf}.
\newblock Predicting Results of Social Science Experiments Using Large Language Models.

\bibitem[Luo et~al.(2025)Luo, Jia, Xiong, Li, Guo, Yu, Wei, and Zhang]{luo2025benchmarking}
Erpai Luo, Jinmeng Jia, Yifan Xiong, Xiangyu Li, Xiaobo Guo, Baoqi Yu, Lei Wei, and Xuegong Zhang.
\newblock Benchmarking ai scientists in omics data-driven biological research.
\newblock \emph{arXiv preprint arXiv:2505.08341}, 2025.

\bibitem[Luo et~al.(2024)Luo, Qian, Glass, and Ma]{luo2024clinical}
Junyu Luo, Cheng Qian, Lucas Glass, and Fenglong Ma.
\newblock Clinical trial retrieval via multi-grained similarity learning.
\newblock In \emph{Proceedings of the 47th International ACM SIGIR Conference on Research and Development in Information Retrieval}, pages 2950--2954, Jul 2024.

\bibitem[Luo et~al.(2025{\natexlab{2}})Luo, Xie, Li, Zhang, Cao, Huang, Qu, Zhu, Chen, Jiang, et~al.]{luo2025physics}
Man Luo, Zikai Xie, Huirong Li, Baicheng Zhang, Jiaqi Cao, Yan Huang, Hang Qu, Qing Zhu, Linjiang Chen, Jun Jiang, et~al.
\newblock Physics-informed, dual-objective optimization of high-entropy-alloy nanozymes by a robotic ai chemist.
\newblock \emph{Matter}, 8\penalty0 (4), Apr 2025{\natexlab{2}}.

\bibitem[Luo et~al.(2024{\natexlab{2}})Luo, Tang, Jiang, Feng, Zhang, and Ding]{luo2024generating}
Shunyang Luo, Yuqi Tang, Mingyuan Jiang, Kehua Feng, Qiang Zhang, and Keyan Ding.
\newblock Generating multiple choice questions from scientific literature via large language models.
\newblock In \emph{2024 IEEE International Conference on Knowledge Graph (ICKG)}, pages 219--226, Feb 2024{\natexlab{2}}.
\newblock \doi{10.1109/ICKG63256.2024.00035}.

\bibitem[Luo et~al.(2025{\natexlab{3}})Luo, Rechardt, Sun, Nejad, Y{\'a}{\~n}ez, Yilmaz, Lee, Cohen, Borghesani, Pashkov, et~al.]{luo2025large}
Xiaoliang Luo, Akilles Rechardt, Guangzhi Sun, Kevin~K Nejad, Felipe Y{\'a}{\~n}ez, Bati Yilmaz, Kangjoo Lee, Alexandra~O Cohen, Valentina Borghesani, Anton Pashkov, et~al.
\newblock Large language models surpass human experts in predicting neuroscience results.
\newblock \emph{Nature human behaviour}, 9\penalty0 (2):\penalty0 305--315, Nov 2025{\natexlab{3}}.

\bibitem[Luo et~al.(2023)Luo, Xie, and Ananiadou]{luo2023citationsum}
Zheheng Luo, Qianqian Xie, and Sophia Ananiadou.
\newblock Citationsum: Citation-aware graph contrastive learning for scientific paper summarization.
\newblock In \emph{Proceedings of the ACM web conference 2023}, pages 1843--1852, Apr 2023.

\bibitem[Luo et~al.(2025{\natexlab{4}})Luo, Yang, Xu, Yang, and Du]{luo2025llm4sr}
Ziming Luo, Zonglin Yang, Zexin Xu, Wei Yang, and Xinya Du.
\newblock Llm4sr: A survey on large language models for scientific research.
\newblock \emph{arXiv preprint arXiv:2501.04306}, 2025{\natexlab{4}}.

\bibitem[M.~Bran et~al.(2024)M.~Bran, Cox, Schilter, Baldassari, White, and Schwaller]{m2024augmenting}
Andres M.~Bran, Sam Cox, Oliver Schilter, Carlo Baldassari, Andrew~D White, and Philippe Schwaller.
\newblock Augmenting large language models with chemistry tools.
\newblock \emph{Nature Machine Intelligence}, 6\penalty0 (5):\penalty0 525--535, May 2024.

\bibitem[Ma et~al.(2024)Ma, Wang, Guo, Sun, Tenenbaum, Rus, Gan, and Matusik]{ma2024llm}
Pingchuan Ma, Tsun-Hsuan Wang, Minghao Guo, Zhiqing Sun, Joshua~B Tenenbaum, Daniela Rus, Chuang Gan, and Wojciech Matusik.
\newblock Llm and simulation as bilevel optimizers: A new paradigm to advance physical scientific discovery.
\newblock \emph{arXiv preprint arXiv:2405.09783}, 2024.

\bibitem[Ma et~al.(2025)Ma, Zhou, and Li]{ma2025automated}
Qinyu Ma, Yuhao Zhou, and Jianfeng Li.
\newblock Automated retrosynthesis planning of macromolecules using large language models and knowledge graphs.
\newblock \emph{Macromolecular Rapid Communications}, page 2500065, Feb 2025.

\bibitem[Ma et~al.(2021)Ma, Zhang, Zhang, and Liu]{ma2021chronological}
Shutian Ma, Heng Zhang, Chengzhi Zhang, and Xiaozhong Liu.
\newblock Chronological citation recommendation with time preference.
\newblock \emph{Scientometrics}, 126:\penalty0 2991--3010, Feb 2021.

\bibitem[Ma et~al.(2024{\natexlab{2}})Ma, Gou, Hao, Xu, Wang, Pan, Yang, Cao, and Sun]{ma-etal-2024-sciagent}
Yubo Ma, Zhibin Gou, Junheng Hao, Ruochen Xu, Shuohang Wang, Liangming Pan, Yujiu Yang, Yixin Cao, and Aixin Sun.
\newblock {S}ci{A}gent: Tool-augmented language models for scientific reasoning.
\newblock In Yaser Al-Onaizan, Mohit Bansal, and Yun-Nung Chen, editors, \emph{Proceedings of the 2024 Conference on Empirical Methods in Natural Language Processing}, pages 15701--15736, Miami, Florida, USA, November 2024{\natexlab{2}}. Association for Computational Linguistics.
\newblock \doi{10.18653/v1/2024.emnlp-main.880}.
\newblock URL \url{https://aclanthology.org/2024.emnlp-main.880/}.

\bibitem[Machi et~al.(2025)Machi, Akiyama, Nagata, and Yoshioka]{machi2025framework}
Kojiro Machi, Seiji Akiyama, Yuuya Nagata, and Masaharu Yoshioka.
\newblock A framework for reviewing the results of automated conversion of structured organic synthesis procedures from the literature.
\newblock \emph{Digital Discovery}, 4\penalty0 (1):\penalty0 172--180, Nov 2025.

\bibitem[MacLeod et~al.(2020)MacLeod, Parlane, Morrissey, H{\"a}se, Roch, Dettelbach, Moreira, Yunker, Rooney, Deeth, et~al.]{macleod2020self}
Benjamin~P MacLeod, Fraser~GL Parlane, Thomas~D Morrissey, Florian H{\"a}se, Lo{\"\i}c~M Roch, Kevan~E Dettelbach, Raphaell Moreira, Lars~PE Yunker, Michael~B Rooney, Joseph~R Deeth, et~al.
\newblock Self-driving laboratory for accelerated discovery of thin-film materials.
\newblock \emph{Science Advances}, 6\penalty0 (20):\penalty0 eaaz8867, 2020.

\bibitem[Maharjan(2024)]{maharjan2024benchmark}
Puja Maharjan.
\newblock Benchmark for evaluation and analysis of citation recommendation models.
\newblock \emph{arXiv preprint arXiv:2412.07713}, 2024.

\bibitem[Majumder et~al.(2024)Majumder, Surana, Agarwal, Hazra, Sabharwal, and Clark]{majumder2024position}
Bodhisattwa~Prasad Majumder, Harshit Surana, Dhruv Agarwal, Sanchaita Hazra, Ashish Sabharwal, and Peter Clark.
\newblock Position: data-driven discovery with large generative models.
\newblock In \emph{Forty-first International Conference on Machine Learning}, May 2024.

\bibitem[Majumder et~al.(2024{\natexlab{2}})Majumder, Surana, Agarwal, Mishra, Meena, Prakhar, Vora, Khot, Sabharwal, and Clark]{majumder2024discoverybench}
Bodhisattwa~Prasad Majumder, Harshit Surana, Dhruv Agarwal, Bhavana~Dalvi Mishra, Abhijeetsingh Meena, Aryan Prakhar, Tirth Vora, Tushar Khot, Ashish Sabharwal, and Peter Clark.
\newblock Discoverybench: Towards data-driven discovery with large language models.
\newblock \emph{arXiv preprint arXiv:2407.01725}, 2024{\natexlab{2}}.

\bibitem[Maleki et~al.(2024)Maleki, Huetter, Chuang, Richmond, Scalia, and Biancalani]{maleki2024efficient}
Sepideh Maleki, Jan-Christian Huetter, Kangway~V Chuang, David Richmond, Gabriele Scalia, and Tommaso Biancalani.
\newblock Efficient fine-tuning of single-cell foundation models enables zero-shot molecular perturbation prediction.
\newblock \emph{arXiv preprint arXiv:2412.13478}, 2024.

\bibitem[Mandal et~al.(2024)Mandal, Soni, Zaki, Smedskjaer, Wondraczek, Wondraczek, Gosvami, and Krishnan]{mandal2024autonomous}
Indrajeet Mandal, Jitendra Soni, Mohd Zaki, Morten~M Smedskjaer, Katrin Wondraczek, Lothar Wondraczek, Nitya~Nand Gosvami, and NM~Krishnan.
\newblock Autonomous microscopy experiments through large language model agents.
\newblock \emph{arXiv preprint arXiv:2501.10385}, 2024.

\bibitem[Manning et~al.(2024)Manning, Zhu, and Horton]{manning2024automated}
Benjamin~S Manning, Kehang Zhu, and John~J Horton.
\newblock Automated social science: Language models as scientist and subjects.
\newblock Technical report, National Bureau of Economic Research, Apr 2024.

\bibitem[Mansour et~al.(2025)Mansour, Rahimi, and Alrahabi]{mansour-etal-2025-well}
Nacef~Ben Mansour, Hamed Rahimi, and Motasem Alrahabi.
\newblock How well do large language models extract keywords? a systematic evaluation on scientific corpora.
\newblock In Peter Jansen, Bhavana Dalvi~Mishra, Harsh Trivedi, Bodhisattwa Prasad~Majumder, Tom Hope, Tushar Khot, Doug Downey, and Eric Horvitz, editors, \emph{Proceedings of the 1st Workshop on AI and Scientific Discovery: Directions and Opportunities}, pages 13--21, Albuquerque, New Mexico, USA, May 2025. Association for Computational Linguistics.
\newblock ISBN 979-8-89176-224-4.
\newblock \doi{10.18653/v1/2025.aisd-main.2}.
\newblock URL \url{https://aclanthology.org/2025.aisd-main.2/}.

\bibitem[Markowitz(2024)]{markowitz2024complexity}
David~M Markowitz.
\newblock From complexity to clarity: How ai enhances perceptions of scientists and the public's understanding of science.
\newblock \emph{PNAS nexus}, 3\penalty0 (9):\penalty0 pgae387, Sep 2024.

\bibitem[Martin-Boyle et~al.(2024)Martin-Boyle, Tyagi, Hearst, and Kang]{martin2024shallow}
Anna Martin-Boyle, Aahan Tyagi, Marti~A Hearst, and Dongyeop Kang.
\newblock Shallow synthesis of knowledge in gpt-generated texts: A case study in automatic related work composition.
\newblock \emph{arXiv preprint arXiv:2402.12255}, 2024.

\bibitem[Maruf et~al.(2024)Maruf, Daw, Mehrab, Manogaran, Neog, Sawhney, Khurana, Balhoff, Bakis, Altintas, et~al.]{maruf2024vlm4bio}
M~Maruf, Arka Daw, Kazi~Sajeed Mehrab, Harish~Babu Manogaran, Abhilash Neog, Medha Sawhney, Mridul Khurana, James Balhoff, Yasin Bakis, Bahadir Altintas, et~al.
\newblock Vlm4bio: A benchmark dataset to evaluate pretrained vision-language models for trait discovery from biological images.
\newblock \emph{Advances in Neural Information Processing Systems}, 37:\penalty0 131035--131071, Sep 2024.

\bibitem[Masry et~al.(2022)Masry, Long, Tan, Joty, and Hoque]{masry-etal-2022-chartqa}
Ahmed Masry, Do~Xuan Long, Jia~Qing Tan, Shafiq Joty, and Enamul Hoque.
\newblock {C}hart{QA}: A benchmark for question answering about charts with visual and logical reasoning.
\newblock In Smaranda Muresan, Preslav Nakov, and Aline Villavicencio, editors, \emph{Findings of the Association for Computational Linguistics: ACL 2022}, pages 2263--2279, Dublin, Ireland, May 2022. Association for Computational Linguistics.
\newblock \doi{10.18653/v1/2022.findings-acl.177}.
\newblock URL \url{https://aclanthology.org/2022.findings-acl.177/}.

\bibitem[Masry et~al.(2024)Masry, Shahmohammadi, Parvez, Hoque, and Joty]{masry-etal-2024-chartinstruct}
Ahmed Masry, Mehrad Shahmohammadi, Md~Rizwan Parvez, Enamul Hoque, and Shafiq Joty.
\newblock {C}hart{I}nstruct: Instruction tuning for chart comprehension and reasoning.
\newblock In Lun-Wei Ku, Andre Martins, and Vivek Srikumar, editors, \emph{Findings of the Association for Computational Linguistics: ACL 2024}, pages 10387--10409, Bangkok, Thailand, August 2024. Association for Computational Linguistics.
\newblock \doi{10.18653/v1/2024.findings-acl.619}.
\newblock URL \url{https://aclanthology.org/2024.findings-acl.619/}.

\bibitem[Masry et~al.(2025)Masry, Thakkar, Bajaj, Kartha, Hoque, and Joty]{masry-etal-2025-chartgemma}
Ahmed Masry, Megh Thakkar, Aayush Bajaj, Aaryaman Kartha, Enamul Hoque, and Shafiq Joty.
\newblock {C}hart{G}emma: Visual instruction-tuning for chart reasoning in the wild.
\newblock In Owen Rambow, Leo Wanner, Marianna Apidianaki, Hend Al-Khalifa, Barbara~Di Eugenio, Steven Schockaert, Kareem Darwish, and Apoorv Agarwal, editors, \emph{Proceedings of the 31st International Conference on Computational Linguistics: Industry Track}, pages 625--643, Abu Dhabi, UAE, January 2025. Association for Computational Linguistics.
\newblock URL \url{https://aclanthology.org/2025.coling-industry.54/}.

\bibitem[Mathur et~al.(2025)Mathur, van~der Vleuten, Yager, and Tsai]{mathur2025vision}
Shray Mathur, Noah van~der Vleuten, Kevin~G Yager, and Esther Tsai.
\newblock Vision: A modular ai assistant for natural human-instrument interaction at scientific user facilities.
\newblock \emph{Machine Learning: Science and Technology}, Jun 2025.

\bibitem[McShane et~al.(2025)McShane, Gal, and Duhachek]{mcshane2025artificial}
Blakeley~B McShane, David Gal, and Adam Duhachek.
\newblock Artificial intelligence and dichotomania.
\newblock \emph{Judgment and Decision Making}, 20:\penalty0 e23, Apr 2025.

\bibitem[Meincke et~al.(2024)Meincke, Girotra, Nave, Terwiesch, and Ulrich]{meincke2024innovation}
Lennart Meincke, Karan Girotra, Gideon Nave, Christian Terwiesch, and Karl~T. Ulrich.
\newblock Using large language models for idea generation in innovation, Aug 2024.
\newblock SSRN.

\bibitem[Meincke et~al.(2024{\natexlab{2}})Meincke, Mollick, and Terwiesch]{meincke2024prompting}
Lennart Meincke, Ethan~R. Mollick, and Christian Terwiesch.
\newblock Prompting diverse ideas: Increasing ai idea variance, Feb 2024{\natexlab{2}}.
\newblock SSRN.

\bibitem[Meng et~al.(2025)Meng, Griesemer, Cao, Seo, and Liu]{meng2025physics}
Chuizheng Meng, Sam Griesemer, Defu Cao, Sungyong Seo, and Yan Liu.
\newblock When physics meets machine learning: A survey of physics-informed machine learning.
\newblock \emph{Machine Learning for Computational Science and Engineering}, 1\penalty0 (1):\penalty0 1--23, May 2025.

\bibitem[Meng et~al.(2024)Meng, Shao, Lu, Gao, Zhang, Qiao, and Luo]{meng-etal-2024-chartassistant}
Fanqing Meng, Wenqi Shao, Quanfeng Lu, Peng Gao, Kaipeng Zhang, Yu~Qiao, and Ping Luo.
\newblock {C}hart{A}ssistant: A universal chart multimodal language model via chart-to-table pre-training and multitask instruction tuning.
\newblock In Lun-Wei Ku, Andre Martins, and Vivek Srikumar, editors, \emph{Findings of the Association for Computational Linguistics: ACL 2024}, pages 7775--7803, Bangkok, Thailand, August 2024. Association for Computational Linguistics.
\newblock \doi{10.18653/v1/2024.findings-acl.463}.
\newblock URL \url{https://aclanthology.org/2024.findings-acl.463/}.

\bibitem[Meng et~al.(2024{\natexlab{2}})Meng, Shao, Lu, Gao, Zhang, Qiao, and Luo]{meng2024chartassisstant}
Fanqing Meng, Wenqi Shao, Quanfeng Lu, Peng Gao, Kaipeng Zhang, Yu~Qiao, and Ping Luo.
\newblock Chartassisstant: A universal chart multimodal language model via chart-to-table pre-training and multitask instruction tuning.
\newblock \emph{arXiv preprint arXiv:2401.02384}, 2024{\natexlab{2}}.

\bibitem[Mensah(2023)]{mensah2023artificial}
George~Benneh Mensah.
\newblock Artificial intelligence and ethics: a comprehensive review of bias mitigation, transparency, and accountability in ai systems.
\newblock \emph{Preprint, November}, 10\penalty0 (1), Nov 2023.

\bibitem[Merchant et~al.(2023)Merchant, Batzner, Schoenholz, Aykol, Cheon, and Cubuk]{merchant2023scaling}
Amil Merchant, Simon Batzner, Samuel~S Schoenholz, Muratahan Aykol, Gowoon Cheon, and Ekin~Dogus Cubuk.
\newblock Scaling deep learning for materials discovery.
\newblock \emph{Nature}, 624\penalty0 (7990):\penalty0 80--85, Nov 2023.

\bibitem[Metropolitansky and Larson(2025)]{metropolitansky2025towards}
Dasha Metropolitansky and Jonathan Larson.
\newblock Towards effective extraction and evaluation of factual claims.
\newblock \emph{arXiv preprint arXiv:2502.10855}, 2025.

\bibitem[Miao et~al.(2023)Miao, Teh, and Rainforth]{miao2023selfcheck}
Ning Miao, Yee~Whye Teh, and Tom Rainforth.
\newblock Selfcheck: Using llms to zero-shot check their own step-by-step reasoning.
\newblock \emph{arXiv preprint arXiv:2308.00436}, 2023.

\bibitem[Mieszczanek et~al.(2024)Mieszczanek, Corke, Mehanian, Dalton, and Hutmacher]{mieszczanek2024towards}
Pawel Mieszczanek, Peter Corke, Courosh Mehanian, Paul~D Dalton, and Dietmar~W Hutmacher.
\newblock Towards industry-ready additive manufacturing: Ai-enabled closed-loop control for 3d melt electrowriting.
\newblock \emph{Communications Engineering}, 3\penalty0 (1):\penalty0 158, 2024.

\bibitem[Mou et~al.(2024)Mou, Ding, He, Wang, Liang, Zhang, Sun, Lin, Zhou, Huang, et~al.]{mou2024individual}
Xinyi Mou, Xuanwen Ding, Qi~He, Liang Wang, Jingcong Liang, Xinnong Zhang, Libo Sun, Jiayu Lin, Jie Zhou, Xuanjing Huang, et~al.
\newblock From individual to society: A survey on social simulation driven by large language model-based agents.
\newblock \emph{arXiv preprint arXiv:2412.03563}, 2024.

\bibitem[Moured et~al.(2024)Moured, Alzalabny, Osman, Schwarz, M{\"u}ller, and Stiefelhagen]{moured2024chartformer}
Omar Moured, Sara Alzalabny, Anas Osman, Thorsten Schwarz, Karin M{\"u}ller, and Rainer Stiefelhagen.
\newblock Chartformer: A large vision language model for converting chart images into tactile accessible svgs.
\newblock In \emph{International Conference on Computers Helping People with Special Needs}, pages 299--305. Springer, May 2024.

\bibitem[Mroz et~al.(2025)Mroz, Basford, Hastedt, Jayasekera, Mosquera-Lois, Sedgwick, Ballester, Bocarsly, del R{\'\i}o~Chanona, Evans, et~al.]{mroz2025cross}
Austin~M Mroz, Annabel~R Basford, Friedrich Hastedt, Isuru~Shavindra Jayasekera, Irea Mosquera-Lois, Ruby Sedgwick, Pedro~J Ballester, Joshua~D Bocarsly, Ehecatl~Antonio del R{\'\i}o~Chanona, Matthew~L Evans, et~al.
\newblock Cross-disciplinary perspectives on the potential for artificial intelligence across chemistry.
\newblock \emph{Chemical Society Reviews}, Apr 2025.

\bibitem[Muangkammuen et~al.(2022)Muangkammuen, Fukumoto, Li, and Suzuki]{muangkammuen2022exploiting}
Panitan Muangkammuen, Fumiyo Fukumoto, Jiyi Li, and Yoshimi Suzuki.
\newblock Exploiting labeled and unlabeled data via transformer fine-tuning for peer-review score prediction.
\newblock In \emph{Findings of the Association for Computational Linguistics: EMNLP 2022}, pages 2233--2240, Dec 2022.

\bibitem[Muharram and Purwarianti(2024)]{muharram2024enhancing}
Arief~Purnama Muharram and Ayu Purwarianti.
\newblock Enhancing natural language inference performance with knowledge graph for covid-19 automated fact-checking in indonesian language.
\newblock \emph{arXiv preprint arXiv:2409.00061}, 2024.

\bibitem[Myakala et~al.(2024)Myakala, Jonnalagadda, and Bura]{myakala2024federated}
Praveen~Kumar Myakala, Anil~Kumar Jonnalagadda, and Chiranjeevi Bura.
\newblock Federated learning and data privacy: A review of challenges and opportunities.
\newblock \emph{International Journal of Research Publication and Reviews}, 5\penalty0 (12):\penalty0 10--55248, Jan 2024.

\bibitem[Mysore et~al.(2025)Mysore, Das, Cao, and Sarrafzadeh]{mysore2025prototypical}
Sheshera Mysore, Debarati Das, Hancheng Cao, and Bahareh Sarrafzadeh.
\newblock Prototypical human-ai collaboration behaviors from llm-assisted writing in the wild.
\newblock \emph{arXiv preprint arXiv:2505.16023}, 2025.

\bibitem[Nair et~al.(2024)Nair, Tan, Su, Gere, Wang, and Wang]{nair2024closing}
Inderjeet Nair, Jiaye Tan, Xiaotian Su, Anne Gere, Xu~Wang, and Lu~Wang.
\newblock Closing the loop: Learning to generate writing feedback via language model simulated student revisions.
\newblock \emph{arXiv preprint arXiv:2410.08058}, 2024.

\bibitem[Narayanan et~al.(2024)Narayanan, Braza, Griffiths, Ponnapati, Bou, Laurent, Kabeli, Wellawatte, Cox, Rodriques, et~al.]{narayanan2024aviary}
Siddharth Narayanan, James~D Braza, Ryan-Rhys Griffiths, Manu Ponnapati, Albert Bou, Jon Laurent, Ori Kabeli, Geemi Wellawatte, Sam Cox, Samuel~G Rodriques, et~al.
\newblock Aviary: training language agents on challenging scientific tasks.
\newblock \emph{arXiv preprint arXiv:2412.21154}, 2024.

\bibitem[Nathani et~al.(2025)Nathani, Madaan, Roberts, Bashlykov, Menon, Moens, Budhiraja, Magka, Vorotilov, Chaurasia, et~al.]{nathani2025mlgym}
Deepak Nathani, Lovish Madaan, Nicholas Roberts, Nikolay Bashlykov, Ajay Menon, Vincent Moens, Amar Budhiraja, Despoina Magka, Vladislav Vorotilov, Gaurav Chaurasia, et~al.
\newblock Mlgym: A new framework and benchmark for advancing ai research agents.
\newblock \emph{arXiv preprint arXiv:2502.14499}, 2025.

\bibitem[Nature(2023)]{nature}
Springer Nature.
\newblock Snapp: Springer nature's next-generation peer review system.
\newblock https://www.springernature.com/gp/snapp, Dec 2023.

\bibitem[Naumov et~al.(2025)Naumov, Zagirova, Lin, Xie, Gou, Urban, Tikhonova, Alawi, Durymanov, Galkin, et~al.]{naumov2025dora}
Vladimir Naumov, Diana Zagirova, Sha Lin, Yupeng Xie, Wenhao Gou, Anatoly Urban, Nina Tikhonova, Khadija Alawi, Mike Durymanov, Fedor Galkin, et~al.
\newblock Dora ai scientist: Multi-agent virtual research team for scientific exploration discovery and automated report generation.
\newblock \emph{bioRxiv}, Mar 2025.

\bibitem[Newman et~al.(2024)Newman, Lee, Naik, Siangliulue, Fok, Kim, Weld, Chang, and Lo]{newman-etal-2024-arxivdigestables}
Benjamin Newman, Yoonjoo Lee, Aakanksha Naik, Pao Siangliulue, Raymond Fok, Juho Kim, Daniel~S Weld, Joseph~Chee Chang, and Kyle Lo.
\newblock {A}rxiv{DIGEST}ables: Synthesizing scientific literature into tables using language models.
\newblock In Yaser Al-Onaizan, Mohit Bansal, and Yun-Nung Chen, editors, \emph{Proceedings of the 2024 Conference on Empirical Methods in Natural Language Processing}, pages 9612--9631, Miami, Florida, USA, November 2024. Association for Computational Linguistics.
\newblock \doi{10.18653/v1/2024.emnlp-main.538}.
\newblock URL \url{https://aclanthology.org/2024.emnlp-main.538/}.

\bibitem[Newsham et~al.(2025)Newsham, Kova{\v{c}}evi{\'c}, Moulange, Ke, and Mukherjee]{newsham2025large}
Izzy Newsham, Luka Kova{\v{c}}evi{\'c}, Richard Moulange, Nan~Rosemary Ke, and Sach Mukherjee.
\newblock Large language models for zero-shot inference of causal structures in biology.
\newblock \emph{arXiv preprint arXiv:2503.04347}, 2025.

\bibitem[Nguyen et~al.(2024)Nguyen, Hong, Dang, and Huang]{nguyen2024human}
Andy Nguyen, Yvonne Hong, Belle Dang, and Xiaoshan Huang.
\newblock Human-ai collaboration patterns in ai-assisted academic writing.
\newblock \emph{Studies in Higher Education}, 49\penalty0 (5):\penalty0 847--864, Oct 2024.

\bibitem[Ni and Buehler(2024)]{ni2024mechagents}
Bo~Ni and Markus~J Buehler.
\newblock Mechagents: Large language model multi-agent collaborations can solve mechanics problems, generate new data, and integrate knowledge.
\newblock \emph{Extreme Mechanics Letters}, 67:\penalty0 102131, Mar 2024.

\bibitem[Ni et~al.(2024)Ni, Li, Hu, Han, Xu, Chen, Liu, Ye, and Bai]{ni2024matpilot}
Ziqi Ni, Yahao Li, Kaijia Hu, Kunyuan Han, Ming Xu, Xingyu Chen, Fengqi Liu, Yicong Ye, and Shuxin Bai.
\newblock Matpilot: an llm-enabled ai materials scientist under the framework of human-machine collaboration.
\newblock \emph{arXiv preprint arXiv:2411.08063}, 2024.

\bibitem[Nigam et~al.(2024)Nigam, Patwardhan, Vig, and Shroff]{nigam-etal-2024-interactive}
Harshit Nigam, Manasi Patwardhan, Lovekesh Vig, and Gautam Shroff.
\newblock An interactive co-pilot for accelerated research ideation.
\newblock In Su~Lin Blodgett, Amanda Cercas~Curry, Sunipa Dev, Michael Madaio, Ani Nenkova, Diyi Yang, and Ziang Xiao, editors, \emph{Proceedings of the Third Workshop on Bridging Human--Computer Interaction and Natural Language Processing}, pages 60--73, Mexico City, Mexico, June 2024. Association for Computational Linguistics.
\newblock \doi{10.18653/v1/2024.hcinlp-1.6}.
\newblock URL \url{https://aclanthology.org/2024.hcinlp-1.6/}.

\bibitem[Nigam et~al.(2024{\natexlab{2}})Nigam, Patwardhan, Vig, and Shroff]{nigam2024acceleron}
Harshit Nigam, Manasi Patwardhan, Lovekesh Vig, and Gautam Shroff.
\newblock Acceleron: A tool to accelerate research ideation.
\newblock \emph{arXiv preprint arXiv:2403.04382}, 2024{\natexlab{2}}.

\bibitem[Nijkamp et~al.(2022)Nijkamp, Pang, Hayashi, Tu, Wang, Zhou, Savarese, and Xiong]{nijkamp2022codegen}
Erik Nijkamp, Bo~Pang, Hiroaki Hayashi, Lifu Tu, Huan Wang, Yingbo Zhou, Silvio Savarese, and Caiming Xiong.
\newblock Codegen: An open large language model for code with multi-turn program synthesis.
\newblock \emph{arXiv preprint arXiv:2203.13474}, 2022.

\bibitem[Ning et~al.(2025)Ning, Yang, Liu, Yao, Liu, Tian, Song, and Yuan]{ning2025pico}
Kun-Peng Ning, Shuo Yang, Yuyang Liu, Jia-Yu Yao, Zhenhui Liu, Yonghong Tian, Yibing Song, and Li~Yuan.
\newblock Pi{CO}: Peer review in {LLM}s based on consistency optimization.
\newblock In \emph{The Thirteenth International Conference on Learning Representations}, Jan 2025.
\newblock URL \url{https://openreview.net/forum?id=sfQ6XpApfS}.

\bibitem[Nishimura et~al.(2024)Nishimura, Saito, Hirasawa, and Ushiku]{nishimura2024toward}
Kazuya Nishimura, Kuniaki Saito, Tosho Hirasawa, and Yoshitaka Ushiku.
\newblock Toward related work generation with structure and novelty statement.
\newblock In \emph{Proceedings of the Fourth Workshop on Scholarly Document Processing (SDP 2024)}, pages 38--57, Aug 2024.

\bibitem[Niu et~al.(2024)Niu, Hu, Zhou, and Zhan]{niu2024comprehensive}
Haoyi Niu, Jianming Hu, Guyue Zhou, and Xianyuan Zhan.
\newblock A comprehensive survey of cross-domain policy transfer for embodied agents.
\newblock \emph{arXiv preprint arXiv:2402.04580}, 2024.

\bibitem[Niu et~al.(2023)Niu, Xue, and P{\"o}pper]{niu2023unveiling}
Liang Niu, Nian Xue, and Christina P{\"o}pper.
\newblock Unveiling the sentinels: Assessing ai performance in cybersecurity peer review.
\newblock \emph{arXiv preprint arXiv:2309.05457}, 2023.

\bibitem[Novikov et~al.(2025)Novikov, Vu, Eisenberger, Dupont, Huang, Wagner, Shirobokov, Kozlovskii, Ruiz, Mehrabian, et~al.]{novikov2025alphaevolve}
Alexander Novikov, Ng{\^a}n Vu, Marvin Eisenberger, Emilien Dupont, Po-Sen Huang, Adam~Zsolt Wagner, Sergey Shirobokov, Borislav Kozlovskii, Francisco~JR Ruiz, Abbas Mehrabian, et~al.
\newblock Alphaevolve: A coding agent for scientific and algorithmic discovery.
\newblock \emph{Google DeepMind}, Jun 2025.

\bibitem[O'Neill et~al.(2025)O'Neill, Ghosal, R{\u{a}}ileanu, Walmsley, Bui, Schawinski, and Ciuc{\u{a}}]{o2025sparks}
Charles O'Neill, Tirthankar Ghosal, Roberta R{\u{a}}ileanu, Mike Walmsley, Thang Bui, Kevin Schawinski, and Ioana Ciuc{\u{a}}.
\newblock Sparks of science: Hypothesis generation using structured paper data.
\newblock \emph{arXiv preprint arXiv:2504.12976}, 2025.

\bibitem[{OpenAI}(2025)]{openai2025deepresearch}
{OpenAI}.
\newblock Introducing deep research, Feb 2025.
\newblock URL \url{https://openai.com/index/introducing-deep-research/}.

\bibitem[{OpenAI}(2025{\natexlab{2}})]{openai2025gpt4osearch}
{OpenAI}.
\newblock Gpt-4o search preview, May 2025{\natexlab{2}}.
\newblock URL \url{https://platform.openai.com/docs/models/gpt-4o-search-preview}.

\bibitem[Ortega and G{\'o}mez-P{\'e}rez(2025)]{ortega2025sciclaims}
Ra{\'u}l Ortega and Jos{\'e}~Manuel G{\'o}mez-P{\'e}rez.
\newblock Sciclaims: An end-to-end generative system for biomedical claim analysis.
\newblock \emph{arXiv preprint arXiv:2503.18526}, 2025.

\bibitem[Othman et~al.(2025)Othman, Ahmed, Okesanya, Ibrahim, Musa, Hassan, Saeed, and Lucero-Prisno~III]{othman2025advancing}
Zhinya~Kawa Othman, Mohamed~Mustaf Ahmed, Olalekan~John Okesanya, Adamu~Muhammad Ibrahim, Shuaibu~Saidu Musa, Bryar~A Hassan, Lanja~Ibrahim Saeed, and Don~Eliseo Lucero-Prisno~III.
\newblock Advancing drug discovery and development through gpt models: a review on challenges, innovations and future prospects.
\newblock \emph{Intelligence-Based Medicine}, page 100233, Mar 2025.

\bibitem[Ou et~al.(2025)Ou, Luo, Zheng, Wei, Qiao, Zhang, Zheng, Chen, and Zhang]{ou2025automind}
Yixin Ou, Yujie Luo, Jingsheng Zheng, Lanning Wei, Shuofei Qiao, Jintian Zhang, Da~Zheng, Huajun Chen, and Ningyu Zhang.
\newblock Automind: Adaptive knowledgeable agent for automated data science.
\newblock \emph{arXiv preprint arXiv:2506.10974}, 2025.

\bibitem[Ovelman et~al.(2024)Ovelman, Kugley, Gartlehner, and Viswanathan]{ovelman2024use}
Colleen Ovelman, Shannon Kugley, Gerald Gartlehner, and Meera Viswanathan.
\newblock The use of a large language model to create plain language summaries of evidence reviews in healthcare: A feasibility study.
\newblock \emph{Cochrane Evidence Synthesis and Methods}, 2\penalty0 (2):\penalty0 e12041, Feb 2024.

\bibitem[Ozkaya(2023)]{ozkaya2023application}
Ipek Ozkaya.
\newblock Application of large language models to software engineering tasks: Opportunities, risks, and implications.
\newblock \emph{IEEE Software}, 40\penalty0 (3):\penalty0 4--8, Apr 2023.

\bibitem[Padmakumar and He(2021)]{padmakumar2021machine}
Vishakh Padmakumar and He~He.
\newblock Machine-in-the-loop rewriting for creative image captioning.
\newblock \emph{arXiv preprint arXiv:2111.04193}, 2021.

\bibitem[Padmakumar and He(2023)]{padmakumar2023does}
Vishakh Padmakumar and He~He.
\newblock Does writing with language models reduce content diversity?
\newblock \emph{arXiv preprint arXiv:2309.05196}, 2023.

\bibitem[Pal et~al.(2022)Pal, Umapathi, and Sankarasubbu]{pal2022medmcqa}
Ankit Pal, Logesh~Kumar Umapathi, and Malaikannan Sankarasubbu.
\newblock Medmcqa: A large-scale multi-subject multi-choice dataset for medical domain question answering.
\newblock In \emph{Conference on health, inference, and learning}, pages 248--260. PMLR, Apr 2022.

\bibitem[Pan et~al.(2023)Pan, Wu, Lu, Luu, Wang, Kan, and Nakov]{pan2023fact}
Liangming Pan, Xiaobao Wu, Xinyuan Lu, Anh~Tuan Luu, William~Yang Wang, Min-Yen Kan, and Preslav Nakov.
\newblock Fact-checking complex claims with program-guided reasoning.
\newblock \emph{arXiv preprint arXiv:2305.12744}, 2023.

\bibitem[Pan et~al.(2023{\natexlab{2}})Pan, Zhang, and Kan]{pan2023investigating}
Liangming Pan, Yunxiang Zhang, and Min-Yen Kan.
\newblock Investigating zero-and few-shot generalization in fact verification.
\newblock \emph{arXiv preprint arXiv:2309.09444}, 2023{\natexlab{2}}.

\bibitem[Pan et~al.(2025)Pan, Liu, Chen, Zhou, Yu, and Jia]{pan2025hidden}
Wenbo Pan, Zhichao Liu, Qiguang Chen, Xiangyang Zhou, Haining Yu, and Xiaohua Jia.
\newblock The hidden dimensions of llm alignment: A multi-dimensional safety analysis.
\newblock \emph{arXiv preprint arXiv:2502.09674}, 2025.

\bibitem[Pang et~al.(2024)Pang, Fan, Wang, Xiao, Tang, Yang, Jia, Huang, and Yu]{pang2024empowering}
Jing-Cheng Pang, Heng-Bo Fan, Pengyuan Wang, Jia-Hao Xiao, Nan Tang, Si-Hang Yang, Chengxing Jia, Sheng-Jun Huang, and Yang Yu.
\newblock Empowering language models with active inquiry for deeper understanding.
\newblock \emph{arXiv preprint arXiv:2402.03719}, 2024.

\bibitem[Pang et~al.(2025)Pang, Lin, Jian, He, and Torr]{pang2025paper2poster}
Wei Pang, Kevin~Qinghong Lin, Xiangru Jian, Xi~He, and Philip Torr.
\newblock Paper2poster: Towards multimodal poster automation from scientific papers.
\newblock \emph{arXiv preprint arXiv:2505.21497}, 2025.

\bibitem[Park et~al.(2025)Park, Taneja, Wang, and Kang]{park2025stealing}
Jong~Inn Park, Maanas Taneja, Qianwen Wang, and Dongyeop Kang.
\newblock Stealing creator's workflow: A creator-inspired agentic framework with iterative feedback loop for improved scientific short-form generation.
\newblock \emph{arXiv preprint arXiv:2504.18805}, 2025.

\bibitem[Pei et~al.(2024)Pei, Wu, Gao, Zhu, Wang, Wang, Qin, and Yan]{pei2024leveraging}
Qizhi Pei, Lijun Wu, Kaiyuan Gao, Jinhua Zhu, Yue Wang, Zun Wang, Tao Qin, and Rui Yan.
\newblock Leveraging biomolecule and natural language through multi-modal learning: A survey.
\newblock \emph{arXiv preprint arXiv:2403.01528}, 2024.

\bibitem[Pendyala et~al.(2025)Pendyala, Kamdar, and Mulchandani]{pendyala2025automated}
Vishnu~S Pendyala, Karnavee Kamdar, and Kapil Mulchandani.
\newblock Automated research review support using machine learning, large language models, and natural language processing.
\newblock \emph{Electronics}, 14\penalty0 (2):\penalty0 256, Jan 2025.

\bibitem[Peng et~al.(2025)Peng, Zhou, Chen, Liu, Chen, and Qin]{peng2025dlpo}
Dengyun Peng, Yuhang Zhou, Qiguang Chen, Jinhao Liu, Jingjing Chen, and Libo Qin.
\newblock Dlpo: Towards a robust, efficient, and generalizable prompt optimization framework from a deep-learning perspective.
\newblock \emph{arXiv preprint arXiv:2503.13413}, 2025.

\bibitem[Peretz et~al.(2023)Peretz, Arraf, and Radinsky]{peretz2023if}
Gal Peretz, Mousa Arraf, and Kira Radinsky.
\newblock What if: Generating code to answer simulation questions in chemistry texts.
\newblock In \emph{Proceedings of the 46th International ACM SIGIR Conference on Research and Development in Information Retrieval}, pages 1335--1344, Jul 2023.

\bibitem[Perez et~al.(2024)Perez, L{\'e}ger, Ovando-Tellez, Foulon, Dussauld, Oudeyer, and Moulin-Frier]{perez2024cultural}
J{\'e}r{\'e}my Perez, Corentin L{\'e}ger, Marcela Ovando-Tellez, Chris Foulon, Joan Dussauld, Pierre-Yves Oudeyer, and Cl{\'e}ment Moulin-Frier.
\newblock Cultural evolution in populations of large language models.
\newblock \emph{arXiv preprint arXiv:2403.08882}, 2024.

\bibitem[Phan et~al.(2024)Phan, Nguyen, Nguyen, and Bui]{phan2024hyperagent}
Huy~Nhat Phan, Tien~N Nguyen, Phong~X Nguyen, and Nghi~DQ Bui.
\newblock Hyperagent: Generalist software engineering agents to solve coding tasks at scale.
\newblock \emph{arXiv preprint arXiv:2409.16299}, 2024.

\bibitem[Pinedo et~al.(2024)Pinedo, Larra{\~n}aga, and Arruarte]{pinedo2024arzigo}
Iratxe Pinedo, Mikel Larra{\~n}aga, and Ana Arruarte.
\newblock Arzigo: A recommendation system for scientific articles.
\newblock \emph{Information Systems}, 122:\penalty0 102367, May 2024.

\bibitem[Plank and van Dalen(2019)]{plank2019citetracked}
Barbara Plank and Reinard van Dalen.
\newblock Citetracked: A longitudinal dataset of peer reviews and citations.
\newblock In \emph{Proceedings of the 4th Joint Workshop on Bibliometric-enhanced Information Retrieval and Natural Language Processing for Digital Libraries (BIRNDL 2019) co-located with the 42nd International ACM SIGIR Conference on Research and Development in Information Retrieval (SIGIR 2019)}, pages 116--122. CEUR Workshop Proceedings (CEUR-WS. org), Jul 2019.

\bibitem[Polu and Sutskever(2020)]{polu2020generative}
Stanislas Polu and Ilya Sutskever.
\newblock Generative language modeling for automated theorem proving.
\newblock \emph{arXiv preprint arXiv:2009.03393}, 2020.

\bibitem[Pradhan et~al.(2020)Pradhan, Chakraborty, Choudhary, and Nandi]{pradhan2020automated}
Dinesh~K Pradhan, Joyita Chakraborty, Prasenjit Choudhary, and Subrata Nandi.
\newblock An automated conflict of interest based greedy approach for conference paper assignment system.
\newblock \emph{Journal of Informetrics}, 14\penalty0 (2):\penalty0 101022, May 2020.

\bibitem[Pradier et~al.(2025)Pradier, C{\'e}spedes, and Larivi{\`e}re]{pradier2025smack}
Carolina Pradier, Luc{\'\i}a C{\'e}spedes, and Vincent Larivi{\`e}re.
\newblock A smack of all neighbouring languages: How multilingual is scholarly communication?
\newblock \emph{arXiv preprint arXiv:2504.21100}, 2025.

\bibitem[Pramanick et~al.(2024)Pramanick, Chellappa, and Venugopalan]{pramanickspiqa}
Shraman Pramanick, Rama Chellappa, and Subhashini Venugopalan.
\newblock Spiqa: A dataset for multimodal question answering on scientific papers.
\newblock In \emph{The Thirty-eight Conference on Neural Information Processing Systems Datasets and Benchmarks Track}, Jul 2024.
\newblock URL \url{https://openreview.net/forum?id=h3lddsY5nf}.

\bibitem[Pratapa and Mitamura(2025)]{pratapa2025estimating}
Adithya Pratapa and Teruko Mitamura.
\newblock Estimating optimal context length for hybrid retrieval-augmented multi-document summarization.
\newblock \emph{arXiv preprint arXiv:2504.12972}, 2025.

\bibitem[Praveen et~al.(2024)Praveen, Gajjar, Ray, and Dutt]{PRAVEEN2024103975}
S.V. Praveen, Pranshav Gajjar, Rajeev~Kumar Ray, and Ashutosh Dutt.
\newblock Crafting clarity: Leveraging large language models to decode consumer reviews.
\newblock \emph{Journal of Retailing and Consumer Services}, 81:\penalty0 103975, Nov 2024.
\newblock ISSN 0969-6989.
\newblock \doi{https://doi.org/10.1016/j.jretconser.2024.103975}.
\newblock URL \url{https://www.sciencedirect.com/science/article/pii/S0969698924002716}.

\bibitem[Pu and Demberg(2024)]{pu2024rst}
Dongqi Pu and Vera Demberg.
\newblock Rst-lora: A discourse-aware low-rank adaptation for long document abstractive summarization.
\newblock In \emph{Proceedings of the 2024 Conference of the North American Chapter of the Association for Computational Linguistics: Human Language Technologies (Volume 1: Long Papers)}, pages 2200--2220, May 2024.

\bibitem[Pu et~al.(2025)Pu, Feng, Grossman, Hope, Dalvi~Mishra, Latzke, Bragg, Chang, and Siangliulue]{pu2025ideasynth}
Kevin Pu, KJ~Kevin Feng, Tovi Grossman, Tom Hope, Bhavana Dalvi~Mishra, Matt Latzke, Jonathan Bragg, Joseph~Chee Chang, and Pao Siangliulue.
\newblock Ideasynth: Iterative research idea development through evolving and composing idea facets with literature-grounded feedback.
\newblock In \emph{Proceedings of the 2025 CHI Conference on Human Factors in Computing Systems}, pages 1--31, Apr 2025.

\bibitem[Pu et~al.(2025{\natexlab{2}})Pu, Lin, and Chen]{pu2025piflow}
Yingming Pu, Tao Lin, and Hongyu Chen.
\newblock Piflow: Principle-aware scientific discovery with multi-agent collaboration.
\newblock \emph{arXiv preprint arXiv:2505.15047}, 2025{\natexlab{2}}.

\bibitem[Purkayastha et~al.(2025)Purkayastha, Li, Lauscher, Qu, and Gurevych]{purkayastha2025lazyreview}
Sukannya Purkayastha, Zhuang Li, Anne Lauscher, Lizhen Qu, and Iryna Gurevych.
\newblock Lazyreview a dataset for uncovering lazy thinking in nlp peer reviews.
\newblock \emph{arXiv preprint arXiv:2504.11042}, 2025.

\bibitem[Putrama and Martinek(2024)]{putrama2024heterogeneous}
I~Made Putrama and P{\'e}ter Martinek.
\newblock Heterogeneous data integration: Challenges and opportunities.
\newblock \emph{Data in Brief}, page 110853, Oct 2024.

\bibitem[Pyzer-Knapp et~al.(2022)Pyzer-Knapp, Pitera, Staar, Takeda, Laino, Sanders, Sexton, Smith, and Curioni]{pyzer2022accelerating}
Edward~O Pyzer-Knapp, Jed~W Pitera, Peter~WJ Staar, Seiji Takeda, Teodoro Laino, Daniel~P Sanders, James Sexton, John~R Smith, and Alessandro Curioni.
\newblock Accelerating materials discovery using artificial intelligence, high performance computing and robotics.
\newblock \emph{npj Computational Materials}, 8\penalty0 (1):\penalty0 84, Apr 2022.

\bibitem[Qian et~al.(2023)Qian, Dang, Li, Liu, Xie, Wang, Chen, Yang, Cong, Che, et~al.]{qian2023experiential}
Chen Qian, Yufan Dang, Jiahao Li, Wei Liu, Zihao Xie, Yifei Wang, Weize Chen, Cheng Yang, Xin Cong, Xiaoyin Che, et~al.
\newblock Experiential co-learning of software-developing agents.
\newblock \emph{arXiv preprint arXiv:2312.17025}, 2023.

\bibitem[Qian et~al.(2023{\natexlab{2}})Qian, Liu, Liu, Chen, Dang, Li, Yang, Chen, Su, Cong, et~al.]{qian2023chatdev}
Chen Qian, Wei Liu, Hongzhang Liu, Nuo Chen, Yufan Dang, Jiahao Li, Cheng Yang, Weize Chen, Yusheng Su, Xin Cong, et~al.
\newblock Chatdev: Communicative agents for software development.
\newblock \emph{arXiv preprint arXiv:2307.07924}, 2023{\natexlab{2}}.

\bibitem[Qin et~al.(2020)Qin, Xu, Che, Zhang, and Liu]{qin2020dynamic}
Libo Qin, Xiao Xu, Wanxiang Che, Yue Zhang, and Ting Liu.
\newblock Dynamic fusion network for multi-domain end-to-end task-oriented dialog.
\newblock pages 6344--6354, July 2020.
\newblock \doi{10.18653/v1/2020.acl-main.565}.
\newblock URL \url{https://aclanthology.org/2020.acl-main.565/}.

\bibitem[Qin et~al.(2023)Qin, Huang, Chen, Cai, Zhang, Liang, Che, and Xu]{qin2023mmsd2}
Libo Qin, Shijue Huang, Qiguang Chen, Chenran Cai, Yudi Zhang, Bin Liang, Wanxiang Che, and Ruifeng Xu.
\newblock Mmsd2. 0: Towards a reliable multi-modal sarcasm detection system.
\newblock \emph{arXiv preprint arXiv:2307.07135}, 2023.

\bibitem[Qin et~al.(2024)Qin, Chen, Fei, Chen, Li, and Che]{qin2024factors}
Libo Qin, Qiguang Chen, Hao Fei, Zhi Chen, Min Li, and Wanxiang Che.
\newblock What factors affect multi-modal in-context learning? an in-depth exploration.
\newblock \emph{arXiv preprint arXiv:2410.20482}, 2024.

\bibitem[Qin et~al.(2024{\natexlab{2}})Qin, Chen, Feng, Wu, Zhang, Li, Li, Che, and Yu]{qin2024large}
Libo Qin, Qiguang Chen, Xiachong Feng, Yang Wu, Yongheng Zhang, Yinghui Li, Min Li, Wanxiang Che, and Philip~S Yu.
\newblock Large language models meet nlp: A survey.
\newblock \emph{arXiv preprint arXiv:2405.12819}, 2024{\natexlab{2}}.

\bibitem[Qin et~al.(2025)Qin, Chen, Zhou, Chen, Li, Liao, Li, Che, and Yu]{qin2025survey}
Libo Qin, Qiguang Chen, Yuhang Zhou, Zhi Chen, Yinghui Li, Lizi Liao, Min Li, Wanxiang Che, and Philip~S Yu.
\newblock A survey of multilingual large language models.
\newblock \emph{Patterns}, 6\penalty0 (1), Jan 2025.
\newblock URL \url{https://www.cell.com/patterns/fulltext/S2666-3899(24)00290-3}.

\bibitem[Qin et~al.(2023{\natexlab{2}})Qin, Liang, Ye, Zhu, Yan, Lu, Lin, Cong, Tang, Qian, et~al.]{qin2023toolllm}
Yujia Qin, Shihao Liang, Yining Ye, Kunlun Zhu, Lan Yan, Yaxi Lu, Yankai Lin, Xin Cong, Xiangru Tang, Bill Qian, et~al.
\newblock Toolllm: Facilitating large language models to master 16000+ real-world apis.
\newblock \emph{arXiv preprint arXiv:2307.16789}, 2023{\natexlab{2}}.

\bibitem[Qiu and Lan(2024)]{qiu2024interactive}
Huachuan Qiu and Zhenzhong Lan.
\newblock Interactive agents: Simulating counselor-client psychological counseling via role-playing llm-to-llm interactions.
\newblock \emph{arXiv preprint arXiv:2408.15787}, 2024.

\bibitem[Qiu et~al.(2025)Qiu, Zhang, Xu, Li, Song, Wang, and Zhang]{qiu2025ai}
Yansheng Qiu, Haoquan Zhang, Zhaopan Xu, Ming Li, Diping Song, Zheng Wang, and Kaipeng Zhang.
\newblock Ai idea bench 2025: Ai research idea generation benchmark.
\newblock \emph{arXiv preprint arXiv:2504.14191}, 2025.

\bibitem[Qu et~al.(2020)Qu, Yang, Chen, Qiu, Croft, and Iyyer]{qu2020open}
Chen Qu, Liu Yang, Cen Chen, Minghui Qiu, W~Bruce Croft, and Mohit Iyyer.
\newblock Open-retrieval conversational question answering.
\newblock In \emph{Proceedings of the 43rd International ACM SIGIR conference on research and development in Information Retrieval}, pages 539--548, Jul 2020.

\bibitem[Qu et~al.(2024)Qu, Zhang, Garg, and Kumar]{qu2024recursive}
Yuxiao Qu, Tianjun Zhang, Naman Garg, and Aviral Kumar.
\newblock Recursive introspection: Teaching language model agents how to self-improve.
\newblock \emph{Advances in Neural Information Processing Systems}, 37:\penalty0 55249--55285, Dec 2024.

\bibitem[Radensky et~al.(2024)Radensky, Shahid, Fok, Siangliulue, Hope, and Weld]{radensky2024scideator}
Marissa Radensky, Simra Shahid, Raymond Fok, Pao Siangliulue, Tom Hope, and Daniel~S Weld.
\newblock Scideator: Human-llm scientific idea generation grounded in research-paper facet recombination.
\newblock \emph{arXiv preprint arXiv:2409.14634}, 2024.

\bibitem[Radosavovic et~al.(2024)Radosavovic, Xiao, Zhang, Darrell, Malik, and Sreenath]{radosavovic2024real}
Ilija Radosavovic, Tete Xiao, Bike Zhang, Trevor Darrell, Jitendra Malik, and Koushil Sreenath.
\newblock Real-world humanoid locomotion with reinforcement learning.
\newblock \emph{Science Robotics}, 9\penalty0 (89):\penalty0 eadi9579, 2024.

\bibitem[Raeini(2025)]{raeini2025rise}
Mohammad Raeini.
\newblock On the rise of new mathematical spaces and towards ai-driven scientific discovery.
\newblock \emph{Available at SSRN}, Mar 2025.

\bibitem[Raghunathan et~al.(2024)Raghunathan, Shah, et~al.]{raghunathan2024vulnerability}
Aditi Raghunathan, Nihar~B Shah, et~al.
\newblock Vulnerability of text-matching in ml/ai conference reviewer assignments to collusions.
\newblock \emph{arXiv preprint arXiv:2412.06606}, 2024.

\bibitem[Rai et~al.(2024)Rai, Zhou, Feng, Saparov, and Yao]{rai2024practical}
Daking Rai, Yilun Zhou, Shi Feng, Abulhair Saparov, and Ziyu Yao.
\newblock A practical review of mechanistic interpretability for transformer-based language models.
\newblock \emph{arXiv preprint arXiv:2407.02646}, 2024.

\bibitem[Raissi et~al.(2019)Raissi, Perdikaris, and Karniadakis]{raissi2019physics}
Maziar Raissi, Paris Perdikaris, and George~E Karniadakis.
\newblock Physics-informed neural networks: A deep learning framework for solving forward and inverse problems involving nonlinear partial differential equations.
\newblock \emph{Journal of Computational physics}, 378:\penalty0 686--707, Feb 2019.

\bibitem[Raissi et~al.(2024)Raissi, Perdikaris, Ahmadi, and Karniadakis]{raissi2024physics}
Maziar Raissi, Paris Perdikaris, Nazanin Ahmadi, and George~Em Karniadakis.
\newblock Physics-informed neural networks and extensions.
\newblock \emph{arXiv preprint arXiv:2408.16806}, 2024.

\bibitem[Rajabi-Kochi et~al.(2025)Rajabi-Kochi, Mahboubi, Gill, and Moosavi]{rajabi2025adaptive}
Mahyar Rajabi-Kochi, Negareh Mahboubi, Aseem Partap~Singh Gill, and Seyed~Mohamad Moosavi.
\newblock Adaptive representation of molecules and materials in bayesian optimization.
\newblock \emph{Chemical Science}, 16\penalty0 (13):\penalty0 5464--5474, Feb 2025.

\bibitem[Ranga et~al.(2025)Ranga, Mao, Cambria, and Chattopadhyay]{ranga2025plagiarism}
Sriram Ranga, Rui Mao, Erik Cambria, and Anupam Chattopadhyay.
\newblock The plagiarism singularity conjecture.
\newblock In \emph{Proceedings of the 2025 Conference of the Nations of the Americas Chapter of the Association for Computational Linguistics: Human Language Technologies (Volume 1: Long Papers)}, pages 10245--10255, Apr 2025.

\bibitem[Rao et~al.(2024)Rao, Young, Dietterich, and Callison-Burch]{rao2024withdrarxiv}
Delip Rao, Jonathan Young, Thomas Dietterich, and Chris Callison-Burch.
\newblock Withdrarxiv: A large-scale dataset for retraction study.
\newblock \emph{arXiv preprint arXiv:2412.03775}, 2024.

\bibitem[Rao et~al.(2025)Rao, You, Wong, and Callison-Burch]{rao2025nsf}
Delip Rao, Weiqiu You, Eric Wong, and Chris Callison-Burch.
\newblock Nsf-scify: Mining the nsf awards database for scientific claims.
\newblock \emph{arXiv preprint arXiv:2503.08600}, 2025.

\bibitem[Rao et~al.(2025{\natexlab{2}})Rao, Kumar, Lakkaraju, and Shah]{rao2025detecting}
Vishisht Rao, Aounon Kumar, Himabindu Lakkaraju, and Nihar~B Shah.
\newblock Detecting llm-written peer reviews.
\newblock \emph{arXiv preprint arXiv:2503.15772}, 2025{\natexlab{2}}.

\bibitem[Rashid et~al.(2022)Rashid, Jahan, Huzzat, Rahul, Zakir, Meem, Mukta, and Shatabda]{rashid2022text2chart}
Md~Mahinur Rashid, Hasin~Kawsar Jahan, Annysha Huzzat, Riyasaat~Ahmed Rahul, Tamim~Bin Zakir, Farhana Meem, Md~Saddam~Hossain Mukta, and Swakkhar Shatabda.
\newblock Text2chart: A multi-staged chart generator from natural language text.
\newblock In \emph{Pacific-Asia Conference on Knowledge Discovery and Data Mining}, pages 3--16. Springer, May 2022.

\bibitem[Raza et~al.(2022)Raza, Schwartz, and Rosella]{raza2022coquad}
Shaina Raza, Brian Schwartz, and Laura~C Rosella.
\newblock Coquad: a covid-19 question answering dataset system, facilitating research, benchmarking, and practice.
\newblock \emph{BMC bioinformatics}, 23\penalty0 (1):\penalty0 210, Jun 2022.

\bibitem[Reddy and Shojaee(2025)]{reddy2025towards}
Chandan~K Reddy and Parshin Shojaee.
\newblock Towards scientific discovery with generative ai: Progress, opportunities, and challenges.
\newblock In \emph{Proceedings of the AAAI Conference on Artificial Intelligence}, volume~39, pages 28601--28609, Jan 2025.

\bibitem[Rehman et~al.(2025)Rehman, Sanyal, and Chattopadhyay]{rehman2025can}
Tohida Rehman, Debarshi~Kumar Sanyal, and Samiran Chattopadhyay.
\newblock Can pre-trained language models generate titles for research papers?
\newblock In \emph{International Conference on Asian Digital Libraries}, pages 154--170. Springer, Dec 2025.

\bibitem[Ren et~al.(2025)Ren, Jian, Ren, Leng, Xie, and Zhang]{ren2025towards}
Shuo Ren, Pu~Jian, Zhenjiang Ren, Chunlin Leng, Can Xie, and Jiajun Zhang.
\newblock Towards scientific intelligence: A survey of llm-based scientific agents.
\newblock \emph{arXiv preprint arXiv:2503.24047}, 2025.

\bibitem[Roberts et~al.(2024)Roberts, Han, Houlsby, and Albanie]{roberts2024scifibench}
Jonathan Roberts, Kai Han, Neil Houlsby, and Samuel Albanie.
\newblock Scifibench: Benchmarking large multimodal models for scientific figure interpretation.
\newblock \emph{arXiv preprint arXiv:2405.08807}, 2024.

\bibitem[Robertson(2023)]{robertson2023gpt4}
Zachary Robertson.
\newblock Gpt4 is slightly helpful for peer-review assistance: A pilot study.
\newblock \emph{arXiv preprint arXiv:2307.05492}, 2023.

\bibitem[Robinson et~al.(2023)Robinson, Thorne, Wu, Pandor, Essat, Stevenson, and Song]{robinson2023bio}
Ambrose Robinson, William Thorne, Ben~P Wu, Abdullah Pandor, Munira Essat, Mark Stevenson, and Xingyi Song.
\newblock Bio-sieve: exploring instruction tuning large language models for systematic review automation.
\newblock \emph{arXiv preprint arXiv:2308.06610}, 2023.

\bibitem[Rodriguez et~al.(2023)Rodriguez, Vazquez, Laradji, Pedersoli, and Rodriguez]{rodriguez2023figgen}
Juan~A Rodriguez, David Vazquez, Issam Laradji, Marco Pedersoli, and Pau Rodriguez.
\newblock Figgen: Text to scientific figure generation.
\newblock \emph{arXiv preprint arXiv:2306.00800}, 2023.

\bibitem[Rodriguez et~al.(2025)Rodriguez, Jian, Panigrahi, Zhang, Feizi, Puri, Suresh, Savard, Masry, Nayak, et~al.]{rodriguez2025bigdocs}
Juan~A Rodriguez, Xiangru Jian, Siba~Smarak Panigrahi, Tianyu Zhang, Aarash Feizi, Abhay Puri, Akshay~Kalkunte Suresh, Fran{\c{c}}ois Savard, Ahmed Masry, Shravan Nayak, et~al.
\newblock Bigdocs: An open dataset for training multimodal models on document and code tasks.
\newblock In \emph{The Thirteenth International Conference on Learning Representations}, Jan 2025.

\bibitem[Rodriguez et~al.(2025{\natexlab{2}})Rodriguez, Puri, Agarwal, Laradji, Rodriguez, Rajeswar, Vazquez, Pal, and Pedersoli]{rodriguez2025starvector}
Juan~A Rodriguez, Abhay Puri, Shubham Agarwal, Issam~H Laradji, Pau Rodriguez, Sai Rajeswar, David Vazquez, Christopher Pal, and Marco Pedersoli.
\newblock Starvector: Generating scalable vector graphics code from images and text.
\newblock In \emph{Proceedings of the Computer Vision and Pattern Recognition Conference}, pages 16175--16186, Apr 2025{\natexlab{2}}.

\bibitem[Rohman(1965)]{rohman1965pre}
D~Gordon Rohman.
\newblock Pre-writing: The stage of discovery in the writing process.
\newblock \emph{College Composition \& Communication}, 16\penalty0 (2):\penalty0 106--112, May 1965.

\bibitem[Rood et~al.(2024)Rood, Hupalowska, and Regev]{rood2024toward}
Jennifer~E Rood, Anna Hupalowska, and Aviv Regev.
\newblock Toward a foundation model of causal cell and tissue biology with a perturbation cell and tissue atlas.
\newblock \emph{Cell}, 187\penalty0 (17):\penalty0 4520--4545, Aug 2024.

\bibitem[Roohani et~al.(2024)Roohani, Lee, Huang, Vora, Steinhart, Huang, Marson, Liang, and Leskovec]{roohani2024biodiscoveryagent}
Yusuf Roohani, Andrew Lee, Qian Huang, Jian Vora, Zachary Steinhart, Kexin Huang, Alexander Marson, Percy Liang, and Jure Leskovec.
\newblock Biodiscoveryagent: An ai agent for designing genetic perturbation experiments.
\newblock \emph{arXiv preprint arXiv:2405.17631}, 2024.

\bibitem[Rosbach et~al.(2025)Rosbach, Ganz, Ammeling, Riener, and Aubreville]{rosbach2025automation}
Emely Rosbach, Jonathan Ganz, Jonas Ammeling, Andreas Riener, and Marc Aubreville.
\newblock Automation bias in ai-assisted medical decision-making under time pressure in computational pathology.
\newblock In \emph{BVM Workshop}, pages 129--134. Springer, Mar 2025.

\bibitem[Rostam and Kert{\'e}sz(2024)]{rostam2024fine}
Zhyar Rzgar~K Rostam and G{\'a}bor Kert{\'e}sz.
\newblock Fine-tuning large language models for scientific text classification: A comparative study.
\newblock In \emph{2024 IEEE 6th International Symposium on Logistics and Industrial Informatics (LINDI)}, pages 000233--000238. IEEE, Oct 2024.

\bibitem[Rousmaniere et~al.(2025)Rousmaniere, Li, Zhang, and Shah]{rousmaniere2025large}
Tony Rousmaniere, Xu~Li, Yimeng Zhang, and Siddharth Shah.
\newblock Large language models as mental health resources: Patterns of use in the united states, Mar 2025.

\bibitem[Roy and Datta(2025)]{roy2025ai}
Pritam Roy and Dhananjoy Datta.
\newblock Ai-driven discovery: The transformative impact of machine learning on research and development.
\newblock In \emph{Evolving Landscapes of Research and Development: Trends, Challenges, and Opportunities}, pages 29--52. IGI Global Scientific Publishing, Jun 2025.

\bibitem[Roy and Han(2024)]{roy2024ilciter}
Sayar~Ghosh Roy and Jiawei Han.
\newblock Ilciter: Evidence-grounded interpretable local citation recommendation.
\newblock \emph{arXiv preprint arXiv:2403.08737}, 2024.

\bibitem[Roziere et~al.(2023)Roziere, Gehring, Gloeckle, Sootla, Gat, Tan, Adi, Liu, Sauvestre, Remez, et~al.]{roziere2023code}
Baptiste Roziere, Jonas Gehring, Fabian Gloeckle, Sten Sootla, Itai Gat, Xiaoqing~Ellen Tan, Yossi Adi, Jingyu Liu, Romain Sauvestre, Tal Remez, et~al.
\newblock Code llama: Open foundation models for code.
\newblock \emph{arXiv preprint arXiv:2308.12950}, 2023.

\bibitem[Ruan et~al.(2024)Ruan, Wang, Hong, and Sun]{ruan2024liveideabench}
Kai Ruan, Xuan Wang, Jixiang Hong, and Hao Sun.
\newblock Liveideabench: Evaluating llms' scientific creativity and idea generation with minimal context.
\newblock \emph{arXiv preprint arXiv:2412.17596}, 2024.

\bibitem[Ruan et~al.(2024{\natexlab{2}})Ruan, Lu, Xu, He, Chen, Zhang, Xuan, Pan, Fang, Gao, et~al.]{ruan2024automatic}
Yixiang Ruan, Chenyin Lu, Ning Xu, Yuchen He, Yixin Chen, Jian Zhang, Jun Xuan, Jianzhang Pan, Qun Fang, Hanyu Gao, et~al.
\newblock An automatic end-to-end chemical synthesis development platform powered by large language models.
\newblock \emph{Nature communications}, 15\penalty0 (1):\penalty0 10160, Nov 2024{\natexlab{2}}.

\bibitem[Sabet et~al.(2020)Sabet, Dufter, Yvon, and Sch{\"u}tze]{sabet2020simalign}
Masoud~Jalili Sabet, Philipp Dufter, Fran{\c{c}}ois Yvon, and Hinrich Sch{\"u}tze.
\newblock Simalign: High quality word alignments without parallel training data using static and contextualized embeddings.
\newblock \emph{arXiv preprint arXiv:2004.08728}, 2020.

\bibitem[S{\ae}tra(2025)]{saetra2025rise}
Henrik~Skaug S{\ae}tra.
\newblock The rise of the research automaton: Science as process or product in the era of generative ai?
\newblock \emph{Available at SSRN 5219722}, Apr 2025.

\bibitem[Saikh et~al.(2022)Saikh, Ghosal, Mittal, Ekbal, and Bhattacharyya]{saikh2022scienceqa}
Tanik Saikh, Tirthankar Ghosal, Amish Mittal, Asif Ekbal, and Pushpak Bhattacharyya.
\newblock Scienceqa: A novel resource for question answering on scholarly articles.
\newblock \emph{International Journal on Digital Libraries}, 23\penalty0 (3):\penalty0 289--301, Jul 2022.

\bibitem[Santu et~al.(2024)Santu, Sinha, Bansal, Knipper, Sarkar, Salvador, Mahajan, Guttikonda, Akter, Freestone, et~al.]{santu2024prompting}
Shubhra Kanti~Karmaker Santu, Sanjeev~Kumar Sinha, Naman Bansal, Alex Knipper, Souvika Sarkar, John Salvador, Yash Mahajan, Sri Guttikonda, Mousumi Akter, Matthew Freestone, et~al.
\newblock Prompting llms to compose meta-review drafts from peer-review narratives of scholarly manuscripts.
\newblock \emph{arXiv preprint arXiv:2402.15589}, 2024.

\bibitem[Sanyal et~al.(2025)Sanyal, Schapiro, Shashidhar, Moon, Varshney, and Hakkani-Tur]{sanyal2025spark}
Aishik Sanyal, Samuel Schapiro, Sumuk Shashidhar, Royce Moon, Lav~R Varshney, and Dilek Hakkani-Tur.
\newblock Spark: A system for scientifically creative idea generation.
\newblock \emph{arXiv preprint arXiv:2504.20090}, 2025.

\bibitem[Sayin et~al.(2025)Sayin, Schlicht, Hong, Allievi, Staiano, Minervini, and Passerini]{sayin2025medsyn}
Burcu Sayin, Ipek~Baris Schlicht, Ngoc~Vo Hong, Sara Allievi, Jacopo Staiano, Pasquale Minervini, and Andrea Passerini.
\newblock Medsyn: Enhancing diagnostics with human-ai collaboration.
\newblock \emph{arXiv preprint arXiv:2506.14774}, 2025.

\bibitem[Schick et~al.(2023)Schick, Dwivedi-Yu, Dess{\`\i}, Raileanu, Lomeli, Hambro, Zettlemoyer, Cancedda, and Scialom]{schick2023toolformer}
Timo Schick, Jane Dwivedi-Yu, Roberto Dess{\`\i}, Roberta Raileanu, Maria Lomeli, Eric Hambro, Luke Zettlemoyer, Nicola Cancedda, and Thomas Scialom.
\newblock Toolformer: Language models can teach themselves to use tools.
\newblock \emph{Advances in Neural Information Processing Systems}, 36:\penalty0 68539--68551, Dec 2023.

\bibitem[Schleiger et~al.(2024)Schleiger, Mason, Naughtin, Reeson, and Paris]{schleiger2024collaborative}
Emma Schleiger, Claire Mason, Claire Naughtin, Andrew Reeson, and Cecile Paris.
\newblock Collaborative intelligence: A scoping review of current applications.
\newblock \emph{Applied Artificial Intelligence}, 38\penalty0 (1):\penalty0 2327890, Mar 2024.

\bibitem[Schmidgall and Moor(2025)]{schmidgall2025agentrxiv}
Samuel Schmidgall and Michael Moor.
\newblock Agentrxiv: Towards collaborative autonomous research.
\newblock \emph{arXiv preprint arXiv:2503.18102}, 2025.

\bibitem[Schmidgall et~al.(2025)Schmidgall, Su, Wang, Sun, Wu, Yu, Liu, Liu, and Barsoum]{schmidgall2025agent}
Samuel Schmidgall, Yusheng Su, Ze~Wang, Ximeng Sun, Jialian Wu, Xiaodong Yu, Jiang Liu, Zicheng Liu, and Emad Barsoum.
\newblock Agent laboratory: Using llm agents as research assistants.
\newblock \emph{arXiv preprint arXiv:2501.04227}, 2025.

\bibitem[Seabra et~al.(2024)Seabra, Cavalcante, Nepomuceno, Lago, Ruberg, and Lifschitz]{seabra2024dynamic}
Antony Seabra, Claudio Cavalcante, Joao Nepomuceno, Lucas Lago, Nicolaas Ruberg, and Sergio Lifschitz.
\newblock Dynamic multi-agent orchestration and retrieval for multi-source question-answer systems using large language models.
\newblock \emph{arXiv preprint arXiv:2412.17964}, 2024.

\bibitem[Senior et~al.(2020)Senior, Evans, Jumper, Kirkpatrick, Sifre, Green, Qin, {\v{Z}}{\'\i}dek, Nelson, Bridgland, et~al.]{senior2020improved}
Andrew~W Senior, Richard Evans, John Jumper, James Kirkpatrick, Laurent Sifre, Tim Green, Chongli Qin, Augustin {\v{Z}}{\'\i}dek, Alexander~WR Nelson, Alex Bridgland, et~al.
\newblock Improved protein structure prediction using potentials from deep learning.
\newblock \emph{Nature}, 577\penalty0 (7792):\penalty0 706--710, Jan 2020.

\bibitem[Seo et~al.(2025)Seo, Kim, Jeong, Park, and Min]{seo-2025-flavordiffusion}
Junpyo Seo, Dongwan Kim, Jaewook Jeong, Inkyu Park, and Junho Min.
\newblock {F}lavor{D}iffusion: Modeling food-chemical interactions with diffusion.
\newblock In Peter Jansen, Bhavana Dalvi~Mishra, Harsh Trivedi, Bodhisattwa Prasad~Majumder, Tom Hope, Tushar Khot, Doug Downey, and Eric Horvitz, editors, \emph{Proceedings of the 1st Workshop on AI and Scientific Discovery: Directions and Opportunities}, pages 70--77, Albuquerque, New Mexico, USA, May 2025. Association for Computational Linguistics.
\newblock ISBN 979-8-89176-224-4.
\newblock \doi{10.18653/v1/2025.aisd-main.7}.
\newblock URL \url{https://aclanthology.org/2025.aisd-main.7/}.

\bibitem[Serrano et~al.(2024)Serrano, Martinez-Carranza, and Sucar]{serrano2024knowledge}
Sergio~A Serrano, Jose Martinez-Carranza, and L~Enrique Sucar.
\newblock Knowledge transfer for cross-domain reinforcement learning: a systematic review.
\newblock \emph{IEEE Access}, Jul 2024.

\bibitem[Shahid et~al.(2020)Shahid, Afzal, Abdar, Basiri, Zhou, Yen, and Chang]{shahid2020insights}
Abdul Shahid, Muhammad~Tanvir Afzal, Moloud Abdar, Mohammad~Ehsan Basiri, Xujuan Zhou, Neil~Y Yen, and Jia-Wei Chang.
\newblock Insights into relevant knowledge extraction techniques: a comprehensive review.
\newblock \emph{The Journal of Supercomputing}, 76:\penalty0 1695--1733, Oct 2020.

\bibitem[Shahin et~al.(2025)Shahin, Goswami, Lobentanzer, and Corrigan]{shahin2025agents}
Mohamed~H Shahin, Srijib Goswami, Sebastian Lobentanzer, and Brian~W Corrigan.
\newblock Agents for change: Artificial intelligent workflows for quantitative clinical pharmacology and translational sciences.
\newblock \emph{Clinical and Translational Science}, 18\penalty0 (3):\penalty0 e70188, Mar 2025.

\bibitem[Shao et~al.(2024)Shao, Jiang, Kanell, Xu, Khattab, and Lam]{shao2024assisting}
Yijia Shao, Yucheng Jiang, Theodore~A Kanell, Peter Xu, Omar Khattab, and Monica~S Lam.
\newblock Assisting in writing wikipedia-like articles from scratch with large language models.
\newblock \emph{arXiv preprint arXiv:2402.14207}, 2024.

\bibitem[Sheldon and Kumar(2024)]{sheldon2024economic}
Zachary Sheldon and Peeyush Kumar.
\newblock Economic anthropology in the era of generative artificial intelligence.
\newblock \emph{arXiv preprint arXiv:2410.15238}, 2024.

\bibitem[Shi et~al.(2024)Shi, Liu, Liu, Cheng, and Lu]{shi2024every}
Xiang Shi, Jiawei Liu, Yinpeng Liu, Qikai Cheng, and Wei Lu.
\newblock Every part matters: Integrity verification of scientific figures based on multimodal large language models.
\newblock \emph{arXiv preprint arXiv:2407.18626}, 2024.

\bibitem[Shi et~al.(2023)Shi, Gao, Zhang, Chen, Chen, Ren, and Ren]{shi2023towards}
Zhengliang Shi, Shen Gao, Zhen Zhang, Xiuying Chen, Zhumin Chen, Pengjie Ren, and Zhaochun Ren.
\newblock Towards a unified framework for reference retrieval and related work generation.
\newblock In \emph{Findings of the Association for Computational Linguistics: EMNLP 2023}, pages 5785--5799, Dec 2023.

\bibitem[Shi et~al.(2025)Shi, Yan, Yin, Verberne, de~Rijke, and Ren]{shi2025iterative}
Zhengliang Shi, Lingyong Yan, Dawei Yin, Suzan Verberne, Maarten de~Rijke, and Zhaochun Ren.
\newblock Iterative self-incentivization empowers large language models as agentic searchers.
\newblock \emph{arXiv preprint arXiv:2505.20128}, 2025.

\bibitem[Shin et~al.(2025)Shin, Tang, Lee, Kim, Lim, Cho, Hong, Lee, and Kim]{shin2025automatically}
Hyungyu Shin, Jingyu Tang, Yoonjoo Lee, Nayoung Kim, Hyunseung Lim, Ji~Yong Cho, Hwajung Hong, Moontae Lee, and Juho Kim.
\newblock Automatically evaluating the paper reviewing capability of large language models.
\newblock \emph{arXiv preprint arXiv:2502.17086}, 2025.

\bibitem[Shin et~al.(2025{\natexlab{2}})Shin, Tang, Lee, Kim, Lim, Cho, Hong, Lee, and Kim]{shin2025mind}
Hyungyu Shin, Jingyu Tang, Yoonjoo Lee, Nayoung Kim, Hyunseung Lim, Ji~Yong Cho, Hwajung Hong, Moontae Lee, and Juho Kim.
\newblock Mind the blind spots: A focus-level evaluation framework for llm reviews.
\newblock \emph{arXiv preprint arXiv:2502.17086}, 2025{\natexlab{2}}.

\bibitem[Shinn et~al.(2023)Shinn, Cassano, Gopinath, Narasimhan, and Yao]{shinn2023reflexion}
Noah Shinn, Federico Cassano, Ashwin Gopinath, Karthik Narasimhan, and Shunyu Yao.
\newblock Reflexion: Language agents with verbal reinforcement learning.
\newblock \emph{Advances in Neural Information Processing Systems}, 36:\penalty0 8634--8652, Dec 2023.

\bibitem[Shojaee et~al.(2025)Shojaee, Meidani, Gupta, Farimani, and Reddy]{shojaee2025llmsr}
Parshin Shojaee, Kazem Meidani, Shashank Gupta, Amir~Barati Farimani, and Chandan~K. Reddy.
\newblock {LLM}-{SR}: Scientific equation discovery via programming with large language models.
\newblock In \emph{The Thirteenth International Conference on Learning Representations}, Jan 2025.
\newblock URL \url{https://openreview.net/forum?id=m2nmp8P5in}.

\bibitem[Shojaee et~al.(2025{\natexlab{2}})Shojaee, Nguyen, Meidani, Farimani, Doan, and Reddy]{shojaee2025llm}
Parshin Shojaee, Ngoc-Hieu Nguyen, Kazem Meidani, Amir~Barati Farimani, Khoa~D Doan, and Chandan~K Reddy.
\newblock Llm-srbench: A new benchmark for scientific equation discovery with large language models.
\newblock \emph{arXiv preprint arXiv:2504.10415}, 2025{\natexlab{2}}.

\bibitem[Shojaee et~al.(2025{\natexlab{3}})Shojaee, Nguyen, Meidani, Farimani, Doan, and Reddy]{shojaee2025llmsrbench}
Parshin Shojaee, Ngoc-Hieu Nguyen, Kazem Meidani, Amir~Barati Farimani, Khoa~D Doan, and Chandan~K. Reddy.
\newblock {LLM}-{SRB}ench: A new benchmark for scientific equation discovery with large language models.
\newblock In \emph{Forty-second International Conference on Machine Learning}, May 2025{\natexlab{3}}.
\newblock URL \url{https://openreview.net/forum?id=SyQPiZJVWY}.

\bibitem[Si et~al.(2024)Si, Yang, and Hashimoto]{si2024can}
Chenglei Si, Diyi Yang, and Tatsunori Hashimoto.
\newblock Can llms generate novel research ideas? a large-scale human study with 100+ nlp researchers.
\newblock \emph{arXiv preprint arXiv:2409.04109}, 2024.

\bibitem[Siddiqui et~al.(2025)Siddiqui, Chen, Heo, Xia, and Weller]{siddiqui-etal-2025-evaluating}
Shoaib~Ahmed Siddiqui, Yanzhi Chen, Juyeon Heo, Menglin Xia, and Adrian Weller.
\newblock On evaluating {LLM}s' capabilities as functional approximators: A {B}ayesian evaluation framework.
\newblock In Owen Rambow, Leo Wanner, Marianna Apidianaki, Hend Al-Khalifa, Barbara~Di Eugenio, and Steven Schockaert, editors, \emph{Proceedings of the 31st International Conference on Computational Linguistics}, pages 5826--5835, Abu Dhabi, UAE, January 2025. Association for Computational Linguistics.
\newblock URL \url{https://aclanthology.org/2025.coling-main.388/}.

\bibitem[Singh et~al.(2025)Singh, Ehtesham, Kumar, and Khoei]{singh2025agentic}
Aditi Singh, Abul Ehtesham, Saket Kumar, and Tala~Talaei Khoei.
\newblock Agentic retrieval-augmented generation: A survey on agentic rag.
\newblock \emph{arXiv preprint arXiv:2501.09136}, 2025.

\bibitem[Singh et~al.(2025{\natexlab{2}})Singh, Chang, Anastasiades, Haddad, Naik, Tanaka, Zamarron, Nguyen, Hwang, Dunkleberger, et~al.]{singh2025ai2}
Amanpreet Singh, Joseph~Chee Chang, Chloe Anastasiades, Dany Haddad, Aakanksha Naik, Amber Tanaka, Angele Zamarron, Cecile Nguyen, Jena~D Hwang, Jason Dunkleberger, et~al.
\newblock Ai2 scholar qa: Organized literature synthesis with attribution.
\newblock \emph{arXiv preprint arXiv:2504.10861}, 2025{\natexlab{2}}.

\bibitem[Singh et~al.(2023)Singh, Agarwal, Huang, Singh, Yu, Kim, Bursztyn, Vlassis, and Rossi]{singh2023figcaps}
Ashish Singh, Prateek Agarwal, Zixuan Huang, Arpita Singh, Tong Yu, Sungchul Kim, Victor Bursztyn, Nikos Vlassis, and Ryan~A Rossi.
\newblock Figcaps-hf: A figure-to-caption generative framework and benchmark with human feedback.
\newblock \emph{arXiv preprint arXiv:2307.10867}, 2023.

\bibitem[Singh and Chakrabarti(2024)]{singh2024supporting}
Sanjay Singh and Amaresh Chakrabarti.
\newblock Supporting assessment of novelty of design problems using concept of problem sapphire.
\newblock \emph{arXiv preprint arXiv:2410.18629}, 2024.

\bibitem[Singh et~al.(2021)Singh, Singh, and Goyal]{singh2021compare}
Shruti Singh, Mayank Singh, and Pawan Goyal.
\newblock Compare: a taxonomy and dataset of comparison discussions in peer reviews.
\newblock In \emph{2021 ACM/IEEE Joint Conference on Digital Libraries (JCDL)}, pages 238--241. IEEE, Aug 2021.

\bibitem[Singh et~al.(2024)Singh, Sarkar, and Cohan]{singh2024scidqa}
Shruti Singh, Nandan Sarkar, and Arman Cohan.
\newblock Scidqa: A deep reading comprehension dataset over scientific papers.
\newblock \emph{arXiv preprint arXiv:2411.05338}, 2024.

\bibitem[Singhal et~al.(2023)Singhal, Azizi, Tu, Mahdavi, Wei, Chung, Scales, Tanwani, Cole-Lewis, Pfohl, Payne, Seneviratne, Gamble, Kelly, Babiker, Schärli, Chowdhery, Mansfield, Demner-Fushman, Agüera~y Arcas, Webster, Corrado, Matias, Chou, Gottweis, Tomasev, Liu, Rajkomar, Barral, Semturs, Karthikesalingam, and Natarajan]{Singhal2023}
Karan Singhal, Shekoofeh Azizi, Tao Tu, S.~Sara Mahdavi, Jason Wei, Hyung~Won Chung, Nathan Scales, Ajay Tanwani, Heather Cole-Lewis, Stephen Pfohl, Perry Payne, Martin Seneviratne, Paul Gamble, Chris Kelly, Abubakr Babiker, Nathanael Schärli, Aakanksha Chowdhery, Philip Mansfield, Dina Demner-Fushman, Blaise Agüera~y Arcas, Dale Webster, Greg~S. Corrado, Yossi Matias, Katherine Chou, Juraj Gottweis, Nenad Tomasev, Yun Liu, Alvin Rajkomar, Joelle Barral, Christopher Semturs, Alan Karthikesalingam, and Vivek Natarajan.
\newblock Large language models encode clinical knowledge.
\newblock \emph{Nature}, 620\penalty0 (7972):\penalty0 172--180, Aug 2023.
\newblock ISSN 1476-4687.
\newblock \doi{10.1038/s41586-023-06291-2}.
\newblock URL \url{https://doi.org/10.1038/s41586-023-06291-2}.

\bibitem[Singhal et~al.(2025)Singhal, Tu, Gottweis, Sayres, Wulczyn, Amin, Hou, Clark, Pfohl, Cole-Lewis, et~al.]{singhal2025toward}
Karan Singhal, Tao Tu, Juraj Gottweis, Rory Sayres, Ellery Wulczyn, Mohamed Amin, Le~Hou, Kevin Clark, Stephen~R Pfohl, Heather Cole-Lewis, et~al.
\newblock Toward expert-level medical question answering with large language models.
\newblock \emph{Nature Medicine}, pages 1--8, Jan 2025.

\bibitem[Sinha et~al.(2015)Sinha, Shen, Song, Ma, Eide, Hsu, and Wang]{sinha2015overview}
Arnab Sinha, Zhihong Shen, Yang Song, Hao Ma, Darrin Eide, Bo-June Hsu, and Kuansan Wang.
\newblock An overview of microsoft academic service (mas) and applications.
\newblock In \emph{Proceedings of the 24th international conference on world wide web}, pages 243--246, May 2015.

\bibitem[Sinha et~al.(2025)Sinha, Goel, Kumaraguru, Geiping, Bethge, and Prabhu]{sinha2025can}
Shiven Sinha, Shashwat Goel, Ponnurangam Kumaraguru, Jonas Geiping, Matthias Bethge, and Ameya Prabhu.
\newblock Can language models falsify? evaluating algorithmic reasoning with counterexample creation.
\newblock \emph{arXiv preprint arXiv:2502.19414}, 2025.

\bibitem[Skarlinski et~al.(2024)Skarlinski, Cox, Laurent, Braza, Hinks, Hammerling, Ponnapati, Rodriques, and White]{skarlinski2024language}
Michael~D Skarlinski, Sam Cox, Jon~M Laurent, James~D Braza, Michaela Hinks, Michael~J Hammerling, Manvitha Ponnapati, Samuel~G Rodriques, and Andrew~D White.
\newblock Language agents achieve superhuman synthesis of scientific knowledge.
\newblock \emph{arXiv preprint arXiv:2409.13740}, 2024.

\bibitem[Smbatyan et~al.(2025)Smbatyan, Ghukasyan, Aghajanyan, Dabaghyan, Adamyan, Bughdaryan, Altunyan, Navasardyan, Davtyan, Hakobyan, et~al.]{smbatyan2025can}
Khachik Smbatyan, Tsolak Ghukasyan, Tigran Aghajanyan, Hovhannes Dabaghyan, Sergey Adamyan, Aram Bughdaryan, Vahagn Altunyan, Gagik Navasardyan, Aram Davtyan, Anush Hakobyan, et~al.
\newblock Can ai agents design and implement drug discovery pipelines?
\newblock \emph{arXiv preprint arXiv:2504.19912}, 2025.

\bibitem[Society()]{ieeecomputersociety}
IEEE~Computer Society.
\newblock How to make peer review recommendations and decisions.
\newblock https://www.computer.org/publications/making-peer-review-recommendations.

\bibitem[Solovev et~al.(2024)Solovev, Zhidkovskaya, Orlova, Vepreva, Ilya, Golovinskii, Gubina, Chistiakov, Aliev, Poddiakov, et~al.]{solovevtowards}
Gleb~Vitalevich Solovev, Alina~Borisovna Zhidkovskaya, Anastasia Orlova, Anastasia Vepreva, Tonkii Ilya, Rodion Golovinskii, Nina Gubina, Denis Chistiakov, Timur~A Aliev, Ivan Poddiakov, et~al.
\newblock Towards llm-driven multi-agent pipeline for drug discovery: Neurodegenerative diseases case study.
\newblock In \emph{2nd AI4Research Workshop: Towards a Knowledge-grounded Scientific Research Lifecycle}, Dec 2024.

\bibitem[Son et~al.(2025)Son, Hong, Fan, Nam, Ko, Lim, Song, Choi, Paulo, Yu, et~al.]{son2025ai}
Guijin Son, Jiwoo Hong, Honglu Fan, Heejeong Nam, Hyunwoo Ko, Seungwon Lim, Jinyeop Song, Jinha Choi, Gon{\c{c}}alo Paulo, Youngjae Yu, et~al.
\newblock When ai co-scientists fail: Spot-a benchmark for automated verification of scientific research.
\newblock \emph{arXiv preprint arXiv:2505.11855}, 2025.

\bibitem[Song and Song(2023)]{song2023enhancing}
Cuiping Song and Yanping Song.
\newblock Enhancing academic writing skills and motivation: assessing the efficacy of chatgpt in ai-assisted language learning for efl students.
\newblock \emph{Frontiers in Psychology}, 14:\penalty0 1260843, Dec 2023.

\bibitem[Song et~al.(2024)Song, Song, Zhou, Fu, Zhang, and Zan]{song2024enhancing}
Jinwang Song, Yanxin Song, Guangyu Zhou, Wenhui Fu, Kunli Zhang, and Hongying Zan.
\newblock Enhancing chinese essay discourse logic evaluation through optimized fine-tuning of large language models.
\newblock In \emph{CCF International Conference on Natural Language Processing and Chinese Computing}, pages 342--352. Springer, Nov 2024.

\bibitem[Song et~al.(2024{\natexlab{2}})Song, Yang, and Anandkumar]{song2024towards}
Peiyang Song, Kaiyu Yang, and Anima Anandkumar.
\newblock Towards large language models as copilots for theorem proving in lean.
\newblock \emph{arXiv preprint arXiv:2404.12534}, 2024{\natexlab{2}}.

\bibitem[Song et~al.(2025)Song, Luo, Zhang, Chen, Huang, Cao, Zhu, Liu, Zhang, Zou, et~al.]{song2025multiagent}
Tao Song, Man Luo, Xiaolong Zhang, Linjiang Chen, Yan Huang, Jiaqi Cao, Qing Zhu, Daobin Liu, Baicheng Zhang, Gang Zou, et~al.
\newblock A multiagent-driven robotic ai chemist enabling autonomous chemical research on demand.
\newblock \emph{Journal of the American Chemical Society}, 147\penalty0 (15):\penalty0 12534--12545, Mar 2025.

\bibitem[Song et~al.(2024{\natexlab{3}})Song, Zhang, Eisenach, Kakade, Foster, and Ghai]{song2024mind}
Yuda Song, Hanlin Zhang, Carson Eisenach, Sham Kakade, Dean Foster, and Udaya Ghai.
\newblock Mind the gap: Examining the self-improvement capabilities of large language models.
\newblock \emph{arXiv preprint arXiv:2412.02674}, 2024{\natexlab{3}}.

\bibitem[Song et~al.(2025{\natexlab{2}})Song, Ju, Ren, Li, Li, Zhou, and Wang]{song2025llm}
Zhilong Song, Minggang Ju, Chunjin Ren, Qiang Li, Chongyi Li, Qionghua Zhou, and Jinlan Wang.
\newblock Llm-feynman: Leveraging large language models for universal scientific formula and theory discovery.
\newblock \emph{arXiv preprint arXiv:2503.06512}, 2025{\natexlab{2}}.

\bibitem[Spotte-Smith(2025)]{spotte2025considering}
Evan Walter~Clark Spotte-Smith.
\newblock Considering the ethics of large machine learning models in the chemical sciences.
\newblock Mar 2025.

\bibitem[Sprueill et~al.(2024)Sprueill, Edwards, Agarwal, Olarte, Sanyal, Johnston, Liu, Ji, and Choudhury]{sprueill2024chemreasoner}
Henry~W Sprueill, Carl Edwards, Khushbu Agarwal, Mariefel~V Olarte, Udishnu Sanyal, Conrad Johnston, Hongbin Liu, Heng Ji, and Sutanay Choudhury.
\newblock Chemreasoner: Heuristic search over a large language model's knowledge space using quantum-chemical feedback.
\newblock \emph{arXiv preprint arXiv:2402.10980}, 2024.

\bibitem[Spytska(2025)]{spytska2025use}
Liana Spytska.
\newblock The use of artificial intelligence in psychotherapy: development of intelligent therapeutic systems.
\newblock \emph{BMC psychology}, 13\penalty0 (1):\penalty0 175, Feb 2025.

\bibitem[Sreedhar et~al.(2025)Sreedhar, Cai, Ma, Nickerson, and Chilton]{sreedhar2025simulating}
Karthik Sreedhar, Alice Cai, Jenny Ma, Jeffrey~V Nickerson, and Lydia~B Chilton.
\newblock Simulating cooperative prosocial behavior with multi-agent llms: Evidence and mechanisms for ai agents to inform policy decisions.
\newblock In \emph{Proceedings of the 30th International Conference on Intelligent User Interfaces}, pages 1272--1286, Mar 2025.

\bibitem[Sridhar et~al.(2024)Sridhar, Dutta, Jayaraman, and Lee]{sridhar2024regent}
Kaustubh Sridhar, Souradeep Dutta, Dinesh Jayaraman, and Insup Lee.
\newblock Regent: A retrieval-augmented generalist agent that can act in-context in new environments.
\newblock In \emph{NeurIPS 2024 Workshop on Open-World Agents}, Dec 2024.

\bibitem[Staudinger et~al.(2024)Staudinger, Kusa, Piroi, and Hanbury]{staudinger2024analysis}
Moritz Staudinger, Wojciech Kusa, Florina Piroi, and Allan Hanbury.
\newblock An analysis of tasks and datasets in peer reviewing.
\newblock In \emph{Proceedings of the Fourth Workshop on Scholarly Document Processing (SDP 2024)}, pages 257--268, Aug 2024.

\bibitem[Stergiopoulos et~al.(2024)Stergiopoulos, Vassilakopoulos, Tousidou, and Corral]{stergiopoulos2024academic}
Vaios Stergiopoulos, Michael Vassilakopoulos, Eleni Tousidou, and Antonio Corral.
\newblock An academic recommender system on large citation data based on clustering, graph modeling and deep learning.
\newblock \emph{Knowledge and Information Systems}, 66\penalty0 (8):\penalty0 4463--4496, Apr 2024.

\bibitem[Sternlicht and Hope(2025)]{sternlicht2025chimera}
Noy Sternlicht and Tom Hope.
\newblock Chimera: A knowledge base of idea recombination in scientific literature.
\newblock \emph{arXiv preprint arXiv:2505.20779}, 2025.

\bibitem[Stokes et~al.(2020)Stokes, Yang, Swanson, Jin, Cubillos-Ruiz, Donghia, MacNair, French, Carfrae, Bloom-Ackermann, et~al.]{stokes2020deep}
Jonathan~M Stokes, Kevin Yang, Kyle Swanson, Wengong Jin, Andres Cubillos-Ruiz, Nina~M Donghia, Craig~R MacNair, Shawn French, Lindsey~A Carfrae, Zohar Bloom-Ackermann, et~al.
\newblock A deep learning approach to antibiotic discovery.
\newblock \emph{Cell}, 180\penalty0 (4):\penalty0 688--702, Feb 2020.

\bibitem[Strachan et~al.(2024)Strachan, Albergo, Borghini, Pansardi, Scaliti, Gupta, Saxena, Rufo, Panzeri, Manzi, Graziano, and Becchio]{strachan2024theory}
James W.~A. Strachan, Dalila Albergo, Giulia Borghini, Oriana Pansardi, Eugenio Scaliti, Saurabh Gupta, Krati Saxena, Alessandro Rufo, Stefano Panzeri, Guido Manzi, Michael S.~A. Graziano, and Cristina Becchio.
\newblock Testing theory of mind in large language models and humans.
\newblock \emph{Nature Human Behaviour}, 8:\penalty0 1285--1295, May 2024.
\newblock \doi{10.1038/s41562-024-01895-1}.
\newblock URL \url{https://www.nature.com/articles/s41562-024-01895-1}.

\bibitem[Su et~al.(2024)Su, Chen, Tang, Zheng, Li, Yin, Ouyang, and Dong]{su2024two}
Haoyang Su, Renqi Chen, Shixiang Tang, Xinzhe Zheng, Jingzhe Li, Zhenfei Yin, Wanli Ouyang, and Nanqing Dong.
\newblock Two heads are better than one: A multi-agent system has the potential to improve scientific idea generation.
\newblock \emph{arXiv preprint arXiv:2410.09403}, 2024.

\bibitem[Sui et~al.(2024)Sui, Zhou, Zhou, Han, and Zhang]{si2024table}
Yuan Sui, Mengyu Zhou, Mingjie Zhou, Shi Han, and Dongmei Zhang.
\newblock Table meets llm: Can large language models understand structured table data? a benchmark and empirical study.
\newblock In \emph{Proceedings of the 17th ACM International Conference on Web Search and Data Mining}, WSDM '24, page 645–654, New York, NY, USA, Mar 2024. Association for Computing Machinery.
\newblock ISBN 9798400703713.
\newblock \doi{10.1145/3616855.3635752}.
\newblock URL \url{https://doi.org/10.1145/3616855.3635752}.

\bibitem[Sui et~al.(2024{\natexlab{2}})Sui, Zhou, Zhou, Han, and Zhang]{sui2024table}
Yuan Sui, Mengyu Zhou, Mingjie Zhou, Shi Han, and Dongmei Zhang.
\newblock Table meets llm: Can large language models understand structured table data? a benchmark and empirical study.
\newblock In \emph{Proceedings of the 17th ACM International Conference on Web Search and Data Mining}, pages 645--654, Mar 2024{\natexlab{2}}.

\bibitem[Sukpanichnant et~al.(2024)Sukpanichnant, Rapberger, and Toni]{sukpanichnant2024peerarg}
Purin Sukpanichnant, Anna Rapberger, and Francesca Toni.
\newblock Peerarg: Argumentative peer review with llms.
\newblock \emph{arXiv preprint arXiv:2409.16813}, 2024.

\bibitem[Sun et~al.(2023)Sun, Wang, Wang, Che, Wu, Wang, and Liu]{sun2023csed}
Bo~Sun, Baoxin Wang, Yixuan Wang, Wanxiang Che, Dayong Wu, Shijin Wang, and Ting Liu.
\newblock Csed: A chinese semantic error diagnosis corpus.
\newblock \emph{arXiv preprint arXiv:2305.05183}, 2023.

\bibitem[Sun et~al.(2024)Sun, Lin, Wang, Wu, Fu, and Wang]{sun2024lalaeval}
Chongyan Sun, Ken Lin, Shiwei Wang, Hulong Wu, Chengfei Fu, and Zhen Wang.
\newblock Lalaeval: A holistic human evaluation framework for domain-specific large language models.
\newblock In \emph{First Conference on Language Modeling}, Aug 2024.

\bibitem[Sun et~al.(2025)Sun, Shen, and van~der Schaar]{sun2025openreview}
Hao Sun, Yunyi Shen, and Mihaela van~der Schaar.
\newblock Openreview should be protected and leveraged as a community asset for research in the era of large language models.
\newblock \emph{arXiv preprint arXiv:2505.21537}, 2025.

\bibitem[Sun et~al.(2023{\natexlab{2}})Sun, Zheng, Xie, Liu, Chu, Qiu, Xu, Ding, Li, Geng, et~al.]{sun2023survey}
Jiankai Sun, Chuanyang Zheng, Enze Xie, Zhengying Liu, Ruihang Chu, Jianing Qiu, Jiaqi Xu, Mingyu Ding, Hongyang Li, Mengzhe Geng, et~al.
\newblock A survey of reasoning with foundation models.
\newblock \emph{arXiv preprint arXiv:2312.11562}, 2023{\natexlab{2}}.

\bibitem[Sun et~al.(2024{\natexlab{2}})Sun, Tao, Hu, and Dow]{sun2024metawriter}
Lu~Sun, Stone Tao, Junjie Hu, and Steven~P Dow.
\newblock Metawriter: Exploring the potential and perils of ai writing support in scientific peer review.
\newblock \emph{Proceedings of the ACM on Human-Computer Interaction}, 8\penalty0 (CSCW1):\penalty0 1--32, Apr 2024{\natexlab{2}}.

\bibitem[Sun et~al.(2025{\natexlab{2}})Sun, Luo, Zhang, Li, Li, Niu, Kong, and Liu]{sun2025docs2kg}
Qiang Sun, Yuanyi Luo, Wenxiao Zhang, Sirui Li, Jichunyang Li, Kai Niu, Xiangrui Kong, and Wei Liu.
\newblock Docs2kg: A human-llm collaborative approach to unified knowledge graph construction from heterogeneous documents.
\newblock In \emph{Companion Proceedings of the ACM on Web Conference 2025}, pages 801--804, May 2025{\natexlab{2}}.

\bibitem[Sun et~al.(2025{\natexlab{3}})Sun, Liu, Ma, Ding, Xu, Yin, Zhao, Wu, Cheng, Liu, et~al.]{sun2025scienceboard}
Qiushi Sun, Zhoumianze Liu, Chang Ma, Zichen Ding, Fangzhi Xu, Zhangyue Yin, Haiteng Zhao, Zhenyu Wu, Kanzhi Cheng, Zhaoyang Liu, et~al.
\newblock Scienceboard: Evaluating multimodal autonomous agents in realistic scientific workflows.
\newblock \emph{arXiv preprint arXiv:2505.19897}, 2025{\natexlab{3}}.

\bibitem[Sun et~al.(2025{\natexlab{4}})Sun, Pan, Yang, Sui, Shi, Cheng, Li, Huang, Zhang, Yang, et~al.]{sun2025p2p}
Tao Sun, Enhao Pan, Zhengkai Yang, Kaixin Sui, Jiajun Shi, Xianfu Cheng, Tongliang Li, Wenhao Huang, Ge~Zhang, Jian Yang, et~al.
\newblock P2p: Automated paper-to-poster generation and fine-grained benchmark.
\newblock \emph{arXiv preprint arXiv:2505.17104}, 2025{\natexlab{4}}.

\bibitem[Sundararaj et~al.(2024)Sundararaj, Vyas, and Gonzalez-Maldonado]{sundararaj2024automated}
Jayaprakash Sundararaj, Akhil Vyas, and Benjamin Gonzalez-Maldonado.
\newblock Automated latex code generation from handwritten math expressions using vision transformer.
\newblock \emph{arXiv preprint arXiv:2412.03853}, 2024.

\bibitem[Suri et~al.(2024)Suri, Slater, Ziaee, and Nguyen]{suri2023largelanguagemodelsdecision}
Gaurav Suri, Lily~R Slater, Ali Ziaee, and Morgan Nguyen.
\newblock Do large language models show decision heuristics similar to humans? a case study using gpt-3.5.
\newblock \emph{Journal of Experimental Psychology: General}, 153\penalty0 (4):\penalty0 1066, May 2024.

\bibitem[Susnjak et~al.(2025)Susnjak, Hwang, Reyes, Barczak, McIntosh, and Ranathunga]{susnjak2025automating}
Teo Susnjak, Peter Hwang, Napoleon Reyes, Andre~LC Barczak, Timothy McIntosh, and Surangika Ranathunga.
\newblock Automating research synthesis with domain-specific large language model fine-tuning.
\newblock \emph{ACM Transactions on Knowledge Discovery from Data}, 19\penalty0 (3):\penalty0 1--39, Mar 2025.

\bibitem[Swanson et~al.(2025)Swanson, Wu, Bulaong, Pak, and Zou]{swanson2025virtual}
Kyle Swanson, Wesley Wu, Nash~L Bulaong, John~E Pak, and James Zou.
\newblock The virtual lab of ai agents designs new sars-cov-2 nanobodies.
\newblock \emph{Nature}, pages 1--3, Aug 2025.

\bibitem[Szumega et~al.(2023)Szumega, Bougueroua, Gkotse, Jouvelot, and Ravotti]{szumega2023open}
Jaroslaw Szumega, Lamine Bougueroua, Blerina Gkotse, Pierre Jouvelot, and Federico Ravotti.
\newblock The open review-based (orb) dataset: Towards automatic assessment of scientific papers and experiment proposals in high-energy physics.
\newblock \emph{arXiv preprint arXiv:2312.04576}, 2023.

\bibitem[Szymanski et~al.(2023)Szymanski, Rendy, Fei, Kumar, He, Milsted, McDermott, Gallant, Cubuk, Merchant, et~al.]{szymanski2023autonomous}
Nathan~J Szymanski, Bernardus Rendy, Yuxing Fei, Rishi~E Kumar, Tanjin He, David Milsted, Matthew~J McDermott, Max Gallant, Ekin~Dogus Cubuk, Amil Merchant, et~al.
\newblock An autonomous laboratory for the accelerated synthesis of novel materials.
\newblock \emph{Nature}, 624\penalty0 (7990):\penalty0 86--91, Nov 2023.

\bibitem[Taechoyotin and Acuna(2025)]{taechoyotin2025remor}
Pawin Taechoyotin and Daniel Acuna.
\newblock Remor: Automated peer review generation with llm reasoning and multi-objective reinforcement learning.
\newblock \emph{arXiv preprint arXiv:2505.11718}, 2025.

\bibitem[Taechoyotin et~al.(2024)Taechoyotin, Wang, Zeng, Sides, and Acuna]{taechoyotin2024mamorx}
Pawin Taechoyotin, Guanchao Wang, Tong Zeng, Bradley Sides, and Daniel Acuna.
\newblock Mamorx: Multi-agent multi-modal scientific review generation with external knowledge.
\newblock In \emph{Neurips 2024 Workshop Foundation Models for Science: Progress, Opportunities, and Challenges}, Oct 2024.

\bibitem[Tahamtan et~al.(2016)Tahamtan, Safipour~Afshar, and Ahamdzadeh]{tahamtan2016factors}
Iman Tahamtan, Askar Safipour~Afshar, and Khadijeh Ahamdzadeh.
\newblock Factors affecting number of citations: a comprehensive review of the literature.
\newblock \emph{Scientometrics}, 107:\penalty0 1195--1225, Feb 2016.

\bibitem[Tahmid and Notomista(2025)]{tahmid2025value}
Sheikh~A Tahmid and Gennaro Notomista.
\newblock Value iteration for learning concurrently executable robotic control tasks.
\newblock \emph{arXiv preprint arXiv:2504.01174}, 2025.

\bibitem[Tan et~al.(2024)Tan, Lyu, Li, Gao, Wei, Ma, Liu, and Li]{tan2024peer}
Cheng Tan, Dongxin Lyu, Siyuan Li, Zhangyang Gao, Jingxuan Wei, Siqi Ma, Zicheng Liu, and Stan~Z Li.
\newblock Peer review as a multi-turn and long-context dialogue with role-based interactions.
\newblock \emph{arXiv preprint arXiv:2406.05688}, 2024.

\bibitem[Tan et~al.(2025)Tan, Zhan, Jia, Zheng, and Chan]{tan2025hierarchical}
Hongming Tan, Shaoxiong Zhan, Fengwei Jia, Hai-Tao Zheng, and Wai~Kin Chan.
\newblock A hierarchical framework for measuring scientific paper innovation via large language models.
\newblock \emph{arXiv preprint arXiv:2504.14620}, 2025.

\bibitem[Tang et~al.(2024)Tang, Dai, Knight, Wu, Li, Li, and Gerstein]{tang2024survey}
Xiangru Tang, Howard Dai, Elizabeth Knight, Fang Wu, Yunyang Li, Tianxiao Li, and Mark Gerstein.
\newblock A survey of generative ai for de novo drug design: new frontiers in molecule and protein generation.
\newblock \emph{Briefings in Bioinformatics}, 25\penalty0 (4):\penalty0 bbae338, Jul 2024.

\bibitem[Tang et~al.(2024{\natexlab{2}})Tang, Zhang, Shao, Wu, Zhao, Cohan, Gong, Zhang, and Gerstein]{tang2024step}
Xiangru Tang, Xingyao Zhang, Yanjun Shao, Jie Wu, Yilun Zhao, Arman Cohan, Ming Gong, Dongmei Zhang, and Mark Gerstein.
\newblock Step-back profiling: Distilling user history for personalized scientific writing.
\newblock \emph{arXiv preprint arXiv:2406.14275}, 2024{\natexlab{2}}.

\bibitem[Tang et~al.(2024{\natexlab{3}})Tang, Zhou, Li, Ji, Hou, and Zhang]{tang2024citeeval}
Zecheng Tang, Keyan Zhou, Juntao Li, Baibei Ji, Jianye Hou, and Min Zhang.
\newblock L-citeeval: Do long-context models truly leverage context for responding?
\newblock \emph{arXiv preprint arXiv:2410.02115}, 2024{\natexlab{3}}.

\bibitem[Team et~al.(2023)Team, Anil, Borgeaud, Alayrac, Yu, Soricut, Schalkwyk, Dai, Hauth, Millican, et~al.]{team2023gemini}
Gemini Team, Rohan Anil, Sebastian Borgeaud, Jean-Baptiste Alayrac, Jiahui Yu, Radu Soricut, Johan Schalkwyk, Andrew~M Dai, Anja Hauth, Katie Millican, et~al.
\newblock Gemini: a family of highly capable multimodal models.
\newblock \emph{arXiv preprint arXiv:2312.11805}, 2023.

\bibitem[Team et~al.(2024)Team, Georgiev, Lei, Burnell, Bai, Gulati, Tanzer, Vincent, Pan, Wang, et~al.]{team2024gemini}
Gemini Team, Petko Georgiev, Ving~Ian Lei, Ryan Burnell, Libin Bai, Anmol Gulati, Garrett Tanzer, Damien Vincent, Zhufeng Pan, Shibo Wang, et~al.
\newblock Gemini 1.5: Unlocking multimodal understanding across millions of tokens of context.
\newblock \emph{arXiv preprint arXiv:2403.05530}, 2024.

\bibitem[Team et~al.(2024{\natexlab{2}})Team, Riviere, Pathak, Sessa, Hardin, Bhupatiraju, Hussenot, Mesnard, Shahriari, Ram{\'e}, et~al.]{team2024gemma}
Gemma Team, Morgane Riviere, Shreya Pathak, Pier~Giuseppe Sessa, Cassidy Hardin, Surya Bhupatiraju, L{\'e}onard Hussenot, Thomas Mesnard, Bobak Shahriari, Alexandre Ram{\'e}, et~al.
\newblock Gemma 2: Improving open language models at a practical size.
\newblock \emph{arXiv preprint arXiv:2408.00118}, 2024{\natexlab{2}}.

\bibitem[Team(2025)]{illustrae2025how}
Illustrae Team.
\newblock How to create accurate scientific illustrations with ai in 2025.
\newblock May 2025.
\newblock URL \url{https://illustrae.co/blog/how-to-create-accurate-scientific-illustrations-ai}.

\bibitem[Team et~al.(2025)Team, Du, Gao, Xing, Jiang, Chen, Li, Xiao, Du, Liao, et~al.]{team2025kimi}
Kimi Team, Angang Du, Bofei Gao, Bowei Xing, Changjiu Jiang, Cheng Chen, Cheng Li, Chenjun Xiao, Chenzhuang Du, Chonghua Liao, et~al.
\newblock Kimi k1.5: Scaling reinforcement learning with llms.
\newblock \emph{arXiv preprint arXiv:2501.12599}, 2025.

\bibitem[Team et~al.(2025{\natexlab{2}})Team, Zhang, Feng, Yan, Yuan, Yu, He, Huang, Hou, Nie, et~al.]{team2025novelseek}
NovelSeek Team, Bo~Zhang, Shiyang Feng, Xiangchao Yan, Jiakang Yuan, Zhiyin Yu, Xiaohan He, Songtao Huang, Shaowei Hou, Zheng Nie, et~al.
\newblock Novelseek: When agent becomes the scientist--building closed-loop system from hypothesis to verification.
\newblock \emph{arXiv preprint arXiv:2505.16938}, 2025{\natexlab{2}}.

\bibitem[Team(2024)]{qwen2024qwq}
Qwen Team.
\newblock Qwq: Reflect deeply on the boundaries of the unknown.
\newblock \url{https://qwenlm.github.io/blog/qwq-32b-preview/}, Dec 2024.
\newblock Accessed: 2024-12-16.

\bibitem[Team(2025{\natexlab{2}})]{qwq2025qwq}
Qwen Team.
\newblock Qwq-32b: Embracing the power of reinforcement learning, March 2025{\natexlab{2}}.
\newblock URL \url{https://qwenlm.github.io/blog/qwq-32b/}.

\bibitem[Tedford(2015)]{tedford2015helping}
Adrian Tedford.
\newblock Helping editors find reviewers.
\newblock https://www.elsevier.com/connect/helping-editors-find-reviewers, Sep 2015.

\bibitem[Thelwall and Yaghi(2025)]{thelwall2025evaluating}
Mike Thelwall and Abdallah Yaghi.
\newblock Evaluating the predictive capacity of chatgpt for academic peer review outcomes across multiple platforms.
\newblock \emph{Scientometrics}, pages 1--23, Mar 2025.

\bibitem[Tian et~al.(2024)Tian, Li, Zhang, Chen, Zou, Zhao, and Zeng]{tian2024benchmarking}
Tingzhong Tian, Shuya Li, Ziting Zhang, Lin Chen, Ziheng Zou, Dan Zhao, and Jianyang Zeng.
\newblock Benchmarking compound activity prediction for real-world drug discovery applications.
\newblock \emph{Communications Chemistry}, 7\penalty0 (1):\penalty0 127, Jun 2024.

\bibitem[Tobin et~al.(2017)Tobin, Fong, Ray, Schneider, Zaremba, and Abbeel]{tobin2017domain}
Josh Tobin, Rachel Fong, Alex Ray, Jonas Schneider, Wojciech Zaremba, and Pieter Abbeel.
\newblock Domain randomization for transferring deep neural networks from simulation to the real world.
\newblock In \emph{2017 IEEE/RSJ international conference on intelligent robots and systems (IROS)}, pages 23--30. IEEE, 2017.

\bibitem[Tom et~al.(2024)Tom, Schmid, Baird, Cao, Darvish, Hao, Lo, Pablo-Garc{\'\i}a, Rajaonson, Skreta, et~al.]{tom2024self}
Gary Tom, Stefan~P Schmid, Sterling~G Baird, Yang Cao, Kourosh Darvish, Han Hao, Stanley Lo, Sergio Pablo-Garc{\'\i}a, Ella~M Rajaonson, Marta Skreta, et~al.
\newblock Self-driving laboratories for chemistry and materials science.
\newblock \emph{Chemical Reviews}, 124\penalty0 (16):\penalty0 9633--9732, Aug 2024.

\bibitem[Tong et~al.(2024)Tong, Mao, Huang, Zhao, and Peng]{tong2024automating}
Song Tong, Kai Mao, Zhen Huang, Yukun Zhao, and Kaiping Peng.
\newblock Automating psychological hypothesis generation with ai: when large language models meet causal graph.
\newblock \emph{Humanities and Social Sciences Communications}, 11\penalty0 (1):\penalty0 1--14, Jul 2024.

\bibitem[Tornede et~al.(2023)Tornede, Deng, Eimer, Giovanelli, Mohan, Ruhkopf, Segel, Theodorakopoulos, Tornede, Wachsmuth, et~al.]{tornede2023automl}
Alexander Tornede, Difan Deng, Theresa Eimer, Joseph Giovanelli, Aditya Mohan, Tim Ruhkopf, Sarah Segel, Daphne Theodorakopoulos, Tanja Tornede, Henning Wachsmuth, et~al.
\newblock Automl in the age of large language models: Current challenges, future opportunities and risks.
\newblock \emph{arXiv preprint arXiv:2306.08107}, 2023.

\bibitem[Trifonov et~al.(2025)Trifonov, Kononov, Sherki, Svidchenko, Shpilman, and Muravleva]{trifonov2025ai}
Vladislav Trifonov, Iaroslav Kononov, Daniil Sherki, Oleg Svidchenko, Aleksei Shpilman, and Ekaterina Muravleva.
\newblock Ai-powered platform for scientific discovery.
\newblock In \emph{AI4X 2025 International Conference}, Jul 2025.

\bibitem[Truex et~al.(2019)Truex, Baracaldo, Anwar, Steinke, Ludwig, Zhang, and Zhou]{truex2019hybrid}
Stacey Truex, Nathalie Baracaldo, Ali Anwar, Thomas Steinke, Heiko Ludwig, Rui Zhang, and Yi~Zhou.
\newblock A hybrid approach to privacy-preserving federated learning.
\newblock In \emph{Proceedings of the 12th ACM workshop on artificial intelligence and security}, pages 1--11, Aug 2019.

\bibitem[Tu et~al.(2024)Tu, Hadan, Wang, Sgandurra, Mogavi, and Nacke]{tu2024augmenting}
Joseph Tu, Hilda Hadan, Derrick~M Wang, Sabrina~A Sgandurra, Reza~Hadi Mogavi, and Lennart~E Nacke.
\newblock Augmenting the author: Exploring the potential of ai collaboration in academic writing.
\newblock \emph{arXiv preprint arXiv:2404.16071}, 2024.

\bibitem[Tyser et~al.(2024)Tyser, Segev, Longhitano, Zhang, Meeks, Lee, Garg, Belsten, Shporer, Udell, et~al.]{tyser2024ai}
Keith Tyser, Ben Segev, Gaston Longhitano, Xin-Yu Zhang, Zachary Meeks, Jason Lee, Uday Garg, Nicholas Belsten, Avi Shporer, Madeleine Udell, et~al.
\newblock Ai-driven review systems: evaluating llms in scalable and bias-aware academic reviews.
\newblock \emph{arXiv preprint arXiv:2408.10365}, 2024.

\bibitem[Uddin et~al.(2025)Uddin, Vaidya, Choudhary, Chen, Salib, Huang, Englund, and Solja{\v{c}}i{\'c}]{uddin2025ai}
Shiekh~Zia Uddin, Sachin Vaidya, Shrish Choudhary, Zhuo Chen, Raafat~K Salib, Luke Huang, Dirk~R Englund, and Marin Solja{\v{c}}i{\'c}.
\newblock Ai-driven robotics for free-space optics.
\newblock \emph{arXiv preprint arXiv:2505.17985}, 2025.

\bibitem[Vischia(2025)]{vischia2025ai}
Pietro Vischia.
\newblock Ai-assisted design of experiments at the frontiers of computation: methods and new perspectives.
\newblock \emph{arXiv preprint arXiv:2501.04448}, 2025.

\bibitem[Vladika and Matthes(2024)]{vladika2024comparing}
Juraj Vladika and Florian Matthes.
\newblock Comparing knowledge sources for open-domain scientific claim verification.
\newblock \emph{arXiv preprint arXiv:2402.02844}, 2024.

\bibitem[Vladika and Matthes(2024{\natexlab{2}})]{vladika2024improving}
Juraj Vladika and Florian Matthes.
\newblock Improving health question answering with reliable and time-aware evidence retrieval.
\newblock \emph{arXiv preprint arXiv:2404.08359}, 2024{\natexlab{2}}.

\bibitem[Vre{\v{c}}ar et~al.(2024)Vre{\v{c}}ar, Wells, and Kamareddine]{vrevcar2024towards}
Luka Vre{\v{c}}ar, Joe Wells, and Fairouz Kamareddine.
\newblock Towards semantic markup of mathematical documents via user interaction.
\newblock In \emph{International Conference on Intelligent Computer Mathematics}, pages 223--240. Springer, Jul 2024.

\bibitem[Wadden et~al.(2021)Wadden, Lo, Wang, Cohan, Beltagy, and Hajishirzi]{wadden2021multivers}
David Wadden, Kyle Lo, Lucy~Lu Wang, Arman Cohan, Iz~Beltagy, and Hannaneh Hajishirzi.
\newblock Multivers: Improving scientific claim verification with weak supervision and full-document context.
\newblock \emph{arXiv preprint arXiv:2112.01640}, 2021.

\bibitem[Wadden et~al.(2024)Wadden, Shi, Morrison, Naik, Singh, Barzilay, Lo, Hope, Soldaini, Shen, et~al.]{wadden2024sciriff}
David Wadden, Kejian Shi, Jacob Morrison, Aakanksha Naik, Shruti Singh, Nitzan Barzilay, Kyle Lo, Tom Hope, Luca Soldaini, Zejiang Shen, et~al.
\newblock Sciriff: A resource to enhance language model instruction-following over scientific literature.
\newblock In \emph{Neurips 2024 Workshop Foundation Models for Science: Progress, Opportunities, and Challenges}, Dec 2024.

\bibitem[Wallace et~al.(2021)Wallace, Saha, Soboczenski, and Marshall]{wallace2021generating}
Byron~C Wallace, Sayantan Saha, Frank Soboczenski, and Iain~J Marshall.
\newblock Generating (factual?) narrative summaries of rcts: Experiments with neural multi-document summarization.
\newblock \emph{AMIA Summits on Translational Science Proceedings}, 2021:\penalty0 605, May 2021.

\bibitem[Wan et~al.(2024)Wan, Liu, Ajith, Grazian, Hoex, Zhang, Kit, Xie, and Foster]{wan2024sciqag}
Yuwei Wan, Yixuan Liu, Aswathy Ajith, Clara Grazian, Bram Hoex, Wenjie Zhang, Chunyu Kit, Tong Xie, and Ian Foster.
\newblock Sciqag: A framework for auto-generated science question answering dataset with fine-grained evaluation.
\newblock \emph{arXiv preprint arXiv:2405.09939}, 2024.

\bibitem[Wang et~al.(2025)Wang, Kim, Vriza, Batra, Baskaran, Shan, Li, Darancet, Ward, Liu, et~al.]{wang2025autonomous}
Chengshi Wang, Yeon-Ju Kim, Aikaterini Vriza, Rohit Batra, Arun Baskaran, Naisong Shan, Nan Li, Pierre Darancet, Logan Ward, Yuzi Liu, et~al.
\newblock Autonomous platform for solution processing of electronic polymers.
\newblock \emph{Nature communications}, 16\penalty0 (1):\penalty0 1498, 2025.

\bibitem[Wang et~al.(2025{\natexlab{2}})Wang, Shen, Kuang, Cohan, and Zhao]{wang-etal-2025-sciver}
Chengye Wang, Yifei Shen, Zexi Kuang, Arman Cohan, and Yilun Zhao.
\newblock {S}ci{V}er: Evaluating foundation models for multimodal scientific claim verification.
\newblock In \emph{Proceedings of the 63rd Annual Meeting of the Association for Computational Linguistics (Volume 1: Long Papers)}, pages 8562--8579, 2025{\natexlab{2}}.

\bibitem[Wang et~al.(2022)Wang, Dou, and Che]{wang2022survey}
Dingzirui Wang, Longxu Dou, and Wanxiang Che.
\newblock A survey on table-and-text hybridqa: Concepts, methods, challenges and future directions.
\newblock \emph{arXiv preprint arXiv:2212.13465}, 2022.

\bibitem[Wang et~al.(2024)Wang, Dou, Zhang, Zhu, and Che]{wang2024improving_a}
Dingzirui Wang, Longxu Dou, Xuanliang Zhang, Qingfu Zhu, and Wanxiang Che.
\newblock Improving demonstration diversity by human-free fusing for text-to-sql.
\newblock \emph{arXiv preprint arXiv:2402.10663}, 2024.

\bibitem[Wang et~al.(2023)Wang, Xin, Zheng, Li, Liu, Cao, Huang, Xiong, Shi, Xie, et~al.]{wang2023lego}
Haiming Wang, Huajian Xin, Chuanyang Zheng, Lin Li, Zhengying Liu, Qingxing Cao, Yinya Huang, Jing Xiong, Han Shi, Enze Xie, et~al.
\newblock Lego-prover: Neural theorem proving with growing libraries.
\newblock \emph{arXiv preprint arXiv:2310.00656}, 2023.

\bibitem[Wang et~al.(2023{\natexlab{2}})Wang, Yuan, Liu, Shen, Yin, Xiong, Xie, Shi, Li, Li, et~al.]{wang2023dt}
Haiming Wang, Ye~Yuan, Zhengying Liu, Jianhao Shen, Yichun Yin, Jing Xiong, Enze Xie, Han Shi, Yujun Li, Lin Li, et~al.
\newblock Dt-solver: Automated theorem proving with dynamic-tree sampling guided by proof-level value function.
\newblock In \emph{Proceedings of the 61st Annual Meeting of the Association for Computational Linguistics (Volume 1: Long Papers)}, pages 12632--12646, Jul 2023{\natexlab{2}}.

\bibitem[Wang et~al.(2024{\natexlab{2}})Wang, Xin, Liu, Li, Huang, Lu, Yang, Tang, Yin, Li, et~al.]{wang2024proving}
Haiming Wang, Huajian Xin, Zhengying Liu, Wenda Li, Yinya Huang, Jianqiao Lu, Zhicheng Yang, Jing Tang, Jian Yin, Zhenguo Li, et~al.
\newblock Proving theorems recursively.
\newblock \emph{arXiv preprint arXiv:2405.14414}, 2024{\natexlab{2}}.

\bibitem[Wang(2024)]{wang2024content}
Haining Wang.
\newblock A content-based novelty measure for scholarly publications: A proof of concept.
\newblock In \emph{International Conference on Information}, pages 409--420. Springer, Apr 2024.

\bibitem[Wang et~al.(2023{\natexlab{3}})Wang, Fu, Du, Gao, Huang, Liu, Chandak, Liu, Van~Katwyk, Deac, et~al.]{wang2023scientific}
Hanchen Wang, Tianfan Fu, Yuanqi Du, Wenhao Gao, Kexin Huang, Ziming Liu, Payal Chandak, Shengchao Liu, Peter Van~Katwyk, Andreea Deac, et~al.
\newblock Scientific discovery in the age of artificial intelligence.
\newblock \emph{Nature}, 620\penalty0 (7972):\penalty0 47--60, Aug 2023{\natexlab{3}}.

\bibitem[Wang et~al.(2025{\natexlab{3}})Wang, He, Coelho, Bucci, Nazir, Chen, Trinh, Zhang, Huang, Chandrasekar, et~al.]{wang2025spatialagent}
Hanchen Wang, Yichun He, Paula~P Coelho, Matthew Bucci, Abbas Nazir, Bob Chen, Linh Trinh, Serena Zhang, Kexin Huang, Vineethkrishna Chandrasekar, et~al.
\newblock Spatialagent: An autonomous ai agent for spatial biology.
\newblock \emph{bioRxiv}, pages 2025--04, Apr 2025{\natexlab{3}}.

\bibitem[Wang et~al.(2024{\natexlab{3}})Wang, Skreta, Ser, Gao, Kong, Strieth-Kalthoff, Duan, Zhuang, Yu, Zhu, et~al.]{wang2024efficient}
Haorui Wang, Marta Skreta, Cher-Tian Ser, Wenhao Gao, Lingkai Kong, Felix Strieth-Kalthoff, Chenru Duan, Yuchen Zhuang, Yue Yu, Yanqiao Zhu, et~al.
\newblock Efficient evolutionary search over chemical space with large language models.
\newblock \emph{arXiv preprint arXiv:2406.16976}, 2024{\natexlab{3}}.

\bibitem[Wang et~al.(2025{\natexlab{4}})Wang, Fu, Zhang, Wang, Ren, Wang, Li, He, An, Liu, et~al.]{wang2025llm}
Haoyu Wang, Yujia Fu, Zhu Zhang, Shuo Wang, Zirui Ren, Xiaorong Wang, Zhili Li, Chaoqun He, Bo~An, Zhiyuan Liu, et~al.
\newblock Llm $\times$ mapreduce-v2: Entropy-driven convolutional test-time scaling for generating long-form articles from extremely long resources.
\newblock \emph{arXiv preprint arXiv:2504.05732}, 2025{\natexlab{4}}.

\bibitem[Wang et~al.(2024{\natexlab{4}})Wang, Wu, Coates, Zeng, Kuang, Liu, Qiu, and Park]{wang2024neural}
Izia~Xiaoxiao Wang, Xihan Wu, Edith Coates, Min Zeng, Jiexin Kuang, Siliang Liu, Mengyang Qiu, and Jungyeul Park.
\newblock Neural automated writing evaluation with corrective feedback.
\newblock \emph{arXiv preprint arXiv:2402.17613}, 2024{\natexlab{4}}.

\bibitem[Wang et~al.(2024{\natexlab{5}})Wang, Xiao, Li, Song, Xu, Tan, and Li]{wang2024clientcenteredassessmentllmtherapists}
Jiashuo Wang, Yang Xiao, Yanran Li, Changhe Song, Chunpu Xu, Chenhao Tan, and Wenjie Li.
\newblock Towards a client-centered assessment of llm therapists by client simulation.
\newblock \emph{arXiv preprint arXiv:2406.12266}, 2024{\natexlab{5}}.

\bibitem[Wang et~al.(2023{\natexlab{4}})Wang, Hu, He, Xu, Liu, Liu, and Shen]{wang2023t}
Lei Wang, Yi~Hu, Jiabang He, Xing Xu, Ning Liu, Hui Liu, and Heng~Tao Shen.
\newblock T-sciq: Teaching multimodal chain-of-thought reasoning via mixed large language model signals for science question answering.
\newblock \emph{arXiv preprint arXiv:2305.03453}, 2023{\natexlab{4}}.

\bibitem[Wang et~al.(2025{\natexlab{5}})Wang, Lee, Volkov, Chau, and Kang]{wang2025scholawrite}
Linghe Wang, Minhwa Lee, Ross Volkov, Luan~Tuyen Chau, and Dongyeop Kang.
\newblock Scholawrite: A dataset of end-to-end scholarly writing process.
\newblock \emph{arXiv preprint arXiv:2502.02904}, 2025{\natexlab{5}}.

\bibitem[Wang et~al.(2022{\natexlab{2}})Wang, DeYoung, and Wallace]{wang2022overview}
Lucy~Lu Wang, Jay DeYoung, and Byron Wallace.
\newblock Overview of mslr2022: A shared task on multi-document summarization for literature reviews.
\newblock In \emph{Proceedings of the third workshop on scholarly document processing}, Oct 2022{\natexlab{2}}.

\bibitem[Wang et~al.(2019)Wang, Li, Zhou, Tang, and Wang]{wang2019toc}
Pancheng Wang, Shasha Li, Haifang Zhou, Jintao Tang, and Ting Wang.
\newblock Toc-rwg: Explore the combination of topic model and citation information for automatic related work generation.
\newblock \emph{Ieee Access}, 8:\penalty0 13043--13055, Dec 2019.

\bibitem[Wang et~al.(2022{\natexlab{3}})Wang, Li, Pang, He, Li, Tang, and Wang]{wang2022multi}
Pancheng Wang, Shasha Li, Kunyuan Pang, Liangliang He, Dong Li, Jintao Tang, and Ting Wang.
\newblock Multi-document scientific summarization from a knowledge graph-centric view.
\newblock \emph{arXiv preprint arXiv:2209.04319}, 2022{\natexlab{3}}.

\bibitem[Wang et~al.(2024{\natexlab{6}})Wang, Li, Li, Long, Tang, and Wang]{wang2024disentangling}
Pancheng Wang, Shasha Li, Dong Li, Kehan Long, Jintao Tang, and Ting Wang.
\newblock Disentangling instructive information from ranked multiple candidates for multi-document scientific summarization.
\newblock In \emph{Proceedings of the 47th International ACM SIGIR Conference on Research and Development in Information Retrieval}, pages 2028--2037, Jul 2024{\natexlab{6}}.

\bibitem[Wang et~al.(2024{\natexlab{7}})Wang, Zhang, Fei, Chen, Wang, Si, Lu, Li, and Qin]{wang2024s3}
Peng Wang, Yongheng Zhang, Hao Fei, Qiguang Chen, Yukai Wang, Jiasheng Si, Wenpeng Lu, Min Li, and Libo Qin.
\newblock S3 agent: Unlocking the power of vllm for zero-shot multi-modal sarcasm detection.
\newblock \emph{ACM Transactions on Multimedia Computing, Communications and Applications}, Aug 2024{\natexlab{7}}.

\bibitem[Wang et~al.(2025{\natexlab{6}})Wang, Tao, Chen, Hu, and Qin]{wang2025x}
Peng Wang, Ruihan Tao, Qiguang Chen, Mengkang Hu, and Libo Qin.
\newblock X-webagentbench: A multilingual interactive web benchmark for evaluating global agentic system.
\newblock \emph{arXiv preprint arXiv:2505.15372}, 2025{\natexlab{6}}.

\bibitem[Wang et~al.(2024{\natexlab{8}})Wang, Downey, Ji, and Hope]{wang2024scimon}
Qingyun Wang, Doug Downey, Heng Ji, and Tom Hope.
\newblock Scimon: Scientific inspiration machines optimized for novelty.
\newblock In \emph{Proceedings of the 62nd Annual Meeting of the Association for Computational Linguistics (Volume 1: Long Papers)}, pages 279--299, Aug 2024{\natexlab{8}}.

\bibitem[Wang et~al.(2025{\natexlab{7}})Wang, Wang, Li, Zhang, and Cheng]{wang2025drsr}
Runxiang Wang, Boxiao Wang, Kai Li, Yifan Zhang, and Jian Cheng.
\newblock Drsr: Llm based scientific equation discovery with dual reasoning from data and experience.
\newblock \emph{arXiv preprint arXiv:2506.04282}, 2025{\natexlab{7}}.

\bibitem[Wang et~al.(2025{\natexlab{8}})Wang, Foulds, Gani, and Pan]{wang2025llmbased}
Siyuan Wang, James~R Foulds, Md~Osman Gani, and Shimei Pan.
\newblock Llm-based corroborating and refuting evidence retrieval for scientific claim verification.
\newblock \emph{arXiv preprint arXiv:2503.07937}, 2025{\natexlab{8}}.

\bibitem[Wang et~al.(2024{\natexlab{9}})Wang, Gu, Zhang, Luo, Dai, Shen, Xie, Lin, He, and Ye]{wang2024scipip}
Wenxiao Wang, Lihui Gu, Liye Zhang, Yunxiang Luo, Yi~Dai, Chen Shen, Liang Xie, Binbin Lin, Xiaofei He, and Jieping Ye.
\newblock Scipip: An llm-based scientific paper idea proposer.
\newblock \emph{arXiv preprint arXiv:2410.23166}, 2024{\natexlab{9}}.

\bibitem[Wang et~al.(2025{\natexlab{9}})Wang, Ma, Wang, Wu, Chen, Li, and Yuan]{wang2025survey}
Wenxuan Wang, Zizhan Ma, Zheng Wang, Chenghan Wu, Wenting Chen, Xiang Li, and Yixuan Yuan.
\newblock A survey of llm-based agents in medicine: How far are we from baymax?
\newblock \emph{arXiv preprint arXiv:2502.11211}, 2025{\natexlab{9}}.

\bibitem[Wang et~al.(2025{\natexlab{10}})Wang, Tan, Jin, Xiong, Hu, Zhang, Lu, and Zhang]{wang2025medcite}
Xiao Wang, Mengjue Tan, Qiao Jin, Guangzhi Xiong, Yu~Hu, Aidong Zhang, Zhiyong Lu, and Minjia Zhang.
\newblock Medcite: Can language models generate verifiable text for medicine?
\newblock \emph{arXiv preprint arXiv:2506.06605}, 2025{\natexlab{10}}.

\bibitem[Wang et~al.(2023{\natexlab{5}})Wang, Hu, Lu, Zhu, Zhang, Subramaniam, Loomba, Zhang, Sun, and Wang]{wang2023scibench}
Xiaoxuan Wang, Ziniu Hu, Pan Lu, Yanqiao Zhu, Jieyu Zhang, Satyen Subramaniam, Arjun~R Loomba, Shichang Zhang, Yizhou Sun, and Wei Wang.
\newblock Scibench: Evaluating college-level scientific problem-solving abilities of large language models.
\newblock \emph{arXiv preprint arXiv:2307.10635}, 2023{\natexlab{5}}.

\bibitem[Wang et~al.(2024{\natexlab{10}})Wang, Li, Song, Xu, Tang, Zhuge, Pan, Song, Li, Singh, et~al.]{wang2024opendevin}
Xingyao Wang, Boxuan Li, Yufan Song, Frank~F Xu, Xiangru Tang, Mingchen Zhuge, Jiayi Pan, Yueqi Song, Bowen Li, Jaskirat Singh, et~al.
\newblock Opendevin: An open platform for ai software developers as generalist agents.
\newblock \emph{arXiv preprint arXiv:2407.16741}, 2024{\natexlab{10}}.

\bibitem[Wang et~al.(2024{\natexlab{11}})Wang, Li, Song, Xu, Tang, Zhuge, Pan, Song, Li, Singh, et~al.]{wang2024openhands}
Xingyao Wang, Boxuan Li, Yufan Song, Frank~F Xu, Xiangru Tang, Mingchen Zhuge, Jiayi Pan, Yueqi Song, Bowen Li, Jaskirat Singh, et~al.
\newblock Openhands: An open platform for ai software developers as generalist agents.
\newblock \emph{arXiv preprint arXiv:2407.16741}, 2024{\natexlab{11}}.

\bibitem[Wang et~al.(2025{\natexlab{11}})Wang, Cui, and Jiang]{wang2025enabling}
Yao Wang, Mingxuan Cui, and Arthur Jiang.
\newblock Enabling ai scientists to recognize innovation: A domain-agnostic algorithm for assessing novelty.
\newblock \emph{arXiv preprint arXiv:2503.01508}, 2025{\natexlab{11}}.

\bibitem[Wang et~al.(2024{\natexlab{12}})Wang, Guo, Yao, Zhang, Zhang, Wu, Zhang, Dai, Wen, Ye, et~al.]{wang2024autosurvey}
Yidong Wang, Qi~Guo, Wenjin Yao, Hongbo Zhang, Xin Zhang, Zhen Wu, Meishan Zhang, Xinyu Dai, Qingsong Wen, Wei Ye, et~al.
\newblock Autosurvey: Large language models can automatically write surveys.
\newblock \emph{Advances in Neural Information Processing Systems}, 37:\penalty0 115119--115145, Dec 2024{\natexlab{12}}.

\bibitem[Wang et~al.(2024{\natexlab{13}})Wang, Stevens, Shah, Jiang, Liu, Chen, Kuo, Li, Gong, Lee, et~al.]{wang2024model}
Yifan Wang, David Stevens, Pranay Shah, Wenwen Jiang, Miao Liu, Xu~Chen, Robert Kuo, Na~Li, Boying Gong, Daniel Lee, et~al.
\newblock Model-in-the-loop (milo): Accelerating multimodal ai data annotation with llms.
\newblock \emph{arXiv preprint arXiv:2409.10702}, 2024{\natexlab{13}}.

\bibitem[Wang et~al.(2024{\natexlab{14}})Wang, Wang, Liu, Wu, and Che]{wang2024lm}
Yixuan Wang, Baoxin Wang, Yijun Liu, Dayong Wu, and Wanxiang Che.
\newblock Lm-combiner: A contextual rewriting model for chinese grammatical error correction.
\newblock \emph{arXiv preprint arXiv:2403.17413}, 2024{\natexlab{14}}.

\bibitem[Wang et~al.(2024{\natexlab{15}})Wang, Wang, Liu, Zhu, Wu, and Che]{wang2024improving_b}
Yixuan Wang, Baoxin Wang, Yijun Liu, Qingfu Zhu, Dayong Wu, and Wanxiang Che.
\newblock Improving grammatical error correction via contextual data augmentation.
\newblock \emph{arXiv preprint arXiv:2406.17456}, 2024{\natexlab{15}}.

\bibitem[Wang et~al.(2024{\natexlab{16}})Wang, Bai, Lin, Wang, Anitescu, Sun, Eshaghi, Gu, Feng, Zhuang, et~al.]{wang2024artificial}
Yizheng Wang, Jinshuai Bai, Zhongya Lin, Qimin Wang, Cosmin Anitescu, Jia Sun, Mohammad~Sadegh Eshaghi, Yuantong Gu, Xi-Qiao Feng, Xiaoying Zhuang, et~al.
\newblock Artificial intelligence for partial differential equations in computational mechanics: A review.
\newblock \emph{arXiv preprint arXiv:2410.19843}, 2024{\natexlab{16}}.

\bibitem[Wang et~al.(2019{\natexlab{2}})Wang, Liu, and Gao]{wang2019neural}
Yongzhen Wang, Xiaozhong Liu, and Zheng Gao.
\newblock Neural related work summarization with a joint context-driven attention mechanism.
\newblock \emph{arXiv preprint arXiv:1901.09492}, 2019{\natexlab{2}}.

\bibitem[Wang et~al.(2025{\natexlab{12}})Wang, Ma, Nie, Zeng, Lyu, Zhang, Schneider, Lu, Yue, and Chen]{wang2025scholarcopilot}
Yubo Wang, Xueguang Ma, Ping Nie, Huaye Zeng, Zhiheng Lyu, Yuxuan Zhang, Benjamin Schneider, Yi~Lu, Xiang Yue, and Wenhu Chen.
\newblock Scholarcopilot: Training large language models for academic writing with accurate citations.
\newblock \emph{arXiv preprint arXiv:2504.00824}, 2025{\natexlab{12}}.

\bibitem[Wang et~al.(2025{\natexlab{13}})Wang, Danek, and Sun]{wang2025biodsa}
Zifeng Wang, Benjamin Danek, and Jimeng Sun.
\newblock Biodsa-1k: Benchmarking data science agents for biomedical research.
\newblock \emph{arXiv preprint arXiv:2505.16100}, 2025{\natexlab{13}}.

\bibitem[Wang et~al.(2024{\natexlab{17}})Wang, Zhang, Li, Eisenschlos, Perot, Wang, Miculicich, Fujii, Shang, Lee, et~al.]{wang2024chain}
Zilong Wang, Hao Zhang, Chun-Liang Li, Julian~Martin Eisenschlos, Vincent Perot, Zifeng Wang, Lesly Miculicich, Yasuhisa Fujii, Jingbo Shang, Chen-Yu Lee, et~al.
\newblock Chain-of-table: Evolving tables in the reasoning chain for table understanding.
\newblock In \emph{The Twelfth International Conference on Learning Representations}, Jan 2024{\natexlab{17}}.
\newblock URL \url{https://openreview.net/forum?id=4L0xnS4GQM}.

\bibitem[Wang et~al.(2023{\natexlab{6}})Wang, Hou, Lu, Wu, Li, Yu, and Ji]{wang2023enabling}
Ziqi Wang, Le~Hou, Tianjian Lu, Yuexin Wu, Yunxuan Li, Hongkun Yu, and Heng Ji.
\newblock Enabling language models to implicitly learn self-improvement.
\newblock \emph{arXiv preprint arXiv:2310.00898}, 2023{\natexlab{6}}.

\bibitem[Wang et~al.(2024{\natexlab{18}})Wang, Xia, He, Chen, Liu, Zhu, Liang, Wu, Liu, Malladi, Chevalier, Arora, and Chen]{wang2024charxiv}
Zirui Wang, Mengzhou Xia, Luxi He, Howard Chen, Yitao Liu, Richard Zhu, Kaiqu Liang, Xindi Wu, Haotian Liu, Sadhika Malladi, Alexis Chevalier, Sanjeev Arora, and Danqi Chen.
\newblock Charxiv: Charting gaps in realistic chart understanding in multimodal llms.
\newblock In A.~Globerson, L.~Mackey, D.~Belgrave, A.~Fan, U.~Paquet, J.~Tomczak, and C.~Zhang, editors, \emph{Advances in Neural Information Processing Systems}, volume~37, pages 113569--113697. Curran Associates, Inc., Dec 2024{\natexlab{18}}.
\newblock URL \url{https://proceedings.neurips.cc/paper_files/paper/2024/file/cdf6f8e9fd9aeaf79b6024caec24f15b-Paper-Datasets_and_Benchmarks_Track.pdf}.

\bibitem[Watson and Yong(2024)]{watson2024directed}
William Watson and Lawrence Yong.
\newblock Directed criteria citation recommendation and ranking through link prediction.
\newblock \emph{arXiv preprint arXiv:2403.18855}, 2024.

\bibitem[Webster and Watson(2002)]{webster2002analyzing}
Jane Webster and Richard~T Watson.
\newblock Analyzing the past to prepare for the future: Writing a literature review.
\newblock \emph{MIS quarterly}, pages xiii--xxiii, Jun 2002.

\bibitem[Wei et~al.(2025)Wei, Sun, Papay, McKinney, Han, Fulford, Chung, Passos, Fedus, and Glaese]{wei2025browsecomp}
Jason Wei, Zhiqing Sun, Spencer Papay, Scott McKinney, Jeffrey Han, Isa Fulford, Hyung~Won Chung, Alex~Tachard Passos, William Fedus, and Amelia Glaese.
\newblock Browsecomp: A simple yet challenging benchmark for browsing agents.
\newblock \emph{arXiv preprint arXiv:2504.12516}, 2025.

\bibitem[Wen et~al.(2024)Wen, Wei, Lin, Wang, Liang, and Wan]{wen2024overleafcopilot}
Haomin Wen, Zhenjie Wei, Yan Lin, Jiyuan Wang, Yuxuan Liang, and Huaiyu Wan.
\newblock Overleafcopilot: Empowering academic writing in overleaf with large language models.
\newblock \emph{arXiv preprint arXiv:2403.09733}, 2024.

\bibitem[Wen et~al.(2025)Wen, Si, Chen, He, and Feng]{wen2025predicting}
Jiaxin Wen, Chenglei Si, Yueh-han Chen, He~He, and Shi Feng.
\newblock Predicting empirical ai research outcomes with language models.
\newblock \emph{arXiv preprint arXiv:2506.00794}, 2025.

\bibitem[Wen and Yi(2025)]{wen2025plain}
Ju~Wen and Lan Yi.
\newblock Are plain language summaries more readable than scientific abstracts? evidence from six biomedical and life sciences journals.
\newblock \emph{Public Understanding of Science}, 34\penalty0 (1):\penalty0 114--126, May 2025.

\bibitem[Wen and Laporte(2025)]{doi:10.1177/07439156241297973}
Yingting Wen and Sandra Laporte.
\newblock Experiential narratives in marketing: A comparison of generative ai and human content.
\newblock \emph{Journal of Public Policy \& Marketing}, 44\penalty0 (3):\penalty0 392--410, Oct 2025.
\newblock \doi{10.1177/07439156241297973}.
\newblock URL \url{https://doi.org/10.1177/07439156241297973}.

\bibitem[Weng et~al.(2024)Weng, Zhu, Bao, Zhang, Wang, Zhang, and Yang]{weng2024cycleresearcher}
Yixuan Weng, Minjun Zhu, Guangsheng Bao, Hongbo Zhang, Jindong Wang, Yue Zhang, and Linyi Yang.
\newblock Cycleresearcher: Improving automated research via automated review.
\newblock \emph{arXiv preprint arXiv:2411.00816}, 2024.

\bibitem[Wheeler(2025)]{wheeler2025responsible}
Nicole~E Wheeler.
\newblock Responsible ai in biotechnology: balancing discovery, innovation and biosecurity risks.
\newblock \emph{Frontiers in Bioengineering and Biotechnology}, 13:\penalty0 1537471, Feb 2025.

\bibitem[Wright and Augenstein(2021)]{wright2021citeworth}
Dustin Wright and Isabelle Augenstein.
\newblock Citeworth: Cite-worthiness detection for improved scientific document understanding.
\newblock In \emph{Findings of the Association for Computational Linguistics: ACL-IJCNLP 2021}, pages 1796--1807, May 2021.

\bibitem[Wu et~al.(2023)Wu, Lei, Zheng, Zhao, Lin, Zhang, Zhou, Zhao, Zhang, Wang, et~al.]{wu2023can}
Chaoyi Wu, Jiayu Lei, Qiaoyu Zheng, Weike Zhao, Weixiong Lin, Xiaoman Zhang, Xiao Zhou, Ziheng Zhao, Ya~Zhang, Yanfeng Wang, et~al.
\newblock Can gpt-4v (ision) serve medical applications? case studies on gpt-4v for multimodal medical diagnosis.
\newblock \emph{arXiv preprint arXiv:2310.09909}, Oct 2023.

\bibitem[Wu et~al.(2025)Wu, Li, Fang, Yin, Zhang, Tao, Zhang, Xi, Jiang, Xie, et~al.]{wu2025webdancer}
Jialong Wu, Baixuan Li, Runnan Fang, Wenbiao Yin, Liwen Zhang, Zhengwei Tao, Dingchu Zhang, Zekun Xi, Yong Jiang, Pengjun Xie, et~al.
\newblock Webdancer: Towards autonomous information seeking agency.
\newblock \emph{arXiv preprint arXiv:2505.22648}, 2025.

\bibitem[Wu et~al.(2025{\natexlab{2}})Wu, Zhang, Li, Yang, Zhong, Jiang, Wen, and Zhang]{wu2025lag}
Jian Wu, Jiayu Zhang, Dongyuan Li, Linyi Yang, Aoxiao Zhong, Renhe Jiang, Qingsong Wen, and Yue Zhang.
\newblock Lag: Llm agents for leaderboard auto generation on demanding.
\newblock \emph{arXiv preprint arXiv:2502.18209}, 2025{\natexlab{2}}.

\bibitem[Wu et~al.(2023{\natexlab{2}})Wu, Chao, Zhou, and Luo]{wu2023characterizing}
Jinxuan Wu, Wenhan Chao, Xian Zhou, and Zhunchen Luo.
\newblock Characterizing and verifying scientific claims: Qualitative causal structure is all you need.
\newblock In \emph{Proceedings of the 2023 Conference on Empirical Methods in Natural Language Processing}, pages 13428--13439, Dec 2023{\natexlab{2}}.

\bibitem[Wu et~al.(2025{\natexlab{3}})Wu, He, Liu, Zheng, Cao, Chen, Zou, Zou, Zhou, Sturgess, et~al.]{wu2025literature}
Junfeng Wu, Jing He, Hai Liu, Zhaoqi Zheng, Yichen Cao, Xingguo Chen, Bingjie Zou, Ruiping Zou, Guohua Zhou, David Sturgess, et~al.
\newblock From literature to lab: Hardware-independent autonomous chemical synthesis with reinforcement learning.
\newblock In \emph{Companion Proceedings of the ACM on Web Conference 2025}, pages 2923--2926, May 2025{\natexlab{3}}.

\bibitem[Wu et~al.(2024)Wu, Ma, Luo, Li, Shi, Chang, Lin, Luo, Pei, Du, et~al.]{wu2024automated}
Shican Wu, Xiao Ma, Dehui Luo, Lulu Li, Xiangcheng Shi, Xin Chang, Xiaoyun Lin, Ran Luo, Chunlei Pei, Changying Du, et~al.
\newblock Automated review generation method based on large language models.
\newblock \emph{arXiv preprint arXiv:2407.20906}, 2024.

\bibitem[Wu et~al.(2025{\natexlab{4}})Wu, Bao, Kunievsky, and Evans]{wu2025introspective}
Siyang Wu, Honglin Bao, Nadav Kunievsky, and James~A Evans.
\newblock Introspective growth: Automatically advancing llm expertise in technology judgment.
\newblock \emph{arXiv preprint arXiv:2505.12452}, 2025{\natexlab{4}}.

\bibitem[Wu et~al.(2018)Wu, U, Bhowmick, and Gatterbauer]{wu2018pistis}
Siyuan Wu, Leong~Hou U, Sourav~S Bhowmick, and Wolfgang Gatterbauer.
\newblock Pistis: A conflict of interest declaration and detection system for peer review management.
\newblock In \emph{Proceedings of the 2018 International Conference on Management of Data}, pages 1713--1716, May 2018.

\bibitem[Wu et~al.(2025{\natexlab{5}})Wu, Zhang, Bao, and Zhao]{wu2025sc4anm}
Wenqing Wu, Chengzhi Zhang, Tong Bao, and Yi~Zhao.
\newblock Sc4anm: Identifying optimal section combinations for automated novelty prediction in academic papers.
\newblock \emph{Expert Systems with Applications}, page 126778, May 2025{\natexlab{5}}.

\bibitem[Wu et~al.(2025{\natexlab{6}})Wu, Yang, Chai, Zhang, Liu, Du, Liang, Shu, Cheng, Sun, et~al.]{wu2025tablebench}
Xianjie Wu, Jian Yang, Linzheng Chai, Ge~Zhang, Jiaheng Liu, Xeron Du, Di~Liang, Daixin Shu, Xianfu Cheng, Tianzhen Sun, et~al.
\newblock Tablebench: A comprehensive and complex benchmark for table question answering.
\newblock In \emph{Proceedings of the AAAI Conference on Artificial Intelligence}, volume~39, pages 25497--25506, Apr 2025{\natexlab{6}}.

\bibitem[Wu et~al.(2024{\natexlab{2}})Wu, Zhang, Zhou, Wu, Sunchu, Hu, Chen, Zhao, Liu, Sun, et~al.]{wu2024deepcre}
Yushuai Wu, Ting Zhang, Hao Zhou, Hainan Wu, Hanwen Sunchu, Lei Hu, Xiaofang Chen, Suyuan Zhao, Gaochao Liu, Chao Sun, et~al.
\newblock Deepcre: Transforming drug r\&d via ai-driven cross-drug response evaluation.
\newblock \emph{arXiv preprint arXiv:2403.03768}, 2024{\natexlab{2}}.

\bibitem[W{\"u}hrl et~al.(2024)W{\"u}hrl, Resendiz, Grimminger, and Klinger]{wuhrl2024makes}
Amelie W{\"u}hrl, Yarik~Menchaca Resendiz, Lara Grimminger, and Roman Klinger.
\newblock What makes medical claims (un) verifiable? analyzing entity and relation properties for fact verification.
\newblock \emph{arXiv preprint arXiv:2402.01360}, 2024.

\bibitem[W{\"u}hrl et~al.(2024{\natexlab{2}})W{\"u}hrl, Wright, Klinger, and Augenstein]{wuhrl2024understanding}
Amelie W{\"u}hrl, Dustin Wright, Roman Klinger, and Isabelle Augenstein.
\newblock Understanding fine-grained distortions in reports of scientific findings.
\newblock \emph{arXiv preprint arXiv:2402.12431}, 2024{\natexlab{2}}.

\bibitem[X.AI(2024)]{xai2024grok2}
X.AI.
\newblock Grok-2 beta release.
\newblock \url{https://x.ai/blog/grok-2}, Dec 2024.
\newblock Accessed: 2024-12-16.

\bibitem[Xia et~al.(2024)Xia, Zhang, Ye, Yan, Liu, Zhou, Chen, Ye, Dou, Shi, et~al.]{xia2024chartx}
Renqiu Xia, Bo~Zhang, Hancheng Ye, Xiangchao Yan, Qi~Liu, Hongbin Zhou, Zijun Chen, Peng Ye, Min Dou, Botian Shi, et~al.
\newblock Chartx \& chartvlm: A versatile benchmark and foundation model for complicated chart reasoning.
\newblock \emph{arXiv preprint arXiv:2402.12185}, 2024.

\bibitem[Xia et~al.(2025)Xia, Qu, Zheng, Tang, Zhuang, Liang, Wang, Wu, Sun, Zimmermann, et~al.]{xia2025reimagining}
Yutong Xia, Ao~Qu, Yunhan Zheng, Yihong Tang, Dingyi Zhuang, Yuxuan Liang, Shenhao Wang, Cathy Wu, Lijun Sun, Roger Zimmermann, et~al.
\newblock Reimagining urban science: Scaling causal inference with large language models.
\newblock \emph{arXiv preprint arXiv:2504.12345}, 2025.

\bibitem[Xiang et~al.(2025)Xiang, Yan, Ouyang, Gui, and He]{xiang2025scireplicate}
Yanzheng Xiang, Hanqi Yan, Shuyin Ouyang, Lin Gui, and Yulan He.
\newblock Scireplicate-bench: Benchmarking llms in agent-driven algorithmic reproduction from research papers.
\newblock \emph{arXiv preprint arXiv:2504.00255}, 2025.

\bibitem[Xiao et~al.(2024)Xiao, Liu, Zheng, Xie, Hao, Li, Wang, Ni, Li, Luo, et~al.]{xiao2024cellagent}
Yihang Xiao, Jinyi Liu, Yan Zheng, Xiaohan Xie, Jianye Hao, Mingzhi Li, Ruitao Wang, Fei Ni, Yuxiao Li, Jintian Luo, et~al.
\newblock Cellagent: An llm-driven multi-agent framework for automated single-cell data analysis.
\newblock \emph{arXiv preprint arXiv:2407.09811}, 2024.

\bibitem[Xin et~al.(2024)Xin, Guo, Shao, Ren, Zhu, Liu, Ruan, Li, and Liang]{xin2024deepseek}
Huajian Xin, Daya Guo, Zhihong Shao, Zhizhou Ren, Qihao Zhu, Bo~Liu, Chong Ruan, Wenda Li, and Xiaodan Liang.
\newblock Deepseek-prover: Advancing theorem proving in llms through large-scale synthetic data.
\newblock \emph{arXiv preprint arXiv:2405.14333}, 2024.

\bibitem[Xing et~al.(2024)Xing, He, Zhou, Dong, Han, Zhang, and Chaudhuri]{xing2024table}
Junjie Xing, Yeye He, Mengyu Zhou, Haoyu Dong, Shi Han, Dongmei Zhang, and Surajit Chaudhuri.
\newblock Table-llm-specialist: Language model specialists for tables using iterative generator-validator fine-tuning.
\newblock \emph{arXiv preprint arXiv:2410.12164}, 2024.

\bibitem[Xiong et~al.(2025)Xiong, Xie, Williams, Kim, Shariatmadari, Guo, Bekiranov, and Zhang]{xiong2025toward}
Guangzhi Xiong, Eric Xie, Corey Williams, Myles Kim, Amir~Hassan Shariatmadari, Sikun Guo, Stefan Bekiranov, and Aidong Zhang.
\newblock Toward reliable biomedical hypothesis generation: Evaluating truthfulness and hallucination in large language models.
\newblock \emph{arXiv preprint arXiv:2505.14599}, 2025.

\bibitem[Xiong et~al.(2025{\natexlab{2}})Xiong, Chen, Khizbullin, Zhuge, and Schmidhuber]{xiong2025beyond}
Ruibin Xiong, Yimeng Chen, Dmitrii Khizbullin, Mingchen Zhuge, and J{\"u}rgen Schmidhuber.
\newblock Beyond outlining: Heterogeneous recursive planning for adaptive long-form writing with language models.
\newblock \emph{arXiv preprint arXiv:2503.08275}, 2025{\natexlab{2}}.

\bibitem[Xu et~al.(2025)Xu, Zhao, Zhou, Yue, Fei, Ling, Zhang, and Bai]{xu2025earthse}
Wanghan Xu, Xiangyu Zhao, Yuhao Zhou, Xiaoyu Yue, Ben Fei, Fenghua Ling, Wenlong Zhang, and Lei Bai.
\newblock Earthse: A benchmark evaluating earth scientific exploration capability for large language models.
\newblock \emph{arXiv preprint arXiv:2505.17139}, 2025.

\bibitem[Xu et~al.(2023)Xu, Shi, Hu, Yu, Li, Zhang, and Wu]{xu2023towards}
Zhenran Xu, Senbao Shi, Baotian Hu, Jindi Yu, Dongfang Li, Min Zhang, and Yuxiang Wu.
\newblock Towards reasoning in large language models via multi-agent peer review collaboration.
\newblock \emph{arXiv preprint arXiv:2311.08152}, 2023.

\bibitem[Xu(2025)]{xu2025patterns}
Ziyang Xu.
\newblock Patterns and purposes: A cross-journal analysis of ai tool usage in academic writing.
\newblock \emph{arXiv preprint arXiv:2502.00632}, 2025.

\bibitem[Yager(2024)]{yager2024towards}
Kevin~G Yager.
\newblock Towards a science exocortex.
\newblock \emph{Digital Discovery}, 3\penalty0 (10):\penalty0 1933--1957, Aug 2024.

\bibitem[Yamada et~al.(2025)Yamada, Lange, Lu, Hu, Lu, Foerster, Clune, and Ha]{yamada2025ai}
Yutaro Yamada, Robert~Tjarko Lange, Cong Lu, Shengran Hu, Chris Lu, Jakob Foerster, Jeff Clune, and David Ha.
\newblock The ai scientist-v2: Workshop-level automated scientific discovery via agentic tree search.
\newblock \emph{arXiv preprint arXiv:2504.08066}, 2025.

\bibitem[Yan et~al.(2025)Yan, Feng, Yuan, Xia, Wang, Zhang, and Bai]{yan2025surveyforge}
Xiangchao Yan, Shiyang Feng, Jiakang Yuan, Renqiu Xia, Bin Wang, Bo~Zhang, and Lei Bai.
\newblock Surveyforge: On the outline heuristics, memory-driven generation, and multi-dimensional evaluation for automated survey writing.
\newblock \emph{arXiv preprint arXiv:2503.04629}, 2025.

\bibitem[Yan et~al.(2025{\natexlab{2}})Yan, Wang, Huo, Ye, Chu, Hu, Yu, Gomes, Selman, and Wen]{yan2025position}
Yibo Yan, Shen Wang, Jiahao Huo, Jingheng Ye, Zhendong Chu, Xuming Hu, Philip~S Yu, Carla Gomes, Bart Selman, and Qingsong Wen.
\newblock Position: Multimodal large language models can significantly advance scientific reasoning.
\newblock \emph{arXiv preprint arXiv:2502.02871}, 2025{\natexlab{2}}.

\bibitem[Yang et~al.(2024)Yang, Yang, Zhang, Hui, Zheng, Yu, Li, Liu, Huang, Wei, Lin, Yang, Tu, Zhang, Yang, Yang, Zhou, Lin, Dang, Lu, Bao, Yang, Yu, Li, Xue, Zhang, Zhu, Men, Lin, Li, Tang, Xia, Ren, Ren, Fan, Su, Zhang, Wan, Liu, Cui, Zhang, and Qiu]{qwen25}
An~Yang, Baosong Yang, Beichen Zhang, Binyuan Hui, Bo~Zheng, Bowen Yu, Chengyuan Li, Dayiheng Liu, Fei Huang, Haoran Wei, Huan Lin, Jian Yang, Jianhong Tu, Jianwei Zhang, Jianxin Yang, Jiaxi Yang, Jingren Zhou, Junyang Lin, Kai Dang, Keming Lu, Keqin Bao, Kexin Yang, Le~Yu, Mei Li, Mingfeng Xue, Pei Zhang, Qin Zhu, Rui Men, Runji Lin, Tianhao Li, Tianyi Tang, Tingyu Xia, Xingzhang Ren, Xuancheng Ren, Yang Fan, Yang Su, Yichang Zhang, Yu~Wan, Yuqiong Liu, Zeyu Cui, Zhenru Zhang, and Zihan Qiu.
\newblock Qwen2.5 technical report.
\newblock \emph{arXiv preprint arXiv:2412.15115}, 2024.

\bibitem[Yang et~al.(2025)Yang, Khosravi, Walt, Krishnan, and Sarkar]{yang2025zero}
Hsin-Jung Yang, Mahsa Khosravi, Benjamin Walt, Girish Krishnan, and Soumik Sarkar.
\newblock Zero-shot sim-to-real transfer for reinforcement learning-based visual servoing of soft continuum arms.
\newblock \emph{arXiv preprint arXiv:2504.16916}, 2025.

\bibitem[Yang et~al.(2025{\natexlab{2}})Yang, Bhat, Hu, Cao, Dehmamy, Walters, and Yu]{yang2025discovering}
Jianke Yang, Manu Bhat, Bryan Hu, Yadi Cao, Nima Dehmamy, Robin Walters, and Rose Yu.
\newblock Discovering symbolic differential equations with symmetry invariants.
\newblock \emph{arXiv preprint arXiv:2505.12083}, 2025{\natexlab{2}}.

\bibitem[Yang and Hwang(2025)]{yang2025transforming}
John~Jeongseok Yang and Sang-Hyun Hwang.
\newblock Transforming hematological research documentation with large language models: an approach to scientific writing and data analysis.
\newblock \emph{Blood research}, 60\penalty0 (1):\penalty0 1--11, Mar 2025.

\bibitem[Yang et~al.(2024{\natexlab{2}})Yang, Xu, Sellergren, Kohlberger, Zhou, Ktena, Kiraly, Ahmed, Hormozdiari, Jaroensri, et~al.]{yang2024advancing}
Lin Yang, Shawn Xu, Andrew Sellergren, Timo Kohlberger, Yuchen Zhou, Ira Ktena, Atilla Kiraly, Faruk Ahmed, Farhad Hormozdiari, Tiam Jaroensri, et~al.
\newblock Advancing multimodal medical capabilities of gemini.
\newblock \emph{arXiv preprint arXiv:2405.03162}, 2024{\natexlab{2}}.

\bibitem[Yang et~al.(2024{\natexlab{3}})Yang, Yang, Feng, Ouyang, Blum, She, Jiang, Lecue, Lu, and Li]{yang2024graphusion}
Rui Yang, Boming Yang, Aosong Feng, Sixun Ouyang, Moritz Blum, Tianwei She, Yuang Jiang, Freddy Lecue, Jinghui Lu, and Irene Li.
\newblock Graphusion: a rag framework for knowledge graph construction with a global perspective.
\newblock \emph{arXiv preprint arXiv:2410.17600}, 2024{\natexlab{3}}.

\bibitem[Yang et~al.(2024{\natexlab{4}})Yang, Shi, Li, Jiang, Yang, Lu, and Zhao]{yang2024batgpt}
Yifei Yang, Runhan Shi, Zuchao Li, Shu Jiang, Yang Yang, Bao-Liang Lu, and Hai Zhao.
\newblock Batgpt-chem: A foundation large model for chemical engineering.
\newblock Apr 2024{\natexlab{4}}.

\bibitem[Yang et~al.(2025{\natexlab{3}})Yang, Chen, Liu, Hu, Chen, Zhang, Hu, Zhai, Qiao, Wang, et~al.]{yang2025truly}
Yue Yang, MingKang Chen, Qihua Liu, Mengkang Hu, Qiguang Chen, Gengrui Zhang, Shuyue Hu, Guangtao Zhai, Yu~Qiao, Yu~Wang, et~al.
\newblock Truly assessing fluid intelligence of large language models through dynamic reasoning evaluation.
\newblock \emph{arXiv preprint arXiv:2506.02648}, 2025{\natexlab{3}}.

\bibitem[Yang et~al.(2025{\natexlab{4}})Yang, Pan, Wang, Wang, Liu, Zhu, Zhang, and Chen]{yang2025multimodal}
Zhaorui Yang, Bo~Pan, Han Wang, Yiyao Wang, Xingyu Liu, Minfeng Zhu, Bo~Zhang, and Wei Chen.
\newblock Multimodal deepresearcher: Generating text-chart interleaved reports from scratch with agentic framework.
\newblock \emph{arXiv preprint arXiv:2506.02454}, 2025{\natexlab{4}}.

\bibitem[Yang et~al.(2023)Yang, Du, Li, Zheng, Poria, and Cambria]{yang2023large}
Zonglin Yang, Xinya Du, Junxian Li, Jie Zheng, Soujanya Poria, and Erik Cambria.
\newblock Large language models for automated open-domain scientific hypotheses discovery.
\newblock \emph{arXiv preprint arXiv:2309.02726}, 2023.

\bibitem[Yang et~al.(2025{\natexlab{5}})Yang, Liu, Gao, Xie, Li, Ouyang, Poria, Cambria, and Zhou]{yanglarge}
Zonglin Yang, Wanhao Liu, Ben Gao, Tong Xie, Yuqiang Li, Wanli Ouyang, Soujanya Poria, Erik Cambria, and Dongzhan Zhou.
\newblock {MOOSE}-chem: Large language models for rediscovering unseen chemistry scientific hypotheses.
\newblock In \emph{The Thirteenth International Conference on Learning Representations}, Jan 2025{\natexlab{5}}.
\newblock URL \url{https://openreview.net/forum?id=X9OfMNNepI}.

\bibitem[Yang et~al.(2025{\natexlab{6}})Yang, Zhu, and Zhu]{yang2025curiousllm}
Zukang Yang, Zixuan Zhu, and Jennifer Zhu.
\newblock Curiousllm: Elevating multi-document question answering with llm-enhanced knowledge graph reasoning.
\newblock In \emph{Proceedings of the 2025 Conference of the Nations of the Americas Chapter of the Association for Computational Linguistics: Human Language Technologies (Volume 3: Industry Track)}, pages 274--286, Jan 2025{\natexlab{6}}.

\bibitem[Yax et~al.(2024)Yax, Anll{\'o}, and Palminteri]{yax2024studying}
Nicolas Yax, Hern{\'a}n Anll{\'o}, and Stefano Palminteri.
\newblock Studying and improving reasoning in humans and machines.
\newblock \emph{Communications Psychology}, 2\penalty0 (1):\penalty0 51, Jun 2024.

\bibitem[Ye et~al.(2023)Ye, Fang, Du, Yuen, and Tao]{ye2023heterogeneous}
Mang Ye, Xiuwen Fang, Bo~Du, Pong~C Yuen, and Dacheng Tao.
\newblock Heterogeneous federated learning: State-of-the-art and research challenges.
\newblock \emph{ACM Computing Surveys}, 56\penalty0 (3):\penalty0 1--44, Oct 2023.

\bibitem[Ye et~al.(2024)Ye, Pang, Chai, Chen, Yin, Xiang, Dong, Shao, and Chen]{ye2024we}
Rui Ye, Xianghe Pang, Jingyi Chai, Jiaao Chen, Zhenfei Yin, Zhen Xiang, Xiaowen Dong, Jing Shao, and Siheng Chen.
\newblock Are we there yet? revealing the risks of utilizing large language models in scholarly peer review.
\newblock \emph{arXiv preprint arXiv:2412.01708}, 2024.

\bibitem[Ye et~al.(2024{\natexlab{2}})Ye, Ren, Wang, Wan, Razzak, Hoex, Wang, Xie, and Zhang]{ye2024construction}
Yanpeng Ye, Jie Ren, Shaozhou Wang, Yuwei Wan, Imran Razzak, Bram Hoex, Haofen Wang, Tong Xie, and Wenjie Zhang.
\newblock Construction and application of materials knowledge graph in multidisciplinary materials science via large language model.
\newblock \emph{Advances in Neural Information Processing Systems}, 37:\penalty0 56878--56897, Dec 2024{\natexlab{2}}.

\bibitem[Yin et~al.(2025)Yin, Hsu, Min, Kim, Rossi, Yu, Jung, Huang, et~al.]{yin2025understanding}
Ho~Yin, Ting-Yao Hsu, Jiyoo Min, Sungchul Kim, Ryan~A Rossi, Tong Yu, Hyunggu Jung, Ting-Hao'Kenneth' Huang, et~al.
\newblock Understanding how paper writers use ai-generated captions in figure caption writing.
\newblock \emph{arXiv preprint arXiv:2501.06317}, 2025.

\bibitem[Yin et~al.(2025{\natexlab{2}})Yin, Qu, Liu, Yang, Cong, and Wang]{yin2025genome}
Ming Yin, Yuanhao Qu, Dyllan Liu, Ling Yang, Le~Cong, and Mengdi Wang.
\newblock Genome-bench: A scientific reasoning benchmark from real-world expert discussions.
\newblock \emph{bioRxiv}, pages 2025--06, 2025{\natexlab{2}}.

\bibitem[Yu et~al.(2025)Yu, Zhang, Liu, Ding, Sun, and Jin]{yu2025frame}
Chengzhang Yu, Yiming Zhang, Zhixin Liu, Zenghui Ding, Yining Sun, and Zhanpeng Jin.
\newblock Frame: Feedback-refined agent methodology for enhancing medical research insights.
\newblock \emph{arXiv preprint arXiv:2505.04649}, 2025.

\bibitem[Yu et~al.(2024)Yu, Zhang, Tiwari, and Wang]{yu2024natural}
Fei Yu, Hongbo Zhang, Prayag Tiwari, and Benyou Wang.
\newblock Natural language reasoning, a survey.
\newblock \emph{ACM Comput. Surv.}, 56\penalty0 (12), October 2024.
\newblock ISSN 0360-0300.
\newblock \doi{10.1145/3664194}.
\newblock URL \url{https://doi.org/10.1145/3664194}.

\bibitem[Yu et~al.(2024{\natexlab{2}})Yu, Hong, Cheng, Zhu, Xuan, Yao, Feng, and You]{yu2024researchtown}
Haofei Yu, Zhaochen Hong, Zirui Cheng, Kunlun Zhu, Keyang Xuan, Jinwei Yao, Tao Feng, and Jiaxuan You.
\newblock Researchtown: Simulator of human research community.
\newblock \emph{arXiv preprint arXiv:2412.17767}, 2024{\natexlab{2}}.

\bibitem[Yu and Jin(2025)]{yu2025unlocking}
Hengjie Yu and Yaochu Jin.
\newblock Unlocking the potential of ai researchers in scientific discovery: What is missing?
\newblock \emph{arXiv preprint arXiv:2503.05822}, 2025.

\bibitem[Yu et~al.(2024{\natexlab{3}})Yu, Tan, Ding, Zhu, Li, Cheng, Cui, Lan, and Li]{yu2024seagraph}
Jianxiang Yu, Jiaqi Tan, Zichen Ding, Jiapeng Zhu, Jiahao Li, Yao Cheng, Qier Cui, Yunshi Lan, and Xiang Li.
\newblock Seagraph: Unveiling the whole story of paper review comments.
\newblock \emph{arXiv preprint arXiv:2412.11939}, 2024{\natexlab{3}}.

\bibitem[Yu et~al.(2025{\natexlab{2}})Yu, Tang, Feng, Rao, Liang, Zhang, Sun, Zhang, Zhang, Ding, et~al.]{yu2025scicueval}
Jing Yu, Yuqi Tang, Kehua Feng, Mingyang Rao, Lei Liang, Zhiqiang Zhang, Mengshu Sun, Wen Zhang, Qiang Zhang, Keyan Ding, et~al.
\newblock Scicueval: A comprehensive dataset for evaluating scientific context understanding in large language models.
\newblock \emph{arXiv preprint arXiv:2505.15094}, 2025{\natexlab{2}}.

\bibitem[Yu et~al.(2024{\natexlab{4}})Yu, Zhang, Shi, Lao, and Xiao]{yu2024reinforced}
Luyao Yu, Qi~Zhang, Chongyang Shi, An~Lao, and Liang Xiao.
\newblock Reinforced subject-aware graph neural network for related work generation.
\newblock In \emph{International Conference on Knowledge Science, Engineering and Management}, pages 201--213. Springer, Jul 2024{\natexlab{4}}.

\bibitem[Yu et~al.(2024{\natexlab{5}})Yu, Luo, Madasu, Lal, and Howard]{yu2024your}
Sungduk Yu, Man Luo, Avinash Madasu, Vasudev Lal, and Phillip Howard.
\newblock Is your paper being reviewed by an llm? investigating ai text detectability in peer review.
\newblock \emph{arXiv preprint arXiv:2410.03019}, 2024{\natexlab{5}}.

\bibitem[Yuan et~al.(2025)Yuan, Yan, Shi, Chen, Ouyang, Zhang, Bai, Qiao, and Zhou]{yuan2025dolphin}
Jiakang Yuan, Xiangchao Yan, Botian Shi, Tao Chen, Wanli Ouyang, Bo~Zhang, Lei Bai, Yu~Qiao, and Bowen Zhou.
\newblock Dolphin: Closed-loop open-ended auto-research through thinking, practice, and feedback.
\newblock \emph{arXiv preprint arXiv:2501.03916}, 2025.

\bibitem[Yuan and F{\"a}rber(2025)]{yuan2025hallucinations}
Shuzhou Yuan and Michael F{\"a}rber.
\newblock Hallucinations can improve large language models in drug discovery.
\newblock \emph{arXiv preprint arXiv:2501.13824}, 2025.

\bibitem[Yuan and Liu(2022)]{yuan2022kid}
Weizhe Yuan and Pengfei Liu.
\newblock Kid-review: knowledge-guided scientific review generation with oracle pre-training.
\newblock In \emph{Proceedings of the AAAI Conference on Artificial Intelligence}, volume~36, pages 11639--11647, Jun 2022.

\bibitem[Yuan et~al.(2022)Yuan, Liu, and Neubig]{yuan2022can}
Weizhe Yuan, Pengfei Liu, and Graham Neubig.
\newblock Can we automate scientific reviewing?
\newblock \emph{Journal of Artificial Intelligence Research}, 75:\penalty0 171--212, Sep 2022.

\bibitem[Yuan et~al.(2025{\natexlab{2}})Yuan, Chen, Wang, and You]{yuan2025empowering}
Wenhao Yuan, Guangyao Chen, Zhilong Wang, and Fengqi You.
\newblock Empowering generalist material intelligence with large language models.
\newblock \emph{Advanced Materials}, page 2502771, May 2025{\natexlab{2}}.

\bibitem[Yue et~al.(2025)Yue, Somasekharan, Cao, and Pan]{yue2025foam}
Ling Yue, Nithin Somasekharan, Yadi Cao, and Shaowu Pan.
\newblock Foam-agent: Towards automated intelligent cfd workflows.
\newblock \emph{arXiv preprint arXiv:2505.04997}, 2025.

\bibitem[Zabaleta and Lehman(2024)]{zabaleta2024simulating}
Miguel Zabaleta and Joel Lehman.
\newblock Simulating tabular datasets through llms to rapidly explore hypotheses about real-world entities.
\newblock \emph{arXiv preprint arXiv:2411.18071}, 2024.

\bibitem[Zamudio et~al.(2024)Zamudio, Grigsby, and Michelsen]{zamudio2024raise}
C{\'e}sar Zamudio, Jamie~L Grigsby, and Meg Michelsen.
\newblock Raise: A new method to develop experimental stimuli for advertising research with image generative artificial intelligence.
\newblock \emph{Journal of Advertising}, pages 1--16, Jan 2024.

\bibitem[Zeng and Gao(2023)]{zeng2023prompt}
Fengzhu Zeng and Wei Gao.
\newblock Prompt to be consistent is better than self-consistent? few-shot and zero-shot fact verification with pre-trained language models.
\newblock \emph{arXiv preprint arXiv:2306.02569}, 2023.

\bibitem[Zeng et~al.(2023)Zeng, Sidhu, Chan, Wang, and Ji]{zeng2023meta}
Qi~Zeng, Mankeerat Sidhu, Hou~Pong Chan, Lu~Wang, and Heng Ji.
\newblock Meta-review generation with checklist-guided iterative introspection.
\newblock \emph{arXiv preprint arXiv:2305.14647}, 2023.

\bibitem[Zeng et~al.(2024)Zeng, Sidhu, Blume, Chan, Wang, and Ji]{zeng2024scientific}
Qi~Zeng, Mankeerat Sidhu, Ansel Blume, Hou~Pong Chan, Lu~Wang, and Heng Ji.
\newblock Scientific opinion summarization: Paper meta-review generation dataset, methods, and evaluation.
\newblock In \emph{Artificial Intelligence for Research and Democracy: First International Workshop, AI4Research 2024, and 4th International Workshop, DemocrAI 2024, Held in Conjunction with IJCAI 2024, Jeju, South Korea, August 5, 2024, Proceedings}, page~20. Springer Nature, Jun 2024.

\bibitem[Zeng et~al.(2024{\natexlab{2}})Zeng, Brown, Raymond, Byari, Hotz, and Rounsevell]{zeng2024exploring}
Yongchao Zeng, Calum Brown, Joanna Raymond, Mohamed Byari, Ronja Hotz, and Mark Rounsevell.
\newblock Exploring the opportunities and challenges of using large language models to represent institutional agency in land system modelling.
\newblock \emph{EGUsphere}, 2024:\penalty0 1--35, Mar 2024{\natexlab{2}}.

\bibitem[Zhang et~al.(2021)Zhang, Xie, Bai, Yu, Li, and Gao]{zhang2021survey}
Chen Zhang, Yu~Xie, Hang Bai, Bin Yu, Weihong Li, and Yuan Gao.
\newblock A survey on federated learning.
\newblock \emph{Knowledge-Based Systems}, 216:\penalty0 106775, Mar 2021.

\bibitem[Zhang et~al.(2024)Zhang, Hu, Zhoubian, Du, Yang, Wang, Yue, Dong, and Tang]{zhang2024sciglm}
Dan Zhang, Ziniu Hu, Sining Zhoubian, Zhengxiao Du, Kaiyu Yang, Zihan Wang, Yisong Yue, Yuxiao Dong, and Jie Tang.
\newblock Sciglm: Training scientific language models with self-reflective instruction annotation and tuning.
\newblock \emph{arXiv preprint arXiv:2401.07950}, 2024.

\bibitem[Zhang et~al.(2025)Zhang, Bao, Du, Zhao, Zhang, Bao, and Yang]{zhang2025re}
Daoze Zhang, Zhijian Bao, Sihang Du, Zhiyi Zhao, Kuangling Zhang, Dezheng Bao, and Yang Yang.
\newblock Re $^2$: A consistency-ensured dataset for full-stage peer review and multi-turn rebuttal discussions.
\newblock \emph{arXiv preprint arXiv:2505.07920}, 2025.

\bibitem[Zhang et~al.(2024{\natexlab{2}})Zhang, Shi, Zhu, Chen, Cen, Yu, Chen, Wang, Zhao, Cheng, Han, An, Zhang, Tam, Cao, Pang, Guan, Yuan, Song, Li, Dong, and Tang]{zhang2024oag}
Fanjin Zhang, Shijie Shi, Yifan Zhu, Bo~Chen, Yukuo Cen, Jifan Yu, Yelin Chen, Lulu Wang, Qingfei Zhao, Yuqing Cheng, Tianyi Han, Yuwei An, Dan Zhang, Weng~Lam Tam, Kun Cao, Yunhe Pang, Xinyu Guan, Huihui Yuan, Jian Song, Xiaoyan Li, Yuxiao Dong, and Jie Tang.
\newblock Oag-bench: A human-curated benchmark for academic graph mining.
\newblock \emph{arXiv preprint arXiv:2402.15810}, 2024{\natexlab{2}}.

\bibitem[Zhang et~al.(2024{\natexlab{3}})Zhang, Yu, Li, Yu, Long, Jin, and Li]{guozhen2024human}
Guozhen Zhang, Zihan Yu, Nian Li, Fudan Yu, Qingyue Long, Depeng Jin, and Yong Li.
\newblock Human behavior simulation: Objectives, methodologies, and open problems.
\newblock \emph{arXiv preprint arXiv:2412.07788}, 2024{\natexlab{3}}.

\bibitem[Zhang et~al.(2023)Zhang, Xu, Zhang, Liu, Hooi, and Deng]{zhang2023exploring}
Jintian Zhang, Xin Xu, Ningyu Zhang, Ruibo Liu, Bryan Hooi, and Shumin Deng.
\newblock Exploring collaboration mechanisms for llm agents: A social psychology view.
\newblock \emph{arXiv preprint arXiv:2310.02124}, 2023.

\bibitem[Zhang et~al.(2023{\natexlab{2}})Zhang, Zhang, Ren, Li, and Yang]{zhang2023mlcopilot}
Lei Zhang, Yuge Zhang, Kan Ren, Dongsheng Li, and Yuqing Yang.
\newblock Mlcopilot: Unleashing the power of large language models in solving machine learning tasks.
\newblock \emph{arXiv preprint arXiv:2304.14979}, 2023{\natexlab{2}}.

\bibitem[Zhang et~al.(2024{\natexlab{4}})Zhang, Eger, Cheng, Zhai, Belouadi, Leiter, Ponzetto, Moafian, and Zhao]{zhang2024scimage}
Leixin Zhang, Steffen Eger, Yinjie Cheng, Weihe Zhai, Jonas Belouadi, Christoph Leiter, Simone~Paolo Ponzetto, Fahimeh Moafian, and Zhixue Zhao.
\newblock Scimage: How good are multimodal large language models at scientific text-to-image generation?
\newblock \emph{arXiv preprint arXiv:2412.02368}, 2024{\natexlab{4}}.

\bibitem[Zhang and Choi(2023)]{zhang2023clarify}
Michael~JQ Zhang and Eunsol Choi.
\newblock Clarify when necessary: Resolving ambiguity through interaction with lms.
\newblock \emph{arXiv preprint arXiv:2311.09469}, 2023.

\bibitem[Zhang et~al.(2025{\natexlab{2}})Zhang, Shen, Li, Sha, Hu, Wang, Huang, Liu, Tong, Jiang, et~al.]{zhang2025llmeval}
Ming Zhang, Yujiong Shen, Zelin Li, Huayu Sha, Binze Hu, Yuhui Wang, Chenhao Huang, Shichun Liu, Jingqi Tong, Changhao Jiang, et~al.
\newblock Llmeval-med: A real-world clinical benchmark for medical llms with physician validation.
\newblock \emph{arXiv preprint arXiv:2506.04078}, 2025{\natexlab{2}}.

\bibitem[Zhang et~al.(2023{\natexlab{3}})Zhang, Gong, Wu, Liu, and Zhou]{zhang2023automl}
Shujian Zhang, Chengyue Gong, Lemeng Wu, Xingchao Liu, and Mingyuan Zhou.
\newblock Automl-gpt: Automatic machine learning with gpt.
\newblock \emph{arXiv preprint arXiv:2305.02499}, 2023{\natexlab{3}}.

\bibitem[Zhang et~al.(2025{\natexlab{3}})Zhang, Liu, Xin, and Jiao]{zhang2025mooseagent}
Tao Zhang, Zhenhai Liu, Yong Xin, and Yongjun Jiao.
\newblock Mooseagent: A llm based multi-agent framework for automating moose simulation.
\newblock \emph{arXiv preprint arXiv:2504.08621}, 2025{\natexlab{3}}.

\bibitem[Zhang and Abernethy(2025)]{zhang2025reviewing}
Tianmai~M Zhang and Neil~F Abernethy.
\newblock Reviewing scientific papers for critical problems with reasoning llms: Baseline approaches and automatic evaluation.
\newblock \emph{arXiv preprint arXiv:2505.23824}, 2025.

\bibitem[Zhang et~al.(2024{\natexlab{5}})Zhang, Wang, Lu, Ye, Wang, Bao, Yan, and Su]{zhang2024augmenting}
Xiaocheng Zhang, Xi~Wang, Yifei Lu, Zhuangzhuang Ye, Jianing Wang, Mengjiao Bao, Peng Yan, and Xiaohong Su.
\newblock Augmenting the veracity and explanations of complex fact checking via iterative self-revision with llms.
\newblock \emph{arXiv preprint arXiv:2410.15135}, 2024{\natexlab{5}}.

\bibitem[Zhang et~al.(2024{\natexlab{6}})Zhang, Xie, Huang, Ma, Pan, Liu, Xiong, Ergen, Shim, Lee, et~al.]{zhang2024massw}
Xingjian Zhang, Yutong Xie, Jin Huang, Jinge Ma, Zhaoying Pan, Qijia Liu, Ziyang Xiong, Tolga Ergen, Dongsub Shim, Honglak Lee, et~al.
\newblock Massw: A new dataset and benchmark tasks for ai-assisted scientific workflows.
\newblock \emph{arXiv preprint arXiv:2406.06357}, 2024{\natexlab{6}}.

\bibitem[Zhang et~al.()Zhang, Ni, Fanady, Chen, Chen, Wang, Chen, He, Bai, and Zhao]{zhang5127472chatgpt}
Xinyu Zhang, Zongming Ni, Billy Fanady, Feibei Chen, Zijian Chen, Zixuan Wang, Guofei Chen, Zhengda He, Yang Bai, and Haitao Zhao.
\newblock Chatgpt-assisted rational design for iterative performance optimization of perovskite solar cells.
\newblock \emph{Available at SSRN 5127472}, Feb .

\bibitem[Zhang and Gao(2023)]{zhang2023towards}
Xuan Zhang and Wei Gao.
\newblock Towards llm-based fact verification on news claims with a hierarchical step-by-step prompting method.
\newblock \emph{arXiv preprint arXiv:2310.00305}, 2023.

\bibitem[Zhang et~al.(2025{\natexlab{4}})Zhang, Wang, Dou, Zhu, and Che]{zhang2025survey}
Xuanliang Zhang, Dingzirui Wang, Longxu Dou, Qingfu Zhu, and Wanxiang Che.
\newblock A survey of table reasoning with large language models.
\newblock \emph{Frontiers of Computer Science}, 19\penalty0 (9):\penalty0 199348, Jan 2025{\natexlab{4}}.

\bibitem[Zhang et~al.(2025{\natexlab{5}})Zhang, Khan, Mahmud, Yang, Lavin, Levin, Frey, Dunnmon, Evans, Bundy, et~al.]{zhang2025advancing}
Yanbo Zhang, Sumeer~A Khan, Adnan Mahmud, Huck Yang, Alexander Lavin, Michael Levin, Jeremy Frey, Jared Dunnmon, James Evans, Alan Bundy, et~al.
\newblock Advancing the scientific method with large language models: From hypothesis to discovery.
\newblock \emph{arXiv preprint arXiv:2505.16477}, 2025{\natexlab{5}}.

\bibitem[Zhang et~al.(2023{\natexlab{4}})Zhang, Wang, Wang, Sheng, Yao, Mahmood, Zhang, and Zhao]{zhang2023large}
Yang Zhang, Yufei Wang, Kai Wang, Quan~Z Sheng, Lina Yao, Adnan Mahmood, Wei~Emma Zhang, and Rongying Zhao.
\newblock When large language models meet citation: A survey.
\newblock \emph{arXiv preprint arXiv:2309.09727}, 2023{\natexlab{4}}.

\bibitem[Zhang et~al.(2024{\natexlab{7}})Zhang, Chen, Li, Che, and Qin]{zhang2024autocap}
Yongheng Zhang, Qiguang Chen, Min Li, Wanxiang Che, and Libo Qin.
\newblock Autocap: Towards automatic cross-lingual alignment planning for zero-shot chain-of-thought.
\newblock \emph{arXiv preprint arXiv:2406.13940}, 2024{\natexlab{7}}.

\bibitem[Zhang et~al.(2024{\natexlab{8}})Zhang, Chen, Zhou, Wang, Si, Wang, Lu, and Qin]{zhang2024wrong}
Yongheng Zhang, Qiguang Chen, Jingxuan Zhou, Peng Wang, Jiasheng Si, Jin Wang, Wenpeng Lu, and Libo Qin.
\newblock Wrong-of-thought: An integrated reasoning framework with multi-perspective verification and wrong information.
\newblock \emph{arXiv preprint arXiv:2410.04463}, 2024{\natexlab{8}}.

\bibitem[Zhang et~al.(2025{\natexlab{6}})Zhang, Liu, Zhou, Chen, Fei, Lu, and Qin]{zhang2025cchall}
Yongheng Zhang, Xu~Liu, Ruoxi Zhou, Qiguang Chen, Hao Fei, Wenpeng Lu, and Libo Qin.
\newblock Cchall: A novel benchmark for joint cross-lingual and cross-modal hallucinations detection in large language models.
\newblock \emph{arXiv preprint arXiv:2505.19108}, 2025{\natexlab{6}}.

\bibitem[Zhang et~al.(2024{\natexlab{9}})Zhang, Chen, Jin, Wang, Ji, Wang, and Han]{zhang2024comprehensive}
Yu~Zhang, Xiusi Chen, Bowen Jin, Sheng Wang, Shuiwang Ji, Wei Wang, and Jiawei Han.
\newblock A comprehensive survey of scientific large language models and their applications in scientific discovery.
\newblock \emph{arXiv preprint arXiv:2406.10833}, 2024{\natexlab{9}}.

\bibitem[Zhang et~al.(2025{\natexlab{7}})Zhang, Khalifa, Bhushan, Murphy, Logeswaran, Kim, Lee, Lee, and Wang]{zhang2025mlrc}
Yunxiang Zhang, Muhammad Khalifa, Shitanshu Bhushan, Grant~D Murphy, Lajanugen Logeswaran, Jaekyeom Kim, Moontae Lee, Honglak Lee, and Lu~Wang.
\newblock Mlrc-bench: Can language agents solve machine learning research challenges?
\newblock \emph{arXiv preprint arXiv:2504.09702}, 2025{\natexlab{7}}.

\bibitem[Zhang et~al.(2025{\natexlab{8}})Zhang, Su, Sun, Xi, Xiao, Zheng, Zhang, Liu, Zan, Sun, et~al.]{zhang2025seed}
Yuyu Zhang, Jing Su, Yifan Sun, Chenguang Xi, Xia Xiao, Shen Zheng, Anxiang Zhang, Kaibo Liu, Daoguang Zan, Tao Sun, et~al.
\newblock Seed-coder: Let the code model curate data for itself.
\newblock \emph{arXiv preprint arXiv:2506.03524}, 2025{\natexlab{8}}.

\bibitem[Zhang et~al.(2025{\natexlab{9}})Zhang, Cao, and Liao]{zhang2025enhancing}
Zhihan Zhang, Yixin Cao, and Lizi Liao.
\newblock Enhancing chart-to-code generation in multimodal large language models via iterative dual preference learning.
\newblock \emph{arXiv preprint arXiv:2504.02906}, 2025{\natexlab{9}}.

\bibitem[Zhao et~al.(2024)Zhao, Zhou, Xie, and Zhang]{zhao2024hierarchical}
Chenlong Zhao, Xiwen Zhou, Xiaopeng Xie, and Yong Zhang.
\newblock Hierarchical attention graph for scientific document summarization in global and local level.
\newblock \emph{arXiv preprint arXiv:2405.10202}, 2024.

\bibitem[Zhao et~al.(2025)Zhao, Ma, Xu, Kong, and Deng]{zhao2025biomaze}
Haiteng Zhao, Chang Ma, Fangzhi Xu, Lingpeng Kong, and Zhi-Hong Deng.
\newblock Biomaze: Benchmarking and enhancing large language models for biological pathway reasoning.
\newblock \emph{arXiv preprint arXiv:2502.16660}, 2025.

\bibitem[Zhao et~al.(2024{\natexlab{2}})Zhao, Yang, Shen, Lakkaraju, and Du]{zhao2024towards}
Haiyan Zhao, Fan Yang, Bo~Shen, Himabindu Lakkaraju, and Mengnan Du.
\newblock Towards uncovering how large language model works: An explainability perspective.
\newblock \emph{arXiv preprint arXiv:2402.10688}, 2024{\natexlab{2}}.

\bibitem[Zhao et~al.(2025{\natexlab{2}})Zhao, Pei, Tao, Mei, and Shou]{zhao2025interfeedback}
Henry~Hengyuan Zhao, Wenqi Pei, Yifei Tao, Haiyang Mei, and Mike~Zheng Shou.
\newblock Interfeedback: Unveiling interactive intelligence of large multimodal models via human feedback.
\newblock \emph{arXiv preprint arXiv:2502.15027}, 2025{\natexlab{2}}.

\bibitem[Zhao et~al.(2025{\natexlab{3}})Zhao, Xing, Dou, Tian, Tai, Yang, Cheng, and Li]{zhao2025words}
Penghai Zhao, Qinghua Xing, Kairan Dou, Jinyu Tian, Ying Tai, Jian Yang, Ming-Ming Cheng, and Xiang Li.
\newblock From words to worth: Newborn article impact prediction with llm.
\newblock In \emph{Proceedings of the AAAI Conference on Artificial Intelligence}, volume~39, pages 1183--1191, Jan 2025{\natexlab{3}}.
\newblock URL \url{https://ojs.aaai.org/index.php/AAAI/article/view/32106/34261}.

\bibitem[Zhao et~al.(2023)Zhao, Zhou, Li, Tang, Wang, Hou, Min, Zhang, Zhang, Dong, et~al.]{zhao2023survey}
Wayne~Xin Zhao, Kun Zhou, Junyi Li, Tianyi Tang, Xiaolei Wang, Yupeng Hou, Yingqian Min, Beichen Zhang, Junjie Zhang, Zican Dong, et~al.
\newblock A survey of large language models.
\newblock \emph{arXiv preprint arXiv:2303.18223}, 1\penalty0 (2), 2023.

\bibitem[Zhao et~al.(2025{\natexlab{4}})Zhao, Luo, Shi, Chen, Wang, Liu, and Sun]{zhao2025chartcoder}
Xuanle Zhao, Xianzhen Luo, Qi~Shi, Chi Chen, Shuo Wang, Zhiyuan Liu, and Maosong Sun.
\newblock Chartcoder: Advancing multimodal large language model for chart-to-code generation.
\newblock \emph{arXiv preprint arXiv:2501.06598}, 2025{\natexlab{4}}.

\bibitem[Zhao et~al.(2025{\natexlab{5}})Zhao, Sang, Li, Shi, Wang, Zhang, Han, Liu, and Sun]{zhao2025autoreproduce}
Xuanle Zhao, Zilin Sang, Yuxuan Li, Qi~Shi, Shuo Wang, Duzhen Zhang, Xu~Han, Zhiyuan Liu, and Maosong Sun.
\newblock Autoreproduce: Automatic ai experiment reproduction with paper lineage.
\newblock \emph{arXiv preprint arXiv:2505.20662}, 2025{\natexlab{5}}.

\bibitem[Zhao et~al.(2023{\natexlab{2}})Zhao, Li, and Kong]{zhao2023decomposing}
Xueliang Zhao, Wenda Li, and Lingpeng Kong.
\newblock Decomposing the enigma: Subgoal-based demonstration learning for formal theorem proving.
\newblock \emph{arXiv preprint arXiv:2305.16366}, 2023{\natexlab{2}}.

\bibitem[Zhao et~al.(2025{\natexlab{6}})Zhao, Wang, Li, and Cohan]{zhao-etal-2025-multimodal-foundation}
Yilun Zhao, Chengye Wang, Chuhan Li, and Arman Cohan.
\newblock Can multimodal foundation models understand schematic diagrams? an empirical study on information-seeking {QA} over scientific papers.
\newblock In \emph{Findings of the Association for Computational Linguistics: ACL 2025}, pages 18598--18631, 2025{\natexlab{6}}.

\bibitem[Zhao et~al.(2025{\natexlab{7}})Zhao, Zhang, Hu, Wu, Bras, Anderson, Bragg, Chang, Dodge, Latzke, et~al.]{zhao2025sciarena}
Yilun Zhao, Kaiyan Zhang, Tiansheng Hu, Sihong Wu, Ronan~Le Bras, Taira Anderson, Jonathan Bragg, Joseph~Chee Chang, Jesse Dodge, Matt Latzke, et~al.
\newblock Sciarena: An open evaluation platform for foundation models in scientific literature tasks.
\newblock \emph{arXiv preprint arXiv:2507.01001}, 2025{\natexlab{7}}.

\bibitem[Zhao et~al.(2025{\natexlab{8}})Zhao, Zhao, Wang, and Wang]{zhao2025artificial}
Yiming Zhao, Yongjia Zhao, Jian Wang, and Zhuo Wang.
\newblock Artificial intelligence meets laboratory automation in discovery and synthesis of metal--organic frameworks: A review.
\newblock \emph{Industrial \& Engineering Chemistry Research}, 64\penalty0 (9):\penalty0 4637--4668, Feb 2025{\natexlab{8}}.

\bibitem[Zheng et~al.(2023)Zheng, Li, Wei, Chen, Qin, and Che]{zheng2023hit}
Bo~Zheng, Zhouyang Li, Fuxuan Wei, Qiguang Chen, Libo Qin, and Wanxiang Che.
\newblock Hit-scir at mmnlu-22: Consistency regularization for multilingual spoken language understanding.
\newblock \emph{arXiv preprint arXiv:2301.02010}, 2023.

\bibitem[Zheng et~al.(2021)Zheng, Han, and Polu]{zheng2021minif2f}
Kunhao Zheng, Jesse~Michael Han, and Stanislas Polu.
\newblock Minif2f: a cross-system benchmark for formal olympiad-level mathematics.
\newblock \emph{arXiv preprint arXiv:2109.00110}, 2021.

\bibitem[Zheng et~al.(2024)Zheng, Feng, Si, She, Lin, Jiang, and Wang]{zheng2024multimodal}
Mingyu Zheng, Xinwei Feng, Qingyi Si, Qiaoqiao She, Zheng Lin, Wenbin Jiang, and Weiping Wang.
\newblock Multimodal table understanding.
\newblock In \emph{Proceedings of the 62nd Annual Meeting of the Association for Computational Linguistics (Volume 1: Long Papers)}, pages 9102--9124, Jun 2024.

\bibitem[Zheng et~al.(2023{\natexlab{2}})Zheng, Qu, Cui, Shi, and Yin]{10.1145/3579355}
Ruiqi Zheng, Liang Qu, Bin Cui, Yuhui Shi, and Hongzhi Yin.
\newblock Automl for deep recommender systems: A survey.
\newblock \emph{ACM Trans. Inf. Syst.}, 41\penalty0 (4), March 2023{\natexlab{2}}.
\newblock ISSN 1046-8188.
\newblock \doi{10.1145/3579355}.
\newblock URL \url{https://doi.org/10.1145/3579355}.

\bibitem[Zheng et~al.(2025)Zheng, Cheng, Yao, Wu, Ding, Cheng, Hu, Bai, Zhou, Cui, et~al.]{zheng2025scaling}
Shenghe Zheng, Qianjia Cheng, Junchi Yao, Mengsong Wu, Ning Ding, Yu~Cheng, Shuyue Hu, Lei Bai, Dongzhan Zhou, Ganqu Cui, et~al.
\newblock Scaling physical reasoning with the physics dataset.
\newblock \emph{arXiv preprint arXiv:2506.00022}, 2025.

\bibitem[Zheng et~al.(2025{\natexlab{2}})Zheng, Deng, Tsang, Wang, Bai, Wang, and Song]{zheng2025automation}
Tianshi Zheng, Zheye Deng, Hong~Ting Tsang, Weiqi Wang, Jiaxin Bai, Zihao Wang, and Yangqiu Song.
\newblock From automation to autonomy: A survey on large language models in scientific discovery.
\newblock \emph{arXiv preprint arXiv:2505.13259}, 2025{\natexlab{2}}.

\bibitem[Zheng and Zhang(2025)]{zheng2025usage}
Xiaofeng Zheng and Jian Zhang.
\newblock The usage of a transformer based and artificial intelligence driven multidimensional feedback system in english writing instruction.
\newblock \emph{Scientific Reports}, 15\penalty0 (1):\penalty0 1--22, Jun 2025.

\bibitem[Zhong et~al.(2024)Zhong, Liu, Pan, Zhang, Zhou, Liang, Wu, Lyu, Shu, Yu, et~al.]{zhong2024evaluation}
Tianyang Zhong, Zhengliang Liu, Yi~Pan, Yutong Zhang, Yifan Zhou, Shizhe Liang, Zihao Wu, Yanjun Lyu, Peng Shu, Xiaowei Yu, et~al.
\newblock Evaluation of openai o1: Opportunities and challenges of agi.
\newblock \emph{arXiv preprint arXiv:2409.18486}, 2024.

\bibitem[Zhou et~al.(2025)Zhou, Li, Liao, Zhang, Qi, Wu, and Yang]{zhou2025academicbrowse}
Junting Zhou, Wang Li, Yiyan Liao, Nengyuan Zhang, Tingjia Miaoand~Zhihui Qi, Yuhan Wu, and Tong Yang.
\newblock Academicbrowse: Benchmarking academic browse ability of llms.
\newblock \emph{arXiv preprint arXiv:2506.13784}, 2025.

\bibitem[Zhou et~al.(2025{\natexlab{2}})Zhou, Zhang, Dai, Hershcovich, and Li]{zhou2025large}
Li~Zhou, Ruijie Zhang, Xunlian Dai, Daniel Hershcovich, and Haizhou Li.
\newblock Large language models penetration in scholarly writing and peer review.
\newblock \emph{arXiv preprint arXiv:2502.11193}, 2025{\natexlab{2}}.

\bibitem[Zhou et~al.(2024)Zhou, Chen, and Yu]{zhou2024llm}
Ruiyang Zhou, Lu~Chen, and Kai Yu.
\newblock Is llm a reliable reviewer? a comprehensive evaluation of llm on automatic paper reviewing tasks.
\newblock In \emph{Proceedings of the 2024 Joint International Conference on Computational Linguistics, Language Resources and Evaluation (LREC-COLING 2024)}, pages 9340--9351, May 2024.

\bibitem[Zhou et~al.(2024{\natexlab{2}})Zhou, Liu, Srivastava, Mei, and Tan]{zhou2024hypothesis}
Yangqiaoyu Zhou, Haokun Liu, Tejes Srivastava, Hongyuan Mei, and Chenhao Tan.
\newblock Hypothesis generation with large language models.
\newblock \emph{arXiv preprint arXiv:2404.04326}, 2024{\natexlab{2}}.

\bibitem[Zhou et~al.(2024{\natexlab{3}})Zhou, Xiong, Savarese, and Wu]{zhou2024shared}
Yilun Zhou, Caiming Xiong, Silvio Savarese, and Chien-Sheng Wu.
\newblock Shared imagination: Llms hallucinate alike.
\newblock \emph{arXiv preprint arXiv:2407.16604}, 2024{\natexlab{3}}.

\bibitem[Zhu et~al.(2025)Zhu, Silva, Cadena, Soper, Lisicki, Peetoom, Baranzini, Sundaram, Ray, and Drocco]{zhu2025deep}
Haonan Zhu, Mary Silva, Jose Cadena, Braden Soper, Micha{\l} Lisicki, Braian Peetoom, Sergio~E Baranzini, Shivshankar Sundaram, Priyadip Ray, and Jeff Drocco.
\newblock Deep active learning based experimental design to uncover synergistic genetic interactions for host targeted therapeutics.
\newblock \emph{arXiv preprint arXiv:2502.01012}, 2025.

\bibitem[Zhu et~al.(2023)Zhu, Feng, Feng, Wu, and Qin]{zhu2023hierarchical}
Kun Zhu, Xiaocheng Feng, Xiachong Feng, Yingsheng Wu, and Bing Qin.
\newblock Hierarchical catalogue generation for literature review: a benchmark.
\newblock \emph{arXiv preprint arXiv:2304.03512}, 2023.

\bibitem[Zhu et~al.(2025{\natexlab{2}})Zhu, Zhang, Qi, Shang, Liu, Han, Su, Yu, and You]{zhu2025safescientist}
Kunlun Zhu, Jiaxun Zhang, Ziheng Qi, Nuoxing Shang, Zijia Liu, Peixuan Han, Yue Su, Haofei Yu, and Jiaxuan You.
\newblock Safescientist: Toward risk-aware scientific discoveries by llm agents.
\newblock \emph{arXiv preprint arXiv:2505.23559}, 2025{\natexlab{2}}.

\bibitem[Zhu et~al.(2025{\natexlab{3}})Zhu, Weng, Yang, and Zhang]{zhu2025deepreview}
Minjun Zhu, Yixuan Weng, Linyi Yang, and Yue Zhang.
\newblock Deepreview: Improving llm-based paper review with human-like deep thinking process.
\newblock \emph{arXiv preprint arXiv:2503.08569}, 2025{\natexlab{3}}.

\bibitem[Zhu et~al.(2025{\natexlab{4}})Zhu, Xie, Weng, Wu, Lin, Yang, and Zhang]{zhu2025ai}
Minjun Zhu, Qiujie Xie, Yixuan Weng, Jian Wu, Zhen Lin, Linyi Yang, and Yue Zhang.
\newblock Ai scientists fail without strong implementation capability.
\newblock \emph{arXiv preprint arXiv:2506.01372}, 2025{\natexlab{4}}.

\bibitem[Zhu et~al.(2024)Zhu, Huang, Zhou, Zhao, Guo, Yang, Sun, Luo, Zhang, Xiao, et~al.]{zhu2024automated}
Qing Zhu, Yan Huang, Donglai Zhou, Luyuan Zhao, Lulu Guo, Ruyu Yang, Zixu Sun, Man Luo, Fei Zhang, Hengyu Xiao, et~al.
\newblock Automated synthesis of oxygen-producing catalysts from martian meteorites by a robotic ai chemist.
\newblock \emph{Nature Synthesis}, 3\penalty0 (3):\penalty0 319--328, 2024.

\bibitem[Zhu et~al.(2024{\natexlab{2}})Zhu, Guo, Qi, Chu, Yu, and Li]{zhu2024survey}
Ronghang Zhu, Dongliang Guo, Daiqing Qi, Zhixuan Chu, Xiang Yu, and Sheng Li.
\newblock A survey of trustworthy representation learning across domains.
\newblock \emph{ACM Transactions on Knowledge Discovery from Data}, 18\penalty0 (7):\penalty0 1--53, Jun 2024{\natexlab{2}}.

\bibitem[Zhu et~al.(2015)Zhu, Turney, Lemire, and Vellino]{zhu2015measuring}
Xiaodan Zhu, Peter Turney, Daniel Lemire, and Andr{\'e} Vellino.
\newblock Measuring academic influence: Not all citations are equal.
\newblock \emph{Journal of the Association for Information Science and Technology}, 66\penalty0 (2):\penalty0 408--427, Jan 2015.

\bibitem[Zhuang and Kennington(2024)]{zhuang-kennington-2024-understanding}
Jun Zhuang and Casey Kennington.
\newblock Understanding survey paper taxonomy about large language models via graph representation learning.
\newblock In Tirthankar Ghosal, Amanpreet Singh, Anita Waard, Philipp Mayr, Aakanksha Naik, Orion Weller, Yoonjoo Lee, Shannon Shen, and Yanxia Qin, editors, \emph{Proceedings of the Fourth Workshop on Scholarly Document Processing (SDP 2024)}, pages 58--69, Bangkok, Thailand, August 2024. Association for Computational Linguistics.
\newblock URL \url{https://aclanthology.org/2024.sdp-1.6/}.

\bibitem[Zhuang et~al.(2025)Zhuang, Chen, Xu, Jiang, and Lin]{zhuang2025large}
Zhenzhen Zhuang, Jiandong Chen, Hongfeng Xu, Yuwen Jiang, and Jialiang Lin.
\newblock Large language models for automated scholarly paper review: A survey.
\newblock \emph{arXiv preprint arXiv:2501.10326}, 2025.

\bibitem[Zhuang et~al.(2023)Zhuang, Chen, Ma, Li, Han, Qian, Bai, Feng, Zhang, and Liu]{zhuang2023through}
Ziyu Zhuang, Qiguang Chen, Longxuan Ma, Mingda Li, Yi~Han, Yushan Qian, Haopeng Bai, Zixian Feng, Weinan Zhang, and Ting Liu.
\newblock Through the lens of core competency: Survey on evaluation of large language models.
\newblock \emph{arXiv preprint arXiv:2308.07902}, 2023.

\bibitem[Zimmermann et~al.(2025)Zimmermann, Bazgir, Al-Feghali, Ansari, Brinson, Chiang, Circi, Chiu, Daelman, Evans, et~al.]{zimmermann202534}
Yoel Zimmermann, Adib Bazgir, Alexander Al-Feghali, Mehrad Ansari, L~Catherine Brinson, Yuan Chiang, Defne Circi, Min-Hsueh Chiu, Nathan Daelman, Matthew~L Evans, et~al.
\newblock 34 examples of llm applications in materials science and chemistry: Towards automation, assistants, agents, and accelerated scientific discovery.
\newblock \emph{arXiv preprint arXiv:2505.03049}, 2025.

\bibitem[Zou and Topol(2025)]{zou2025rise}
James Zou and Eric~J Topol.
\newblock The rise of agentic ai teammates in medicine.
\newblock \emph{The Lancet}, 405\penalty0 (10477):\penalty0 457, Feb 2025.

\bibitem[Zournas et~al.(2025)Zournas, Incha, Radivojevic, Blay, Mart{\'\i}, Costello, Schimdt, Chung, Thompson, Pearson, et~al.]{zournas2025machine}
Apostolos Zournas, Matthew~R Incha, Tijana Radivojevic, Vincent Blay, Jose~Manuel Mart{\'\i}, Zak Costello, Matthias Schimdt, Tan Chung, Mitchell~G Thompson, Allison Pearson, et~al.
\newblock Machine learning-led semi-automated medium optimization reveals salt as key for flaviolin production in pseudomonas putida.
\newblock \emph{Communications Biology}, 8\penalty0 (1):\penalty0 630, Apr 2025.

\end{thebibliography}
